\documentclass[a4paper,12pt,times,numbered,print,index ]{Classes/PhDThesisPSnPDF}
\input{Preamble/preamble}

\title{Dilated Convolution with Learnable Spacings}

\author{Ismail Khalfaoui Hassani}

\dept{EDMITT - École Doctorale Mathématiques, Informatique et
Télécommunications de Toulouse}

\supervisor{Prof. Timothée Masquelier}

\advisor{Prof. Nicolas Thome \newline
Prof. Emre Neftci \newline
Prof. Sylvie Chambon \newline
Prof. Gaël Richard \newline
Prof. Fisher Yu \newline
Prof. Timothée Masquelier}
     
\degreetitle{Doctor of Philosophy}

\college{University of Toulouse}

\degreedate{March 2024} 

\subject{LaTeX} \keywords{{LaTeX} {PhD Thesis} {Computer Science} {University of
Toulouse}}

\ifdefineAbstract
\fi

\ifdefineChapter
\fi

\begin{document}

\frontmatter

\pdfbookmark[section]{Title page}{}

\includepdf[pages=-]{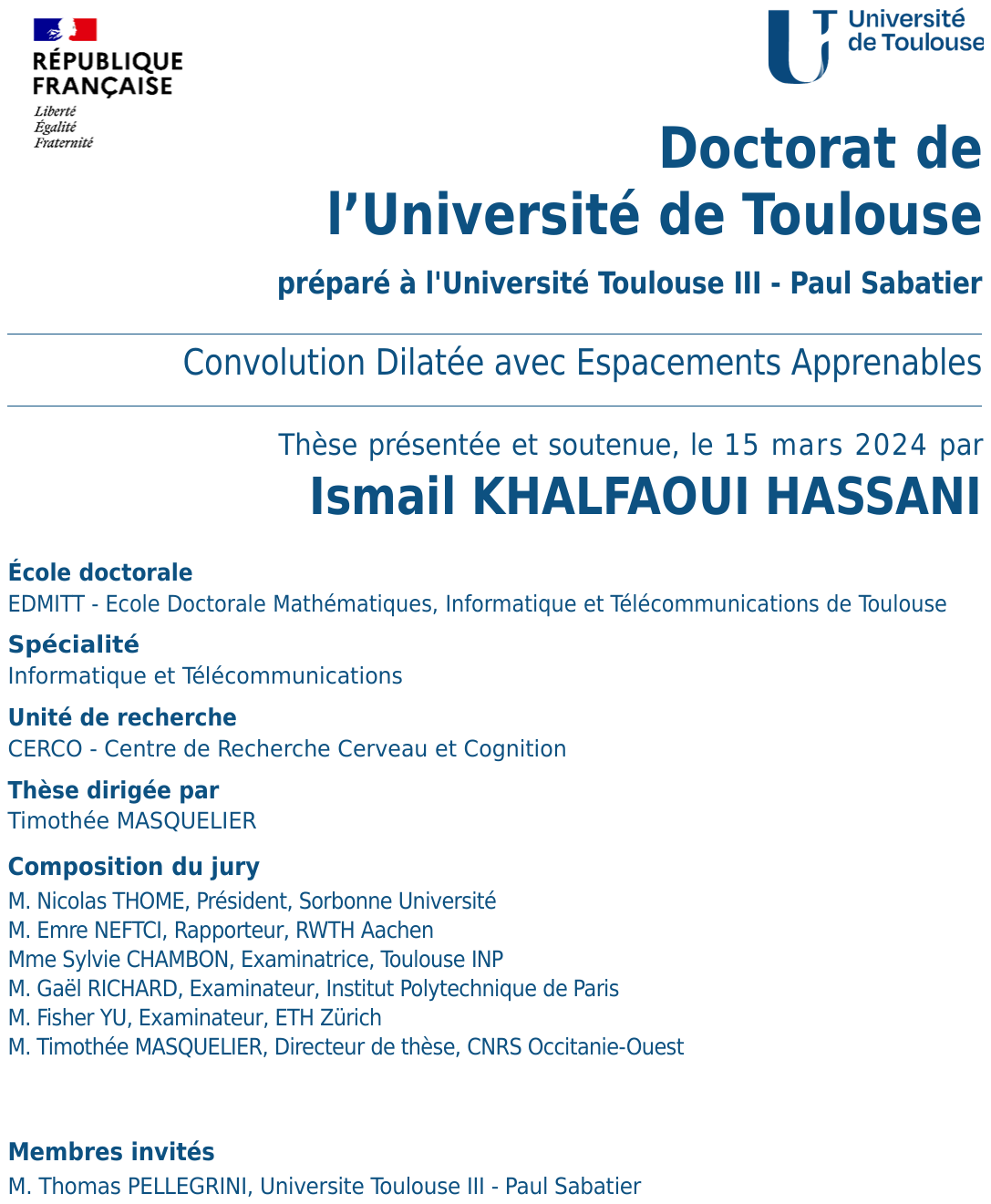}

\pdfbookmark[section]{Dedication}{dedication}
\begin{dedication} 

\vspace*{\fill}
\epigraph{Sometimes the discoveries are simultaneous or almost so; sometimes a scientist will make anew a discovery which, unknown to him, somebody else had made years before.}{Robert K. Merton, \textit{ Resistance to the systematic study of multiple discoveries in science.}}

\end{dedication}

\pdfbookmark[section]{Acknowledgements}{acknowledgements}
\begin{acknowledgements}      

I would like to express my deep gratitude to the institutions and people who made my Ph.D. possible. First and foremost, I would like to thank the French National Research Agency (ANR) and the Artificial and Natural Intelligence Toulouse Institute (ANR-3IA ANITI), as well as the Toulouse Occitanie region, for generously funding my Ph.D. studies. 

I am also grateful to the ``AI for physical models with geometric tools'' chair, headed by Prof. Fabrice Gamboa, for their continuous support.

My work has greatly benefited from the HPC resources of IDRIS and GENCI's Jean Zay computing center (grants 2021-[AD011013219] and 2023-[AD011013219R1]) and CALMIP computing center (grants 2021-[P21052] and 2023-[P22021]). I sincerely appreciate the contributions of all the staff at both computing centers for enabling us researchers and graduate students to achieve our research goals.

Heartfelt gratitude is owed to my thesis supervisor, Timothée Masquelier, for believing in me and proposing this thesis topic, on which we worked together for a little more than 3 years. We have always worked in an atmosphere of trust and scientific exchange.  I particularly appreciated the fact that Timothée's office was always open to me. Whatever ideas I had, good or bad, Tim was there to hear them and to suggest improvements or contradict me if they weren't relevant. I particularly appreciated Timothée's ability to adapt his guidance to my needs. I never felt belittled, but always challenged and supported, and I sincerely thank him for that.

My sincere appreciation goes to my thesis co-supervisor Thomas Pellegrini for his mentorship, for our weekly exchanges, and for being there when I really needed him!

I would like to extend my deepest thanks to the jury of my PhD thesis for their invaluable time, insightful questions, and constructive comments. In particular, I express my gratitude to the rapporteurs of my thesis, Prof. Nicolas Thome and Prof. Emre Neftci, for their thorough examination and valuable feedback.

I would like to thank my co-authors: Ilyass Hammouamri from the NeuroAI team at CerCo and Etienne Labbé from the SAMOVA team at IRIT.

I would also like to thank Wei Fang, although I didn't work with him directly, for developing the SpikingJelly framework that we used in the fourth chapter of my thesis.

I want to express my thanks to the research teams in which I have worked over the last few years, in particular the NeuroAI team at CerCo. I would especially like to thank the permanent members of the team, Timothée Masquelier, Rufin VanRullen for his advice and wise remarks, and Andrea Alamia, whom I knew as a post-doc, then as a researcher at the CNRS and finally as an HDR researcher.

Many thanks to all those with whom I shared a workspace, especially my office colleagues, for their patience and understanding during our sometimes lively exchanges. In particular, I would like to thank Andrea Alamia, Colin Decourt, Leslie Marie-Louise, Aimen Zerroug, Mohit Vaishnav, Jakob Schwenk, Xiaoqi Xu, Martina Pasqualetti, and especially my friends and colleagues Leopold Maytié, Benjamin Devillers, Ulysse Rançon, and Javier Cuadrado Aníbarro. 

I am grateful to Rufin VanRullen's group: Sabine Muzellec, Furkan Ôzçelik, Victor Boutin, the newcomers Mitja Nikolaus, Rolland Bertin-Johannet, Hugo Chateau-Laurent, and Lara Scipio, those who left: Milad Mozafari, Anaïs Servais, Ana Szabo, Bhavin Choksi, and Zhaoyang Pang. Benedikt Zoefel's group: Florian Kasten, Troby Ka-Yan Lui, Jules Erkens, Ram Pari, and Marina Inyutina.

Special appreciation for the technical support team at CerCo, especially Damien Mateo, for his help and for showing me an A40 graphics card for the first time. I owe a debt of gratitude to Prof. Robin Baures for lending us the first GPU that saw the beginnings of DCLS.

I would like to thank the SAMOVA team at IRIT, and in general, I would like to thank the IRIT lab as well as the CerCo lab, the successive directors of the CerCo lab, Prof. Simon Thorpe and then Prof. Isabelle Berry, all the research staff, postdocs, graduate students, engineers and trainees, those I've known for three years as well as those I just met a week ago. 

I'm grateful to everyone who has taught me something, from my teachers to my everyday colleagues, to those who have shared their wisdom and life experiences with me. In particular, I extend my deep gratitude to thank Prof. Daniel Ruiz for opening the door to academic research for me and Philippe Leleux for guiding my first research project. A heartfelt thank you to Maroufatou Salami for her kindness and wisdom.

I am very thankful to my friends: Christophe Sun for the years spent at ENSEEIHT,  Valentin Durante for taking it easy on me during our boxing matches, Anas Dehmani for the video game sessions, and Yassine Kossir for his support and advice.

I would also like to express my deep gratitude to my family - my brother and my parents - whose unfailing support during my studies has been a constant source of motivation since I was very young.

I cannot express enough gratitude to my mother, who taught me how to read, write, and count. My father who taught me spelling and cross-multiplication, sometimes with great pain. I thank them from the bottom of my heart for putting up with me, it's not easy to live with someone like me from one day to the next. 

Last but not least, my warmest thanks go to my girlfriend for her unwavering support and love throughout this course. 
I may not have named everyone, but please know that I am grateful to all of you.

I'll end my acknowledgments with a quote from a wise man:
``Teach people what you know and learn what others know. That way you can master your own knowledge and learn what you didn't know before.''

\end{acknowledgements}

\pdfbookmark[section]{Abstract}{abstract}
\begin{abstract}
In this thesis, we develop and study the Dilated Convolution with Learnable Spacings (DCLS) method. The DCLS method can be considered as an extension of the standard dilated convolution method, but in which the positions of the weights of a neural network are learned during training by the gradient backpropagation algorithm, thanks to an interpolation technique. We empirically demonstrate the effectiveness of the DCLS method by providing concrete evidence from numerous supervised learning experiments. These experiments are drawn from the fields of computer vision, audio, and speech processing, and all show that the DCLS method has a competitive advantage over standard convolution techniques, as well as over several advanced convolution methods.

Our approach is structured in several steps, starting with an analysis of the literature and existing convolution techniques that preceded the development of the DCLS method. We were particularly interested in the methods that are closely related to our own and that remain essential to capture the nuances and uniqueness of our approach.

The cornerstone of our study is the introduction and application of the DCLS method to convolutional neural networks (CNNs), as well as to hybrid architectures that rely on both convolutional and visual attention approaches. The DCLS method is particularly noteworthy for its capabilities in supervised computer vision tasks such as classification, semantic segmentation, and object detection, all of which are essential tasks in the field.

Having originally developed the DCLS method with bilinear interpolation, we explored other interpolation methods that could replace the bilinear interpolation conventionally used in DCLS, and which aim to make the position parameters of the weights in the convolution kernel differentiable. Gaussian interpolation proved to be slightly better in terms of performance.

Our research then led us to apply the DCLS method in the field of spiking neural networks (SNNs) to enable synaptic delay learning within a neural network that could eventually be transferred to so-called neuromorphic chips. The results show that the DCLS method stands out as a new state-of-the-art technique in SNN audio classification for certain benchmark tasks in this field. These tasks involve datasets with a high temporal component. In addition, we show that DCLS can significantly improve the accuracy of artificial neural networks for the multi-label audio classification task, a key achievement in one of the most important audio classification benchmarks.

We conclude with a discussion of the chosen experimental setup, its limitations, the limitations of our method, and our results.

\end{abstract}

\pdfbookmark[section]{Résumé}{Résumé}
\cleardoublepage
\setsinglecolumn
\chapter*{\centering \Large Résumé} 
\thispagestyle{empty}
Dans cette thèse, nous avons développé et étudié la méthode de convolution dilatée avec espacements apprenables (Dilated Convolution with Learnable Spacings en anglais, qu'on abrégera par le sigle DCLS). La méthode DCLS peut être considérée comme une extension de la méthode de convolution dilatée standard, mais dans laquelle les positions des poids d'un réseau de neurones sont apprises grâce à l'algorithme de rétropropagation du gradient, et ce, à l'aide d'une technique d'interpolation. Par suite, nous avons démontré empiriquement l'efficacité de la méthode DCLS en fournissant des preuves concrètes, issues de nombreuses expériences en apprentissage supervisé. Ces expériences sont issues des domaines de la vision par ordinateur, de l'audio et du traitement de la parole et toutes montrent que la méthode DCLS a un avantage compétitif sur les techniques standards de convolution ainsi que sur plusieurs méthodes de convolution avancées.

Notre approche s'est faite en plusieurs étapes, en commençant par une analyse de la littérature et des techniques de convolution existantes qui ont précédé le développement de la méthode DCLS. Nous nous sommes particulièrement intéressés aux méthodes étroitement liées à la nôtre et qui demeurent essentielles pour saisir les nuances ainsi que le caractère unique de notre approche.

La pierre angulaire de notre étude repose sur l'introduction et l'application de la méthode DCLS aux réseaux neuronaux convolutifs (CNN), mais aussi aux architectures hybrides qui se basent à la fois sur des méthodes convolutives et des méthodes d'attention visuelle. La méthode DCLS est particulièrement remarquable pour ses capacités dans les tâches supervisées de vision par ordinateur telles que la classification, la segmentation et la détection d'objets, qui sont toutes des tâches essentielles dans ce domaine.

Ayant développé la méthode DCLS à l'origine avec une interpolation bilinéaire, nous avons entrepris l'exploration d'autres méthodes d'interpolation susceptibles de remplacer l'interpolation bilinéaire, traditionnellement utilisée dans DCLS, ainsi que d'autres méthodes de convolution, et qui visent à rendre différentiables les paramètres de positions des poids dans le noyau de convolution. L'interpolation gaussienne s'est avérée être légèrement meilleure en termes de performances.

Notre recherche nous a amené par la suite à appliquer la méthode DCLS dans le domaine des réseaux de neurones à spikes (SNN) afin de permettre l'apprentissage des délais synaptiques à l'intérieur d'un réseau de neurones qui pourrait être éventuellement transféré à des puces dites neuromorphiques. Les résultats montrent que la méthode DCLS se tient comme nouvel état de l'art des SNNs en classification audio pour certaines tâches de référence dans ce domaine. Ces dernières tâches portent sur des ensembles de données connus pour avoir une composante temporelle importante. En outre, nous montrons aussi que DCLS permet d'améliorer de manière significative la précision des réseaux neuronaux artificiels pour la tâche de classification audio multi-label, un aboutissement clé dans l'un des benchmarks de classification audio les plus importants.

Enfin, nous concluons par une discussion sur le dispositif expérimental choisi, ses limites, les limites de notre méthode et nos résultats.

\pdfbookmark[section]{List of publications}{publications}
\begin{publications} 

\subsection*{As first author}
\begin{itemize}
    \item Ismail Khalfaoui-Hassani, Timothée Masquelier, and Thomas Pellegrini. Audio classification with Dilated Convolution with Learnable Spacings. In \textbf{NeurIPS 2023} \textit{Workshop on Machine Learning for Audio}, New Orleans, USA. \citep{khalfaoui2023audio}.    
    \item Ismail Khalfaoui-Hassani, Thomas Pellegrini, and Timothée Masquelier. Dilated convolution with learnable spacings: beyond bilinear interpolation. In \textbf{ICML 2023} \textit{Workshop on Differentiable Almost Everything: Differentiable Relaxations, Algorithms, Operators, and Simulators}, Honolulu, Hawaii, USA. \citep{khalfaouihassani2023dilated}.    
    \item Ismail Khalfaoui-Hassani, Thomas Pellegrini, and Timothée Masquelier. Dilated convolution with learnable spacings. In \textbf{ICLR 2023}, Kigali, Rwanda. \citep{hassani2023dilated}.    
\end{itemize}
\subsection*{Collaborations}
\begin{itemize}
    \item Ilyass Hammouamri, Ismail Khalfaoui-Hassani, Timothée Masquelier. Learning Delays in Spiking Neural Networks using Dilated Convolutions with Learnable Spacings. Set to appear in \textbf{ICLR 2024},  Vienna, Austria. \citep{hammouamri2023learning}. 
    \item Thomas Pellegrini, Ismail Khalfaoui-Hassani, Etienne Labbé, and Timothée Masquelier. Adapting a ConvNeXt Model to Audio Classification on AudioSet. In \textbf{INTERSPEECH 2023}, pages 4169–4173, 2023. doi: 10.21437/Interspeech.2 023-1564. \citep{pellegrini2023adapting}. 
\end{itemize}
\subsubsection*{Submission}
\begin{itemize}
    \item Alireza Azadbakht, Saeed Reza Kheradpisheh, Ismail Khalfaoui-Hassani, Timothée Masquelier. Drastically Reducing the Number of Trainable Parameters in Deep CNNs by Inter-layer Kernel-sharing. arXiv preprint arXiv:2210.14151, 2022. \citep{azadbakht2022drastically}.
\end{itemize}
\end{publications}


\tableofcontents

\listoffigures

\listoftables


\printnomenclature

\mainmatter


\chapter{Introduction}  

\ifpdf
    \graphicspath{{Chapter1/Figs/Raster/}{Chapter1/Figs/PDF/}{Chapter1/Figs/}}
\else
    \graphicspath{{Chapter1/Figs/Vector/}{Chapter1/Figs/}}
\fi
\lettrine[lines=2]{W}{ hether} in mathematics, physics, deep learning, or science in general, the re-examination of certain values long considered constant has been the source of many scientific advances and breakthroughs.

In mathematics, for example, more specifically in the field of ordinary and partial differential equations, the concept of constant variation intuitively led to the method of variation of parameters, also known as variation of constants \cite{lagrange1868oeuvres}. The goal of this method is to find solutions to a non-homogeneous differential equation. By considering a particular solution that has the same form as the one found for the homogeneous case, and then by making its multiplicative constants variable, the variation of constants method allows in many cases to find the analytical solutions of the non-homogeneous differential equation without much effort.

In classical mechanics, physical quantities such as position, momentum, and energy are traditionally treated as fixed, deterministic values. However, quantum mechanics introduced a fundamental departure from this classical perspective by embracing the inherent variability and probabilistic nature of some physical phenomena.

In deep learning, incorporating learnable parameters or adaptive elements into the model architecture is a very common outlook. Some examples, such as adaptive pooling techniques \cite{lecun1989handwritten, lecun1998gradient} and normalization \cite{ioffe2015batch}, illustrate the use of this research approach. Adaptive learning rates \cite{duchi2011adaptive} are an umpteenth example of this same principle, where learning rates are adjusted over time based on the performance of the model.

Convolution and dilated convolution are fundamental operations at the heart of modern deep learning architectures. Convolution, a cornerstone of signal processing and image analysis, involves applying a filter or kernel to an input signal or image to extract features and detect patterns. Dilated convolution, expands upon traditional convolution by inflating the receptive field without compromising resolution. By incorporating gaps, or dilations, between kernel elements, dilated convolution captures a broader context, enabling the model to detect more global features while preserving fine-grained details. Both convolution and dilated convolution have proven indispensable in a myriad of applications, ranging from computer vision tasks like object detection and semantic segmentation to natural language processing challenges like text classification and language modeling.

Now it is our turn to put this principle of constant value re-examination into practice to call into question the possibility that the fixed grid imposed by default by the standard dilated convolution is something to improve and that positions of elements inside the dilated kernel could be learned by backpropagation throughout the learning process. 

As you will have gathered, this thesis aims to study the learning of the disposition of the weights in a dilated convolution kernel using gradient backpropagation, thus leading to a real departure from standard dilated convolution where the weights are arranged in a regular grid and what we have called Dilated Convolution with Learnable Spacings, or DCLS for short.

The principles of Occam's razor and scientific parsimony often favor the most simple explanations that fit the empirical data or in Einstein's elegant words: ``The supreme goal of any theory is to make the irreducible basic elements as simple and as few as possible without having to surrender the adequate representation of a single datum of experience''. Therefore, any deviation from the constant values of the fundamental constants would require strong evidence and extensive testing before being widely accepted by the scientific community. It is this same principle that forced Einstein to remove the cosmological constant term $\Lambda$ from his field equations of general relativity after Edwin Hubble's confirmation of the accelerated expansion of the universe. The fact remains, however, that this term was revisited in the 1990s after recent observations in cosmology, and that it would be the simplest explanation for the so-called dark energy in the $\Lambda$-CDM model of cosmology.

It is to this scientific rigor that we modestly aspire in this thesis, by demonstrating on the basis of several experimental results in computer vision and audio and speech processing that the DCLS method has an advantage over the standard dilated convolution method, as well as over several state-of-the-art convolution and classification methods. Therefore, this will be done in stages, first presenting in the current chapter a substantial state-of-the-art of convolution methods that predate the DCLS method, as well as some that are contemporary with it. This first chapter is not intended to be an exhaustive enumeration of all scientific contributions to the field of convolutions, but rather a description of the methods that are closest to our own, as well as those that will later facilitate our understanding of DCLS. Then, in the second chapter, by introducing the method and showing how it has led to state-of-the-art results using convolutional neural networks (CNNs) in classification, segmentation, object detection, and robustness tasks major benchmarks in the field.

The DCLS method relies on interpolation to overcome the problem of learning non-differentiable integer positions. The third chapter of this thesis will focus on possible interpolations that could supplant bilinear interpolation. The empirical finding will be that Gaussian interpolation is slightly better than the bilinear one.

While the second and third chapters are mainly concerned with vision tasks, the fourth and fifth chapters of this thesis will focus on applications of the DCLS method with Gaussian interpolation to audio classification tasks. First, in the fourth chapter, using Spiking Neural Networks (SNNs), akin to neural networks in living organisms, to enable synaptic delay learning, leading to a new state-of-the-art in two important benchmarks in this literature. Then, in the fifth chapter, we will show how the DCLS method can improve the accuracy of neural networks on the audio classification task on one of the most important benchmarks in audio classification.

The final chapter of this thesis will be a discussion of the observations and results found in the previous chapters, as well as a general conclusion and an outline of potential implications and improvements of the DCLS method. 

Without any further elaboration, we will begin by explaining the fundamental concepts behind the convolution methods used in deep learning, starting with standard convolution.

\section{Standard convolution}
Convolution is a fundamental operation extensively used in deep learning, particularly in Convolutional Neural Networks (CNNs), which have revolutionized various domains such as computer vision and natural language processing. At its core, convolution involves a sliding window approach, where a small matrix known as a kernel is moved across the input data. This kernel contains learnable parameters that determine its behavior. As the kernel slides over the input, it performs element-wise multiplication with the local data in the input grid. The products of these multiplications are summed up to create a single value in the output feature map. This process is repeated across the entire input, enabling the network to capture local patterns, features, and spatial relationships present within the data.

Convolution plays a crucial role in tasks like image recognition. For instance, in the image classification task, the initial layers of a CNN might detect simple features like edges or corners. As the network progresses deeper, subsequent layers combine these basic features to recognize more complex structures like shapes or textures. This hierarchical feature extraction is achieved through the convolution operation. Additionally, convolutional networks leverage parameter sharing – the same kernel is used at different positions across the input. This sharing of parameters enables the network to learn and recognize the same feature regardless of its location in the input data, which is particularly useful for achieving translational invariance.

Again in vision, more specifically in downstream tasks such as object detection or image segmentation, convolutions could be used to fine-tune a neural network with the help of features learned by a pre-trained backbone. The backbone usually consists of a bigger neural network that has been trained on a classification task. The features used for the downstream task are then selected at a given depth of the backbone. Convolutional techniques are also used to estimate optical flow \cite{cuadrado2023optical}, which is the pattern of apparent motion of objects between consecutive frames in image sequences. This is essential in various applications like video stabilization, object tracking, and analyzing fluid dynamics.

As with images, convolution proves crucial for video understanding tasks such as action recognition, tracking objects, and motion estimation.  It is also proving to be central to autonomous driving \cite{decourt2022recurrent} by providing aid in detecting pedestrians, vehicles, and road signs in real-time, enabling autonomous vehicles to navigate safely.

Convolutional operations extend beyond images. In natural language processing, for example, one-dimensional convolutions can be applied to sequential data, like text \cite{conneau2017very}. By treating words or characters as discrete data points, convolutional filters can scan through sequences to identify patterns and relationships. This approach has been applied to tasks like text classification and sentiment analysis. 

Furthermore, convolutions can be used in time-series analysis to predict stock prices, identify market trends, and detect anomalies. They are also used in speech recognition, music analysis, and sound classification where they could help identify acoustic features and temporal patterns in audio data.

These and many more applications that we haven't covered here due to space and time constraints. In summary, convolution has been used extensively in machine learning for increasingly complex tasks over time.

In this thesis, we will limit our experiments to the case of feed-forward neural networks and convolutional operations of dimension two at most. This does not mean that the results presented here cannot be applied to a higher dimension case, but rather that this has not been explored yet and represents a very promising avenue of research. This last remark also applies to non-feed-forward network types (such as recurrent neural networks (RNN) or graph neural networks (GNN)), where convolutional methods are widely used.

\subsection{Parameters, sizes and terminology}
In general, standard convolution has two learnable parameters: the weights ($W$), which serve as synaptic weights, and the bias parameter ($b$). In addition, the standard convolution is parameterized by several non-learnable hyperparameters. These are positive integer hyperparameters that control the way the convolution is applied, the dimensionality of the learnable parameters, and/or the dimensionality of the output, the latter of which are listed below: 

\begin{itemize}
    \item Dimension of the convolution ($d \in \mathbb{N}^*$). This corresponds to the dimensionality of the used convolution.
    \item Input channels ($C_{in} \in \mathbb{N}^*$), or input neurons. They correspond to the input channels of the signal or the input neurons of a previous feature map in the feed-forward network.
    \item Output channels ($C_{out} \in \mathbb{N}^*$), or output neurons. They correspond to the output channels/neurons created by the convolution.
    \item Kernel size ($K =(k_1, k_2, \dots, k_d) \in {\mathbb{N}^*}^d$). The kernel size of the convolution.
    \item Padding ($Pad =(p_1, p_2, \dots, p_d) \in {\mathbb{N}}^d$) is the number of elements to add to each side of the input. This can be done in several ways such as adding zeros or replicating the edge values for example.
    \item Stride ($S = (s_1, s_2, \dots, s_d) \in {\mathbb{N}^*}^d$). This parameter quantifies the hop length that the kernel will take with respect to the input signal when convolution is applied.
    \item Dilation rate or factor ($Df = (df_1, df_2, \dots, df_d) \in {\mathbb{N}^*}^d$) stands for the spacings between kernel elements. This allows the kernel to be inflated without adding extra elements in the case of what is called the standard dilated convolution (as the elements are evenly spaced at a \textit{fixed} rate on the grid) and will be explained in section~\ref{sec:dilated}. 
    \item Groups ($G \in \mathbb{N}^*$). This parameter is a count of the number of separate channels to which the convolution will be applied. This number must be a divisor of $C_{in}$ and $C_{out}$. The extreme case of $G = C_{in}$ is what is known as the depthwise convolution that will be covered in Section~\ref{sec:depthwise}. 
\end{itemize}

Despite not being a true hyperparameter, we can include the batch size ($B$) alongside the previously mentioned hyperparameters. The batch size refers to the number of parallel inputs that the convolution can operate on independently. The processing of data in batches has been made much easier in the modern era of deep learning, thanks to parallel computing frameworks and the exponential growth of hardware and infrastructures that accelerate these calculations. Additionally, artificial neural network optimization heavily depends on stochastic gradient descent methods, which require performing gradient steps on mini-batches of size $B$ chosen randomly among the data.

Now that we've defined the classical hyperparameters of the standard convolution, we can specify the sizes of its learnable parameters, which we'll denote by n-tuples. The weights $W$ are a tensor of size $(C_{out}, \frac{C_{in}}{G}, k_1, k_2, \dots, k_d)$ , while the bias $b$ is a vector of size $C_{out}$.
When applied to a batch of inputs of size $(B, C_{in}, H^{in}_1, H^{in}_2, \dots, H^{in}_d) \in {\mathbb{N}^*}^{d+2}$ The convolution operation produces a tensor of size $(B, C_{out}, H^{out}_1, H^{out}_2, \dots, H^{out}_d)$ with
\begin{equation}
\label{eq:output}
\forall i \in \llbracket 1 \ .. \ d \rrbracket \colon H^{out}_i = \left \lfloor{\frac{H^{in}_i + 2 \times pad_i - df_i \times (k_i - 1 ) - 1}{s_i} + 1} \right \rfloor
\end{equation}

In deep learning, the convolution operation performs the following calculation:
 $\forall i\in \llbracket 1 \ .. \ B \rrbracket$, $\forall j\in \llbracket 1 \ .. \ C_{out} \rrbracket  \colon $
\begin{equation}
    \text{output}(i, j) = \text{bias}(j) + \sum_{k=1}^{C_{in}} W(j,k) \star \text{input}(i,k)
\end{equation}

With the $\star$ operator standing for the $d$-dimensional discrete cross-correlation operator which is  defined in the 1D case for every two functions $f$ and $g$ as:
\begin{equation}
   \forall n \in \mathbb{N} \colon (f \star g) [n] \overset{\text{def}}{=}  \sum_{m=-\infty}^{+\infty}\overline{f[m]}g[n+m]
\end{equation}

Where $\overline{\phantom{X}}$ denotes the complex conjugate, which can be omitted in this context since all the signals considered are real numbers. The 1D cross-correlation operator is easily generalized to higher dimensions, as it is applied independently across them. Note that the convolution operation in deep learning is actually a cross-correlation, although the name ``convolution'' has been retained in this literature.

A stage in the context of a feed-forward neural network designates a sequence of blocks processing feature maps of the same resolution. A stage is thus delimited by two downsampling layers. A block designates a succession of layers. A layer, in the context of neural networks, is a generic term that could designate a convolution, a nominalization, an activation, etc. It is a basic element of a neural network that cannot be written as a sequence of other elements. For example, a convolution layer refers to a convolution operation occurring at a given level of the network, while a batch norm layer \cite{ioffe2015batch} refers to a batch norm operation occurring at another given layer. A ResNet50 model \cite{he2016deep} for example has 4 stages, 16 residual blocks, and 50 layers.

\subsection{Complexity}
\label{ex:standardconv}
In the following, we give the time and space complexity of the standard 2D convolution in dimension $d=2$ as an example. There are several ways of implementing standard convolution, some of which have more advantageous time complexity, but at the expense of memory. This is the case with the \textsc{im2col} algorithm \cite{chellapilla2006high}. This algorithm transforms the input image into a matrix which, when multiplied by the weight matrix, gives the convolution result. A very simple example is given below by way of demonstration.
\subsubsection{A minimal example}
\label{ex:minimal}

The dimension of the input chosen for this example is $d=2$ with a number of channels $C_{in}=1$ and spatial size of $4 \times 4$. We call this kind of input an image or a 2D matrix. The dimension of the input tensor is thus  $ (B,C_{in},H_1,H_2) = (1,1,4,4) $.
We put
\begin{equation}
\label{eq:image}
\renewcommand\arraystretch{1.3}
image = \left[
\begin{array}{c c c c}
 im_1 & im_2 & im_3 & im_4 \\
 im_5 & im_6 & im_7 & im_8 \\
 im_9 & im_{10} & im_{11} & im_{12} \\
 im_{13} & im_{14} & im_{15} & im_{16}
\end{array}
\right]
\end{equation}

The weights in this example are of kernel size $2 \times 2$ defined by $(C_{out},C_{in},k_1,k_2) = (1,1,2,2)$.
\begin{equation}
\label{eq:weight}
W = \left[
\begin{array}{c c c c}
 w_{11} & w_{12} \\
 w_{21} & w_{22}
\end{array}
\right]
\end{equation}

In this example, we chose the stride $S = (1,1)$, the dilation $Df = (1,1)$, the padding $Pad = (0,0)$ and the groups $G=1$.

The \textsc{im2col} produces an output of size $(C_{in} \cdot k_1 \cdot k_2, \  B \cdot H^{out}_1 \cdot H^{out}_2) = (4,9)$.
$H^{out}_1$ and $ H^{out}_2$ are calculated following \ref{eq:output}. And we have
\begin{equation}
\label{eq:image2col}
\renewcommand\arraystretch{1.3}
image2col = \left[
\begin{array}{c c c c c c c c c}
 im_1 & im_2 & im_3 & im_5    & im_6    & im_7    & im_9    & im_{10} & im_{11} \\
 im_2 & im_3 & im_4 & im_6    & im_7    & im_8    & im_{10} & im_{11} & im_{12} \\
 im_5 & im_6 & im_7 & im_9    & im_{10} & im_{11} & im_{13} & im_{14} & im_{15} \\
 im_6 & im_7 & im_8 & im_{10} & im_{11} & im_{12} & im_{14} & im_{15} & im_{116} 
 \end{array}
\right]
\end{equation}
\subsubsection{Space complexity}
The space complexity of the \textsc{im2col} algorithm depends on $B, C_{in}, K$ and $H^{out}$ and is given by the size of the \textsc{im2col} matrix. 
$$C_{in} \cdot k_1 \cdot k_2, \  B \cdot H^{out}_1 \cdot H^{out}_2$$

This is distinct from the number of learnable parameters in the convolution, which is simply the summation of the weight dimension and the bias dimension.
$$C_{out} \cdot C_{in} \cdot k_1 \cdot k_2 + C_{out} $$

For particular choices of $C_{in}$ and $H^{out}$, the space required to construct the convolution matrix may be larger than the space required to store the parameters. 


\subsubsection{Time complexity}
The time complexity of the 2D convolution operation using the \textsc{im2col} algorithm in the worst case is the sum of two complexities: the complexity of the actual \textsc{im2col} algorithm that produces the convolution matrix plus the complexity of the matrix multiplication between the weights and the matrix resulting from the \textsc{im2col} operation. These two operations are present in both the forward and backward passes of the convolution operation, assuming that the weights of the convolution are trained by a gradient backpropagation algorithm. Usually, the complexity of the matrix multiplication is higher than that of forming the \textsc{im2col} matrix. Thus, we can say that the time complexity of a convolution operation is dominated by the matrix multiplication time. Adding a convolution bias amounts to a simple vector addition and could be omitted from the complexity calculation for the same reason. Using our previous example, the output of a 2D convolution operation is 
$$
out = \text{vec}(W) \times image2col
$$
with vec($W$) the vectorization of the tensor $W$ that flattens its $(C_{out},C_{in},k_1,k_2)$ dimension to $(C_{out}, \ C_{in} \cdot k_1 \cdot k_2)$. Thus the two matrices become compatible for matrix multiplication, and we have in our example:
\begingroup
\setlength{\tabcolsep}{2pt}
\renewcommand\arraystretch{1}
$$
out = \left[\begin{tabular}{cccc} $w_{11}$ & $w_{12}$ & $w_{21}$ & $w_{22}$ \end{tabular}\right] 
\left[
\begin{array}{c c c c c c c c c}
 im_1 & im_2 & im_3 & im_5    & im_6    & im_7    & im_9    & im_{10} & im_{11} \\
 im_2 & im_3 & im_4 & im_6    & im_7    & im_8    & im_{10} & im_{11} & im_{12} \\
 im_5 & im_6 & im_7 & im_9    & im_{10} & im_{11} & im_{13} & im_{14} & im_{15} \\
 im_6 & im_7 & im_8 & im_{10} & im_{11} & im_{12} & im_{14} & im_{15} & im_{116} 
 \end{array}
\right] 
$$
\endgroup

The result of this last multiplication is a matrix of size $(C_{out}, \  B \cdot H^{out}_1 \cdot H^{out}_2)$ that is reshaped into a $(B, C_{out}, H^{out}_1, H^{out}_2) = (1,1,3,3)$ tensor which the reader will verify is indeed the convolution of the kernel $W$ and the matrix $image$.
From this, we can conclude that the time complexity of this matrix product is simply 
$$C_{out} \cdot C_{in} \cdot k_1 \cdot k_2 \cdot  B \cdot H^{out}_1 \cdot H^{out}_2  $$

\label{simplification}
It is very common in the literature, and only for the sake of simplicity, to consider a batch of a single image, since batch parallelization is usually highly optimized on hardware accelerators; to consider square kernels, hence $k_1 = k_2 = k$; to consider square output sizes, thus $H^{out}_1 = H^{out}_2 = H$ and to consider the most frequent case of an intermediate layer in a network having the same number of input and output neurons, leading to $C_{in} = C_{out} = C$. The time complexity is then simplified into
$$C^2 \cdot k^2 \cdot  H^2  $$

Similarly, when using the backpropagation algorithm to compute the gradients of the weights in the 2D case, the \textsc{im2col}  matrix is formed. The gradient of the output, which is of size $(B, C_{out}, H^{out}_1, H^{out}_2$) and which we denote $G_{out}$ is provided by the downstream layers of the network. The gradient of the weight noted $\nabla W$ is computed using the $image2col$ and $G_{out}$ tensors as follows

$$\nabla W =  \text{vec}'(G_{out}) \times image2col^T $$
with $\text{vec}'$ the vectorization of the tensor $G_{out}$ that flattens its $(B, C_{out}, H^{out}_1, H^{out}_2)$ dimension to $(C_{out}, \ B \cdot H^{out}_1 \cdot H^{out}_2)$.

The computation of the gradient with respect to the input is performed using an operation known as \textsc{col2im}, which is applied to the gradient of the output. \textsc{col2im} can be thought of as an inverse operation of \textsc{im2col}. \textsc{col2im} is a true inverse function of \textsc{im2col} if there is no overlap between the kernel tiles formed by \textsc{im2col}. When there is overlap, the inputs are accumulated at those points of overlap. The calculation of the input gradient is not discussed here, but we encourage the reader to consult the following short tutorial of our own (\href{https://medium.com/@khalfaoui.ismail/cuda-kernels-in-pytorch-made-easy-with-numba-using-python-only-74012bab23ba}{An implementation of the 2D convolution}) for more details on the implementation of the gradient of the input using \textsc{col2im}.

Theoretically, the time complexity of the 2D convolution in the backward pass is of the same order as in the forward pass. In practice, however, the backward pass is much slower on GPU devices because the batch dimension in the backward case is within the inner dimension of the matrix multiplication, which makes it harder to parallelize over the batch dimension in matrix multiplication GPU kernels. 

In general, we can conclude that for a batch containing a single input, the time complexity of the $d$-dimensional convolution method utilizing the \textsc{im2col} algorithm is dominated by
\begin{equation}
\label{complexity:1}
C^2 \cdot (k \cdot  H)^d
\end{equation}
and that the space complexity for this same method is dominated by
\begin{equation}
\label{complexity:2}
C \cdot (k \cdot  H)^d  
\end{equation}
while its number of parameters is dominated by
\begin{equation}
\label{complexity:3}
C^2 \cdot k^d
\end{equation}

The \textsc{im2col} algorithm is only one of many possible convolution implementations. Note that for other cases (small or very large kernels, small channels, special hardware, etc.) other algorithms exist and are more suitable (such as the product in the Fourier domain or the Winograd algorithm \cite{lavin2016fast}).

Some algorithms are an approximation of the standard convolution and aim to reduce its cost in terms of time and memory, such as the separable convolution, whose aspects will be discussed in detail in section \ref{sec:depthwise}, and for which several high-performance, innovative algorithms have been developed, such as the implicit depthwise gem \cite{ding2022scaling}, which we'll discuss in section \ref{sec:implicit}. 

\section{Strided convolution}  
Strided convolution refers to a convolution with a stride parameter strictly greater than one. Again using our minimal 2D example \ref{ex:minimal}, this time with a stride $S= (2,2)$, a dilation factor $Df= (1,1)$, padding $Pad = (0,0)$ and the groups $G=1$. The 2D image \ref{eq:image} to which \textsc{im2col} is applied becomes:

\begin{equation}
\renewcommand\arraystretch{1.3}
image2col\_stride2 = \left[
\begin{array}{c c c c}
 im_1 & im_3 & im_9 & im_{11} \\   
 im_2 & im_4 & im_{10} & im_{12} \\ 
 im_5 & im_{7} & im_{13}    & im_{15} \\
 im_{6} & im_{8} & im_{14} & im_{16} \\
 \end{array}
\right]
\end{equation}

\subsection{Downsampling}

By applying larger strides during convolution, strided convolution reduces multiplicatively the spatial dimensions of feature maps (see equation \ref{eq:output}), thereby creating a more compact representation that significantly mitigates computational demands. This technique is instrumental in balancing computational efficiency while preserving salient features, thus enabling neural networks to extract hierarchical features from high-resolution input data. 

Several other downsampling techniques are based on the same principle of stride enlargement, but with different operations: max-pooling and average-pooling are two examples.

\subsection{The convolution stem}
The stem designates the first layer or series of layers in a feed-forward neural network. In particular, it is responsible for processing the input signal to produce the first feature maps. Strided convolutions have been used as a stem for several convolutional neural networks \cite{he2016deep}. Strides control the overlap and the coverage of the convolution kernel with respect to the input feature map. When the stride of a convolution layer is equal to the size of its kernel, the convolution layer partitions the input map. These input tiles/partitions are called patches. Patches are non-empty subsets of the input, their union forms the entire input and their intersection is empty. Drawing inspiration from the patch embedding technique used in vision transformers \cite{dosovitskiyimage, liu2021swin}, a growing number of convolutional neural networks have adopted a strided convolution with a kernel size equal to its stride as a stem. The advantage of this particular type of stem is that it doesn't waste any of the input signal while keeping the complexity as low as possible. 
\subsection{DiffStride}  
To address the problem of determining appropriate stride parameters, an innovative approach called DiffStride has been proposed \cite{riad2021learning}. DiffStride introduces a downsampling layer that uses spectral pooling to learn its stride values through backpropagation. In the context of downsampling, DiffStride performs cropping operations in the Fourier domain. This cropping operation in the frequency domain is generally encountered in the spectral pooling methods \cite{rippel2015spectral}. However, what distinguishes DiffStride is its strategy for defining cropping boundaries. Instead of relying on predefined bounding boxes, the method uses backpropagation to learn the dimensions of the box, denoted $\mathcal{W}$. The factors that influence the dimensions of $\mathcal{W}$ include the shape of the input data, a smoothness factor denoted by $\mathcal{R}$, and the chosen steps.

The mask represented by $\mathcal{W}$ is derived by multiplying two differentiable 1D masking functions, each addressing different axes. These masking functions are inspired by the concept of adaptive attention spans introduced by \cite{sukhbaatar2019adaptive} in the context of self-attention models used in natural language processing. 


\section{Receptive field}
\label{sec:rf}
In general terms, a receptive field refers to the region or area within a system that can influence a particular point, element, or component within that system.  In the context of neural networks, a receptive field refers to the region of the input to which a particular neuron within a layer is sensitive. The receptive field of a layer $l \in \mathbb{N}^*$ of neurons in a feed-forward network is defined as the union of the receptive fields associated with each neuron belonging to that layer. The size of the receptive field associated with the layer $l$, which we will call $r_l$, of a fully convolutional neural network depends on the receptive field size of the previous layers in the network and their respective kernel sizes $k_l$ and strides $s_l$. The latter follows the subsequent recurrence equation:
\begin{equation}
\label{eq:rf}
    \left\{\begin{aligned}
    \forall l \in \mathbb{N}^* & \quad r_{l+1} = r_l + (k_l - 1) \prod_{i=0}^{l} s_i \\
    & \quad r_0 = k_0
\end{aligned}\right.
\end{equation}

Note that the stride increases the receptive field size multiplicatively, while the kernel size increases it additively. The equation \ref{eq:rf} could be solved in the case of a network composed of $L \in \mathbb{N}^*$ layers to obtain the following solution:

\begin{equation}
    \label{eq:receptive_field}
    \begin{aligned}
    r_L = r_0 + \sum_{l=1}^L (k_l - 1) \prod_{i=0}^{l} s_i
\end{aligned}
\end{equation}

It then becomes more intuitive to see that the size of the receptive field also grows linearly with the number of stacked layers. Increasing the number of layers doesn't reduce resolution, which is not the case when you increase the strides. Increasing the kernel size linearly enlarges the receptive field. This effect is even more pronounced in the case of the strides, which increase the receptive field multiplicatively but also diminish resolution multiplicatively.

Decreasing resolution is accompanied by a sharp reduction in complexity, making it a popular tool for models striving for the best accuracy-throughput trade-off. However, loss of resolution also means loss of feature detail, often leading to false classification/detection/segmentation of objects or features of small relative size. Sometimes, for certain dense prediction tasks in particular, it is necessary to have both fine-grained information on small objects (which only high resolution can provide) and at the same time an understanding of the global context. Downsampling methods result in a significant loss of resolution, which hinders this objective, hence the need for a method that can both increase the receptive field without reducing resolution.

\section{Effective receptive field}
The concept presented in section \ref{sec:rf} allows us to determine the size of the receptive field and to a greater extent delimit the region to which a specific neuron or group of neurons is sensitive. While this notion already provides a fairly good insight into what can influence the results of a neuron within a feed-forward network, it does not quantify this influence. Let us imagine that some of the activations of a convolution layer are zero or of negligible magnitude, these activations will still have a non-zero receptive field corresponding to the equation \ref{eq:receptive_field}, even though they play no part in the subsequent neurons output. This is where the notion of effective receptive field comes in. Introduced in \cite{luo2016understanding}, the effective receptive field is determined by taking the partial derivative of a specific pixel in the output map with respect to another pixel in the input. To put this more rigorously into equations, let's consider without loss of generality, a 2D input image $x$ of size $N^2$ where $N \in \mathbb{N}^*$,  indexed by $(i, j)$. The concept of effective receptive field can be directly generalized to 1D and 3D cases.  Let's also denote the outputs of each layer of a feed-forward neural network of size $L \in \mathbb{N}^*$ as $x^l$ with $l \in \llbracket 0 \ .. \ L \rrbracket$. By convention, $x^0$ is the input signal. The effective receptive field of the central output pixel $x^L_{0,0}$ is the matrix computed as:

\begin{equation}
\begin{aligned}
        \forall (i,j) \in \llbracket 1 \ .. \ N \rrbracket & & \text{ERF(x)}_{i,j} = \frac{ \partial x^L_{0,0}}{ \partial x^0_{i,j}}
\end{aligned}
\end{equation}

This last differential formulation could be computed by gradient backpropagation. Note that the effective receptive field is input-dependant: the calculations are made with respect to a specific input or batch of inputs.

The \cite{luo2016understanding} study was able to demonstrate mathematically and then confirm empirically certain important properties of the effective receptive field, in particular the fact that the effective receptive field occupies a smaller portion of the theoretical receptive field and that it is in the form of a Gaussian centered on the input center. Not all pixels in a receptive field contribute equally to an output response, and the center has a particular importance that decreases in a Gaussian distribution as you move away from the center. Another remarkable fact demonstrated by this work is that the effective receptive field decreases in $O(1/\sqrt{L})$ with the number of layers $L$, which is surprising compared to the receptive field, which increases linearly with the number of layers $O(L)$.

According to the same study, the effective receptive field expands throughout the learning process. This is the case for several of the feed-forward networks tested. Unsurprisingly, both downsampling layers and dilated convolution performed well in expanding the effective receptive field. 

In a precursory impulse, the work of \cite{luo2016understanding} already raised the possibility that certain architectural changes on CNNs could change ERFs in a more fundamental way. Examples include sparse kernel convolution and dilated convolution \cite{yu2015multi}. The article also mentioned that we might go even further and use sparse connections that are not grid-like, a point that served as motivation for the dilated convolution with learnable spacings that is the main focus of this thesis.

\section{Dilated convolution}  
\label{sec:dilated}
One of the most innovative convolution methods for simultaneously increasing the receptive field without reducing the resolution of the feature maps nor increasing the number of parameters is known as dilated convolution \cite{yu2015multi} (or ``à trous'' convolution). The dilated convolution enlarges the convolution kernel by evenly spacing the weights in the form of a regular grid. The fixed size of the space introduced by the method is referred to as the dilation rate or dilation factor $df \in \mathbb{N}^*$. The main advantage of this method is that it allows for aggregating multi-scale contextual information without losing resolution or analyzing rescaled inputs. As a result, the receptive field is enlarged without losing input details. 

The original work \cite{yu2015multi} demonstrates that the dilated convolution expands the receptive field exponentially if the dilation rates chosen in the successive layers also grow exponentially. More generally, dilated convolution causes the receptive field to grow multiplicatively with respect to the dilations used and linearly decreases the output dimension (see equation \ref{eq:output}). This decrease can be compensated for by an appropriate padding to maintain the same dimension as the input. In the case of dilated convolution, the equation established in \ref{eq:rf}, and that determines the size of the receptive field, becomes:
\begin{equation}
\label{eq:rfd}
    \left\{\begin{aligned}
    \forall l \in \mathbb{N}^* & \quad r_{l+1} = r_l + df_l(k_l - 1) \prod_{i=0}^{l} s_i \\
    & \quad r_0 = df_0(k_0 -1) + 1
\end{aligned}\right.
\end{equation}
With $df_l \in \mathbb{N}^*$ the dilated rate of the convolution at layer $l \in \mathbb{N}^*$.

It is thus evident that dilated convolution serves as a compelling technique for expanding the receptive field, thereby enhancing the ability to capture long-range spatial dependencies within the data, all the while preserving the spatial resolution crucial to detect fine-grained objects/elements in the input. This concept has prompted the consideration of substituting conventional downsampling layers with dilated convolutions, as seen in architectures like Dilated ResNet \cite{yu2017dilated}. 

It is noteworthy that Dilated ResNet has demonstrated better achievements when compared to its baseline counterpart, the traditional ResNet \cite{he2016deep} architecture in the image classification task. However, a significant challenge that arises in the removal of downsampling layers such as strided convolutions and pooling layers lies in the time complexity it introduces to the model. The computational demands associated with the preservation of a high image resolution can be considerable, potentially impeding real-time or resource-constrained applications. The Dilated ResNet had the same number of layers and parameters as the original ResNet. While the original ResNet downsamples the input by a factor of 32, the Dilated ResNet downsamples the input only by a factor of 8, since several of the ResNet's pooling layers are replaced by dilated convolutions in the second.

In our humble opinion, this is the reason why modern vision architectures (whether fully convolutional \cite{liu2022convnet} or hybrid networks \cite{yu2022metaformer, vasu2023fastvit} with both convolution layers and multi-head self-attention layers) have conserved downsampling layers that reduce the resolution at the end of each stage. That said, the choice of a slower architecture with better resolution over a faster architecture with poorer resolution remains a question of need, depending on the task (image classification versus image segmentation, for example), and the search for architectures that retain good resolution (as in the case of Dilated Resnet) while being fast in inference is still very much underway.

Originally developed for vision applications, where it is particularly effective for dense prediction tasks such as semantic segmentation \cite{yu2015multi}, the dilated convolution method has been generalized to a variety of tasks ranging from audio \cite{oord2016wavenet, dhariwal2020jukebox, chen2020wavegrad} and video processing \cite{pavllo20193d} to speech and natural language processing \cite{ren2020fastspeech, kalchbrenner2016neural}.

Moreover, the dilated convolution has extended the capabilities of convolutional networks by incorporating more contextual information without increasing computational complexity (both in space and time), since the ``dilated'' kernel is never constructed per se. A very efficient way to implement dilated convolution is by adapting the \textsc{im2col} algorithm. Let's take our minimal example \ref{ex:minimal} again, this time with a dilation factor $Df= (2,2)$, stride $S = (1,1)$, padding $Pad = (0,0)$ and the groups $G=1$. In this case, the 2D image \ref{eq:image} to which \textsc{im2col} is applied becomes:

\begin{equation}
\renewcommand\arraystretch{1.3}
image2col\_dil2 = \left[
\begin{array}{c c c c}
 im_1 & im_2 & im_5 & im_6 \\   
 im_3 & im_4 & im_7 & im_8 \\ 
 im_9 & im_{10} & im_{13}    & im_{14} \\
 im_{11} & im_{12} & im_{15} & im_{16} \\
 \end{array}
\right]
\end{equation}
The multiplication with vec($W$) gives the expected forward result.

Mathematically, this operation is equivalent to performing the convolution with a dilation factor $Df= (1,1)$ over the following kernel that has been inflated with evenly spaced zeros:
\begin{equation}
W' = \left[
\begin{array}{c c c }
 w_{11} & 0 & w_{12} \\
 0 & 0 & 0 \\
 w_{21} & 0 & w_{22}
\end{array}
\right]
\end{equation}

 However, readers will agree that this introduces an unnecessary overhead in terms of memory (because of the need to store a much larger $image2col$ matrix, and a very sparse kernel with physically implemented zeros), apart from the time overhead it introduces.

Dilated convolution, with its original concept of fixed dilation factors, laid the groundwork for the development of Dilated Convolution with learnable spacings \cite{hassani2023dilated}.  This was undoubtedly the main source of inspiration for our work. In early attempts to implement DCLS, we tried to implement it by modifying the \textsc{im2col} algorithm. With hindsight, we know that this is feasible with reasonable overhead in the context of 2D convolution, only if we limit ourselves to learnable positions of size ($k_1, k_2$) or even ($C_{in},k_1, k_2$). DCLS allows you to learn the positions of all the kernel's learnable parameters ($C_{out}, C_{in},k_1, k_2$) through the construction (admittedly inefficient) of a sparse kernel of large size.

This GPU inefficiency problem is solved for 2D images using two techniques: first, depthwise separable convolution \ref{sec:depthwise}, which greatly reduces the space complexity of convolutions as well as their time complexity, and second, the implicit GeMM method \ref{sec:implicit}, which makes large kernel depthwise separable convolutions even more time competitive. 

\section{Separable convolution} 
Now that we have seen what the standard convolution is, its parameters, and some of the classical techniques associated with it, let us take a look at some of its approximations.  In what follows, we'll present some techniques that are low-rank approximations of standard convolution. These techniques, although less expressive than standard convolution, have the merit of greatly reducing its parameter and time costs, sometimes leading to advantageous trade-offs. Since current neural networks are often over-parameterized for the task they are designed to solve, the constraint or re-parameterization introduced by methods such as separable convolution turns out to be quite advantageous in most cases in terms of accuracy, throughput, and number of parameters. In what follows, we will only consider the case of 2D convolution $d=2$.

\label{sec:depthwise}

\subsection{Spatially separable convolution}  
Spatially separable convolution is the simplest and most intuitive type of separable convolution aimed at reducing the rank of the method. It was first proposed in the computer vision field by \cite{szegedy2016rethinking}. The whole idea of spatially separable convolution is to approximate the kernel $W$ of size ($C_{out}, C_{in},k_1, k_2$) by a product of two kernels of lower rank that are of size ($C_{out}, C_{in},k_1$) and ($C_{out}, C_{in},k_2$) respectively. 
\begin{equation}
\label{approx:spatial}
    W \approx W_1 \cdot W_2
\end{equation}
The spatially separable convolution consists of applying a first convolution with the $W_1$ kernel, followed by a second convolution with the $W_2$ kernel. Hence its somewhat abusive name of spatially separable convolution, since it acts separately on the spatial dimensions of the 2D input.

Here we can appreciate the interest of the method, which lies in the fact that it reduces the complexities in terms of time \ref{complexity:1}, memory \ref{complexity:2}, and parameters \ref{complexity:3} from $k_1 \times k_2$ to $k_1 + k_2$. Naturally, the case of strict equality in the \ref{approx:spatial} approximation is established only for kernels of rank one, which imposes a strong constraint on the kernel to be learned since it severely limits the rank of the 2D convolution by forcing it to be simply the sequence of two 1D convolutions. However, this limitation caused by spatially separable convolution can be used to good advantage in tasks that are well suited to it, such as tasks involving audio spectrograms, where the time (x-axis) and frequency (y-axis) axes could be treated independently by the respective separate kernels.

Spatially separable convolution, combined with a low kernel size standard convolution can be very effective in reducing flops and thus reducing computation time while maintaining a good accuracy. In \cite{liu2023more}, the authors explore very large kernel size in depthwise separable convolutions using the implicit GeMM \ref{sec:implicit} method. Furthermore, they add branches containing spatially separable convolutions. This same idea is further developed and improved by  \cite{yu2023inceptionnext}, where the authors use spatially separable convolutions as branches alongside a small kernel depthwise convolution and an identity branch. This is done by splitting the input channels and passing each chunk through a branch direction before concatenating the outputs. \citet{xu2023parcnetv2}, in their ParCv2 branch, use the same idea except that they place in succession the two spatially separable convolutions all in parallel with a depthwise $7 \times 7$ convolution.

\subsection{Grouped convolution}  
\label{grouped_conv}
Grouped convolution or grouped separable convolution is a convolution method where the weight kernels $W$ are reduced from ($C_{out}, C_{in}, k_1, k_2$) in the standard convolution method, to ($C_{out}, C_{in}/g, k_1, k_2$) shared groups, $g \in \mathbb{N}^*$. Those weights are then applied separately to different groups of the input channels. 

In practice, this involves reshaping the $W$ weights of size ($C_{out}, C_{in}/g, k_1, k_2$) into a tensor of size ($g, C_{out}/g, C_{in}/g, k_1, k_2$) and the input of size ($B,C_{in}, H^{in}_1, H^{in}_2$) into a tensor of size ($g,B,C_{in}/g, H^{in}_1, H^{in}_2$). 

For the rest, convolution is applied as in the standard method for each group separately. This can be seen as applying the standard convolution $g$ times using kernels of size ($C_{out}/g, C_{in}/g, k_1, k_2$) on inputs of size ($B,C_{in}/g, H^{in}_1, H^{in}_2$), with addition between channels taking place only within the same group.

Finally, this gives us outputs of size ($g, B,C_{out}/g, H^{out}_1, H^{out}_2$), that we reshape into ($B,C_{out}, H^{out}_1, H^{out}_2$).
Note that $g$ must be a common divisor of $C_{in}$ and $C_{out}$ and that the $g$ convolutions involved in the grouped convolutions are parallelized as part of the batched matrix multiplication following the \textsc{im2col} algorithm, which is easily adapted for groups.
\subsection{Depthwise convolution}  
The depthwise convolution is a special case of grouped convolution, where the groups are equal to input channels $C_{in} = g$, thus leading to a complete separation between channels. This method is sometimes referred to as the spatial mixing layer or the token mixing layer, as opposed to the channel mixing layer, which refers to pointwise convolution.
\subsection{Pointwise convolution}  
The pointwise convolution is a special case of standard convolution, where the spatial dimensions of the kernel are all equal to one $k_1 = k_2 = 1$. The pointwise convolution can also be referred to in the literature as a linear layer, a multilayer perceptron (MLP), or a channel mixing layer. Multiple linear layers could be used in a pointwise convolution block, sometimes with middle channels increased with a positive integer ratio as part of a so-called inverted bottleneck \cite{sandler2018mobilenetv2} while maintaining output channels equal to input ones.
\subsection{Depthwise separable convolution}  
\label{sec:depthwise_sep}
Depthwise separable convolution is a generic term that refers to the combination of a depthwise convolution with $C$ channels followed by a pointwise convolution with $C^2$ parameters. It was first proposed in \cite{sifre2014rigid} under the name ``rigid motion scattering'', then made popular by \cite{sandler2018mobilenetv2} and \cite{chollet2017xception}. An activation function, which is a non-linear function such as a sigmoid function or a rectified linear unit (ReLU), is often placed after the pointwise convolution or between the depthwise convolution and the pointwise convolution. Depthwise separable convolution is an efficient and inexpensive approximation to standard convolution, and is in practice a perfectly viable alternative to standard convolution, making it the method of choice for most state-of-the-art neural networks in computer vision and many other sub-fields of deep learning.
Depthwise separable convolutions and spatially separable convolutions could be combined, we refer to this as depthwise-spatially separable convolution. Examples of this combination, which do not employ the preceding denomination, can be found in \cite{yu2023inceptionnext, xu2023parcnetv2, liu2023more}. 
Note that with equal kernel size strides and padding, a depthwise convolution has the same receptive field as a corresponding standard convolution, since the separation is performed only on channels.
\subsection{Complexity}

In what follows, we summarize in Table \ref{tab:complexity} the number of parameters, the time complexity, and the space complexity in the worst case of the previous convolution methods. As for \ref{simplification},  we consider square kernels $k_1 = k_2 = k$, square output sizes $H^{out}_1 = H^{out}_2 = H$ and the same number of input and output channels $C_{in} = C_{out} = C$. In the following, $\Delta = \text{max}(C, \ H^2)$.

\begin{table}[htbp]

\resizebox{\textwidth}{!}{
$
\begin{array}{c c c c c c c}
\text {Conv2D} & \text {Standard} & \text {Spatial Sep.} & \text {Grouped}  & \text {Depthwise} & \text {Pointwise} & \text {Depthwise Sep.}  \\
\hline \text {\#params} & C^2 k^2 & 2 C^2 k & C^2 k^2 / g & C k^2 & C^2 & C\left(C+k^2\right)   \\
\hline \text {time}  & C^2 k^2 H^2 & 2 C^2 k H^2 & C^2 k^2 H^2 / g & C k^2 H^2 & C^2 H^2 & CH^2 \left(C+k^2\right) \\
\hline \text {space} & C k^2 \Delta & 2 C k \Delta & C k^2 \Delta / g & k^2 \Delta & C \Delta & \Delta \left(C+k^2\right) 
\end{array}
$
}
\caption{A comparison of the number of parameters, the time complexity, and the space complexity of classical convolution approaches.
}
\label{tab:complexity}
\end{table}
To the question: ``\textit{Are there optimal numbers of groups (with respect to parameters count) in a sequence of separable convolutions that conserve the same representational power as an initially fixed standard convolution?}'', the answer had been given in \cite{wei2022optimized}. While maintaining an equal volumetric receptive field, \cite{wei2022optimized} suggests a method to find optimal groups for a set of separable convolutions that best approximate a standard convolution. For a set of two separable convolutions (depthwise-spatially separable convolutions), the theoretical optimal number of parameters is found to be $2C^{\frac{2}{3}}k$.

\subsection{Implicit GeMM}  
\label{sec:implicit}

The Implicit GeMM method represents a sophisticated approach to enhancing matrix multiplication efficiency, especially relevant within convolutional neural networks (CNNs). This method is tied to the CUDA programming language specificities and capitalizes on modern GPU architectures' strengths.

This Implicit GeMM technique synergizes with the investigation into large kernel design in CNNs. As we've seen before, 2D convolution may be mapped to matrix multiplication (General Matrix Multiplication) by forming the \textsc{im2col}.  The implicit GeMM algorithm is a variation on the blocked, hierarchical GeMM computation in CUDA that instead forms tiles of the convolution matrix on the fly as data is loaded from global memory into shared memory by carefully updating pointers. Once the convolution matrix is formed in shared memory, the CUDA warp-level computing components accumulate the result of convolution and update the output tensor. A similar tiling mechanism is leveraged in the Flash Attention method \cite{dao2022flashattention} for example. 

The Implicit GeMM method has been made functional for depthwise separable convolutions with square kernel sizes in \cite{ding2022scaling} and integrated into \cite{MegEngine}. This has had the effect of reviving large kernel convolutions, which can now be implemented cost-effectively through depthwise separable convolutions and the Implicit GeMM method, making them competitive with the multi-head self-attention module \cite{vaswani2017attention}.

\section{Advanced convolutions}  
Next, we turn to the more advanced cases of convolution. The methods most closely related to DCLS will be discussed further in Section \ref{chap2:related} after the introduction of the latter.


\subsection{Mixed Depthwise Convolution}  
Unlike traditional depthwise convolutions, MixConv \cite{tan2019mixconv} partitions the input channels into distinct groups having the same spatial dimensions and then employs varying kernel sizes for each group. The output of this operation is obtained by concatenating all the partitioned output tensors. This methodology enhances feature extraction by fusing insights from multiple kernel sizes across different channel groups. This method gives improved performance in tasks requiring spatial information capture. MixConv's strength lies in its ability to combine features from diverse receptive fields, ultimately contributing to more comprehensive representations and enhanced network performance.

\subsection{Selective Kernel Convolution}  
Selective Convolution \cite{li2019selective} introduces a dynamic mechanism that addresses the uniformity of receptive field sizes. In conventional CNNs, neurons in each layer share the same receptive field size, a design that disregards the adaptive modulation of receptive fields observed in biological neural systems. In response, Selective Convolution proposes a method where individual neurons can flexibly adjust their receptive field sizes based on varying scales of input data. This adaptive behavior is facilitated through a core component named as Selective Kernel (SK) unit. Within an SK unit, distinct branches, each employing different kernel sizes, are used within an attention mechanism. The attention matrix is informed by the content within these branches, enabling the layer's neurons to exhibit varying effective receptive field sizes based on the attention distributions. This approach, when integrated into deeper architectures, yields promising results on benchmark datasets like CIFAR10 and ImageNet. 

\subsection{Switchable Atrous Convolution}
Switchable Atrous Convolution (SAC) \cite{qiao2021detectors} introduced a learnable switch mechanism that dynamically selects between multiple atrous (dilated) convolution rates \ref{sec:dilated} during training, allowing the network to automatically adjust receptive fields and information capture based on the characteristics of the input data. This innovation has proven particularly effective in tasks like semantic segmentation and object recognition, where varying context scales are crucial for accurate feature extraction and spatial understanding. 
SAC introduces a dynamic and trainable switch mechanism that lets the network convolve input features with varying dilation rates and then combines these results using switch functions. This clever approach allows the network to adapt its receptive fields and information capture based on the specific characteristics of the input data. SAC has found significant success in computer vision tasks like semantic segmentation and object recognition, where being able to handle different spatial contexts and scales is crucial for precise feature extraction and understanding the spatial layout. In a nutshell, Switchable Atrous Convolution enhances model flexibility and performance across various visual recognition applications.

The SAC architecture consists of three main parts: two global context modules placed on either side of the SAC component and the central SAC component itself. SAC uses dilation rates to enrich the convolution process. This technique enables SAC to capture both fine-grained details and broader context by adjusting the atrous rates dynamically during training. Additionally, SAC introduces a switch function that controls the mixing of these multiple dilated rates, allowing the network to choose the most relevant features for each spatial location. 

\subsection{Continuous Kernel convolution}
Continuous Kernel Convolution \cite{romero2021ckconv} (CKConv) introduced a novel approach to sequential data analysis. CKConv builds upon the foundational concept of convolution, which involves applying a kernel or filter to sequential data to extract relevant features. However, CKCONV takes a distinctive approach by employing continuous kernels that traverse the data continuously, as opposed to traditional discrete, fixed-size kernels.

This continuous kernel movement enables CKCONV to capture patterns at multiple scales and positions along the sequence. It is akin to performing a dynamic zoom-in and zoom-out operation on the data, allowing the model to discern both high-level structural features and fine-grained details. This adaptability proves to be especially advantageous in the context of sequential data, where patterns can manifest at varying temporal granularities. By offering this continuous perspective, CKCONV enhances the model's ability to understand intricate temporal dependencies, making it particularly valuable for tasks like natural language processing (NLP), where language structures unfold over different time scales. 
\section{Convolutional neural networks}  
In the previous section, we reviewed some of the key convolution methods (not all of them, of course) that have played a crucial role in the evolution of convolutional neural networks in fields as diverse as computer vision, audio, and language processing. We're now going to focus on the review of influential neural networks, often associated with the convolution methods presented above. In chapter (\ref{chap:2}), by replacing the depthwise separable convolutions with DCLS ones, we'll see how DCLS improves the accuracy of the majority of these models, without significantly compromising their performance in terms of time and memory.
\subsection{The ResNet model }  

ResNet \cite{he2016deep}, short for Residual Networks, is a major milestone in Convolutional Neural Network (CNN) research, introduced to address the challenge of vanishing gradients during deep network training. ResNet incorporates residual connections, also known as skip connections, which enable the direct propagation of information from one layer to another. These connections act as a preconditioning that simplifies optimization and alleviates the degradation problem that often occurs when deep networks are trained, mitigating the vanishing gradient issue and enabling the training of exceptionally deep architectures. The main finding of ResNet is that increasing the network's depth should not result in decreasing performance due to optimization difficulties, as long as residual connections are employed. This insight opened the door to constructing extraordinarily deep networks with significantly improved training and convergence characteristics. ResNet's contributions have extended to various domains, including image classification, object detection, and image generation, underscoring its foundational role in the evolution of deep learning architectures.

ResNet's introduction built upon prior CNN work, including LeNet \cite{lecun1998gradient}, AlexNet \cite{alexnet}, and the influential VGG \cite{simonyan2014very} network architecture, which emphasized the benefits of deeper networks for improved performance. However, ResNet's innovative integration of residual connections propelled it beyond its predecessors, effectively addressing the challenges posed by gradient vanishing in deeper architectures.
\subsection{The MobileNet model }  

MobileNet \cite{sandler2018mobilenetv2}, a pioneering convolutional neural network architecture, is notable for its emphasis on efficiency and lightweight design tailored for resource-constrained environments. The key innovation within MobileNet lies in the integration of depthwise separable convolutions \ref{sec:depthwise_sep}. By splitting the convolution process into depthwise and pointwise convolutions, MobileNet significantly reduces computational complexity while maintaining a high expressive power. This approach is particularly advantageous for mobile devices and embedded systems where computational resources are limited. Moreover, the Xception network \cite{chollet2017xception} further extended the concept of depthwise separable filters, demonstrating how to scale them up effectively, ultimately surpassing the performance of Inception V3 networks \cite{szegedy2016rethinking}. MobileNet's main findings underline the potential of depthwise separable convolutions to enable efficient yet effective neural network architectures, thus becoming a cornerstone for tasks such as real-time object detection and image classification in constrained environments.

\subsection{The ConvNeXt model }  
ConvNeXt \cite{liu2022convnet} emerged as a notable advancement by synergizing key elements from diverse neural network architectures to create a competitive CNN model. Building upon the foundation laid by ResNet, ConvNeXt integrates the efficiency of depthwise convolutions, similar to those explored in MobileNet and Xception \cite{chollet2017xception}, to alleviate computational demands. Moreover, ConvNeXt assimilates insights from the architecture of Swin and Vision Transformers (ViTs) \cite{liu2021swin, dosovitskiy2020image} to craft a robust and performant framework. This novel amalgamation enables ConvNeXt to rival the capabilities of vision transformers in handling complex image-understanding tasks. Notably, ConvNeXt introduces several modifications compared to ResNet, including:
\begin{itemize}
    \item Training techniques
    \begin{itemize}
        \item Using AdamW optimizer \cite{adamw} instead of Adam \cite{adam}.
        \item Training for longer epochs (90 in ResNet to 300 in ConvNeXt)
        \item Evaluating with both model and model exponential average (EMA) \cite{polyak1992acceleration}. 
        \item Using stochastic depths \cite{huang2017densely}.
        \item Data augmentation methods such as:  Mixup \cite{zhang2018mixup}, Cutmix \cite{yun2019cutmix}, RandAugment \cite{cubuk2020randaugment}, Random Erasing \cite{zhong2020random} ...
    \end{itemize}
    \item Macro design
        \begin{itemize}
            \item Adjusting the number of blocks in each stage from (3, 4, 6, 3) in ResNet-50 to (3, 3, 9, 3) in ConvNeXt-tiny.
            \item Patchify: adding a stem cell composed of a 4 by 4 kernel size convolution with stride 4. This corresponds to the patch size extracted from the input image as in \cite{liu2021swin}.
            \item  ResNeXt-ify: replacing the standard convolutions with depthwise separable ones as in \cite{resnext}.
            \item Using inverted bottlenecks as in \cite{sandler2018mobilenetv2}, and moving the depthwise convolution on top of the pointwise ones.
            \item Enlarging the kernel size of the depthwise convolutions. (from 3 to 7). 
        \end{itemize}
    \item Micro design
        \begin{itemize}
            \item Replacing all the ReLU activations \cite{nair2010rectified} with GELU \cite{fukushima1975cognitron} ones.
            \item Using fewer activation functions: one per block. 
            \item Substituting all batchnorms \cite{ioffe2015batch} with layernorms \cite{ba2016layer}.
            \item Using fewer normalizations: one for each separable convolution used.             
            \item Separating downsampling layers from residual skip connections.
        \end{itemize}
\end{itemize}

\subsection{The MetaFormer model }  

Both \cite{yu2022metaformer, yu2022poolformer} explored the potential of MetaFormer, an abstracted architecture derived from the Transformer model \cite{vaswani2017attention}. Instead of focusing on the specific token mixer design, the authors investigated the broader capabilities of MetaFormer by applying various baseline models using basic or common token mixers. 

Even with the simplest token mixer, identity mapping, the MetaFormer model called IdentityFormer achieves over 80\% accuracy on the ImageNet-1K dataset, demonstrating its solid lower bound of performance. MetaFormer can work effectively with arbitrary token mixers, including random matrices. For instance, the model RandFormer using a random token mixer achieves more than 81\% accuracy, outperforming IdentityFormer. Models instantiated from MetaFormer using conventional token mixers from five years before this work surpass state-of-the-art models. Specifically, ConvFormer, which utilizes common depthwise separable convolutions as token mixers, outperforms the strong CNN model ConvNeXt \cite{liu2022convnet}. CAFormer, created by applying depthwise separable convolutions and self-attention, set a new accuracy record of 85.5\% on ImageNet-1K, even without external data or distillation.


The core idea of this work challenges the belief that the token mixer module, often considered the heart of Transformers, is the primary contributor to their success. Instead, the authors argue that MetaFormer, the overarching architecture, plays a more crucial role in achieving superior results for Transformer and MLP-like models in computer vision tasks. They suggest that future research should focus on enhancing MetaFormer itself rather than dedicating too much attention to token mixer modules. The authors invite further exploration of MetaFormer in different learning settings and domains, emphasizing its importance in vision applications.

\subsection{The RepLKNet model }  
RepLKNet (for Re-parameterized Large Kernels) \cite{ding2022scaling} reevaluates the use of large convolutional kernels in modern convolutional neural networks (CNNs) and presents significant findings. Inspired by recent advancements in vision transformers (ViTs \cite{dosovitskiy2020image} and metaformers \cite{yu2022poolformer}), the study suggests that employing a few large convolutional kernels instead of multiple small ones can be a more potent approach. The authors introduce five guidelines for designing efficient and high-performance large-kernel CNNs, including the use of re-parameterized large depth-wise convolutions.

\begin{itemize}
    \item Very large kernels can still be efficient in practice.
    \item Identity shortcut is vital, especially for networks with very large kernels.
    \item Re-parameterizing \cite{ding2022scaling} with small kernels helps to make up the optimization issue.
    \item Large convolutions boost downstream tasks much more than ImageNet
    \item Large kernel is useful even on small feature maps.
\end{itemize}

\citet{ding2022scaling} propose a novel CNN architecture called RepLKNet, which features exceptionally large kernels, up to 31x31 in size, as opposed to the common 3x3 kernels. RepLKNet effectively narrows the performance gap between CNNs and ViTs \cite{dosovitskiyimage}, achieving results comparable to or superior to the Swin Transformer on ImageNet and various downstream tasks, all while maintaining lower latency. 

The study uncovers that, unlike small-kernel CNNs, large-kernel CNNs exhibit significantly larger effective receptive fields (ERFs) and a higher shape bias rather than a texture bias. This finding contributes to the enhanced performance of CNNs, especially in downstream tasks, and brings them closer to ViTs in terms of performance as both data and model sizes increase.

Furthermore, this work emphasizes the importance of using large convolutional kernels in CNN architecture design, which can efficiently expand the effective receptive field and substantially boost CNN performance, thus narrowing the performance gap between CNNs and ViTs as data and models scale up. This work is expected to advance research in both the CNN and ViT communities, providing insights into the significance of effective receptive fields for high-performance models and shedding light on the underlying mechanisms of self-attention in ViTs.

\subsection{The InternImage model }  
\label{internimage}

InternImage \cite{wang2022internimage} addresses the gap in the development of large-scale convolutional neural networks (CNNs) compared to the progress made with large-scale vision transformers (ViTs). \citet{wang2022internimage} introduced a new CNN-based foundation model called InternImage, designed to compete with ViTs \cite{dosovitskiyimage} in terms of performance and scalability.

InternImage stands out by using deformable convolution \ref{deformable} as its core operator, unlike recent CNNs that emphasize large dense kernels. This approach allows InternImage to have a large effective receptive field needed for tasks like object detection and segmentation, along with adaptive spatial aggregation based on input and task information. As a result, InternImage reduces the strict inductive bias associated with traditional CNNs, enabling it to learn more robust patterns from massive datasets, as is the case with ViTs.

The effectiveness of InternImage is validated through rigorous testing on benchmark datasets such as ImageNet \cite{deng2009imagenet}, COCO \cite{lin2014microsoft}, and ADE20K \cite{zhou2019semantic}. Notably, InternImage-H achieves impressive results, setting a new record with a 65.4 mAP on COCO test-dev and 62.9 mIoU on ADE20K, outperforming both current leading CNNs and ViTs.

In summary, the model InternImage leverages deformable convolution to bridge the gap with ViTs, demonstrating that CNNs remain a viable option for large-scale vision model research. However, challenges such as latency for downstream tasks and the early stage of development for large-scale CNNs still need to be addressed, and InternImage serves as a promising starting point for future advancements in this domain.

\subsection{The FastVit model}  

 FastViT \cite{vasu2023fastvit}, is a hybrid vision transformer architecture that achieves an impressive balance between model accuracy and latency. The model introduces a novel component called RepMixer, which employs structural re-parameterization that can be re-parameterized at inference time to a single depthwise convolution as in \cite{ding2022scaling}, to reduce memory access costs by eliminating skip-connections in the network.

To enhance accuracy without significantly affecting latency, the authors of this work also incorporate large kernel convolutions. They found that using depthwise large kernel convolutions can be highly competitive with models using self-attention while introducing a small increase in latency.

FastViT outperforms several state-of-the-art models, including CMT, EfficientNet, and ConvNeXt, in terms of speed while maintaining comparable accuracy on the ImageNet dataset. Specifically, FastViT is 3.5 times faster than CMT \cite{guo2022cmt} (another competitive hybrid model), 4.9 times faster than EfficientNet \cite{tan2021efficientnetv2}, and 1.9 times faster than ConvNeXt \cite{liu2022convnet} on mobile devices.  The model consistently demonstrates superior performance across various tasks, including image classification, object detection, semantic segmentation, and 3D mesh regression, with significant latency improvements on both mobile devices and desktop GPUs. Additionally, FastViT exhibits robustness to out-of-distribution samples and corruptions, surpassing many robust models.  In short, FastViT provides an efficient hybrid vision transform suitable for a wide range of computing platforms. 

\subsection{Comparison of recent convolutional and hybrid models on image classification task using ImageNet1k}
\begin{figure}[ht]
\centering
   \includegraphics[width=0.9\linewidth]{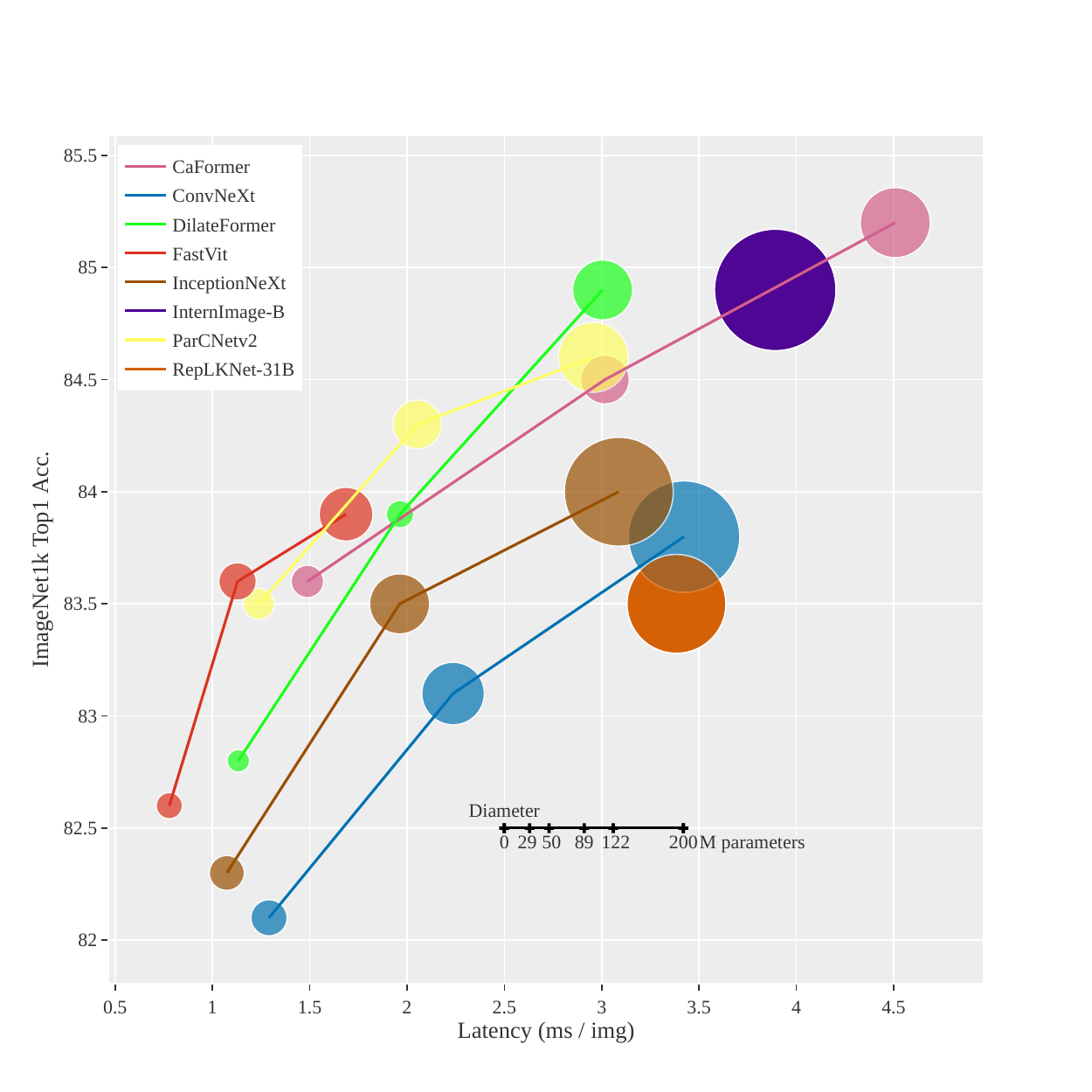}
   \caption{Classification accuracy on ImageNet-1K as a function of latency (i.e. inverse of the throughput). Dot diameter corresponds to the number of parameters. Models represented are: CaFormer \cite{yu2022metaformer}, ConvNeXt \cite{liu2022convnet}, DilateFormer \cite{jiao2023dilateformer}, FastVit \cite{vasu2023fastvit}, InceptioNext \cite{yu2023inceptionnext}, InternImage \cite{wang2022internimage}, ParCNetv2 \cite{xu2023parcnetv2}, ReplKNet-31B \cite{ding2022scaling}.}
\label{fig:general}
\end{figure}

\chapter{Dilated convolution with learnable spacings}
\label{chap:2}
\section{Disclaimer}
This chapter is strongly inspired by the article: \cite{hassani2023dilated}. Most parts of this chapter have been taken verbatim from the paper of which we are the principal author, with the content and wording largely our own. 
A short blog post that summarizes the DCLS method along with a tutorial has been published in \href{https://medium.com/@khalfaoui.ismail/what-is-dilated-convolution-with-learnable-spacings-dcls-and-how-to-use-it-dea93c490a82}{What is Dilated Convolution with Learnable Spacings (DCLS) and how to use it?}

\section{Introduction}
\label{sec:intro}
The receptive field of a deep convolutional network is a crucial element to consider when dealing with recognition and downstream tasks in computer vision. For instance, a logarithmic relationship between classification accuracy and receptive field size was observed in \cite{araujo2019computing}. This tells us that large receptive fields are necessary for high-level vision tasks, but with logarithmically decreasing rewards and thus a higher computational cost to reach them.

Recent advances in vision transformers \citep{dosovitskiy2020image} and in CNNs \citep{liu2022convnet, ding2022scaling,trockman2022patches,liu2022more} highlight the beneficial effect that a large convolution kernel can have, compared to the $3 \times 3$ kernels traditionally used in previous state-of-the-art CNN  models \citep{he2016deep}. However, when naively increasing the kernel size, the accuracy rapidly plateaus or even decreases. For example, in ConvNeXt, the best accuracy was achieved by a $7 \times 7$ kernel~\citep{liu2022convnet,liu2022more}. Using a structural re-parameterization trick, \cite{ding2022scaling} demonstrated the benefit of increasing the kernel size up to 31 by 31. Thereafter, \cite{liu2022more} showed that there was still room for improvement by moving to 51 by 51, using the \textit{depthwise implicit matrix multiplication (gemm)} method developed by \cite{ding2022scaling} and for which the implementation has been integrated into the open-sourced framework MegEngine \citep{MegEngine}, in addition to a spatial separation of the depthwise kernel followed by an accumulation of the resulting activations. Yet, all these improvements have a cost in terms of memory and computation, and it does not seem possible to increase the size of the kernels indefinitely.

\begin{figure}[ht]
  \centering

   \includegraphics[width=1\linewidth]{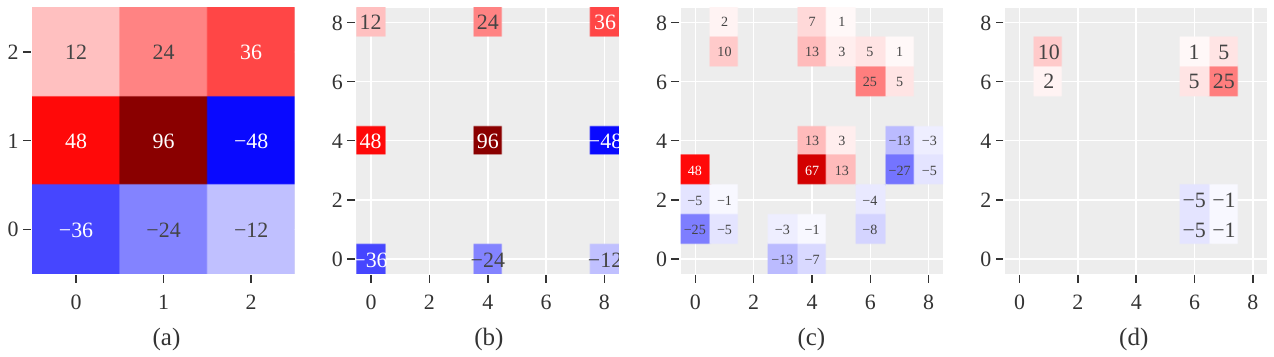}
   \caption{(a): a standard $3\times 3$ kernel. (b): a dilated $3\times 3$ kernel with dilation rate 4. (c): a 2D-DCLS kernel with 9 kernel elements and a dilated kernel size of 9. Each weight is spread over up to four adjacent pixels. (d): a 2D-DCLS kernel with 3 kernel elements and still a dilated kernel size of 9.} 
   \label{fig:1}

\end{figure}

One of the first approaches that allow inflating the receptive field of a convolutional layer without increasing the number of learnable parameters nor the computational cost is called dilated convolution (DC). DC or “atrous convolution” was first described in \cite{holschneider1990real} and \cite{shensa1992discrete}, under the name “convolution with a dilated filter” before being referred to as “dilated convolution” in \cite{yu2015multi}. The purpose of this approach is to inflate the convolutional kernel by regularly inserting spaces (\textit{i.e.} zeros) between the kernel elements, as depicted in Figure \ref{fig:1}b. The spacing between elements is thus constant, it is a hyper-parameter usually referred to as “dilation” or “dilation rate”.
\begin{figure}[ht]
\centering
   \includegraphics[width=0.7\linewidth]{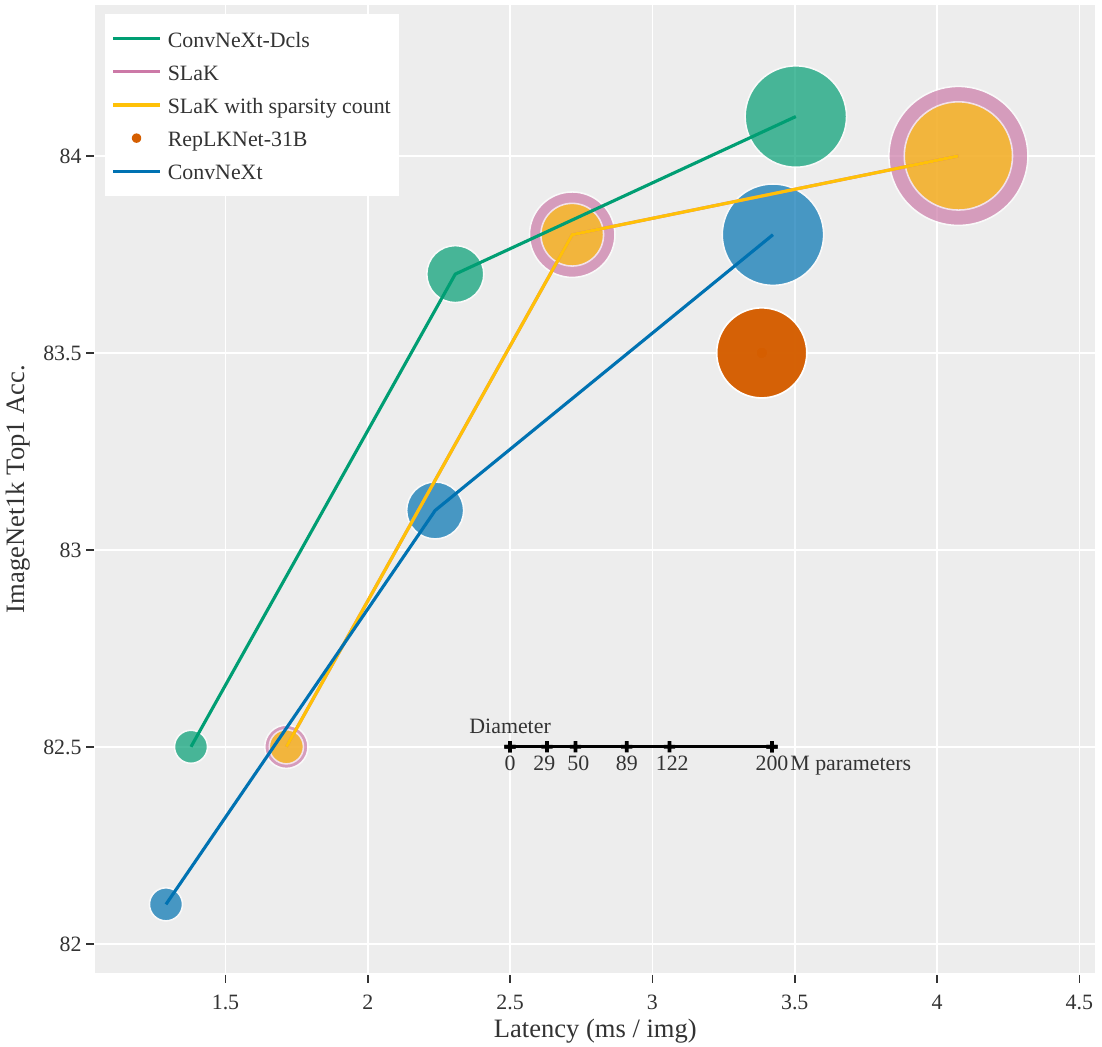}
   \caption{Classification accuracy on ImageNet-1K as a function of latency (i.e. inverse of the throughput). Dot diameter corresponds to the number of parameters.}
\label{fig:0}
\end{figure}
Despite its early successes in classification since \cite{yu2017dilated}, and its even more convincing results in semantic segmentation \cite{sandler2018mobilenetv2,chen2017deeplab,chen2018encoder} and object detection \cite{lu2019grid}, DC has gradually fallen out of favor and has been confined to downstream tasks such as those described above. Without much success, \cite{ding2022scaling} tried to implement DC in their ReplKNet architecture. Our investigation on ResNet and ConvNeXt with standard dilated convolution (Section~\ref{sec:res_resnet50}) will lead to a similar conclusion. The failure of this method for classification tasks could be attributed to the great rigidity imposed by its regular grid as discussed in \cite{wang2018smoothed}.

In this context, we propose DCLS (Dilated Convolution with Learnable Spacings), a new convolution method. In DCLS, the positions of the non-zero elements within the convolutional kernels are learned in a gradient-based manner. The inherent problem of non-differentiability due to the integer nature of the positions in the kernel is circumvented by interpolation (Fig.~\ref{fig:1}c). DCLS is a differentiable method that only constructs the convolutional kernel. To actually apply the method, we could either use the native convolution provided by PyTorch or a more advanced one such as the \textit{depthwise implicit gemm} convolution method \citep{ding2022scaling}, using the constructed kernel. DCLS comes in six sub-versions: 1D, 2D, 3D, and what we call N-MD methods, namely: “2-1D, 3-1D and 3-2D”  where a N-dimension kernel is used but positions are learned only along M dimension(s). The main focus of this chapter will be the 2D version for which we detail mathematical proofs, implementation specificities, and results on image classification, downstream, and robustness tasks.

The principal motivation of DCLS is to explore the possibility of improving the fixed grid imposed by the standard DC via learning the spacings in an input-independent way. Instead of having a grid of kernel elements like in standard and dilated convolutions, DCLS allows an arbitrary number of kernel elements (Fig.~\ref{fig:1}d). We refer to this free tunable hyper-parameter as “kernel count”. In this chapter, we set it in order to be at iso or fewer parameters than the baselines we will compare ourselves to. Conversely, we refer to the size of the kernel or rather the maximum size in which the kernel elements are allowed to move inside the dilated kernel, as the “dilated kernel size”. It is also a tunable hyper-parameter.

The positions of kernel elements in DCLS are randomly initialized and are allowed to move throughout the learning process within the dilated kernel size limit. We will then show how sharing positions across multiple blocks of the same convolution stage could further increase accuracy while reducing the number of learnable parameters. This, together with other learning techniques, empirically and consistently improve the overall performance of the method. They are summarized in Section \ref{sec:usage}.

\section{Kernel construction in DCLS}
Our method entails that we learn the float coordinates ($\vp^1$ and $\vp^2$ in the 2D case for example) for every weight $\vw$ of the dilated kernel (in addition to the actual weights themselves). The positions or coordinates of the weights within the convolution kernel are conceptually integer parameters, yet in order to compute their derivatives, we should consider them as float parameters. This problem of integer positions learning is smoothed by making use of a bilinear interpolation that will be described in equations \ref{eq:2-2}, \ref{eq:2-3} and \ref{eq:2-4}.

\subsection{Notation and preliminaries}
We denote by $\floor{ \ }$ the floor function and we define its derivative by the zero function.
\begin{equation}
    \forall x \in \mathbb{R},  \floor{x}' \overset{\text{def}}{=} 0
\end{equation} 

We denote by $m \in \mathbb{N}^{*}$ the number of kernel elements inside the constructed kernel and we refer to it as the “kernel count”. Moreover, we denote respectively by $s_1, s_2 \in \mathbb{N}^{*} \times \mathbb{N}^{* } $, the sizes of the constructed kernel along the x-axis and the y-axis. The latter could be seen as the limits of the dilated kernel, and we refer to them as the “dilated kernel size”.

The $s_1 \times s_2$ matrix space over $\mathbb{R}$ is defined as the set of all $s_1 \times s_2$ matrices over $\mathbb{R}$, and is denoted $\mathcal{M}_{s_1 , s_2}(\mathbb{R})$.

 The characters $w$, $p^1$ and $p^2$ respectively stand for the weight, the position of that weight along the x-axis (width) and its position along the y-axis (height) in the scalar case while the bold $ \vw = (w_{i})_{1 \leq i \leq m}$, $ \vp^1 = (p^{1}_{i})_{1 \leq i \leq m}$ and $ \vp^2 = (p^{2}_{i})_{1 \leq i \leq m}$ respectively stand for the weight, the width-position of that weight and its height-position in the vector case.

\subsection{Mathematical formulation}
The mathematical construction of the forward pass as well as the proofs for the derivations used in the backward pass are in Appendix \ref{appendixA}. This construction relies on bilinear interpolation and could be described by the following function:
\begin{align}
\label{eq:2-2}
  \begin{split}
  F \colon  \quad \mathbb{R}^m \times \mathbb{R}^m \times \mathbb{R}^m   &\to \mathcal{M}_{s_1 , s_2 } (\mathbb{R})\\
  \vw, \vp^1, \vp^2 \mapsto &  \mK = \sum_{k = 1}^{m}  f(w_{k},p^{1}_{k},p^{2}_{k})
  \end{split}
\end{align}

With $f$ defined as follows:
\begin{align}
\label{eq:2-3}
  \begin{split}
  f \colon \mathbb{R} \times \mathbb{R} \times \mathbb{R} &\to \mathcal{M}_{s_1 , s_2 } (\mathbb{R})\\
  w, p^1, p^2  & \mapsto \quad \mK
  \end{split}
\end{align}
where $\forall i\in \llbracket 1 \ .. \ s_1 \rrbracket$, $\forall j\in \llbracket 1 \ .. \ s_2 \rrbracket  \colon $\\
\begin{equation}
\label{eq:2-4}
\arraycolsep=1.3pt\def\arraystretch{1}
\displaystyle \mK_{ij} =\left\{\begin{array}{cl}
w \ (1 - r^1)\ (1 - r^2) & \text {if } i = \floor{p^1}, \ j = \floor{p^2} \\
w \ r^1 \ (1 - r^2) & \text {if }  i = \floor{p^1} + 1, \ j = \floor{p^2} \\
w \ (1 - r^1) \ r^2 & \text {if }  i = \floor{p^1}, \ j = \floor{p^2} + 1\\
w \ r^1 \ r^2  & \text {if }  i = \floor{p^1} {+} 1, \ j = \floor{p^2} {+} 1 \\
0 & \text {otherwise } 
\end{array}\right.
\end{equation}
and where the fractional parts are:
\begin{equation}
    \begin{array}{ccc}
    r^1 = \{p^1\} = p^1 - \floor{p^1} & \text{and} & r^2 = \{p^2\} = p^2 - \floor{p^2}
    \end{array}
\end{equation}

We invite the reader to look at Appendix \ref{appendixA} for more details on the mathematical proofs and derivatives that permit to calculate the gradients of a differentiable loss function with respect to the weights and their positions. Those derivations lead to the DCLS kernel construction algorithm described for the 2D case in pseudo-code in Appendix \ref{appendix:algo}. Furthermore, the real code for the kernel construction in 1D, 2D, and 3D cases is included in Appendix \ref{appendix:algo_torch}. This code is written in native PyTorch language, with classical modules, and does not require any compilation or adaptation.

\section{Learning techniques}
\label{sec:usage}
So far, we have seen how to implement the DCLS method. We now turn to the techniques that allow us to get the most out of the method. In what follows, we list the training techniques that we have retained and for which we have noticed a consistent and empirical advantage on validation accuracy.
\begin{itemize}[leftmargin=*]
    \item \textbf{Weight decay:} weight decay is a regularization method widely used in a variety of deep learning models. Though its beneficial effect on generalization, we noticed that when applied to the kernel positions in DCLS method, weight decay tends to “artificially” over-concentrate the positions around the center of the kernel, resulting in poorer accuracy. Therefore, we set this hyperparameter to 0 for the kernel positions and kept it unchanged for all the other parameters. 
    \item \textbf{Positions initialization:} the DCLS positions tend to cluster around the center of the RF throughout the learning process (see Appendix~\ref{appendix:hist}). In an attempt to facilitate learning, we chose an initial distribution close to the one obtained at the end of the training, which is a centered normal law of standard deviation 0.5. Yet in practice, the uniform law gives a similar performance.
    \item \textbf{Positions clamping/overlapping :} kernel elements that reach the dilated kernel size limit are clamped. This is done at the end of every batch step to force kernel positions to stay within limits. Agglutination around those limits can sometimes be observed and this indicates that the dilated kernel size is too small and should be enlarged. Positions of different kernel weights (or their interpolations) could also overlap. They are added together in such a case.
    \item \textbf{Dilated kernel size tuning:} we empirically tuned it using the remark above. For simplicity, we used the same dilated kernel size in all the model layers (7 for ResNet-50-dcls and 17 for ConvNeXt-dcls; larger values did not bring any gain in accuracy). Note that increasing the dilated kernel size has no cost on the number of trainable parameters, but has a cost on throughput, especially when using non-separable convolutions. Besides, the convolution algorithm (which is the most time-consuming part of DCLS) that is used after the kernel construction (whether it is the native one or the \textit{depthwise implicit gemm} one) does not leverage the kernel sparsity. Thus, the kernel count does not impact the throughput; only the dilated kernel size does. The development of a convolution method that takes into account sparse kernels, such as \cite{kundu2019pre}, could further help to reduce the throughput gap between DCLS convolution and standard dilated convolution.
    \item \textbf{Kernel count tuning:} as said before, we have set this hyper-parameter to the maximal integer value that allows us to be below the baselines to which we compare ourselves in terms of the number of trainable parameters. Note that adding one element to the 2D-DCLS kernel leads to having three additional learnable parameters: the weight, its vertical and its horizontal position. For simplicity, we used the same kernel count in all the model layers.
    \item \textbf{Positions learning rate scaling:} we found that kernel positions could benefit from a special scaling when it comes to the learning rate.  As their magnitude is different from regular weights, we scaled the learning rate of all the kernel positions by a factor of 5. This is the best factor we found empirically. Custom-made schedulers for positions have been tested, but scaling the learning rate while using the same scheduler as for the kernel weights remained the best choice. Interestingly, we found that throughout learning, the average speed of the positions follows precisely the shape of the learning rate scheduler curve~(see Appendix~\ref{appendix:speed}).      
    \item \textbf{Synchronizing positions:} we shared the kernel positions across convolution layers with the same number of parameters (typically, those belonging to a same stage of the ConvNeXt/ResNet model), without sharing the weights. Positions in this kind of stages were centralized in common parameters that accumulate the gradients. This constraint has surprisingly enhanced the accuracy while reducing the number of extra parameters dedicated to positions (an ablation of this technique led to a 0.13\% accuracy drop on ImageNet1k with ConvNeXt-T-dcls). We think that this is due to the inherent property we find in non-DCLS convolutions (either standard or dilated), where the positions are globally fixed, therefore particularly across layers. This last property is not ensured in DCLS unless we explicitly make use of positions synchronization as the positions are natively unconstrained across layers.
    \item \textbf{Repulsive loss}: following the work of \cite{thomas2019KPConv} on 3D cloud points, we implemented the repulsive loss for the DCLS kernel positions to discourage multiple elements from overlapping. Despite a slight advantage with the ResNet-50-dcls model, this technique did not significantly improve the results with ConvNeXt-dcls.
    \item  \textbf{\textit{Depthwise implicit gemm}:} This method has been developed by \cite{ding2022scaling} and integrated in \cite{MegEngine}, it has for goal to modify the \textit{im2col} algorithm as well as the matrix multiplication that follows, both used in the native 2D convolution method of PyTorch, by a very efficient algorithm which does not build explicitly the \textit{im2col} tensor. The advantage of this method is that it is much faster than the native one for large kernel sizes, without affecting the convolution speed for small kernel ones. In our experiments, the largest dilated kernel size is 17, therefore this method is not absolutely necessary. However, the DCLS user can choose to use it instead of the native method, which will improve the throughput.
\end{itemize}

\section{Results and discussion}
\subsection{Settings}
We started with an exploratory study on ResNet-50, where we drop-in replaced all the $3 \times 3$ convolutions of the model by 2D-DCLS ones. For that, we used a lightweight procedure named the “A3 configuration”, described in \cite{wightman2021resnet}. We then moved to the ConvNeXt models, where we limited our studies to its three first variants namely: the tiny, the small and the base models with input crops of size $224 \times 224$ \citep{liu2022convnet}. Here, we drop-in replaced all the depthwise convolutions of the model by DCLS ones. We reconducted all the experiments, and evaluated the seed sensitivity for the ConvNeXt-dcls model by calculating the standard deviation of the top-1 accuracy on three different seeds, for the tiny variant. We found the standard deviation to be $\pm 0.04$, which is compatible with what was found in \cite{liu2022convnet}. Given this reasonably low variability, the remaining experiments were done on one seed only. Code and scripts to reproduce the training are available at\footnote{\url{https://github.com/K-H-Ismail/ConvNeXt-dcls}}. 

\subsection{Empirical Evaluations on ImageNet1k}
In the following, we report the top-1 accuracies found on the ImageNet1k validation dataset \citep{deng2009imagenet}, using ImageNet1k training dataset only. 

\label{sec:res_resnet50}
\textbf{Using ResNet-50.} Table \ref{tab:1} presents the results obtained for the ResNet-50 experiment using the A3 configuration. The aim of this study is not to exceed the state-of-the-art for this particular configuration, but rather to give a first experimental evidence of the relevance of the DCLS method with non-separable convolutions and with one of the most popular CNN architectures in computer vision. We can observe that when using the standard dilated convolution, the results only get worse as we increase the dilation rate. Moreover, increasing the kernel size ($3 \rightarrow 7$) in the case of standard convolution increases the accuracy but at the expense of not only the throughput, which decreases, but also of the number of parameters, which triples. 

With fewer parameters, the ResNet-50-dcls model surpasses the baseline but at the expense of the throughput. This last disadvantage is due to the fact that ResNet-50 uses non-separable convolutions. We will see in Table \ref{tab:2} that for the ConvNeXt model, this extra cost in throughput is minimized thanks to the depthwise separable convolution.

\begin{table}[htbp]
\resizebox{\textwidth}{!}{
$
\begin{array}{lccccccc}
\text { model } & \begin{array}{l}
\text { kernel size } \\
\text { / count }
\end{array} & \text { dil } & \text { \# param.} & \text { FLOPs } & \begin{array}{l}
\text { throughput } \\
\text { (image / s) }
\end{array} &  \begin{array}{l}
\text { Top-1 acc. } \\
\text { (crop 160) }
\end{array}\\
\hline \text { ResNet-50 } & 3 / 9 & 1 & 25.6 \mathrm{M} & 4.1 \mathrm{G} & \textbf{1021.9} & 75.8 \\
\text { ResNet-50 } & 7 / 49 & 1 & 75.9 \mathrm{M} & 12.3 \mathrm{G} & 642.6 & \textbf{77.0} \\
\text { ResNet-50 } & 3 / 9 & 2 & 25.6 \mathrm{M} & 4.1 \mathrm{G} & 931.8 & 71.7 \\
\text { ResNet-50 } & 3 / 9 & 3 & 25.6 \mathrm{M} & 4.1 \mathrm{G} & 943.4 & 70.1 \\
\hline
\rowcolor{lightcream} \text { ResNet50-dcls \tikzcircle[dcls, fill=dcls]{2pt}} & 7 / 5 & - & 24.0 \mathrm{M} & 12.3 \mathrm{G} & 627.2 & 76.5 \\
\rowcolor{lightcream} \text { ResNet50-dcls \tikzcircle[dcls, fill=dcls]{2pt}} & 7 / 6 & - & 26.0 \mathrm{M} & 12.3 \mathrm{G} & 627.1 & 76.5 \\
\hline
\\
\end{array}
$
}

\caption{\textbf{Classification accuracy on ImageNet-1K using ResNet-50.} The throughput was calculated at inference time, on image crops of size $224 \times 224$ using a single V100-32gb gpu. When the model contains DCLS convolutions, we reported the kernel count and dilated kernel size. Otherwise, the kernel size is reported and thus the kernel count is in fact the square of that parameter.}
\label{tab:1}
\end{table}

\begin{table}[ht]
\resizebox{\textwidth}{!}{
$
\begin{array}{lccccc}

\text { model } & \text { img size }  & \text { \# param.} & \text { FLOPs } & \begin{array}{l}
\text { throughput } \\
\text { (image / s) }
\end{array} & \text { Top-1 acc. } \\
\hline \text { Swin-T } & 224^{2} & 28 \mathrm{M} & 4.5 \mathrm{G} & 757.9 & 81.3 \\
\text { ConvNeXt-T \tikzcircle[convnext, fill=convnext]{2pt} } & 224^{2} & 29 \mathrm{M} & 4.5 \mathrm{G} & \mathbf{774.7} & 82.1 \\
\text { ConvNeXt-T-dil2 } & 224^{2} & 29 \mathrm{M} & 4.5 \mathrm{G} & \mathbf{773.6} & 80.8 \\
\text { ConvNeXt-T-ker17 } & 224^{2} & 30 \mathrm{M} & 5 \mathrm{G} & 560.0 & 82.0 \\
\text { SLaK-T \tikzcircle[slak_sparse, fill=slak_sparse]{2pt} \tikzcircle[slak, fill=slak]{2pt}} & 224^{2} & 30^{\tikzcircle[slak_sparse, fill=slak_sparse]{2pt}} \ / \ 38^{\tikzcircle[slak, fill=slak]{2pt}} \mathrm{M} & 5.0^{\tikzcircle[slak_sparse, fill=slak_sparse]{2pt}} \ / \ 9.4^{\tikzcircle[slak, fill=slak]{2pt}}\mathrm{G} & 583.5 & \mathbf{8 2 . 5} \\
\rowcolor{lightcream} \text { ConvNeXt-T-dcls \tikzcircle[dcls, fill=dcls]{2pt}} & 224^{2} & 29 \mathrm{M} & 5.0 \mathrm{G} & 725.3 & \mathbf{8 2 . 5} \\
\hline
\text { Swin-S } & 224^{2} & 50 \mathrm{M} & 8.7 \mathrm{G} & 436.7 & 83.0 \\
\text { ConvNeXt-S \tikzcircle[convnext, fill=convnext]{2pt}} & 224^{2} & 50 \mathrm{M} & 8.7 \mathrm{G} & \mathbf{447.1} & 83.1 \\
\text { SLaK-S \tikzcircle[slak_sparse, fill=slak_sparse]{2pt} \tikzcircle[slak, fill=slak]{2pt}} & 224^{2} & 55^{\tikzcircle[slak_sparse, fill=slak_sparse]{2pt}} \ / \ 75^{\tikzcircle[slak, fill=slak]{2pt}} \mathrm{M} & 9.8^{\tikzcircle[slak_sparse, fill=slak_sparse]{2pt}} \ / \ 16.6^{\tikzcircle[slak, fill=slak]{2pt}} \mathrm{G} & 367.9 & \mathbf{83.8} \\
\rowcolor{lightcream} \text { ConvNeXt-S-dcls \tikzcircle[dcls, fill=dcls]{2pt}} & 224^{2} & 50 \mathrm{M} & 9.5 \mathrm{G} & 433.4 & \textbf{83.7} \\
\hline
\text { Swin-B } & 224^{2} & 88 \mathrm{M} & 15.4 \mathrm{G} & 286.6 & 83.5 \\
\text { ConvNeXt-B \tikzcircle[convnext, fill=convnext]{2pt}} & 224^{2} & 89 \mathrm{M} & 15.4 \mathrm{G} & \mathbf{292.1} & 83.8 \\
\text { RepLKNet-31B \tikzcircle[replknet, fill=replknet]{2pt}} & 224^{2} & 79 \mathrm{M} & 15.4 \mathrm{G} & \mathbf{295.5} & 83.5 \\
\text { SLaK-B \tikzcircle[slak_sparse, fill=slak_sparse]{2pt} \tikzcircle[slak, fill=slak]{2pt}} & 224^{2} & 95 ^{\tikzcircle[slak_sparse, fill=slak_sparse]{2pt}} \ / \ 122 ^{\tikzcircle[slak, fill=slak]{2pt}} \mathrm{M} & 17.1 ^{\tikzcircle[slak_sparse, fill=slak_sparse]{2pt}} \ / \ 25.9 ^{\tikzcircle[slak, fill=slak]{2pt}} \mathrm{G} & 245.4 & 84.0 \\
\rowcolor{lightcream} \text { ConvNeXt-B-dcls \tikzcircle[dcls, fill=dcls]{2pt}} & 224^{2} & 89 \mathrm{M} & 16.5 \mathrm{G} & 285.4 & \mathbf{84.1} \\
\hline \\
\end{array}
$
}
\caption{\textbf{Classification accuracy on ImageNet-1K.} The inference throughput was calculated at inference using a single V100-32gb gpu and scaled to take into account all the optimizations used in \cite{liu2022convnet}. For the SLaK model, we report both the effective number of parameters and FLOPs returned by PyTorch \tikzcircle[slak, fill=slak]{2pt} and the one reported in \cite{liu2022more} \tikzcircle[slak_sparse, fill=slak_sparse]{2pt}, that takes sparsity into account.}
\label{tab:2}
\end{table}

\textbf{Using ConvNeXt.} We present numerically in Table \ref{tab:2} and graphically in  Fig. \ref{fig:0}, the results obtained for ConvNeXt using the settings for ImageNet1k training with input crops of size $224 \times 224$, as described in \cite{liu2022convnet}: Table 5. Exponential Moving Average (EMA) \citep{polyak1992acceleration} was used, and for all models in Table \ref{tab:2}, we report the accuracy found with this technique. We replaced ConvNeXt's depthwise separable convolutions (of kernel size $7 \times 7$), by 2D-DCLS ones of dilated kernel size $17 \times 17$ and of kernel count equal to 34 for the tiny variant, and 40 for the small and base variants. We used all the techniques previously described in Section~\ref{sec:usage} during training. From Table \ref{tab:2}, we highlight the fact that ConvNeXt with DCLS convolutions always surpasses the ConvNeXt baseline in accuracy (gains ranging from 0.3 to 0.6) with the same number of parameters and only a little cost on throughput. We believe that this gain in accuracy is remarkable, given that we only replaced the depthwise convolutional layers, which represent just about 1\% of the total number of parameters and 2\% of the total number of FLOPs in ConvNeXt models. ConvNeXt model with a standard dilation of rate 2 performed poorly (see ConvNeXt-T-dil2). SLaK model performs about as well as DCLS but with a higher cost on throughput and parameters count.

DCLS can be seen as a kernel reparametrization technique that reduces the number of trainable parameters, and thus regularizes large kernels. For example, in the case of ConvNeXt-T-dcls, a $17 \times 17$ kernel ($289$ parameters) is parameterized by $34$ triplets (x-position, y-position, weight), i.e. 102 parameters. The kernels that could be represented by the DCLS reparametrization constitute a subset of all possible dense kernels. In fact, by learning the suitable weights during training, a dense kernel could implement any DCLS one. It may therefore be counter-intuitive that DCLS leads to higher accuracy than a dense $17 \times 17$ convolution layer (see ConvNeXt-T-ker17). The problem with dense convolutional layers having large kernel sizes is that the number of trainable parameters is huge, which makes learning impractical. 
 Finally, we observe that after training, the DCLS position density is higher around the center of the RF (see Appendix~\ref{appendix:hist}), suggesting that the central region is the most important one (yet we experienced that reducing the dilated kernel size to values $<17$ degrades the accuracy, so the positions far from the center also matter). Conversely, DC samples the whole RF uniformly, which is most likely sub-optimal, which could explain its poor performance (see ConvNeXt-T-dil2).

\subsection{Empirical Evaluation on Downstream and Robustness Tasks}
We now report the results found for semantic segmentation on the ADE20K dataset \citep{zhou2019semantic} and for object detection on the COCO dataset \citep{lin2014microsoft} using ConvNeXt-dcls backbones. Note that the \textit{depthwise implcit gemm} algorithm was not used for those tasks as it led to throughput issues. In addition, we present the results found for robustness tasks consisting of directly testing (without further tuning) the previously obtained ConvNeXt-dcls backbones on the following robustness benchmarks: ImageNet-C/$\overline{\text{C}}$/A/R/Sketch \citep{hendrycks2019benchmarking, mintun2021interaction, hendrycks2021natural, hendrycks2021many, wang2019learning}.

\textbf{Semantic segmentation on ADE20k.} The results obtained in semantic segmentation show an improvement in performance by the ConvNeXt-dcls tiny and base backbones with equal number of parameters and FLOPs (Table~\ref{tab:ade}). As in \cite{liu2021swin} and \cite{bao2021beit}, we evaluated the mIoU with single scale testing and used the exact same configurations as in \cite{liu2022convnet}.

\begin{table}[ht]
\resizebox{\textwidth}{!}{
$
\begin{array}{lccccc}

\text { backbone } & \text { input crop. } & \text { mIoU (ss) } & \text { \# param. } & \text { FLOPs } & \begin{array}{l}
\text { throughput } \\
\text { (image / s) }
\end{array}  \\
\hline
\text { ConvNeXt-T \tikzcircle[convnext, fill=convnext]{2pt}} & 512^{2} & 46.0 & 60 \mathrm{M} & 939 \mathrm{G} & 23.9 \\
\text { SLaK-T \tikzcircle[slak_sparse, fill=slak_sparse]{2pt}} & 512^{2} & \mathbf{47.1} & 65 \mathrm{M} & 945 \mathrm{G} & -  \\
\rowcolor{lightcream} \text { ConvNeXt-T-dcls \tikzcircle[dcls, fill=dcls]{2pt}} & 512^{2} & \mathbf{47.1} & 60 \mathrm{M} & 950 \mathrm{G} & 21.1  \\
\hline
\text { ConvNeXt-S \tikzcircle[convnext, fill=convnext]{2pt}} & 512^{2} & \mathbf{48.7} & 82 \mathrm{M} & 1027 \mathrm{G} & 22.1  \\
\rowcolor{lightcream} \text { ConvNeXt-S-dcls \tikzcircle[dcls, fill=dcls]{2pt}} & 512^{2} & 48.4 & 82 \mathrm{M} & 1045 \mathrm{G} & 19.5  \\
\hline
\text { ConvNeXt-B \tikzcircle[convnext, fill=convnext]{2pt}} & 512^{2} & 49.1 & 122 \mathrm{M} & 1170 \mathrm{G} & 21.7  \\
\rowcolor{lightcream} \text { ConvNeXt-B-dcls \tikzcircle[dcls, fill=dcls]{2pt}} & 512^{2} & \mathbf{49.3} & 122 \mathrm{M} & 1193 \mathrm{G} & 18.6 \\
\hline \\
\end{array}
$
}
\caption{\textbf{ADE20K validation results} using UperNet \citep{xiao2018unified}. We report mIoU results with single-scale testing. FLOPs are based on input sizes of (2048, 512). The inference throughput was calculated at inference using a single A100-80gb gpu and for input sizes of (3, 512, 512). }
\label{tab:ade}
\end{table}
\textbf{Object detection and segmentation on COCO.} All ConvNeXt-dcls backbones have shown a noticeable improvement in average accuracy on both the object detection and segmentation tasks on the COCO dataset, again at iso-parameters and iso-FLOPS (Table~\ref{tab:coco}). We only tested with Cascade Mask-RCNN \citep{cai2018cascade} and used the exact same configurations as in \cite{liu2022convnet}.
\begin{table}[!htbp]
\resizebox{\textwidth}{!}{
$
\begin{array}{lccccccccc}
\text { backbone } & \text { FLOPs }  & \text { TPUT }
& \text { AP }^{\text {box }} & \text { AP }_{50}^{\text {box }} & \text { AP }_{75}^{\text {box }} & \text { AP }^{\text {mask }} & \text { AP}_{50}^{\text {mask }} & \text { AP }_{75}^{\text {mask }} \\
\hline \multicolumn{8}{c}{\text { Cascade Mask-RCNN } 3 \times \text { schedule }} \\
\text { ResNet-50 } & 739 \mathrm{G} & - & 46.3 & 64.3 & 50.5 & 40.1 & 61.7 & 43.4 \\
\text { X101-32 } & 819 \mathrm{G} & - & 48.1 & 66.5 & 52.4 & 41.6 & 63.9 & 45.2 \\
\text { X101-64 } & 972 \mathrm{G} & - & 48.3 & 66.4 & 52.3 & 41.7 & 64.0 & 45.1 \\
\hline 
\text { Swin-T } & 745 \mathrm{G} & - & 50.4 & 69.2 & 54.7 & 43.7 & 66.6 & 47.3 \\
\text { ConvNeXt-T \tikzcircle[convnext, fill=convnext]{2pt}} & 741 \mathrm{G} & 11.6 & 50.4 & 69.1 & 54.8 & 43.7 & 66.5 & 47.3 \\
\rowcolor{lightcream} \text { CNeXt-dcls-T \tikzcircle[dcls, fill=dcls]{2pt}} & 751 \mathrm{G} & 11.2 &  \mathbf{5 1 . 2} & 69.9 & 55.7 & \mathbf{4 4 . 5} & 67.5 & 48.3 \\
\hline
\text { Swin-S } & 838 \mathrm{G} & - & 51.9 & 70.7 & 56.3 & 45.0 & 68.2 & 48.8 \\
\text { ConvNeXt-S \tikzcircle[convnext, fill=convnext]{2pt}} & 827 \mathrm{G} & 11.2 & 51.9 & 70.8 & 56.5 & 45.0 & 68.4 & 49.1 \\
\rowcolor{lightcream} \text { CNeXt-dcls-S \tikzcircle[dcls, fill=dcls]{2pt}} & 844 \mathrm{G} & 10.5 & \mathbf{52.8} & 71.6 & 57.6 & \mathbf{45.6} & 69.0 & 49.3 \\
\hline
\text { Swin-B } & 982 \mathrm{G} & - & 51.9 & 70.5 & 56.4 & 45.0 & 68.1 & 48.9 \\
\text { ConvNeXt-B \tikzcircle[convnext, fill=convnext]{2pt}} & 964 \mathrm{G} & 11.1 & 52.7 & 71.3 & 57.2 & 45.6 & 68.9 & 49.5 \\
\rowcolor{lightcream} \text { CNeXt-dcls-B \tikzcircle[dcls, fill=dcls]{2pt}} & 987 \mathrm{G} & 10.3 & \mathbf{53.0} & 71.5 & 57.7 & \mathbf{46.0} & 69.3 & 50.0 \\ 
\hline \\
\end{array}
$
}
\caption{ \textbf{COCO object detection and segmentation results} using
 Cascade Mask-RCNN. Average Precision of the
ResNet-50 and X101 models are from \citep{liu2021swin}. FLOPs are calculated with image size (3, 1280, 800). The inference throughput (``TPUT'') was calculated at inference using a single A100-80gb gpu and for input sizes of (3, 512, 512).}

\label{tab:coco}
\end{table}

\textbf{Robustness Evaluation on ImageNet-C/$\overline{\text{C}}$/A/R/Sketch.} ConvNeXt-dcls backbones show very good performances when it comes to robustness. This is illustrated by the results obtained for the different benchmarks we have tried and for which we have reconducted the experiments. All of them show a gain in classification accuracy with DCLS, except SK with the S model (Table~\ref{tab:robust}).

\begin{table}[!htbp]
\begin{center}
$$
\begin{array}{l|lccccccc}
\text { Model } & \text { FLOPs / Params } & \text { Clean } & \mathrm{C}(\downarrow) & \overline{\mathrm{C}}(\downarrow) & \mathrm{A} & \mathrm{R} & \mathrm{SK} \\
\hline \text { ResNet-50 } & 4.1 / 25.6 & 76.1 & 76.7 & 57.7 & 0.0 & 36.1 & 24.1 \\
\hline 
\text { ConvNeXt-T \tikzcircle[convnext, fill=convnext]{2pt}} & 4.5 / 28.6 & 82.1 & 41.6 & 41.2 & 23.5 & 47.6 & 33.8 \\
\rowcolor{lightcream} \text { ConvNeXt-dcls-T \tikzcircle[dcls, fill=dcls]{2pt}} &  5.0 / 28.6 &\mathbf{ 82.5} & \mathbf{41.5} & \mathbf{39.7} & \mathbf{23.9} & \mathbf{47.8} & \mathbf{34.7} \\
\hline
\text { ConvNeXt-S \tikzcircle[convnext, fill=convnext]{2pt}} &  8.7 / 50.2 & 83.1 & 38.9 & 37.8 & 30.1 & 50.1 & \mathbf{37.1} \\
\rowcolor{lightcream} \text { ConvNeXt-dcls-S \tikzcircle[dcls, fill=dcls]{2pt}} & 9.5 / 50.2 & \mathbf{83.7} & \mathbf{37.8} & \mathbf{35.2} & \mathbf{33.7} & \mathbf{50.4} & 36.7 \\
\hline
\text { ConvNeXt-B \tikzcircle[convnext, fill=convnext]{2pt}} &  15.4 / 88.6 & 83.8 & 37.0 & 35.7 & 35.5 & 51.7 & 38.2 \\
\rowcolor{lightcream} \text { ConvNeXt-dcls-B \tikzcircle[dcls, fill=dcls]{2pt}} &  16.5 / 88.6 & \mathbf{84.1} & \mathbf{36.3} & \mathbf{34.3} & \mathbf{36.8} & \mathbf{52.6} & \mathbf{38.4} \\
\hline
\end{array}
$$
\end{center}
\caption{\textbf{Robustness evaluation of ConvNeXt-dcls}. We reconducted this study for ConvNeXt. For ImageNet-C and ImageNet-Cbar, the error is reported rather than the accuracy. It was calculated for both datasets by taking the average error over 5 levels of noise severity and over all the noise categories available in the datasets.}
\label{tab:robust}
\end{table}

\section{Related work}
\label{chap2:related}
One of the studies that motivated the DCLS method is that of the \emph{effective receptive field}~ (ERF) \citep{luo2016understanding}, which characterizes how much each input pixel in a receptive field can impact the output of a unit in the downstream convolutional layers of a neural network, leading to the notion of an effective receptive field. Among the findings of this study, the most relevant ones for ours are that not all pixels in a receptive field contribute equally to an output response, the kernel center has a much larger impact, and that the effective receptive field size increases linearly with the square root of convolutional layers. Given these findings, introducing a new degree of freedom by learning the positions of non-zero weights in dilated kernels might increase the expressive power of convolutional neural networks. A visual comparison between DCLS and non-DCLS ERFs is available in Appendix~\ref{appendixF}.

\subsection{Active convolution}
\label{active_convolution}
Active convolution \cite{jeon2017active} was most probably the first work to tackle the question of learning positions inside a convolutional kernel throughout training with backpropagation. Active convolution was a true pioneer in this field, although it received relatively little attention compared to almost contemporary or even more recent approaches such as deformable convolution \cite{dai2017deformable}.

The DCLS method was developed independently from the active convolution method, and unfortunately, it was only after the publication of the DCLS \cite{hassani2021dilated} method that we came across this work, which is the closest to our own. However, it is important to note the differences between the DCLS method and the active convolution one. 

Active convolution uses small kernel sizes $3 \times 3$ in \cite{jeon2017active} and  $9 \times 9 $ in \cite{jeon2020integrating} against dilated kernel sizes of $17 \times 17$ or $23 \times 23$ in DCLS which contains up to 34 kernel elements. For us, this is a real limitation of the active convolution method, as we have found empirically that as the dilated kernel size increases, the performance of models equipped with the DCLS method also increases \ref{fig:kernel_size_effect}. Additionally, the ability to distinguish the dilated kernel size from the kernel count is crucial, as it allows us to maintain the same number of parameters in the model by choosing a suitable kernel count, which can be a non-square number.

Another important limitation of the active convolution method is that the positions of elements in the kernel are shared among channels. This greatly limits the expressivity of the method, since instead of having learnable positions of the same count as twice the weights $W$ (along the x-axis and the y-axis), in active convolution there is only $2 \times k_1 \times k_2 = 18$ learnable parameters for the positions.

An attempt to learn independent positions per channel has been proposed by the same authors of active convolution in \cite{jeon2020integrating}. This last work is closer to DCLS in that it introduces a notion of learnable positions by groups of channels of a grouped convolution \ref{grouped_conv}. Without reaching the maximal group size, i.e. depthwise separable convolution, grouped active convolution is limited to groups of size 16 over channels of size 64 or 128. With hindsight, we now know that performing standard convolution with a DCLS kernel (having ($C_{out}, C_{in},k_1, k_2$) weights thus ($2, C_{out}, C_{in},k_1, k_2$) learnable positions) is very expensive in terms of computation time and that a very good way of amortizing this cost is to build a DCLS kernel for depthwise convolution (with a number of parameters equal to ($C_{out}, 1,k_1, k_2$).

It is also worth noting some of the findings that have been made independently of both approaches and which confirm each other.

The histograms of positions obtained in \cite{jeon2020integrating} emphasize the role of central positions. This was also a main finding of histograms obtained with the DCLS method \ref{appendix:hist} with the distinction that the positions at certain locations of the kernel also play an important role \ref{fig:hist_images}, although less important than the center. We call these privileged locations ``attractors'' \ref{fig:hist_images}, \ref{fig:hist_audio}.

A notable feature of active convolution is that one element of the kernel is always fixed at the center, which makes sense since this element is the most sampled from a receptive field point of view (\ref{sec:rf}). Given the importance of this central position, there's no need to move it.

Active convolution uses bilinear interpolation to smoothen the integer nature of kernel positions, the same applies to the first version of DCLS known as DCLS-bilinear or DCLS v1 (\ref{chap:2}). In the second version of DCLS, DCLS-gauss (\ref{chap:3}), which will be the subject of the next chapter, we use a Gaussian interpolation.

Some training techniques have also been independently investigated by active convolution and by our work such as: scaling or normalizing the gradients of positions. Positions of kernel elements are special learnable parameters that have a different magnitude from convolution weights, therefore a scaling of these parameters is often required. \citet{jeon2017active} proposed to normalize the gradients of the positions. We have tried that independently. We empirically found that using the same one-cycle scheduler for the weights and positions and scaling the scheduled values of the learning rate (by a factor of 5 with respect to the learning rate of the weights) was better.

Training the convolution weights first with frozen positions was also considered by \cite{jeon2017active}. We empirically found that this warming up works but the best practice is still to train the weights and positions conjointly from the beginning. Training the positions only with frozen convolution weights increases accuracy as well, but this ``reversed'' warming up is much less beneficial than the previously cited one.

Besides, DCLS code is simple and very compact \ref{appendix:algo_torch}. This is mainly due to the fact that DCLS's kernel construction is written using high-level vectorized primitives provided by the PyTorch framework. Moreover, DCLS leverages the automatic differentiation implemented in the framework which greatly facilitates its implementation because the backward pass is inferred. 

Our first implementations of DCLS considered a from-scratch implementation of the method by rewriting the forward and backward passes in CUDA. Implementing DCLS without building the dilated kernel is what we call DCLS-im2col, and is an extension of the im2col algorithm. We realized that we could do without it and only use the kernel-construction alternative. Still, the DCLS-im2col code is available and will probably be useful to us in the future.

Finally, we would like to draw the reader's attention to the importance of the deep learning framework used for the research. Although it is independent of the quality of the research, the framework used can influence the popularity of an introduced method. 

\subsection{Deformable convolution}
\label{deformable}
Another work that relates to DCLS is that of deformable convolutions \cite{dai2017deformable} where offsets from the regular dilated grid are optimized. Deformable convolutions are used for several computer vision tasks such as object detection where they excel.

The deformable convolution approach is fundamentally different from DCLS or active convolution. The major difference is that deformable convolution receives the position parameters from an upstream standard convolution operation, and thus the positions of kernel elements are input and location-dependent, while approaches like DCLS, active convolution, or displaced aggregation units \cite{tabernik2020spatially} possess intrinsic position parameters. Even if all these approaches share the use of a rather similar interpolation mechanism, DCLS remains very different from deformable convolution in several aspects: firstly, as said before, deformable convolutions require the application of a regular convolution to get the offsets (which are thus input-dependent) that are then passed to the actual deformable method. Conversely, in DCLS method, a dilated kernel with learnable positions is constructed and then passed to the convolution operation, making the positions input-independent. Secondly, DCLS positions are channel-dependent, unlike deformable convolution ones. Finally, the number of extra learnable parameters in 2D deformable convolutions is the number of kernel elements in the preliminary convolution, precisely ($\frac{C_{in}}{g}, 2 k_1 k_2,  k_1, k_2$), which establishes a strong dependence between the offsets and the input feature map. Contrarily, in 2D-DCLS, the number of extra parameters dedicated to learning positions is simply twice the number of kernel weights ($2, C_{out}, \frac{C_{in}}{g}, k_1, k_2$). 

In its first version \cite{dai2017deformable}, deformable convolution v1 introduced the so-called convolution method along with a deformable region of interest (ROI) pooling derived directly from the deformable convolution. This deformable pooling could be used to further improve the performance of the convolution method. In the second version of the method \cite{zhu2019deformable}, the authors added learnable modulation scalars that could adjust the amplitude of the kernel elements. This is similar to the learning of standard deviations $\sigma$ in DCLS-Triangle and DCLS-Gauss, as we will see in the third chapter of this thesis (\ref{chap:3}). 

Deformable convolutions v1 and v2 have been developed only for standard convolutions, not for depthwise separable ones. We have not been able to compare our work to the deformable convolution v1/v2 in the ConvNeXt architecture that uses depthwise separable convolutions exclusively as 
we noticed that the training loss grew at the first epochs, which means that the early versions of the method were not adapted as a drop-in replacement of the depthwise separable convolution.

To address this problem, deformable convolution v3 has been developed \cite{wang2022internimage}. This third version of deformable convolution can be used as a drop-in replacement for depthwise separable convolution. The authors introduced a multi-group mechanism that permits grouped and depthwise separable convolutions. This latter version is at the heart of the InternImage model \ref{internimage}, which has demonstrated state-of-the-art results in several computer vision tasks.

\subsection{Displaced Aggregation Units}
\label{dau}

In \cite{tabernik2020spatially}, the authors introduce displaced aggregation units (DAUs). DAUs build on previous work by the same authors \cite{tabernik2016towards, tabernik2018spatially} and can be viewed as a type of Gaussian mixture kernels. In chapter (\ref{chap:3}), we discuss the related work concerning Gaussian mixture kernels. DAUs are similar to DCLS with Gaussian interpolation (\ref{chap:3}), yet different. 


DCLS and DAUs differ in their Gaussian density standard deviation adaptation strategies, with DAUs using a fixed, manually tuned standard deviation and DCLS using learnable sigmas for each kernel element (\ref{chap:3}).

Among the findings of this study, the optimal standard deviation of the Gaussian in \cite{tabernik2020spatially} was tuned manually and found empirically to be 0.5. This is very interesting as this is the initial value used in DCLS-Gauss where standard deviations are learned, since using this value gives the closest approximation to bilinear interpolation.

The benefits of DAUs were demonstrated in various computer vision tasks, including image classification, semantic segmentation, and blind image deblurring. Significant improvements in performance and parameter efficiency were observed across different architectures. The study in \cite{tabernik2020spatially} also highlighted the adaptability and self-pruning properties of DAU networks for image segmentation tasks. Additionally, \cite{tabernik2020spatially} explored the use of DAUs in deep residual network architectures and their effectiveness in semantic segmentation tasks, achieving improved performance and reduced parameter allocation.

This last work was implemented using the framework Caffe \cite{jia2014caffe}. We would appreciate it if a more recent version under PyTorch \cite{PyTorch} could be adapted to facilitate comparisons.

\subsection{Learning shifts in convolution kernels}
In \cite{wu2018shift}, “shift”, an operation that doesn't require additional learnable parameters or consumes FLOPs for inference is presented. The concept behind “shift” is to replace depthwise convolution (spatial mixing) with an operation aimed at finding the optimal combination of non-zero kernel elements within a convolution kernel. The shift operation can be viewed as a special case of depthwise convolutions with kernels belonging to the canonical basis. The authors call these canonical matrices shift matrices, and they show that this new operation, used as a drop-in replacement for the convolution method in the ResNet model, outperforms the latter in the image classification task on CIFAR \cite{krizhevsky2009learning} and ImageNet \cite{deng2009imagenet}, and shows furthermore, good resilience to parameter reduction.

Searching for the optimal shift in a kernel of size $C,k_1,k_2$ is very expensive as it involves exploring $(k_1k_2)^{C}$ possible kernels. The first thing that the author considered to reduce this exponential complexity is to use a simple heuristic where the shifts are considered among $k_1k_2$ groups where each group of the $\floor{\frac{C}{k_1k_2}}$ channels adopts the same common shift.

The now relaxed problem amounts to finding the optimal permutation, i.e., how to map each channel to a shift group. The complexity of this last problem is combinatorial. To amortize this complexity, the authors modified the shift operation such that the input and output become invariant to the channel order. Still, permutation operators are discrete and therefore impossible to optimize by gradient backpropagation. To address this problem, an augmented shift operation was created by placing the shift operation between two pointwise convolutions, allowing it to be trained end-to-end with gradient descent without regard to channel order.

Similar to \cite{wu2018shift}, \cite{jeon2018constructing} showed that convolution can be deconstructed into two components: shift operation and pointwise convolution. This reduces the number of parameters and computational complexity of standard convolutions. In \cite{wu2018shift}, shifts are assigned heuristically by grouping input channels, while in \cite{jeon2018constructing} they can be trained from random initialization through backpropagation. The proposed method in \cite{jeon2018constructing}, named active shift layer (ASL), can replace the existing spatial convolutions while reducing the computational complexity and inference time and was used to construct light and fast networks which showed state-of-the-art results with fewer parameters and low inference time.

\citet{hacene2021attention} later introduced Shift Attention Layers (SALs) as a new pruning shift operation method that replaces the convolution with a shift layer \cite{wu2018shift} by the end of training. The proposed methodology results in a network that is exactly as described in \cite{wu2018shift}, with the main difference being that shifts are learned instead of being predetermined. \citet{hacene2021attention} showed that higher accuracy can be obtained with exactly the same complexity and number of parameters.

\subsection{CoordConv}  
CoordConv \cite{liu2018intriguing} is a technique that brings a novel dimension to spatial understanding. The main idea of CoordConv lies in augmenting convolutional filters with an awareness of their position within the Cartesian space. This is achieved by incorporating additional, predefined input channels that convey the coordinates of the data being processed by the filter. \citet{liu2018intriguing} introduces a simplified toy dataset called Not-so-Clevr, containing randomly positioned squares on a canvas. The CoordConv operation provides convolutional filters with an intrinsic awareness of their location. This operation is demonstrated to significantly enhance performance across various tasks, even in complex scenarios involving rendering images from the Not-so-Clevr dataset based on square coordinates. By employing CoordConv, difficulties in tasks such as supervised coordinate classification and regression are notably alleviated. This improvement extends to a range of applications, including Generative Adversarial Networks (GANs), Variational Autoencoders (VAEs), and object detection tasks.  It's essential to underscore that while CoordConv exhibits remarkable promise in its experimental findings, these results are restricted to small-scale datasets. Moreover, the approach's efficacy is, as the deformable convolution method \ref{deformable}, profoundly input-dependent.

\subsection{More related work}
Another input-dependent convolution method that seeks to learn the dilation rates rather than the kernel positions is the one of ADCNN \citep{yao2022adcnn}. But in contrast to DCLS, this method uses regular grids with learnable rates.

DCLS is also similar to other input-independent kernel re-parameterization techniques, yet different from them. For example, in CKConv~\citep{romero2021ckconv} and FlexConv~\citep{romero2021flexconv}, the kernel weights are not learned directly; what is learned is the continuous function that maps the positions to the weights. In~\cite{jacobsen2016structured}, the kernel is modeled as a weighted sum of basis functions, which consist of centered Gaussian filters and their derivatives. \cite{pintea2021resolution} extended the approach by also learning the width of the Gaussians, which is equivalent to learning the optimal resolution. \cite{shelhamer2019blurring} proposed to factorize the kernel as the composition of a standard kernel with a structured Gaussian one. 
Finally, other methods such as \citep{worrall2019deep,sosnovik2019scale,sosnovik2021disco,sosnovik2021transform,bekkers2019b,zhu2019scaling}, where the goal is to build a scale equivariant neural network, could be considered as similar to the approach. One limitation of all these studies is that only small datasets were used (e.g., CIFAR), and whether they scale well to larger and more challenging datasets like ImageNet1k is unknown. In addition, they cannot be used as a drop-in replacement for the depthwise separable convolution in ConvNeXt, at least in their current forms, so we could not benchmark them with DCLS.
\section*{Reproducibility Statement}
\label{reproducibility}
The code of the method is based on PyTorch and available at \url{https://github.com/K-H-Ismail/Dilated-Convolution-with-Learnable-Spacings-PyTorch}. Code and scripts to reproduce the training and the experiments are available at \url{https://github.com/K-H-Ismail/ConvNeXt-dcls}.

\section{Appendix: proofs and derivation for the DCLS method}
\label{appendixA}

In the following, we show how to mathematically describe the DCLS kernel construction and how to explicitly calculate the gradients of the loss function with respect to weights and positions that are used in the backpropagation algorithm. These gradients are useful to implement the DCLS method in a way that is compatible with the automatic differentiation of PyTorch.

\subsection{Notation and preliminaries}
We denote by $\floor{ \ }$ the floor function and we define its derivative by the zero function.
\begin{equation}
    \forall x \in \mathbb{R},  \floor{x}' \overset{\text{def}}{=} 0
\end{equation} 

We denote by $m \in \mathbb{N}^{*}$ the number of kernel elements inside the constructed kernel and we refer to it as the “kernel count”. Moreover, we denote respectively by $s_1, s_2 \in \mathbb{N}^{*} \times \mathbb{N}^{* } $, the sizes of the constructed kernel along the x-axis and the y-axis. The latter could be seen as the limits of the dilated kernel, and we refer to them as the “dilated kernel size”.

The $n \times p$ matrix space over $\mathbb{R}$ is defined as the set of all $n \times p$ matrices over $\mathbb{R}$, and is denoted $\mathcal{M}_{n , p}(\mathbb{R})$.

The Frobenius inner product $\underset{\text{F}}{\times}$ of two matrices $\displaystyle \mA$ and $\displaystyle \mB$ of $\mathcal{M}_{n,p} (\mathbb{R})$ is defined by: 
$$\displaystyle \mA \underset{\text{F}}{\times}  \mB = \text{tr}( \mA^T  \mB)$$

Where “$\text{tr}$” stands for the trace of the square matrix  $\mA^T  \mB$.

 The characters $w$, $p^1$ and $p^2$ respectively stand for the weight, the position of that weight along the x-axis (width) and its position along the y-axis (height) in the scalar case while the bold $ \vw = (w_{i})_{1 \leq i \leq m}$, $ \vp^1 = (p^{1}_{i})_{1 \leq i \leq m}$ and $ \vp^2 = (p^{2}_{i})_{1 \leq i \leq m}$ respectively stand for the weight, the width-position of that weight and its height-position in the vector case.


The proofs and algorithms that will be shown in the next subsections are made for the case of tensors with one input channel and one output channel. Without loss of generality, those proofs and algorithms hold for the general case of 4D tensors and higher by considering and applying them channel-wise.

\subsection{2D-DCLS, scalar weight case}
We begin by the case of a kernel containing only one element. 

The function $f$ that defines the kernel construction in the scalar weight case is as follows:

\begin{align}
\label{eq:f}
  \begin{split}
  f \colon \mathbb{R} \times \mathbb{R} \times \mathbb{R} &\to \mathcal{M}_{s_1 , s_2 } (\mathbb{R})\\
  w, p^1, p^2  & \mapsto \displaystyle \mK
  \end{split}
\end{align}
where $\forall i\in \llbracket 1 \ .. \ s_1 \rrbracket$, $\forall j\in \llbracket 1 \ .. \ s_2 \rrbracket  \colon $\\
\begin{equation}
\arraycolsep=1.3pt\def\arraystretch{1}
\displaystyle \mK_{ij} =\left\{\begin{array}{cl}
w \ (1 - r^1)\ (1 - r^2) & \text {if } i = \floor{p^1}, \ j = \floor{p^2} \\
w \ r^1 \ (1 - r^2) & \text {if }  i = \floor{p^1} + 1, \ j = \floor{p^2} \\
w \ (1 - r^1) \ r^2 & \text {if }  i = \floor{p^1}, \ j = \floor{p^2} + 1\\
w \ r^1 \ r^2  & \text {if }  i = \floor{p^1} {+} 1, \ j = \floor{p^2} {+} 1 \\
0 & \text {otherwise } 
\end{array}\right.
\end{equation}
and where the fractional parts are:
\begin{equation}
    \begin{array}{ccc}
    r^1 = \{p^1\} = p^1 - \floor{p^1} & \text{and} & r^2 = \{p^2\} = p^2 - \floor{p^2}
    \end{array}
\end{equation}

The constructed kernel $\mK$ is zero except for at most the 4 adjacent positions that represent the 2D interpolation of the single weight $w$. Note that $\sum\limits_{j=1}^{s_2}\sum\limits_{i=1}^{s_1} \mK_{ij}=w$. 

We then define the scalar loss function as:
\begin{equation}
\label{eq:loss}
  loss = g(f(w,p^1,p^2))  
\end{equation}
with $g  \colon \mathcal{M}_{s_1 , s_2 } (\mathbb{R}) \to \mathbb{R}$ a differentiable function that models the action of all the layers that will follow $f$ in the model.

By applying the chain rule we obtain:
\begin{align}
\frac{\partial loss}{\partial w} = g'(f(w,p^1,p^2)) \underset{\text{F}}{\times} \frac{\partial f(w,p^1,p^2)}{\partial w}  \label{eq:chain1} \\ 
\frac{\partial loss}{\partial p^1} = g'(f(w,p^1,p^2)) \underset{\text{F}}{\times} \frac{\partial f(w,p^1,p^2)}{\partial p^1}  \label{eq:chain2} \\ 
\frac{\partial loss}{\partial p^2} = g'(f(w,p^1,p^2)) \underset{\text{F}}{\times} \frac{\partial f(w,p^1,p^2)}{\partial p^2} \label{eq:chain3}
\end{align}
with
\begin{equation}
    g'(f(w,p^1,p^2)) = \frac{\partial loss}{\partial \mK} = \frac{\partial loss}{\partial f(w,p^1,p^2)}
\end{equation}

Let us put 
\begin{equation}g'(f(w,p^1,p^2)) = \displaystyle \mG = 
\begin{bmatrix}
g_{11} & g_{12} & \cdots & g_{1 s_2}\\
g_{21} & \ddots &  & g_{2 s_2}\\
\vdots &  & \ddots & \vdots\\
g_{s_1 1} & g_{s_12} & \cdots & g_{s_1 s_2}\\
\end{bmatrix}
\end{equation}
and let us consider two column vectors $\displaystyle \vx = \begin{bmatrix}
x_1 & x_2 & \cdots & x_{s_1}
\end{bmatrix}^T$ of $\mathbb{R}^{s_1}$ and $\displaystyle \vy = \begin{bmatrix}
y_1 & y_2 & \cdots & y_{s_2}
\end{bmatrix}^T$ of $\mathbb{R}^{s_2}$. We have: 
\begin{equation}
\displaystyle \vx^T f(w,p^1,p^2) \displaystyle \vy = \sum_{i=1}^{s_1} \sum_{j=1}^{s_2} \displaystyle \mK_{ij} x_iy_j 
\end{equation}

Since $\mK$ is zero except for the 4 aforementioned positions, we have:
\begin{align}
  \begin{split}
    \displaystyle \vx^T f(w,p^1,p^2) \displaystyle \vy = & \ \mK_{\floor{p^1}\floor{p^2}} \ x_{\floor{p^1}} \ y_{\floor{p^2}}\\ 
        &+ \mK_{\floor{p^1}+1\floor{p^2}} \ x_{\floor{p^1}+1} \ y_{\floor{p^2}} \\
        &+ \mK_{\floor{p^1}\floor{p^2}+1} \ x_{\floor{p^1}} \ y_{\floor{p^2}+1} \\
        &+ \mK_{\floor{p^1}+1\floor{p^2}+1} \ x_{\floor{p^1}+1} \ y_{\floor{p^2}+1}
    \end{split}
\end{align}

By deriving this expression with respect to $w$, $p^1$ and $p^2$ we obtain:
\begin{align}
  \begin{split}
    \frac{\partial( \vx^T f(w,p^1,p^2) \vy)}{\partial w} = \ & \ (1 - r^1)\ (1 - r^2) \ x_{\floor{p^1}} \ y_{\floor{p^2}}\\ 
        &+ r^1\ (1 - r^2) \ x_{\floor{p^1}+1} \ y_{\floor{p^2}} \\
        &+ (1 - r^1)\ r^2 \ x_{\floor{p^1}} \ y_{\floor{p^2}+1} \\
        &+ r^1 \ r^2 \ x_{\floor{p^1}+1} \ y_{\floor{p^2}+1}
    \end{split}
\\
  \begin{split}
    \frac{\partial( \vx^T f(w,p^1,p^2) \vy)}{\partial p^1} = w \ [ & - (1 - r^2) \ x_{\floor{p^1}} \ y_{\floor{p^2}}\\ 
        &+ (1 - r^2) \ x_{\floor{p^1}+1} \ y_{\floor{p^2}} \\
        &-  r^2 \ x_{\floor{p^1}} \ y_{\floor{p^2}+1} \\
        &+  r^2 \ x_{\floor{p^1}+1} \ y_{\floor{p^2}+1}]
    \end{split}
\\
  \begin{split}
    \frac{\partial( \vx^T f(w,p^1,p^2) \vy)}{\partial p^2} = w \ [ & - (1 - r^1) \ x_{\floor{p^1}} \ y_{\floor{p^2}}\\ 
        &- r^1 \ x_{\floor{p^1}+1} \ y_{\floor{p^2}} \\
        &+ (1 - r^1) \ x_{\floor{p^1}} \ y_{\floor{p^2}+1} \\
        &+ r^1 \ x_{\floor{p^1}+1} \ y_{\floor{p^2}+1} ]
    \end{split}
\end{align}

Because of the linearity of the differentiation in the three previous equations, we could write:
\begin{align}
\frac{\partial( \vx^T f(w,p^1,p^2) \vy)}{\partial w} =  \vx^T\frac{\partial f(w,p^1,p^2)}{\partial w}\vy =  \vx^T \mG_w \vy \label{eq:linear1}\\
\frac{\partial( \vx^T f(w,p^1,p^2) \vy)}{\partial p^1} =  \vx^T\frac{\partial f(w,p^1,p^2)}{\partial p^1}\vy =  \vx^T \mG_{p^1} \vy \label{eq:linear2}\\
\frac{\partial( \vx^T f(w,p^1,p^2) \vy)}{\partial p^2} =  \vx^T\frac{\partial f(w,p^1,p^2)}{\partial p^2}\vy =  \vx^T \mG_{p^2} \vy  \label{eq:linear3}   
\end{align}
where $\mG_w$, $\mG_{p^1}$, $\mG_{p^2}$, respectively stand for the $s_1$ by $s_2$ matrices described below and which have zeros everywhere except at the four positions of interpolation.\\

$\mG_w = $
\begin{equation}
\begin{blockarray}{ccclccl}
 &  & \floor{p^2} & \floor{p^2}{+}1 &  & & \\
\begin{block}{(cccccc)l}
  0 & \cdots & 0 & 0 & \cdots & 0 &  \\
  \vdots & \ddots & 0 & 0 & \iddots & \vdots &  \\
  0 & 0 &  (1 {-} r^1) (1 {-} r^2) & r^2 (1 {-} r^1) & 0 & 0 & \floor{p^1} \\
  0 & 0 &  r^1 (1 {-} r^2) & r^1 r^2 & 0 & 0 & \floor{p^1}{+}1 \\
  \vdots & \iddots & 0 & 0 & \ddots & \vdots &  \\
  0 & \cdots & 0 & 0 & \cdots & 0 &  \\
\end{block}
\end{blockarray}
\end{equation}

$\mG_{p^1} = $
\begin{equation}
\begin{blockarray}{ccclccl}
 &  & \floor{p^2} & \floor{p^2}{+}1 &  & & \\
\begin{block}{(cccccc)l}
  0 & \cdots & 0 & 0 & \cdots & 0 &  \\
  \vdots & \ddots & 0 & 0 & \iddots & \vdots &  \\
  0 & 0 &  - w(1 {-} r^2) & -  wr^2 & 0 & 0 & \floor{p^1} \\
  0 & 0 &  w(1 {-} r^2) & wr^2 & 0 & 0 & \floor{p^1}{+}1 \\
  \vdots & \iddots & 0 & 0 & \ddots & \vdots &  \\
  0 & \cdots & 0 & 0 & \cdots & 0 &  \\
\end{block}
\end{blockarray}
\end{equation}

$\mG_{p^2} = $
\begin{equation}
\begin{blockarray}{ccclccl}
 &  & \floor{p^2} & \floor{p^2}{+}1 &  & & \\
\begin{block}{(cccccc)l}
  0 & \cdots & 0 & 0 & \cdots & 0 &  \\
  \vdots & \ddots & 0 & 0 & \iddots & \vdots &  \\
  0 & 0 &  - w(1 {-} r^1) & w(1 {-} r^1) & 0 & 0 & \floor{p^1} \\
  0 & 0 &  - wr^1 & wr^1 & 0 & 0 & \floor{p^1}{+}1 \\
  \vdots & \iddots & 0 & 0 & \ddots & \vdots &  \\
  0 & \cdots & 0 & 0 & \cdots & 0 &  \\
\end{block}
\end{blockarray}
\end{equation}

From the three equations (\ref{eq:linear1}), (\ref{eq:linear2}) and (\ref{eq:linear3}), we can identify
\begin{align}
\frac{\partial f(w,p^1,p^2)}{\partial w} = \mG_w  \\
\frac{\partial f(w,p^1,p^2)}{\partial p^1} = \mG_{p^1} \\
\frac{\partial f(w,p^1,p^2)}{\partial p^2} = \mG_{p^2}   
\end{align}

Finally, we have: 
\begin{align}
\begin{split}
\label{eq:res1}
\frac{\partial loss}{\partial w} =  \mG \underset{\text{F}}{\times}  \mG_w = \ & (1 - r^1)\ (1 - r^2) \ g_{\floor{p^1}\floor{p^2}} \\ 
    & + r^1\ (1 - r^2) \  g_{\floor{p^1}+1\floor{p^2}} \\
    & + (1 - r^1)\ r^2 \  g_{\floor{p^1}\floor{p^2}+1} \\
    & + r^1 \ r^2 \ g_{\floor{p^1}+1\floor{p^2}+1}
\end{split}
\\
\begin{split}
\label{eq:res2}
\frac{\partial loss}{\partial p^1} = \mG \underset{\text{F}}{\times}  \mG_{p^1}=  \ w \ [ & - (1 - r^2) \ g_{\floor{p^1}\floor{p^2}}\\ 
    & + (1 - r^2) \ g_{\floor{p^1}+1\floor{p^2}}\\
    & -  r^2 \ g_{\floor{p^1}\floor{p^2}+1}\\
    & +  r^2 \ g_{\floor{p^1}+1\floor{p^2}+1}]
\end{split}
\\
\begin{split}
\label{eq:res3}
\frac{\partial loss}{\partial p^2} = \mG \underset{\text{F}}{\times}  \mG_{p^2} = \ w \ [ & - (1 - r^1) \ g_{\floor{p^1}\floor{p^2}}\\ 
    & - r^1 \ g_{\floor{p^1}+1\floor{p^2}}\\
    & + (1 - r^1) \ g_{\floor{p^1}\floor{p^2}+1}\\
    & + r^1 \ g_{\floor{p^1}+1\floor{p^2}+1}]
\end{split}
\end{align}

In the next subsection, we will see how this result can be generalized to the vector case.
\subsection{2D-DCLS, general case}
\label{sec:gen_case}
The general case is the one where the weights $\vw = \begin{bmatrix} w_1 & w_2 & \cdots & w_{m} \end{bmatrix}^T$ and the positions $\vp^1 = \begin{bmatrix} p^1_1 & p^1_2 & \cdots & p^1_{m} \end{bmatrix}^T$ , $\vp^2 = \begin{bmatrix} p^2_1 & p^2_2 & \cdots & p^2_{m} \end{bmatrix}^T$  are stored in vectors, with the fractional parts $\vr^1 = \{\vp^1\} = \vp^1 - \floor{\vp^1} = \begin{bmatrix} r^1_1 & r^1_2 & \cdots & r^1_{m} \end{bmatrix}^T$ and $\vr^2 = \{\vp^2\} = \vp^2 - \floor{\vp^2} = \begin{bmatrix} r^2_1 & r^2_2 & \cdots & r^2_{m} \end{bmatrix}^T$ extended as well. 

The function $f$ defined in equation (\ref{eq:f}) is then extended to the function $F$ defined as follows:
\begin{align}
  \begin{split}
  F \colon  \quad \mathbb{R}^m \times \mathbb{R}^m \times \mathbb{R}^m   &\to \mathcal{M}_{s_1 , s_2 } (\mathbb{R})\\
  \vw, \vp^1, \vp^2 \mapsto &  \mK = \sum_{i = 1}^{m}  f(w_{i},p^{1}_{i},p^{2}_{i})
  \end{split}
\end{align}

The constructed kernel $\mK$ here is the result of a summation of the function $f$ defined in (\ref{eq:f}) over the elements of weight and position vectors. We then define the scalar loss function as in (\ref{eq:loss}).

\begin{equation}
\label{eq:lossF}
  loss = g(F(\vw, \vp^1, \vp^2))  
\end{equation}

with $g  \colon \mathcal{M}_{s_1 , s_2 } (\mathbb{R}) \to \mathbb{R}$ a differentiable function that models the action of all the layers that will follow $F$ in the model.

Let us put 
\begin{equation}g'(F(\vw,\vp^1,\vp^2)) =  \mG = 
\begin{bmatrix}
g_{11} & g_{12} & \cdots & g_{1 s_2}\\
g_{21} & \ddots &  & g_{2 s_2}\\
\vdots &  & \ddots & \vdots\\
g_{s_1 1} & g_{s_12} & \cdots & g_{s_1 s_2}\\
\end{bmatrix}
\end{equation}

As in (\ref{eq:chain1}), (\ref{eq:chain2}) and (\ref{eq:chain3}), by applying the chain rule we obtain:

$\forall i\in \llbracket 1 \ .. \ m \rrbracket \colon $
\begin{align}
\frac{\partial loss}{\partial w_i} = g'(F(\vw,\vp^1,\vp^2)) \underset{\text{F}}{\times} \frac{\partial F(\vw,\vp^1,\vp^2)}{\partial w_i} \label{eq:chain_gen1}\\ 
\frac{\partial loss}{\partial p_{i}^1} = g'(F(\vw,\vp^1,\vp^2)) \underset{\text{F}}{\times} \frac{\partial F(\vw,\vp^1,\vp^2)}{\partial p_{i}^1} \label{eq:chain_gen2}\\
\frac{\partial loss}{\partial p_{i}^2} = g'(F(\vw,\vp^1,\vp^2)) \underset{\text{F}}{\times} \frac{\partial F(\vw,\vp^1,\vp^2)}{\partial p_{i}^2} \label{eq:chain_gen3}\
\end{align}
with
\begin{equation}
    \begin{split}
    g'(F(\vw,\vp^1,\vp^2)) = &\frac{\partial loss}{\partial \displaystyle \mK} = \frac{\partial loss}{\partial F(\vw,\vp^1,\vp^2)}
    \end{split}
\end{equation}

Let us put this time
\begin{equation}g'(F(\vw,\vp^1,\vp^2)) =  \mG = 
\begin{bmatrix}
g_{11} & g_{12} & \cdots & g_{1 s_2}\\
g_{21} & \ddots &  & g_{1 s_2}\\
\vdots &  & \ddots & \vdots\\
g_{s_1 1} & g_{12} & \cdots & g_{s_1 s_2}\\
\end{bmatrix}
\end{equation}

Using the definition of $F$, and by substituting the last equation in (\ref{eq:chain_gen1}), (\ref{eq:chain_gen2}) and (\ref{eq:chain_gen3}), we have:

\begin{align}
\label{eq:gen1}
\frac{\partial loss}{\partial w_i} = \mG \underset{\text{F}}{\times} \frac{\partial \sum_{i = 1}^{m}  f(w_{i},p^{1}_{i},p^{2}_{i})}{\partial w_i}\\ 
\label{eq:gen2}
\frac{\partial loss}{\partial p_{i}^1} = \mG \underset{\text{F}}{\times} \frac{\partial\sum_{i = 1}^{m}  f(w_{i},p^{1}_{i},p^{2}_{i})}{\partial p_{i}^1}\\
\label{eq:gen3}
\frac{\partial loss}{\partial p_{i}^2} = \mG\underset{\text{F}}{\times} \frac{\partial \sum_{i = 1}^{m}  f(w_{i},p^{1}_{i},p^{2}_{i})}{\partial p_{i}^2}\
\end{align}

And we know that: 

$ \forall (i,j) \in \llbracket 1 \ .. \ m \rrbracket ^ 2 \colon$  
$$i \neq j \implies \frac{\partial f(w_{j},p^{1}_{j},p^{2}_{j})}{\partial w_{i}} = \frac{\partial f(w_{j},p^{1}_{j},p^{2}_{j})}{\partial p_{i}^1} = \frac{\partial f(w_{j},p^{1}_{j},p^{2}_{j})}{\partial p_{i}^2} = 0 $$

which simplifies equations (\ref{eq:gen1}), (\ref{eq:gen2}) and (\ref{eq:gen3}) to

$ \forall i \in \llbracket 1 \ .. \ m \rrbracket  \colon$  
\begin{align}
\frac{\partial loss}{\partial w_i} = \mG \underset{\text{F}}{\times} \frac{\partial f(w_{i},p^{1}_{i},p^{2}_{i})}{\partial w_i}\\ 
\frac{\partial loss}{\partial p_{i}^1} = \mG \underset{\text{F}}{\times} \frac{\partial  f(w_{i},p^{1}_{i},p^{2}_{i})}{\partial p_{i}^1}\\
\frac{\partial loss}{\partial p_{i}^2} = \mG\underset{\text{F}}{\times} \frac{\partial f(w_{i},p^{1}_{i},p^{2}_{i})}{\partial p_{i}^2}\
\end{align}

We can notice that we are brought back to the scalar case, and we deduce from (\ref{eq:res1}), (\ref{eq:res2}) and (\ref{eq:res3}) the gradients of the loss function with respect to weights and positions in the general case:

$ \forall i \in \llbracket 1 \ .. \ m \rrbracket  \colon$  
\begin{align}
\begin{split}
\label{eq:res4}
\left(\frac{\partial loss}{\partial \vw}\right)_{i} = \ & (1 - r_{i}^1)\ (1 - r_{i}^2) \ g_{\floor{p_{i}^1}\floor{p_{i}^2}} \\ 
    & + r_{i}^1\ (1 - r_{i}^2) \  g_{\floor{p_{i}^1}+1\floor{p_{i}^2}} \\
    & + (1 - r_{i}^1)\ r_{i}^2 \  g_{\floor{p_{i}^1}\floor{p_{i}^2}+1} \\
    & + r_{i}^1 \ r_{i}^2 \ g_{\floor{p_{i}^1}+1\floor{p_{i}^2}+1}
\end{split}
\\
\begin{split}
\label{eq:res5}
\left(\frac{\partial loss}{\partial \vp^1}\right)_{i} = \ w_{i} \ [ & - (1 - r_{i}^2) \ g_{\floor{p_{i}^1}\floor{p_{i}^2}}\\ 
    & + (1 - r_{i}^2) \ g_{\floor{p_{i}^1}+1\floor{p_{i}^2}}\\
    & -  r_{i}^2 \ g_{\floor{p_{i}^1}\floor{p_{i}^2}+1}\\
    & +  r_{i}^2 \ g_{\floor{p_{i}^1}+1\floor{p_{i}^2}+1}]
\end{split}
\\
\begin{split}
\label{eq:res6}
\left(\frac{\partial loss}{\partial \vp^2}\right)_{i} = \ w_{i} \ [ & - (1 - r_{i}^1) \ g_{\floor{p_{i}^1}\floor{p_{i}^2}}\\ 
    & - r_{i}^1 \ g_{\floor{p_{i}^1}+1\floor{p_{i}^2}}\\
    & + (1 - r_{i}^1) \ g_{\floor{p_{i}^1}\floor{p_{i}^2}+1}\\
    & + r_{i}^1 \ g_{\floor{p_{i}^1}+1\floor{p_{i}^2}+1} ]
\end{split}
\end{align}

The results found for the general case are nothing but a component-wise application of the result obtained in the scalar case. In addition, we show in Appendix \ref{appendixC}, the extension to the 1D and 3D convolution cases.

\subsection{1D-DCLS, 3D-DCLS}
\label{appendixC}
We denote respectively by $s_1, s_2, s_3 \in \mathbb{N}^{*} \times \mathbb{N}^{* } \times \mathbb{N}^{*}$, the sizes of the constructed kernel along the x-axis, y-axis and the z-axis. Moreover, the $n \times p \times q$ tensor space of third dimension is denoted $\mathbb{R}^{ n \times p \times q }$.

The function $f$  defined in (\ref{eq:f}) could be adapted in order to construct a suitable kernel for the 1D convolution in the scalar weight case as follows:

\begin{align}
  \begin{split}
  f_{1D} \colon \mathbb{R} \times \mathbb{R}  &\to \mathbb{R}^s\\
  w, p  & \mapsto \displaystyle \vk
  \end{split}
\end{align}
where $\forall i\in \llbracket 1 \ .. \ s \rrbracket \colon $\\
\begin{equation}
\arraycolsep=1.3pt\def\arraystretch{1}
\displaystyle \vk_{i} =\left\{\begin{array}{cl}
w \ (1 - r) & \text {if } i = \floor{p} \\
w \ r & \text {if }  i = \floor{p} + 1 \\

0 & \text {else } 
\end{array}\right.
\end{equation}
and where the fractional part is:
\begin{equation}
    \begin{split}
    r = \{p\} = p - \floor{p}
    \end{split}
\end{equation}

Following the same construction in Appendix \ref{sec:gen_case} we can show that the gradients of the loss function with respect to weights and positions in the general 1D case are:

$ \forall i \in \llbracket 1 \ .. \ m \rrbracket  \colon$  
\begin{align}
\begin{split}
\left(\frac{\partial loss}{\partial \vw}\right)_{i} = \ & (1 - r_i) \ g_{\floor{p_{i}}} + r_{i} \  g_{\floor{p_{i}}+1} 
\end{split}
\\
\begin{split}
\left(\frac{\partial loss}{\partial \vp}\right)_{i} = \ & w_{i} \ (  g_{\floor{p_{i}}+1} - \ g_{\floor{p_{i}}})
\end{split}
\end{align}

Furthermore, we define the function $f_{3D}$, the suitable kernel construction function in the 3D convolution case, as such: 

\begin{align}
  \begin{split}
  f \colon \mathbb{R} \times \mathbb{R} \times \mathbb{R} \times \mathbb{R} &\to \mathbb{R}^{s_1 \times s_2 \times s_3 } \\
  w, p^1, p^2, p^3   & \mapsto  \tK
  \end{split}
\end{align}
where $\forall i\in \llbracket 1 \ .. \ s_1 \rrbracket$, $\forall j\in \llbracket 1 \ .. \ s_2 \rrbracket, \forall l\in \llbracket 1 \ .. \ s_3 \rrbracket  \colon $\\
\begin{equation}
 \tK_{ijl} =\left\{\begin{array}{cl}
w \ (1 - r^1)\ (1 - r^2)\ (1 - r^3) & \text {if } i = \floor{p^1}, \ j = \floor{p^2}, \ l = \floor{p^3} \\
w \ r^1 \ (1 - r^2)\ (1 - r^3) & \text {if }  i = \floor{p^1} + 1, \ j = \floor{p^2}, \ l = \floor{p^3} \\
w \ (1 - r^1) \ r^2 \ (1 - r^3) & \text {if }  i = \floor{p^1}, \ j = \floor{p^2} + 1, \ l = \floor{p^3}\\
w \ r^1 \ r^2  \ (1 - r^3) & \text {if }  i = \floor{p^1} {+} 1, \ j = \floor{p^2} {+} 1, \ l = \floor{p^3} \\
w \ (1 - r^1)\ (1 - r^2) \ r^3 & \text {if } i = \floor{p^1}, \ j = \floor{p^2}, \ l = \floor{p^3} + 1 \\
w \ r^1 \ (1 - r^2) \ r^3 & \text {if }  i = \floor{p^1} + 1, \ j = \floor{p^2}, \ l = \floor{p^3} + 1 \\
w \ (1 - r^1) \ r^2 \ r^3 & \text {if }  i = \floor{p^1}, \ j = \floor{p^2} + 1, \ l = \floor{p^3} + 1 \\
w \ r^1 \ r^2 \ r^3 & \text {if }  i = \floor{p^1} {+} 1, \ j = \floor{p^2} {+} 1, \ l = \floor{p^3} + 1 \\
0 & \text {else } 
\end{array}\right.
\end{equation}
and where the fractional parts are:
\begin{equation}
    \begin{split}
    r^1 = \{p^1\} = p^1 - \floor{p^1}\\
    r^2 = \{p^2\} = p^2 - \floor{p^2}\\
    r^3 = \{p^3\} = p^3 - \floor{p^3}    
    \end{split}
\end{equation}

We can show that the gradients of the loss function with respect to weights and positions in the general 3D case are:

$ \forall i \in \llbracket 1 \ .. \ m \rrbracket  \colon$  
\begin{align}
\begin{split}
\left(\frac{\partial loss}{\partial \vw}\right)_{i} = \ & (1 - r_{i}^1)\ (1 - r_{i}^2)\ (1 - r_{i}^3)  \ g_{\floor{p_{i}^1}\floor{p_{i}^2}\floor{p_{i}^3}} \\ 
    & + r_{i}^1\ (1 - r_{i}^2) \ (1 - r_{i}^3) \  g_{\floor{p_{i}^1}+1\floor{p_{i}^2}\floor{p_{i}^3}} \\
    & + (1 - r_{i}^1)\ r_{i}^2 \ (1 - r_{i}^3)  \  g_{\floor{p_{i}^1}\floor{p_{i}^2}+1\floor{p_{i}^3}} \\
    & + r_{i}^1 \ r_{i}^2 \ (1 - r_{i}^3)  \ g_{\floor{p_{i}^1}+1\floor{p_{i}^2}+1\floor{p_{i}^3}} \\
    & + (1 - r_{i}^1)\ (1 - r_{i}^2)\ r_{i}^3  \ g_{\floor{p_{i}^1}\floor{p_{i}^2}\floor{p_{i}^3} + 1} \\ 
    & + r_{i}^1\ (1 - r_{i}^2) \ r_{i}^3 \  g_{\floor{p_{i}^1}+1\floor{p_{i}^2}\floor{p_{i}^3} + 1} \\
    & + (1 - r_{i}^1)\ r_{i}^2 \ r_{i}^3  \  g_{\floor{p_{i}^1}\floor{p_{i}^2}+1\floor{p_{i}^3} + 1} \\
    & + r_{i}^1 \ r_{i}^2 \ r_{i}^3  \ g_{\floor{p_{i}^1}+1\floor{p_{i}^2}+1\floor{p_{i}^3} + 1}    
\end{split}
\\
\begin{split}
\left(\frac{\partial loss}{\partial \vp^1}\right)_{i} = \ w_{i} \ [ \ & - \ (1 - r_{i}^2)\ (1 - r_{i}^3)  \ g_{\floor{p_{i}^1}\floor{p_{i}^2}\floor{p_{i}^3}} \\ 
    & + \ (1 - r_{i}^2) \ (1 - r_{i}^3) \  g_{\floor{p_{i}^1}+1\floor{p_{i}^2}\floor{p_{i}^3}} \\
    & - \ r_{i}^2 \ (1 - r_{i}^3)  \  g_{\floor{p_{i}^1}\floor{p_{i}^2}+1\floor{p_{i}^3}} \\
    & + \ r_{i}^2 \ (1 - r_{i}^3)  \ g_{\floor{p_{i}^1}+1\floor{p_{i}^2}+1\floor{p_{i}^3}} \\
    & - \ (1 - r_{i}^2)\ r_{i}^3  \ g_{\floor{p_{i}^1}\floor{p_{i}^2}\floor{p_{i}^3} + 1} \\ 
    & + \ (1 - r_{i}^2) \ r_{i}^3 \  g_{\floor{p_{i}^1}+1\floor{p_{i}^2}\floor{p_{i}^3} + 1} \\
    & - \ r_{i}^2 \ r_{i}^3  \  g_{\floor{p_{i}^1}\floor{p_{i}^2}+1\floor{p_{i}^3} + 1} \\
    & + \ r_{i}^2 \ r_{i}^3  \ g_{\floor{p_{i}^1}+1\floor{p_{i}^2}+1\floor{p_{i}^3} + 1}    ]
\end{split}
\end{align}
\begin{align}
\begin{split}
\left(\frac{\partial loss}{\partial \vp^2}\right)_{i} = \ w_{i} \ [  & - \ (1 - r_{i}^1) \ (1 - r_{i}^3)  \ g_{\floor{p_{i}^1}\floor{p_{i}^2}\floor{p_{i}^3}} \\ 
    & - r_{i}^1  \ (1 - r_{i}^3) \  g_{\floor{p_{i}^1}+1\floor{p_{i}^2}\floor{p_{i}^3}} \\
    & + (1 - r_{i}^1) \ (1 - r_{i}^3)  \  g_{\floor{p_{i}^1}\floor{p_{i}^2}+1\floor{p_{i}^3}} \\
    & + r_{i}^1  \ (1 - r_{i}^3)  \ g_{\floor{p_{i}^1}+1\floor{p_{i}^2}+1\floor{p_{i}^3}} \\
    & - (1 - r_{i}^1) \ r_{i}^3  \ g_{\floor{p_{i}^1}\floor{p_{i}^2}\floor{p_{i}^3} + 1} \\ 
    & - r_{i}^1 \ r_{i}^3 \  g_{\floor{p_{i}^1}+1\floor{p_{i}^2}\floor{p_{i}^3} + 1} \\
    & + (1 - r_{i}^1) \ r_{i}^3  \  g_{\floor{p_{i}^1}\floor{p_{i}^2}+1\floor{p_{i}^3} + 1} \\
    & + r_{i}^1 \ r_{i}^3  \ g_{\floor{p_{i}^1}+1\floor{p_{i}^2}+1\floor{p_{i}^3} + 1}    ]
\end{split}
\\
\begin{split}
\left(\frac{\partial loss}{\partial \vp^3}\right)_{i} = \ w_{i} \ [ \ &  - (1 - r_{i}^1)\ (1 - r_{i}^2)\   \ g_{\floor{p_{i}^1}\floor{p_{i}^2}\floor{p_{i}^3}} \\ 
    & - r_{i}^1\ (1 - r_{i}^2) \  \  g_{\floor{p_{i}^1}+1\floor{p_{i}^2}\floor{p_{i}^3}} \\
    & - (1 - r_{i}^1)\ r_{i}^2 \   \  g_{\floor{p_{i}^1}\floor{p_{i}^2}+1\floor{p_{i}^3}} \\
    & - r_{i}^1 \ r_{i}^2 \   \ g_{\floor{p_{i}^1}+1\floor{p_{i}^2}+1\floor{p_{i}^3}} \\
    & + (1 - r_{i}^1)\ (1 - r_{i}^2) \ g_{\floor{p_{i}^1}\floor{p_{i}^2}\floor{p_{i}^3} + 1} \\ 
    & + r_{i}^1\ (1 - r_{i}^2) \  g_{\floor{p_{i}^1}+1\floor{p_{i}^2}\floor{p_{i}^3} + 1} \\
    & + (1 - r_{i}^1)\ r_{i}^2  \  g_{\floor{p_{i}^1}\floor{p_{i}^2}+1\floor{p_{i}^3} + 1} \\
    & + r_{i}^1 \ r_{i}^2 \ g_{\floor{p_{i}^1}+1\floor{p_{i}^2}+1\floor{p_{i}^3} + 1}    ]
\end{split}
\end{align}
\newpage
\section{Appendix: The 2D-DCLS kernel construction algorithm}
\label{appendix:algo}
In the following, we describe with pseudocode the forward and backward passes for kernel construction used in 2D-DCLS. In practice, $\tW$, $\tP^1$ and $\tP^2$ are 3-D tensors of size (\texttt{channels\_out, channels\_in // groups, K\_count}), but the algorithms presented here are easily extended to this case by applying them channel-wise.
\begin{algorithm}[ht]
    \setstretch{1.2}
    \begin{algorithmic}[1]

        \REQUIRE $\tW$, $\tP^1$, $\tP^2$ :  vectors of dimension $m$
        \ENSURE $\tK$ : the constructed kernel, of size ($s_1 \times s_2$)
        \STATE $\tK \leftarrow 0$\label{alg:forward}
        \STATE $\vp^1 \leftarrow \floor{\tP^1} $; $\quad \vp^2 \leftarrow \floor{\tP^2} $ 
        \STATE $\tR^1 \leftarrow \tP^1 - \vp^1$;   $\quad \tR^2 \leftarrow \tP^2 - \vp^2$ 
        \STATE \text{save\_for\_backward ($\vp^1, \vp^2, \tR^1, \tR^2$)}
        
        \FOR{$i=0 \rightarrow m-1 $}
            \STATE $ \tK[\vp^1_{i},\vp^2_{i}] \mathrel{+}= \tW_{i} * (1-\tR^1_{i})* (1-\tR^2_{i})$
            \STATE $ \tK[\vp^1_{i}+1,\vp^2_{i}] \mathrel{+}= \tW_{i} * (\tR^1_{i})* (1-\tR^2_{i})$
            \STATE $ \tK[\vp^1_{i},\vp^2_{i}+1] \mathrel{+}= \tW_{i} * (1-\tR^1_{i})* (\tR^2_{i})$
            \STATE $ \tK[\vp^1_{i}+1,\vp^2_{i}+1] \mathrel{+}= \tW_{i} * (\tR^1_{i})* (\tR^2_{i})$
        \ENDFOR 
    \end{algorithmic}        
    \caption{2D-DCLS kernel construction forward pass}
\end{algorithm}
\begin{algorithm}[ht]
    \begin{algorithmic}[1]

        \REQUIRE $GradK = \frac{\partial Loss}{\partial K}$ : matrix of dimension ($s_1 \times s_2$)
        \ENSURE $\frac{\partial Loss}{\partial W}$, $\frac{\partial Loss}{\partial P^1}$, $\frac{\partial Loss}{\partial P^2}$ : vectors of dimension $m$
        \STATE $\frac{\partial Loss}{\partial W} \leftarrow 0$, $\frac{\partial Loss}{\partial P^1} \leftarrow 0$, $\frac{\partial Loss}{\partial P^2} \leftarrow 0$
        \STATE $\vp^1, \vp^2, \tR^1, \tR^2 \leftarrow \text{load\_saved ( \ )}$\label{alg:backward}
        \FOR{$i=0 \rightarrow m-1 $}      
    		\STATE \small \begin{align*}
    		\frac{\partial Loss}{\partial W}[i] \mathrel{+}= \frac{\partial Loss}{\partial K}[\vp^1_{i},\vp^2_{i}] * (1{-}\tR^1_{i})* (1{-}\tR^2_{i})  + \frac{\partial Loss}{\partial K}[\vp^1_{i}+1,\vp^2_{i}]* \tR^1_{i}* (1{-}\tR^2_{i}) \\
    		+ \frac{\partial Loss}{\partial K}[\vp^1_{i},\vp^2_{i}+1]  * (1{-}\tR^1_{i})* \tR^2_{i} + \frac{\partial Loss}{\partial K}[\vp^1_{i}+1,\vp^2_{i}+1] * \tR^1_{i}* \tR^2_{i}
    		\end{align*} 	
    				
            \STATE \small \begin{align*}
    		\frac{\partial Loss}{\partial P^1}[i] \mathrel{+}=  \tW_{i}* [ -\frac{\partial Loss}{\partial K}[\vp^1_{i},\vp^2_{i}] * (1-\tR^2_{i})  
    		+ \frac{\partial Loss}{\partial K}[\vp^1_{i}+1,\vp^2_{i}] * (1-\tR^2_{i}) \\ 
    		- \frac{\partial Loss}{\partial K}[\vp^1_{i},\vp^2_{i}+1] * \tR^2_{i}  
    		+ \frac{\partial Loss}{\partial K}[\vp^1_{i}+1,\vp^2_{i}+1] * \tR^2_{i} ] 
    		\end{align*} 	
                           
            \STATE \small \begin{align*}
			\frac{\partial Loss}{\partial P^2}[i] \mathrel{+}= \tW_{i}* [-\frac{\partial Loss}{\partial K}[\vp^1_{i},\vp^2_{i}] * (1-\tR^1_{i}) 
			- \frac{\partial Loss}{\partial K}[\vp^1_{i}+1,\vp^2_{i}] * \tR^1_{i}\\
			+ \frac{\partial Loss}{\partial K}[\vp^1_{i},\vp^2_{i}+1] * (1-\tR^1_{i})
			+ \frac{\partial Loss}{\partial K}[\vp^1_{i}+1,\vp^2_{i}+1] * \tR^1_{i}]
			\end{align*} 	
        \ENDFOR        
    \end{algorithmic}
    \caption{2D-DCLS kernel construction backward pass}
\end{algorithm} 

The `for' loops in the algorithms are fully parallelized using GPU threads. The 2D-DCLS convolution with kernel construction is then obtained by applying the classical 2D-convolution provided natively by PyTorch or any other method such as the \textit{depthwise implicit gemm} convolution method \cite{ding2022scaling} using the constructed kernel.

When considering a concurrent execution of this pseudocode, the additions may result, in case of overlapping, in a replacement of the overlapped values instead of the desired accumulation. This problem can be addressed by using atomic addition operations.

\newpage

\section{Appendix: The DCLS kernel construction algorithm in native PyTorch}
In the following, we describe with PyTorch code the DCLS construction module for 1D, 2D and 3D versions. The DCLS convolution method is obtained by using the constructed kernel as a weight for the native torch.nn.Conv\{1,2,3\}d, or another convolution method such as \textit{depthwise implicit gemm} \cite{MegEngine}.
\label{appendix:algo_torch}
\begin{lstlisting}[language=Python]
import torch
from torch.nn import Module
from torch.nn.parameter import Parameter


class ConstructKernel1d(Module):
    def __init__(
        self,
        out_channels,
        in_channels,
        groups,
        kernel_count,
        dilated_kernel_size,
    ):
        super().__init__()
        self.out_channels = out_channels
        self.in_channels = in_channels
        self.groups = groups
        self.dilated_kernel_size = dilated_kernel_size
        self.kernel_count = kernel_count
        I = torch.arange(0, dilated_kernel_size[0])
        I = I.expand(
            out_channels, in_channels // groups, kernel_count, -1
        ).permute(3, 0, 1, 2)
        self.I = Parameter(I, requires_grad=False)

        self.lim = torch.zeros(1)
        self.lim[0] = dilated_kernel_size[0]
        self.lim = self.lim.expand(
            out_channels, in_channels // groups, kernel_count, -1
        ).permute(3, 0, 1, 2)
        self.lim = Parameter(self.lim, requires_grad=False)

    def forward(self, W, P):
        P = P + self.lim // 2
        Pr = P
        P = P.floor()
        R = (Pr - P).expand(self.dilated_kernel_size[0], -1, -1, -1, -1)
        R1 = R.select(1, 0)
        P1 = P.select(0, 0)
        cond1 = self.I == P1
        cond2 = self.I == P1 + 1
        W1 = torch.where(cond1, 1.0, 0.0)
        W2 = torch.where(cond2, 1.0, 0.0)

        K = W1 + R1 * (W2 - W1)
        K = W * K
        K = K.sum(3)
        K = K.permute(1, 2, 0)
        return K


class ConstructKernel2d(Module):
    def __init__(
        self,
        out_channels,
        in_channels,
        groups,
        kernel_count,
        dilated_kernel_size,
    ):
        super().__init__()
        self.out_channels = out_channels
        self.in_channels = in_channels
        self.groups = groups
        self.dilated_kernel_size = dilated_kernel_size
        self.kernel_count = kernel_count
        J = torch.arange(0, dilated_kernel_size[0]).expand(
            dilated_kernel_size[1], -1
        )
        I = torch.arange(0, dilated_kernel_size[1]).expand(
            dilated_kernel_size[0], -1
        )
        I = I.expand(
            out_channels, in_channels // groups, kernel_count, -1, -1
        ).permute(3, 4, 0, 1, 2)
        J = J.expand(
            out_channels, in_channels // groups, kernel_count, -1, -1
        ).permute(4, 3, 0, 1, 2)

        self.I = Parameter(I, requires_grad=False)
        self.J = Parameter(J, requires_grad=False)
        self.lim = torch.zeros(2)
        self.lim[0] = dilated_kernel_size[0]
        self.lim[1] = dilated_kernel_size[1]
        self.lim = self.lim.expand(
            out_channels, in_channels // groups, kernel_count, -1
        ).permute(3, 0, 1, 2)
        self.lim = Parameter(self.lim, requires_grad=False)

    def forward(self, W, P):
        P = P + self.lim // 2
        Pr = P
        P = P.floor()
        R = (Pr - P).expand(
            self.dilated_kernel_size[0],
            self.dilated_kernel_size[1],
            -1,
            -1,
            -1,
            -1,
        )
        R1 = R.select(2, 0)
        P1 = P.select(0, 0)
        R2 = R.select(2, 1)
        P2 = P.select(0, 1)
        R1R2 = R1 * R2
        cond1 = self.I == P1
        cond2 = self.J == P2
        cond3 = self.I == P1 + 1
        cond4 = self.J == P2 + 1
        W1 = torch.where(cond1 * cond2, 1.0, 0.0)
        W2 = torch.where(cond1 * cond4, 1.0, 0.0)
        W3 = torch.where(cond3 * cond2, 1.0, 0.0)
        W4 = torch.where(cond3 * cond4, 1.0, 0.0)
        K = W1 + R1R2 * (W1 - W2 - W3 + W4) + R1 * (W3 - W1) + R2 * (W2 - W1)
        K = W * K
        K = K.sum(4)
        K = K.permute(2, 3, 0, 1)
        return K


class ConstructKernel3d(Module):
    def __init__(
        self,
        out_channels,
        in_channels,
        groups,
        kernel_count,
        dilated_kernel_size,
    ):
        super().__init__()
        self.out_channels = out_channels
        self.in_channels = in_channels
        self.groups = groups
        self.dilated_kernel_size = dilated_kernel_size
        self.kernel_count = kernel_count
        L = torch.arange(0, dilated_kernel_size[0]).expand(
            dilated_kernel_size[1], dilated_kernel_size[2], -1
        )
        J = torch.arange(0, dilated_kernel_size[1]).expand(
            dilated_kernel_size[0], dilated_kernel_size[2], -1
        )
        I = torch.arange(0, dilated_kernel_size[2]).expand(
            dilated_kernel_size[0], dilated_kernel_size[1], -1
        )
        L = L.expand(
            out_channels, in_channels // groups, kernel_count, -1, -1, -1
        ).permute(5, 3, 4, 0, 1, 2)
        I = I.expand(
            out_channels, in_channels // groups, kernel_count, -1, -1, -1
        ).permute(3, 4, 5, 0, 1, 2)
        J = J.expand(
            out_channels, in_channels // groups, kernel_count, -1, -1, -1
        ).permute(3, 5, 4, 0, 1, 2)
        self.L = Parameter(L, requires_grad=False)
        self.I = Parameter(I, requires_grad=False)
        self.J = Parameter(J, requires_grad=False)
        self.lim = torch.zeros(3)
        self.lim[0] = dilated_kernel_size[0]
        self.lim[1] = dilated_kernel_size[1]
        self.lim[2] = dilated_kernel_size[2]
        self.lim = self.lim.expand(
            out_channels, in_channels // groups, kernel_count, -1
        ).permute(3, 0, 1, 2)
        self.lim = Parameter(self.lim, requires_grad=False)

    def forward(self, W, P):
        P = P + self.lim // 2
        Pr = P
        P = P.floor()
        R = (Pr - P).expand(
            self.dilated_kernel_size[0],
            self.dilated_kernel_size[1],
            self.dilated_kernel_size[2],
            -1,
            -1,
            -1,
            -1,
        )
        R1 = R.select(3, 0)
        P1 = P.select(0, 0)
        R2 = R.select(3, 1)
        P2 = P.select(0, 1)
        R3 = R.select(3, 2)
        P3 = P.select(0, 2)

        cond1 = self.L == P1
        cond2 = self.I == P2
        cond3 = self.J == P3
        cond4 = self.L == P1 + 1
        cond5 = self.I == P2 + 1
        cond6 = self.J == P3 + 1
        W1 = torch.where(cond1 * cond2 * cond3, 1.0, 0.0)
        W2 = torch.where(cond4 * cond2 * cond3, 1.0, 0.0)
        W3 = torch.where(cond1 * cond5 * cond3, 1.0, 0.0)
        W4 = torch.where(cond4 * cond5 * cond3, 1.0, 0.0)
        W5 = torch.where(cond1 * cond2 * cond6, 1.0, 0.0)
        W6 = torch.where(cond4 * cond2 * cond6, 1.0, 0.0)
        W7 = torch.where(cond1 * cond5 * cond6, 1.0, 0.0)
        W8 = torch.where(cond4 * cond5 * cond6, 1.0, 0.0)
        # needs a better computing
        K = W1 * (1 - R1) * (1 - R2) * (1 - R3)
        K += W2 * R1 * (1 - R2) * (1 - R3)
        K += W3 * (1 - R1) * R2 * (1 - R3)
        K += W4 * R1 * R2 * (1 - R3)
        K += W5 * (1 - R1) * (1 - R2) * R3
        K += W6 * R1 * (1 - R2) * R3
        K += W7 * (1 - R1) * R2 * R3
        K += W8 * R1 * R2 * R3
        K = W * K
        K = K.sum(5)
        K = K.permute(3, 4, 0, 1, 2)
        return K
\end{lstlisting}
\newpage
\section{Appendix: Histograms of positions}
\label{appendix:hist}
In the following, we show as histograms over training epochs, the distribution of the four kernel positions for the 2D-DCLS convolutions of the ConvNeXt-T-dcls model. Note that there is no agglutination or edge effect around the kernel limits, and that the distributions are relatively stable, with a higher concentration around the center of the kernel. Individual positions, however, are constantly moving; see animation at:

\url{https://github.com/K-H-Ismail/Dilated-Convolution-with-Learnable-Spacings-PyTorch/blob/main/figs/animation.gif}.

\begin{figure}[!htb]
\caption{The distribution over epochs of kernel positions for the four stages of the ConvNeXt-T-dcls model.}
\begin{minipage}{0.45\textwidth}
  \includegraphics[width=\linewidth]{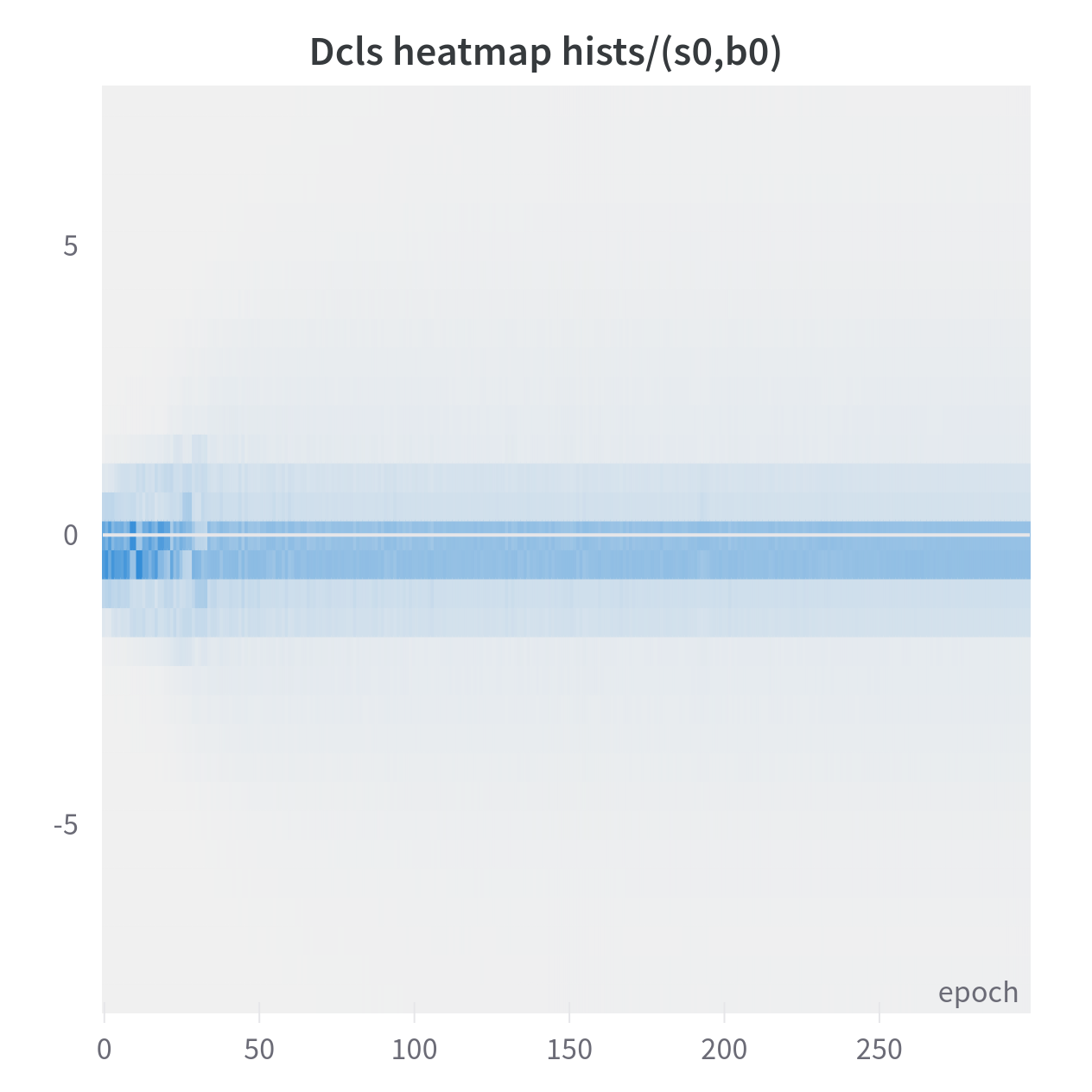}
  \subcaption{Kernel positions distribution - stage 0.}\label{fig:hist_s0b0}
\end{minipage}\hfill
\begin{minipage}{0.45\textwidth}
  \includegraphics[width=\linewidth]{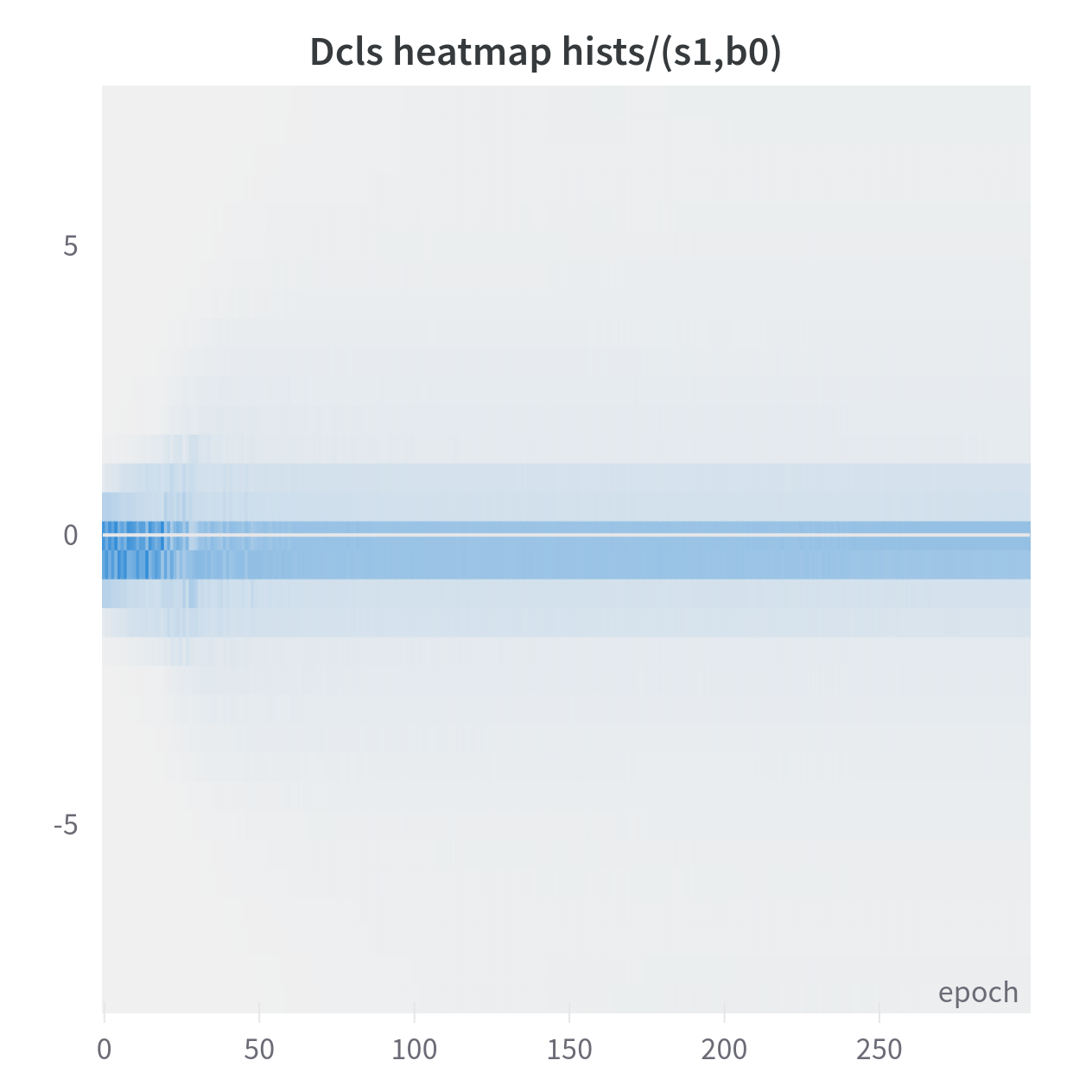}
  \subcaption{Kernel positions distribution - stage 1.}\label{fig:hist_s1b0}
\end{minipage}\hfill
\begin{minipage}{0.45\textwidth}%
  \includegraphics[width=\linewidth]{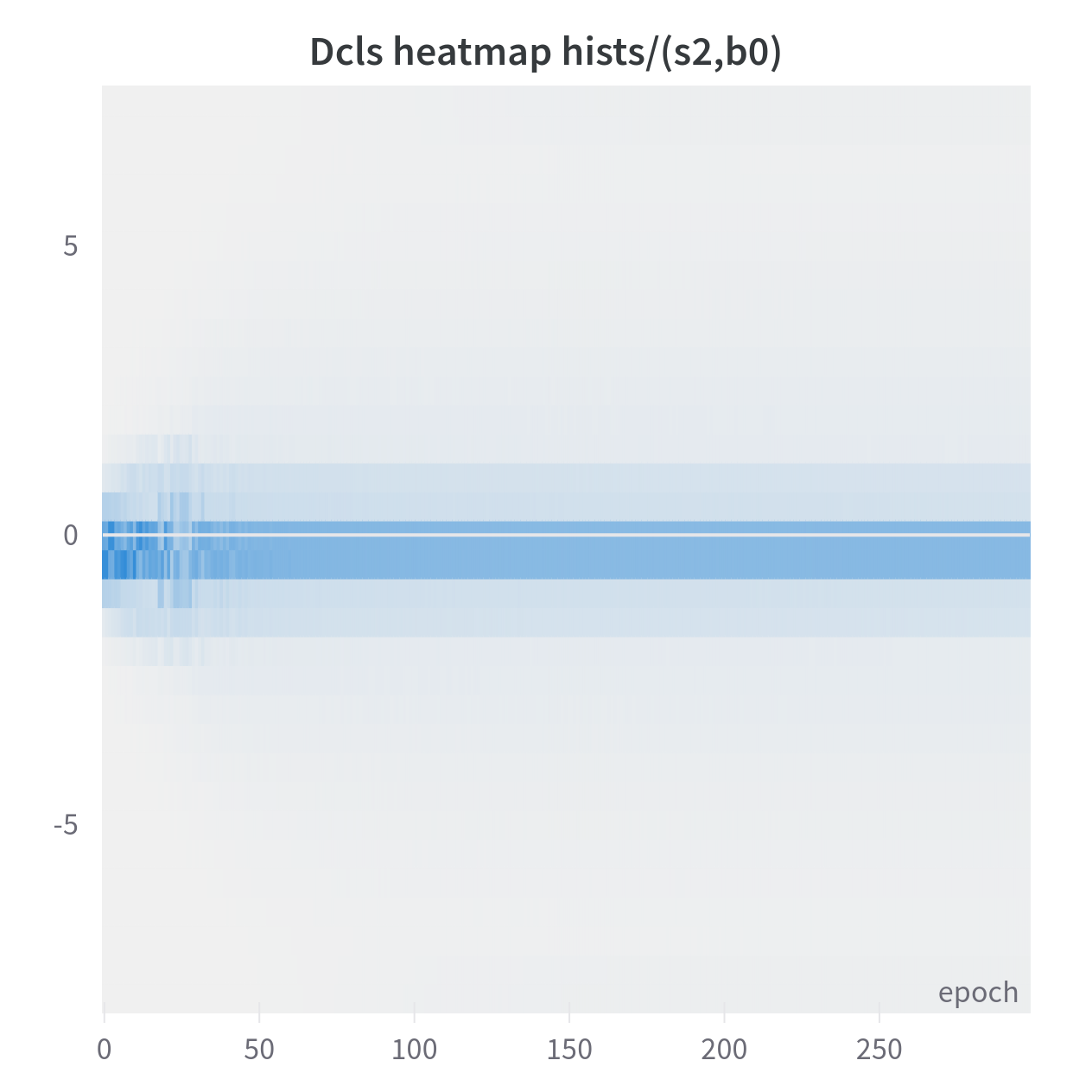}
  \subcaption{Kernel positions distribution - stage 2.}\label{fig:hist_s2b0}
\end{minipage}\hfill
\begin{minipage}{0.45\textwidth}%
  \includegraphics[width=\linewidth]{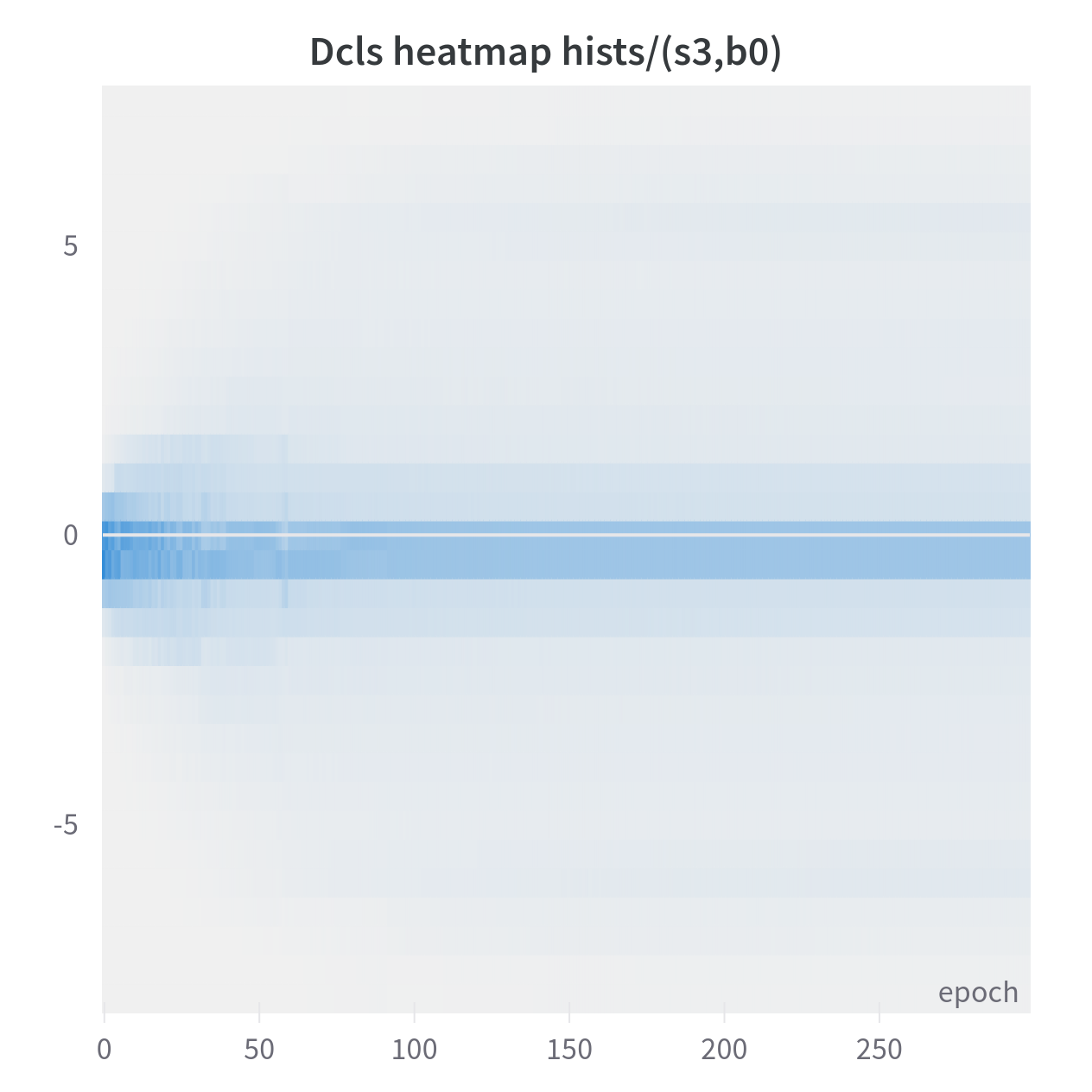}
  \subcaption{Kernel positions distribution - stage 3.}\label{fig:hist_s3b0}
\end{minipage}
\end{figure}

\newpage
\section{Appendix: Speed curves and learning rate schedule}
\label{appendix:speed}
Here, we plot the average speed curves of the four position tensors for the 2D-DCLS convolutions $V_\tP$ of the ConvNeXt-T-dcls model as functions of the training epochs. The general formula for the average speed of a DCLS position tensor of size $(cout, cin, m) \in \mathbb{N^*}^3$,  at epoch $t$, is as follows:
$$\forall t\in \llbracket 1 \ .. \ t_{max} \rrbracket \colon \quad
V_\tP(t) = \frac{1} {cout \cdot cin \cdot m} \sum\limits_{k=1}^{cout}\sum\limits_{j=1}^{cin}\sum\limits_{i=1}^{m} |P^{t}_{ijk} - P^{t-1}_{ijk}|
$$

\begin{figure}[!ht]
  \includegraphics[width=\linewidth, height=0.5\linewidth]{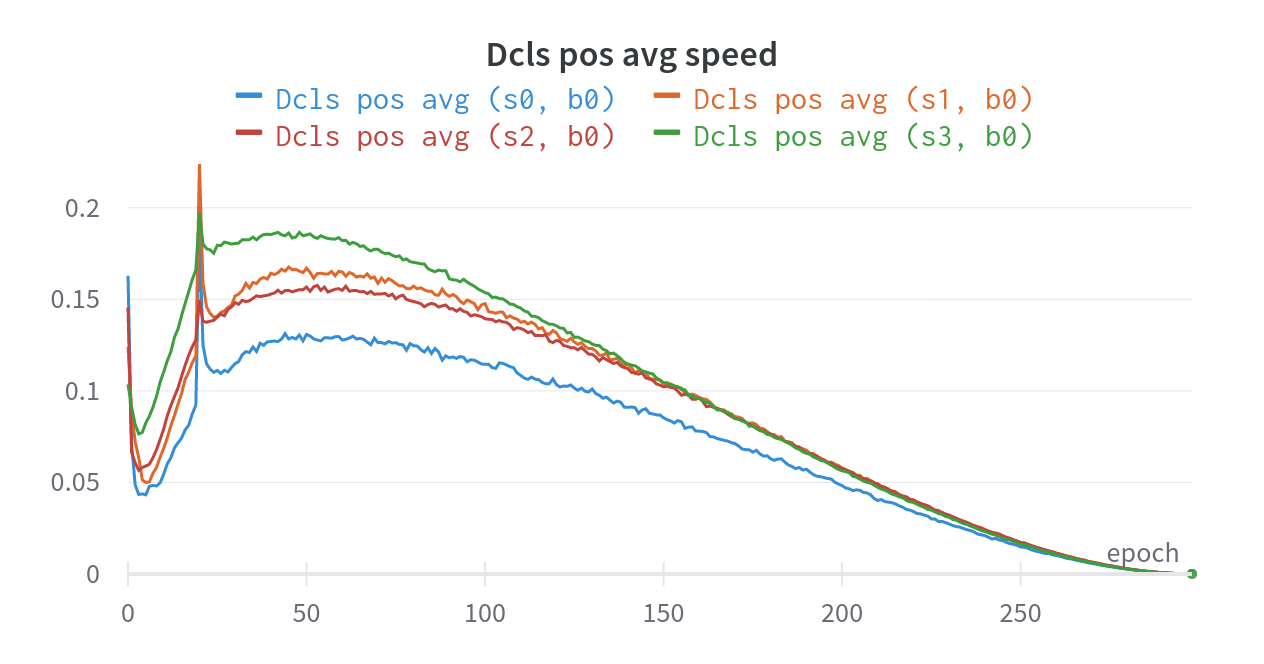}
  \caption{The average speed of the four position tensors for the ConvNeXt-T-dcls model as function of epochs.}\label{fig:speed_s}

\end{figure}
\begin{figure}[!ht]
  \includegraphics[width=\linewidth, height=170pt]{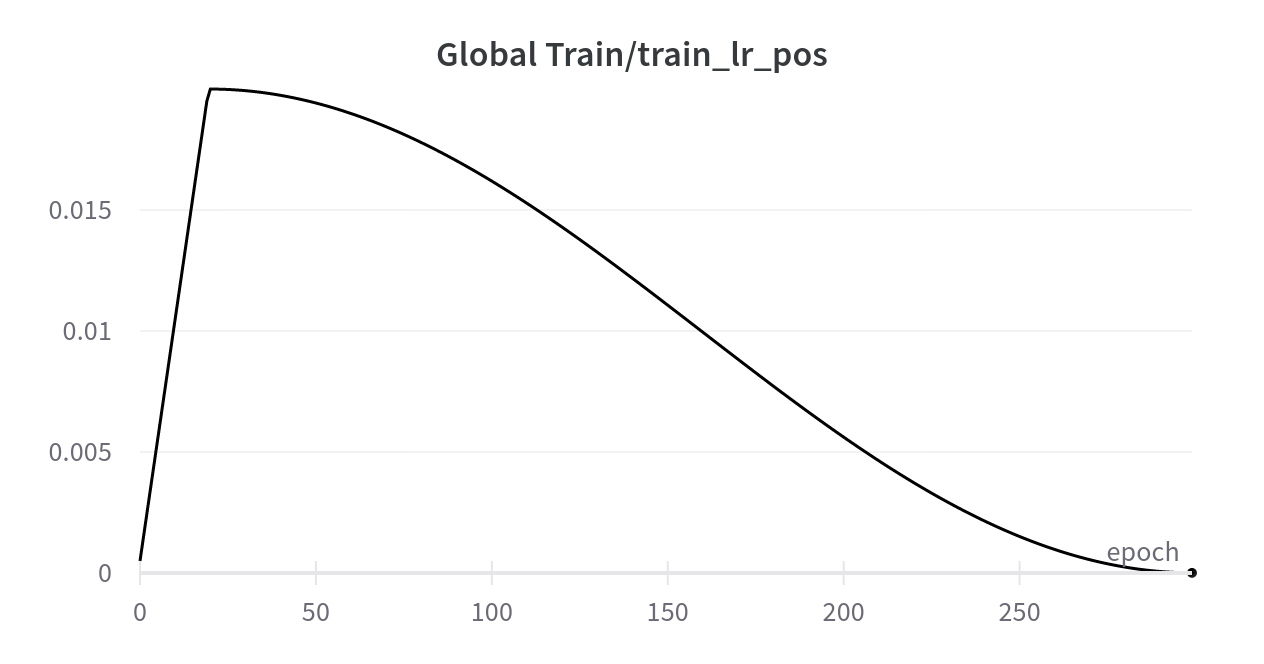}
  \caption{The learning schedule used for training ConvNeXt-T-dcls model.}\label{fig:sched}
\end{figure}

At epoch 0, we can notice that the average speed is abnormally high, this is due to the fact that in the beginning, the positions are initialized randomly and we arbitrarily considered that $V_{\tP}(0)  = 0$, thus the large speed gap at initialization. Another peak can be seen at epoch 20, this one is due to the introduction of the repulsive loss \citep{thomas2019KPConv} at this precise epoch during training. This last causes a momentary increase in the average speed of the DCLS positions. In general, we can say that over epochs, the average DCLS positions follow the shape of the scheduler used in training.

\section{Appendix: Effective receptive fields comparison}
\label{appendixF}
In the following, we show the effective receptive fields (ERF) calculated respectively for ConvNeXt-T-dcls, ConvNeXt-T with a standard dilated kernel of rate 2 and ConvNeXt-T models. The input crops used here are of size $1024 \times 1024$ and the heatmaps are normalized (between 0 and 1). These ERFs are obtained by finding (via backpropagation) the input pixels that most impact the activations of a pretrained model using samples of the dataset. We observe that the ERF of ConvNeXt-T-dcls has a particular shape that resembles a square with more prominent diagonals and medians. The ERF of ConvNeXt-T with a standard dilated kernel is larger but with gridding artifacts. In all plots, it seems that the center has more importance.
\begin{figure}[!htb]
\minipage{0.45\textwidth}
  \includegraphics[width=\linewidth]{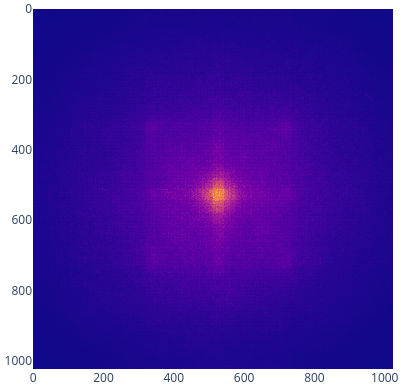}
  \subcaption{The effective receptive field (ERF) of the ConvNeXt-T-dcls model.}\label{fig:erf_dcls}
\endminipage\hfill
\minipage{0.45\textwidth}
  \includegraphics[width=\linewidth]{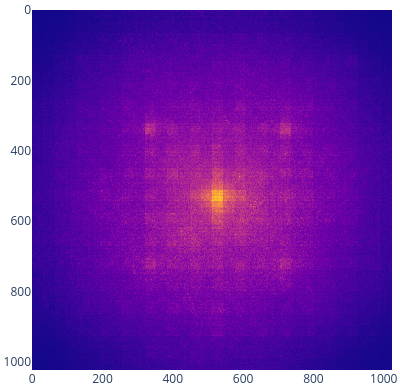}
  \subcaption{The effective receptive field of the ConvNeXt model with dilated kernels (dil rate 2) instead of the dense $7 \times 7$ ones.}\label{fig:erf_dil}
\endminipage\hfill
\minipage{0.45\textwidth}%
  \includegraphics[width=\linewidth]{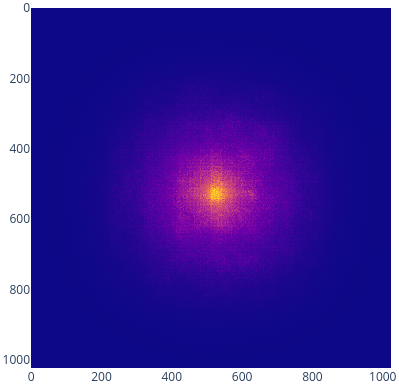}
  \subcaption{The effective receptive field of the ConvNeXt-T model.}\label{fig:erf_baseline}
\endminipage
\end{figure}

\newpage

\section{Appendix: Time measurements}
\label{appendixG}
We notice that when using the \textit{depthwise implicit gemm} algorithm, DCLS forward is slightly slower than the PyTorch native Conv2d forward pass, but the DCLS backward pass is faster, in particular when increasing the model size.
\newpage
\begin{figure}[!htb]
  \centering

  \includegraphics[width=0.7\linewidth]{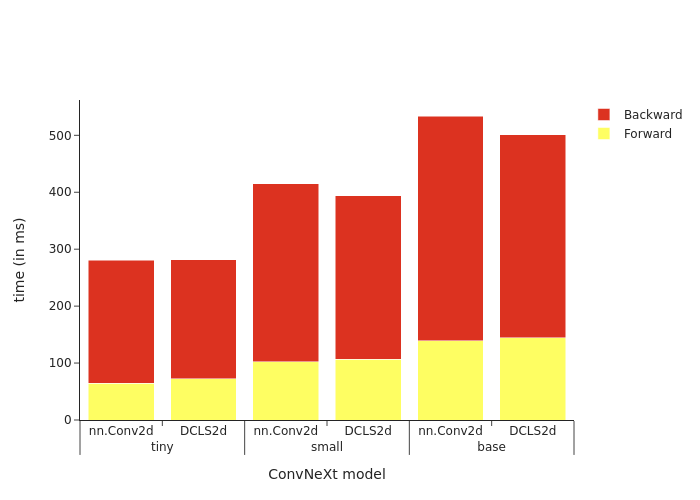}
  \caption{For the 3 ConvNeXt variants (tiny, small and base), we measure the elapsed time in ms for 1 forward + backward pass with a fixed batch size 128 and inputs of size (3,224,224) using DCLS2d convolution \textbf{accelerated by the \textit{depthwise implicit} gemm algorithm}. Measures were carried using a single A100-80gb gpu. We also compare those timings to the 3 ConvNeXt baselines.}

\end{figure}

When not using the \textit{depthwise implicit gemm} algorithm but the PyTorch native Conv2d in DCLS, the forward and backward are about twice slower.

\begin{figure}[!htb]
  \centering
  \includegraphics[width=0.6\linewidth]{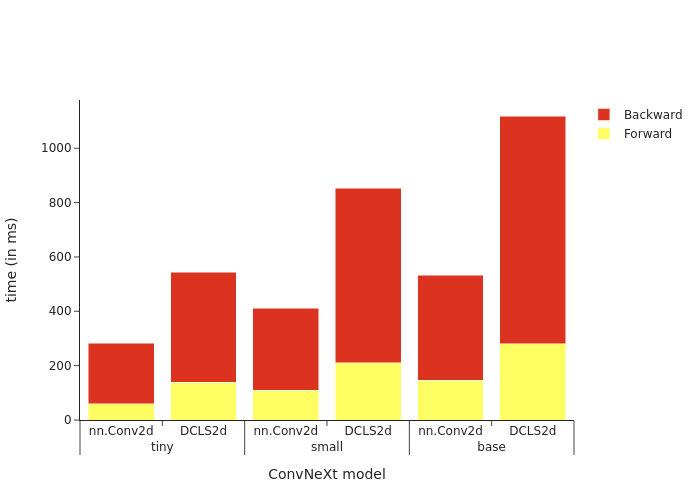}
  \caption{For the 3 ConvNeXt variants (tiny, small and base), we measure the elapsed time in ms for 1 forward + backward pass with a fixed batch size 128 and inputs of size (3,224,224) using DCLS2d convolution \textbf{with the PyTorch native 2D convolution algorithm}. Measures were carried using a single A100-80gb gpu. We compare those timings to the 3 ConvNeXt baselines. We also compare those timings to the 3 ConvNeXt baselines.}

\end{figure}

We notice that when using the \textit{depthwise implicit gemm} algorithm, the DCLS construction algorithm is negligible in time compared to the convolution method for large input map sizes. For small map sizes (example 7x7), the DCLS construction algorithm takes as much time as the convolution.

\begin{figure}[!htb]
  \centering
  \includegraphics[width=\linewidth]{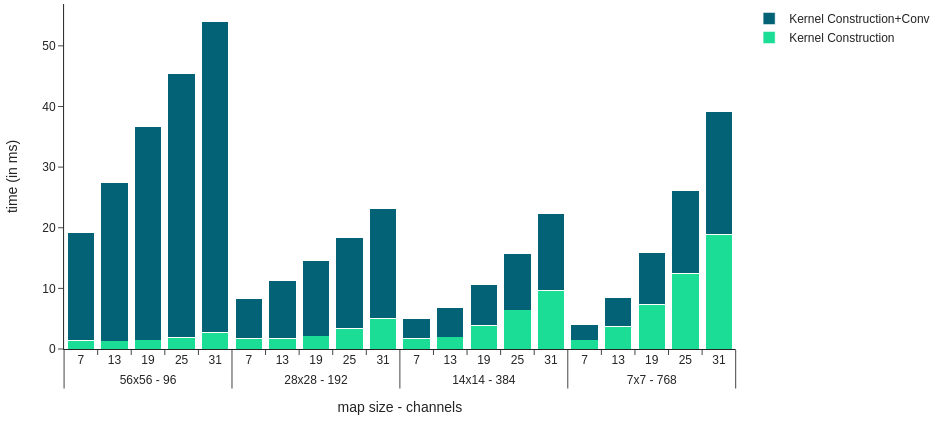}
  \caption{For different dilated kernel sizes (ranging from 7 to 31), and 4 different map sizes, we measure the elapsed time in ms for 1 forward + backward pass with a fixed batch size 128 and a fixed kernel count 34 using a single DCLS2d construct module with a 2D convolution \textbf{accelerated by the \textit{depthwise implicit gemm} algorithm}. Measures were carried using a single Quadro RTX 8000 gpu.}

\end{figure}

When not using the \textit{depthwise implicit gemm} algorithm but the PyTorch native Conv2d in DCLS, the construction algorithm time is significantly lower than the convolution time even for small input map sizes. This is due to the fact the convolution time with the PyTorch native Conv2d is significantly higher than the one with the \textit{depthwise implicit gemm} algorithm.

\begin{figure}[!htb]
  \centering
  \includegraphics[width=\linewidth]{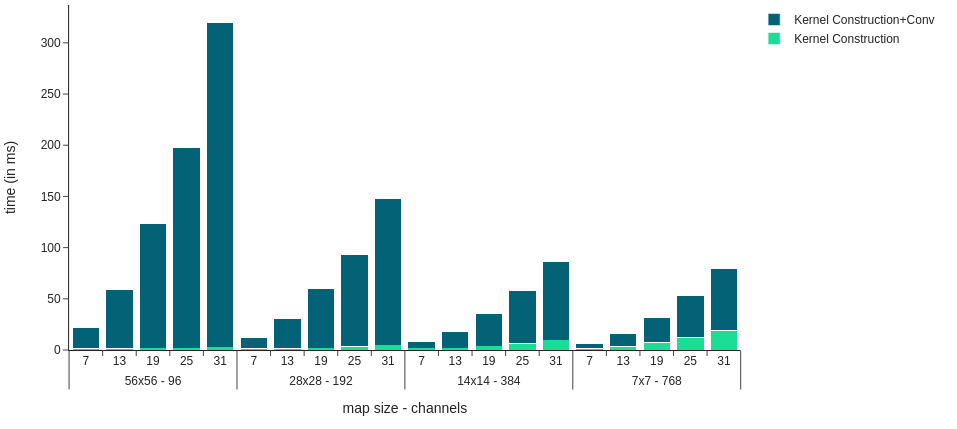}
  \caption{For different dilated kernel sizes (ranging from 7 to 31), and 4 different map sizes/channels, we measure the elapsed time in ms for 1 forward + backward pass with a fixed batch size 128 and a fixed kernel count 34 using a single DCLS2d construct module with \textbf{the PyTorch native 2D convolution algorithm}. Measures were carried using a single Quadro RTX 8000 gpu.}

\end{figure}


\chapter{Dilated convolution with learnable spacings beyond bilinear interpolation}
\label{chap:3}
\section{Disclaimer}
This chapter is strongly inspired by the article: \cite{khalfaouihassani2023dilated}. Most parts of this chapter have been taken verbatim from the paper of which we are the principal author, with the content and wording largely our own. 


\section{Introduction}
 In DCLS, the positions of the non-zero elements within the convolutional kernels are learned in a gradient-based manner. The challenge of non-differentiability caused by the integer nature of the positions is addressed through the application of \textbf{bilinear} interpolation. By doing so, DCLS enables the construction of a differentiable convolutional kernel.

DCLS is a differentiable method that only constructs the convolutional kernel. To implement the whole convolution, one can utilize either the native convolution provided by PyTorch or a more efficient implementation such as the ``depthwise implicit gemm'' convolution method proposed by \citet{ding2022scaling}, which is suitable for large kernels.

\begin{figure*}[!htbp]
     \begin{center}
     \subfloat[][]{\includegraphics[width=0.25\textwidth]{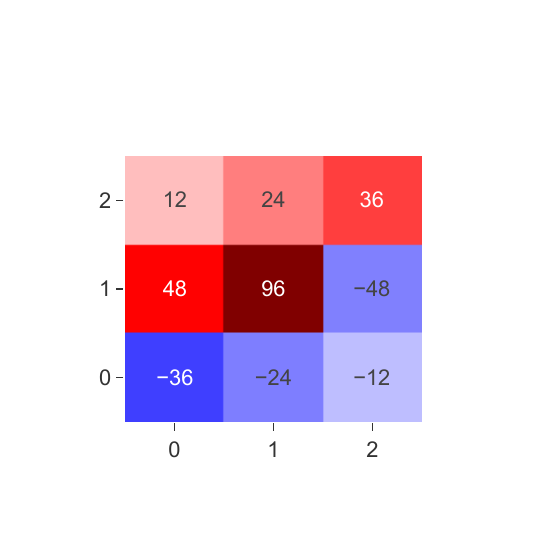}}
     \subfloat[][]{\includegraphics[width=0.5\textwidth]{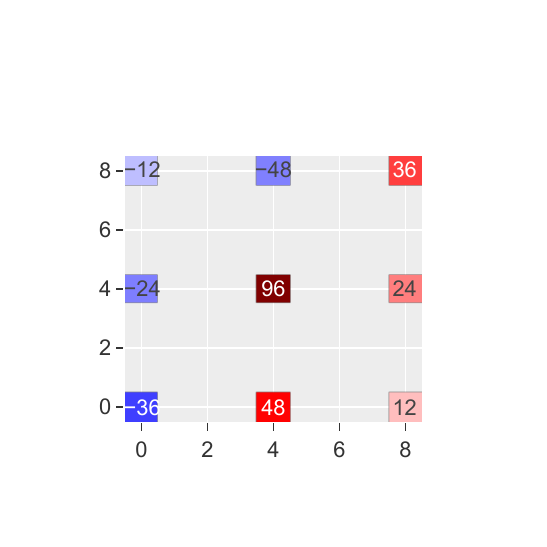}}
     \hfill
     \subfloat[][]{\label{fig:indexc} 
     \includegraphics[width=0.5\textwidth]{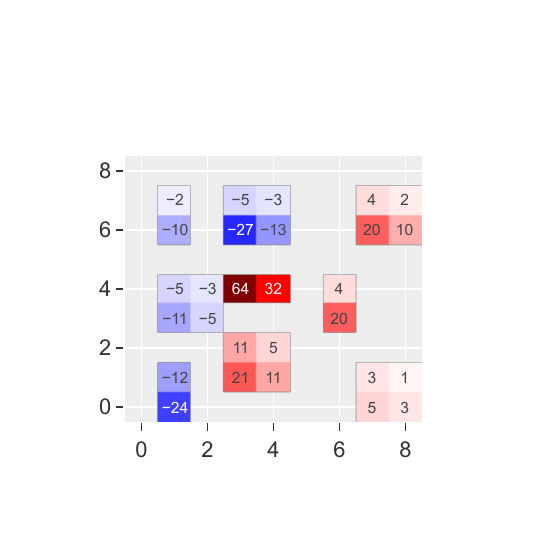}}     
     \subfloat[][]{\includegraphics[width=0.5\textwidth]{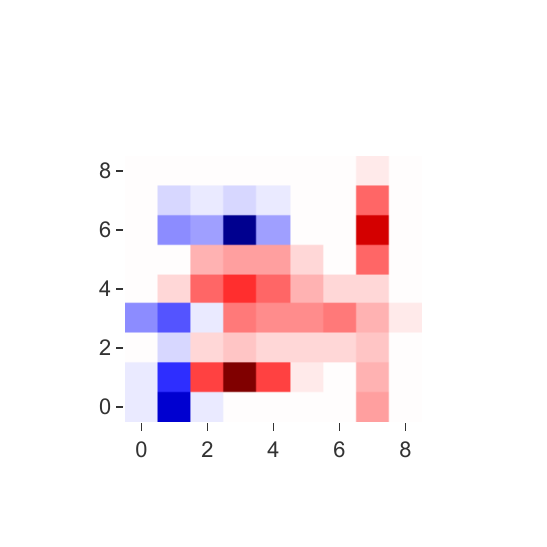}}
  
     \end{center}
    \caption{(a) a standard $3\times 3$ kernel. (b) a standard dilated $3\times 3$ kernel. (c) a 2D-DCLS kernel using bilinear interpolation with 9 kernel elements and a kernel size of 9. (d) the same kernel as (c) with Gaussian interpolation. The numbers have been rounded in all figures and omitted in (d) for readability.}
    \label{fig:index}    
\end{figure*}

The primary motivation behind the development of DCLS was to investigate the potential for enhancing the fixed grid structure imposed by standard dilated convolution in an input-independent way. By allowing an arbitrary number of kernel elements, DCLS introduces a free tunable hyper-parameter called the ``kernel count". Additionally, the ``dilated kernel size'' refers to the maximum extent to which the kernel elements are permitted to move within the dilated kernel (Fig.~\ref{fig:indexc}). Both of these parameters can be adjusted to optimize the performance of DCLS. The positions of the kernel elements in DCLS are initially randomized and subsequently allowed to evolve within the limits of the dilated kernel size during the learning process.
The main focus of this chapter will be to question the choice of \textbf{bilinear} interpolation used by default in DCLS. We tested several interpolations and found in particular that a \textbf{Gaussian} interpolation with learnable standard deviations made the approach more effective.

In fact, the name ``interpolation'' is somewhat too general in this context because we found that the operation that DCLS performs is more of a weighting operation than an interpolation one, even though the term interpolation can be found in the literature. The term weighting has its origins in the so-called multivariate interpolation methods studied in geographic information systems and spatial analysis.

The concept of ``Inverse distance weighting'' was initially introduced by \cite{shepard1968two}, wherein the ``weighting'' designation was adopted instead of the general term ``interpolation'' to specifically denote a technique for generating additional data points within the range of a discrete set of pre-existing data points. The name ``weighting'' is a type of multivariate interpolation applied to a given scattered set of points as in \cite{franke1982scattered} and \cite{lukaszyk2004new}. Nevertheless, we will use the terms weighting and interpolation interchangeably, preferring the term interpolation only to be consistent with recent literature.

To evaluate the effectiveness of DCLS with Gaussian interpolation, we integrate it as a drop-in replacement for the standard depthwise separable convolution in two state-of-the-art convolutional models: the ConvNext-T model~\cite{liu2022convnet} and the ConvFormer-S18 model~\cite{yu2022metaformer}. In Section~\ref{sec:results}, we evaluate the training loss and the classification accuracy of these models on the ImageNet1k dataset \cite{deng2009imagenet}. The remainder of this chapter will present a detailed analysis of the methods, equations, algorithms, and techniques regarding the application of the Gaussian interpolation in DCLS.

\section{Related work}
In the field of convolutional neural networks (CNNs), various approaches have been explored to improve the performance and efficiency of convolutional operations. Gaussian mixture kernels were studied in \cite{tabernik2020spatially} under the name ``Displaced Aggregation Units''. The relationship between DCLS and this work was explained in \ref{dau} Gaussian mixture convolutional networks \cite{celarekgaussian} for their part, have investigated the fit of input channels with Gaussian mixtures, while \citet{chen2023gaussian} utilized Gaussian masks in their work. Additionally, continuous kernel convolution was studied in the context of image processing by \citet{kim2023smpconv}. Their approach is similar to the linear correlation introduced in \citet{thomas2019KPConv}. The interpolation function used in the last two works corresponds to the DCLS-Triangle method described in \ref{sec:methods}. \citet{romero2022ckconv} have also made notable contributions in learning continuous functions that map the positions to the weights~\cite{romero2022flexconv, romero2022ckconv}.

In the work by~\citet{jacobsen2016structured}, the kernel is represented as a weighted sum of basis functions, including centered Gaussian filters and their derivatives. \citet{pintea2021resolution} extended this approach by incorporating the learning of Gaussian width, effectively optimizing the resolution. \citet{shelhamer2019blurring} introduced a kernel factorization method where the kernel is expressed as a composition of a standard kernel and a structured Gaussian one. In these last three works the Gaussians are centered on the kernel.

Furthermore, the utilization of bilinear interpolation within deformable convolution modules has already shown its effectiveness. \citet{dai2017deformable}, \citet{qi2017deformable} and recently \citet{wang2022internimage} leveraged bilinear interpolation to smoothen the non-differentiable regular-grid offsets in the deformable convolution method. Even more recently, in \citet{kim2023understanding}, a Gaussian attention bias with learnable standard deviations has been successfully used in the positional embedding of the attention module of the ViT model \cite{dosovitskiyimage} and leads to reasonable gains on ImageNet1k.


\section{Methods}
\subsection{From bilinear to Gaussian interpolation }
\label{sec:methods} We denote by $m \in \mathbb{N}^{*}$ the number of kernel elements inside the dilated constructed kernel and we refer to it as the “kernel count”. Moreover, we denote respectively by $s_x, s_y \in \mathbb{N}^{*} \times \mathbb{N}^{* } $, the sizes of the constructed kernel along the x-axis and the y-axis. The latter could be seen as the limits of the dilated kernel, and we refer to them as the “dilated kernel size”.

The $s_x \times s_y$ matrix space over $\mathbb{R}$ is defined as the set of all $s_x \times s_y$ matrices over $\mathbb{R}$, and is denoted $\mathcal{M}_{s_x , s_y}(\mathbb{R})$. 
The real numbers $w$, $p^x$, $\sigma^x$, $p^y$  and $\sigma^y$ respectively stand for the weight, the mean position and standard deviation of that weight along the x-axis (width) and its mean position and standard deviation along the y-axis (height).

The mathematical construction of the 2D-DCLS kernel in \citet{hassani2023dilated} relies on bilinear interpolation and is described as follows :
\begin{align}
\label{chap3:eq:2-3}
  \begin{split}
  f \colon \mathbb{R} \times \mathbb{R} \times \mathbb{R} &\to \mathcal{M}_{s_x , s_y } (\mathbb{R})\\
  w, p^x, p^y  & \mapsto \quad \mK
  \end{split}
\end{align}
where $\forall i\in \llbracket 1 \ .. \ s_x \rrbracket$, $\forall j\in \llbracket 1 \ .. \ s_y \rrbracket  \colon $\\
\begin{equation}
\label{chap3:eq:2-4}
\arraycolsep=1.3pt\def\arraystretch{1}
\displaystyle \mK_{ij} =\left\{\begin{array}{cl}
w \ (1 - r^x)\ (1 - r^y) & \text {if } i = \floor{p^x}, \ j = \floor{p^y} \\
w \ r^x \ (1 - r^y) & \text {if }  i = \floor{p^x} + 1, \ j = \floor{p^y} \\
w \ (1 - r^x) \ r^y & \text {if }  i = \floor{p^x}, \ j = \floor{p^y} + 1\\
w \ r^x \ r^y  & \text {if }  i = \floor{p^x} {+} 1, \ j = \floor{p^y} {+} 1 \\
0 & \text {otherwise } 
\end{array}\right.
\end{equation}
and where the fractional parts are:
\begin{equation}
    \begin{array}{ccc}
    r^x = \{p^x\} = p^x - \floor{p^x} & \text{and} & r^y = \{p^y\} = p^y - \floor{p^y}
    \end{array}
\end{equation}

An equivalent way of describing the constructed kernel $K$ in Equation~\ref{chap3:eq:2-4} is:
\begin{equation}
    \label{eq:4}
    \mK_{ij} = w \cdot g(p^x - i) \cdot g(p^y - j) 
\end{equation}
with 
\begin{equation}
    g \colon x \mapsto \text{max}(0, \ 1 - |x|)
\end{equation}
This expression corresponds to the bilinear interpolation as described in \citet[][eq. 4]{dai2017deformable}.

\begin{figure}[bt]
     \centering
     \includegraphics[width=0.3\textwidth]{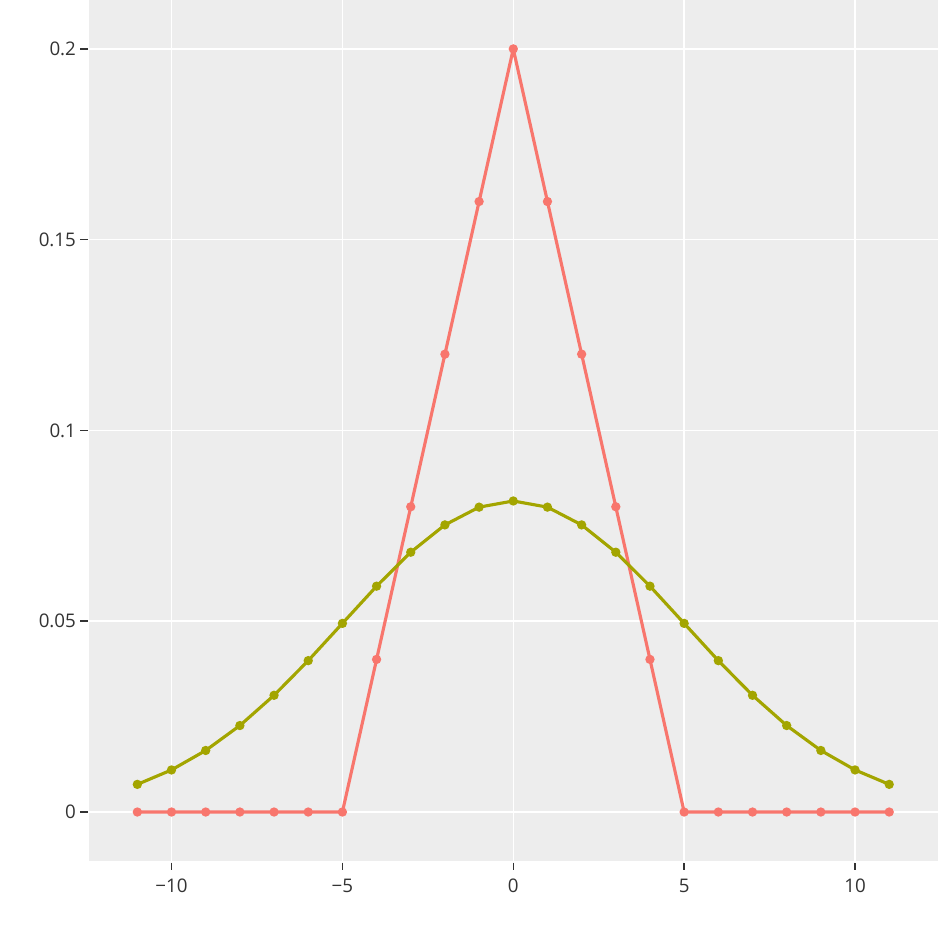}
     \caption{1D view of Gaussian and $\Lambda$ functions with $\sigma = 5$.}
     \label{weightings}
\end{figure}

In fact, this last $g$ function is known as the triangle function (refer to Fig.~\ref{weightings} for a graphic representation), and is widely used in kernel density estimation. From now on, we will note it as
\begin{equation}
    \forall x \in \mathbb{R} \quad \quad \Lambda (x) \overset{\text{def}}{=} \text{max}(0, \ 1 - |x|)
\end{equation}

This chapter aims to empirically show that this default $\Lambda$ function is not the best choice in terms of performance on the ImageNet1k dataset image classification task using convolutional neural networks models that have DCLS as a depthwise convolution layer in each block. 

First, we consider a scaling by a parameter $\sigma \in \mathbb{R}_+$ for the triangle function (the bilinear interpolation corresponds to $\sigma=1$),
\begin{equation}
\label{eq:triangle}
    \forall x \in \mathbb{R}, \quad \forall \sigma \in \mathbb{R}_+ \quad \Lambda_\sigma (x) \overset{\text{def}}{=} \text{max}(0, \ \sigma - |x|)
\end{equation}
We found that this scaling parameter $\sigma$ could be learned by backpropagation and that doing so increases the performance of the DCLS method. As we have different $\sigma$ parameters for the x and y-axes in 2D-DCLS, learning the standard deviations costs two additional learnable parameters and two additional FLOPs (multiplied by the number of the channels of the kernel and the kernel count). We refer to the DCLS method with triangle function interpolation as the DCLS-Triangle method.

Second, we tried a smoother function rather than the piecewise affine triangle function, namely the Gaussian function:
\begin{equation}
\label{eq:gauss}
    \begin{array}{lcrr}
     \forall x \in \mathbb{R}, \ \forall \sigma \in \mathbb{R}^*, & G_\sigma (x) \overset{\text{def}}{=} \text{exp}\left({-  {\dfrac{x^2}{2\sigma^2}}}\right) & &
    \end{array}
\end{equation}
We refer to the DCLS method with Gaussian interpolation as the DCLS-Gauss method. In practice, instead of Equations \ref{eq:triangle} and \ref{eq:gauss}, we respectively use:
\begin{equation}
    \forall x \in \mathbb{R}, \ \forall \sigma \in \mathbb{R}, \enskip \Lambda_{\sigma_0 + \sigma} (x) = \text{max}(0, \ \sigma_0 + |\sigma| - |x|)
\end{equation}
\begin{equation}
     \forall x \in \mathbb{R}, \ \forall \sigma \in \mathbb{R}, \enskip G_{\sigma_0 + \sigma} (x) = \text{exp}\left({-  \dfrac{1}{2} \dfrac{x^2}{(\sigma_0 + |\sigma|)^2}}\right)
\end{equation}
with $\sigma_0 \in \mathbb{R}^*_+$ a constant that determines the minimum standard deviation that the interpolation could reach. For the triangle interpolation, we take $\sigma_0 = 1$ in order to have at least 4 adjacent interpolation values (see Figure~\ref{fig:indexc}). And for the Gaussian interpolation, we set $\sigma_0 = 0.27$. 

Last, to make the sum of the interpolation over the dilated kernel size equal to 1, we divide the interpolations by the following normalization term :
\begin{equation} 
A = \epsilon + \sum_{i=1}^{s_x}\sum_{j=1}^{s_y} \mathcal{I}_{\sigma_0 + \sigma^x}(p^x -i) \cdot \mathcal{I}_{\sigma_0 + \sigma^y}(p^y -j) 
\end{equation} with $\mathcal{I}$ an interpolation function ($\Lambda$ or $G$ in our case) and $\epsilon = 1e-7$ for example, to avoid division by zero.

\textbf{Other interpolations} Based on our tests, other functions such as Lorentz, hyper-Gaussians and sinc functions have been tested with no great success. In addition, learning a correlation parameter $\rho \in [-1,1]$ or equivalently a rotation parameter $\theta \in [0, 2\pi]$ as in the bivariate normal distribution density did not improve performance (maybe because cardinal orientations predominate in natural images).
\subsection{The 2D-DCLS-Gauss kernel construction algorithm}
\label{sec:algo}
In the following, we describe with pseudocode the kernel construction used in 2D-DCLS-Gauss and 2D-DCLS-Triangle. $\mathcal{I}$ is the interpolation function ($\Lambda$ or $G$ in our case) and $\epsilon = 1e-7$. In practice, $w$, $p^x$, $p^y$, $\sigma^x$  and $\sigma^y$ are 3-D tensors of size (\texttt{channels\_out, channels\_in // groups, K\_count}), but the algorithm presented here is easily extended to this case by applying it channel-wise.
\begin{algorithm}[ht]
    \setstretch{1.2}
    \begin{algorithmic}[1]

        \REQUIRE $w$, $p^x$, $p^y$, $\sigma^x$, $\sigma^y$ :  vectors of dimension $m$
        \ENSURE $K$ : the constructed kernel, of size ($s_x \times s_y$)
        \STATE $K \leftarrow 0_{s_x,s_y}$ \COMMENT{zero tensor of size $s_x, s_y$} 
        \label{chap3:alg:forward}        
        \FOR{$k=0$ {\bfseries to} $m-1$}
            \STATE $H \leftarrow 0_{s_x,s_y}$
            \STATE $p_k^x \leftarrow p_k^x + s_x // 2 $; $\quad p_k^y \leftarrow p_k^y + s_y // 2 $ 
            \STATE $\sigma_k^x \leftarrow |\sigma_k^x| + \sigma_0^{\mathcal{I}}$;   $\quad \sigma_k^y \leftarrow|\sigma_k^y| + \sigma_0^{\mathcal{I}}$ 
            
            \FOR{$i=0$ {\bfseries to} $s^x-1 $}
                \FOR{$j=0$ {\bfseries to} $s^y-1 $}
                \STATE $H[i,j] \leftarrow \mathcal{I}_{\sigma_k^x}(p_k^x - i) * \mathcal{I}_{\sigma_k^y}(p_k^y - j)$
                \ENDFOR 
            \ENDFOR        
        \STATE $H[:,:] \leftarrow H[:,:] \ / (\epsilon + \sum\limits_{i=0}^{s^x-1} \sum\limits_{j=0}^{s^y-1} H[i,j])$
        \STATE $K \leftarrow K + H * w_k$      
        \ENDFOR
    \end{algorithmic} 
    \caption{2D-DCLS-interpolation kernel construction}
    \label{algo:1}
\end{algorithm}
 The 2D-DCLS convolution with kernel construction is then obtained by applying the classical 2D-convolution provided natively by PyTorch or any other method such as the \textit{depthwise implicit gemm} convolution method \cite{ding2022scaling} using the constructed kernel.
\begin{table*}[!bt]
\begin{center}
\resizebox{\textwidth}{!}{

$
\begin{array}{lcccccc}
\toprule
\text { model @ 224} & \begin{array}{l}
\text { ker. size } \\
\text { / count  }
\end{array} & \text {interpolation}  & \text { \# param.} & \text { train loss } & \text { Top-5 acc.} & \text { Top-1 acc.} \\
\hline 

\text { ConvNeXt-T }\squareneswfill  & 7^2 \ / \ 49 &  & 28.59 \mathrm{M} & 2.828 & 96.05 & 82.08 \\
\rowcolor{lightcream}\text { ConvNeXt-T } \squarecrossfill & 17^2  \ / \  34 & \text{Bilinear} &  28.59 \mathrm{M} & 2,775 & 96.11 & 82.44 \\
\rowcolor{lightcream}\text { ConvNeXt-T } \odot & 23^2  \ / \  26 & \text{Triangle} &  28.59 \mathrm{M} & 2.787  & 96.09 & 82.34 \\
\rowcolor{lightcream}\text { ConvNeXt-T } \star & 23^2  \ / \  26 & \text{Gaussian} & 28.59 \mathrm{M} & 2.762 & 96.18 & 82.44 \\
\text { ConvNeXt-T } & 17^2  \ / \  26 & \text{Gaussian} &  28.59 \mathrm{M} & 2.773 & 96.17 & 82.40 \\
\text { ConvNeXt-T } & 23^2  \ / \  34 & \text{Gaussian} & 28.69 \mathrm{M} & 2.758 & 96.22 & 82.60
\\
\midrule
\text { ConvFormer-S18 } \squareneswfill &  7^2 \ / \ 49 & & 26.77 \mathrm{M} & 2.807 & 96.17 & 82.84 \\
\rowcolor{lightcream}\text { ConvFormer-S18 } \squarecrossfill & 17^2 \ / \ 40 &  \text{Bilinear}  & 26.76 \mathrm{M} & 2.764 & 96.42 & 83.14\\
\rowcolor{lightcream}\text { ConvFormer-S18 } \odot & 23^2 \ / \ 26 &  \text{Triangle}  & 26.76 \mathrm{M} & 2.761 & 96.38 & 83.09\\
\rowcolor{lightcream}\text { ConvFormer-S18 }  \star & 23^2 \ / \ 26 &  \text{Gaussian} & 26.76 \mathrm{M} & 2.747 & 96.31 & 82.99\\
\bottomrule
\end{array}
$

}
\end{center}
\caption{\textbf{Classification accuracy on the validation set and training loss on ImageNet-1K.} For the 17/34 bilinear, the 23/26 Triangle and Gaussian cases, the results have been averaged over 3 distinct seeds (the corresponding lines are highlighted in yellow).}
\label{tab:results}
\end{table*}
\section{Learning techniques}
\label{chap3:sec:usage}
Having discussed the implementation of the interpolation in the DCLS method, we now shift our focus to the techniques employed to maximize its potential. We retained most of the techniques used in \citet{hassani2023dilated}, and suggest new ones for learning standard deviation parameters. In Appendix~\ref{app:learning_techniques}, we present the training techniques that have been selected based on consistent empirical evidence, yielding improved training loss and validation accuracy.
\section{Results}
\label{sec:results}

\begin{figure}[bt]
     \centering
     \includegraphics[width=0.7\textwidth]{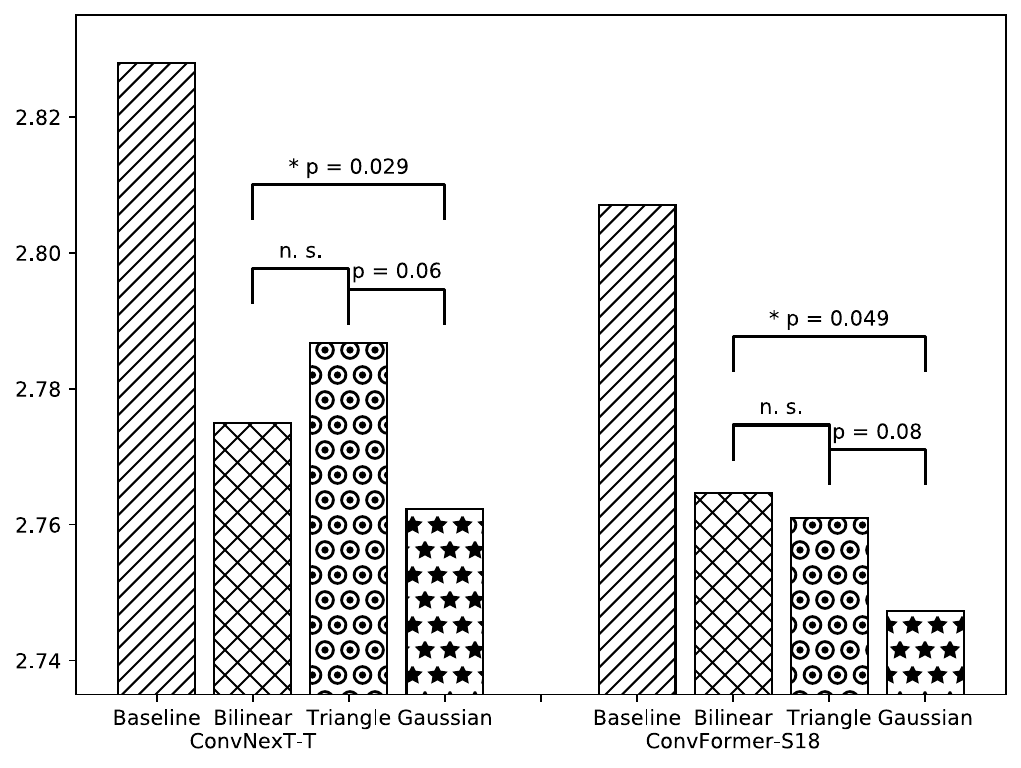}
     \caption{Training loss for ConvNeXt-T and ConvFormer-S18 models with DCLS according to interpolation type (lower is better). The pairwise p-values have been calculated using an independent two-sample Student t-test assuming equal variances. The vertical line segments stand for the standard errors. }
     \label{t-test}
\end{figure}
We took two recent state-of-the-art convolutional architectures, ConvNeXt and ConvFormer, and drop-in replaced all the depthwise convolutions by DCLS ones, using the three different interpolations (bilinear, triangle, or Gauss). Table~\ref{tab:results} reports the results in terms of training loss and validation accuracy.

A first observation is that all the DCLS models perform much better than the baselines, whereas they have the same number of parameters. There are also subtle differences between interpolation functions. As Figure~\ref{t-test} shows, triangle and bilinear interpolations perform similarly, but the Gaussian interpolation performs significantly better. 

Furthermore, the advantage of the Gaussian interpolation w.r.t. bilinear is not only due to the use of a larger kernel, as a 17x17 Gaussian kernel (5th line in Table~\ref{tab:results}) still outperforms the bilinear case (2nd line). Finally, the 6th line in Table~\ref{tab:results} shows that there is still room for improvement by increasing the kernel count, although this slightly increases the number of trainable parameters w.r.t. the baseline.

\section{Appendix: Code and reproducibility}
The code of the method is based on PyTorch and available at \href{https://github.com/K-H-Ismail/Dilated-Convolution-with-Learnable-Spacings-PyTorch}{https://github.com/K-H-Ismail/Dilated-Convolution-with-Learnable-Spacings-PyTorch}.

\section{Appendix: \emph{Pytorch} 2D-DCLS-Gauss / Triangle implementations}
\label{code:pytorch}
\begin{lstlisting}[language=Python]
import torch
from torch.nn import Module
from torch.nn.parameter import Parameter


class ConstructKernel2d(Module):
    def __init__(
        self,
        out_channels,
        in_channels,
        groups,
        kernel_count,
        dilated_kernel_size,
        version,
    ):
        super().__init__()
        self.version = version
        self.out_channels, self.in_channels = out_channels, in_channels
        self.groups = groups
        self.dilated_kernel_size = dilated_kernel_size
        self.kernel_count = kernel_count
        self.IDX, self.lim = None, None

    def __init_tmp_variables__(self, device):
        if self.IDX is None or self.lim is None:
            J = Parameter(
                torch.arange(0, self.dilated_kernel_size[0]),
                requires_grad=False,
            ).to(device)
            I = Parameter(
                torch.arange(0, self.dilated_kernel_size[1]),
                requires_grad=False,
            ).to(device)
            I = I.expand(self.dilated_kernel_size[0], -1)
            J = J.expand(self.dilated_kernel_size[1], -1).t()
            IDX = torch.cat((I.unsqueeze(0), J.unsqueeze(0)), 0)
            IDX = IDX.expand(
                self.out_channels,
                self.in_channels // self.groups,
                self.kernel_count,
                -1,
                -1,
                -1,
            ).permute(4, 5, 3, 0, 1, 2)
            self.IDX = IDX
            lim = torch.tensor(self.dilated_kernel_size).to(device)
            self.lim = lim.expand(
                self.out_channels,
                self.in_channels // self.groups,
                self.kernel_count,
                -1,
            ).permute(3, 0, 1, 2)
        else:
            pass

    def forward_vtriangle(self, W, P, SIG):
        P = P + self.lim // 2
        SIG = SIG.abs() + 1.0
        X = self.IDX - P
        X = ((SIG - X.abs()).relu()).prod(2)
        X = X / (X.sum((0, 1)) + 1e-7)  # normalization
        K = (X * W).sum(-1)
        K = K.permute(2, 3, 0, 1)
        return K

    def forward_vgauss(self, W, P, SIG):
        P = P + self.lim // 2
        SIG = SIG.abs() + 0.27
        X = ((self.IDX - P) / SIG).norm(2, dim=2)
        X = (-0.5 * X ** 2).exp()
        X = X / (X.sum((0, 1)) + 1e-7)  # normalization
        K = (X * W).sum(-1)
        K = K.permute(2, 3, 0, 1)
        return K

    def forward(self, W, P, SIG):
        self.__init_tmp_variables__(W.device)
        if self.version == "triangle":
            return self.forward_vtriangle(W, P, SIG)
        if self.version == "gauss":
            return self.forward_vgauss(W, P, SIG)
        raise
\end{lstlisting}
\section{Appendix: Learning techniques}
\label{app:learning_techniques}
\begin{itemize}[leftmargin=*]
    \item \textbf{Weight decay:} No weight decay was used for positions. We apply the same for standard deviation parameters. 
    \item \textbf{Positions and standard deviations initialization:} position parameters were initialized following a centered normal law of standard deviation 0.5. Standard deviation parameters were initialized to a constant $0.23$ in DCLS-Gauss and to $0$ in DCLS-Triangle in order to have a similar initialization to DCLS with bilinear interpolation at the beginning.
    \item \textbf{Positions clamping :} Previously in DCLS, kernel elements that reach the dilated kernel size limit were clamped. It turns out that this operation is no longer necessary with the Gauss and $\Lambda$ interpolations. 
    \item \textbf{Dilated kernel size tuning:} When utilizing bilinear interpolation in ConvNeXt-dcls, a dilated kernel size of 17 was found to be optimal, as larger sizes did not yield improved accuracy. However, with Gaussian and $\Lambda$ interpolations, there appears to be no strict limit to the dilated kernel size. Accuracy tends to increase logarithmically as the size grows, with improvements observed up to kernel sizes of 51. It is important to note that increasing the dilated kernel size does not impact the number of trainable parameters, but it does affect throughput. Therefore, a compromise between accuracy and throughput was achieved by setting the dilated kernel size to 23.
    \item \textbf{Kernel count tuning:} This hyper-parameter has been configured to the maximum integer value while still remaining below the baselines to which we compare ourselves in terms of trainable parameters. It is worth noting that each additional element in the 2D-DCLS-Gauss or 2D-DCLS-Triangle methods introduces five more learnable parameters: weight, vertical and horizontal position, and their respective standard deviations. To maintain simplicity, the same kernel count was applied across all model layers.
    \item \textbf{Learning rate scaling:} To maintain consistency between positions and standard deviations, we applied the same learning rate scaling ratio of 5 to both. In contrast, the learning rate for weights remained unchanged.
    \item \textbf{Synchronizing positions:} we shared the kernel positions and standard deviations across convolution layers with the same number of parameters, without sharing the weights. Parameters in these stages were centralized in common parameters that accumulate the gradients. 
    
\end{itemize}
\section{Appendix: 1D and 3D convolution cases} 
For the 3D case, Equation~\ref{eq:4} can be generalized as a product along spatial dimensions. We denote respectively by $s_x, s_y, s_z \in \mathbb{N}^{*} \times \mathbb{N}^{* } \times \mathbb{N}^{* } $, the sizes of the constructed kernel along the x-axis, the y-axis and the z-axis. The constructed kernel tensor $K^{3D} \in \mathcal{M}_{s_x , s_y, s_z } (\mathbb{R})$ is therefore:

$\forall i\in \llbracket 1 \ .. \ s_x \rrbracket$, $\forall j\in \llbracket 1 \ .. \ s_y \rrbracket$, $\forall k\in \llbracket 1 \ .. \ s_z \rrbracket  \colon $
\begin{equation}
    K^{3D}_{ijk} = w \cdot \mathcal{I}_{\sigma_0 + \sigma^x}(p^x - i) \cdot \mathcal{I}_{\sigma_0 + \sigma^y}(p^y - j) \cdot \mathcal{I}_{\sigma_0 + \sigma^z}(p^z - k) 
\end{equation}
with $\mathcal{I}$ an interpolation function ($\Lambda$ or $G$), $\sigma_0 = 1$ for the $\Lambda$ interpolation and $\sigma_0 = 0.27$ for the Gaussian one. $w$, $p^x$, $\sigma^x$, $p^y$, $\sigma^y$, $p^z$  and $\sigma^z$ respectively representing the weight, the mean position and standard deviation of that weight along the x-axis (width), the mean position and standard deviation along the y-axis (height) and its mean position and standard deviation along the z-axis (depth).

The constructed kernel vector $K^{1D} \in \mathbb{R}^{s_x}$ for the 1D case is simply:

$\forall i\in \llbracket 1 \ .. \ s_x \rrbracket \colon $
\begin{equation}
    K^{1D}_{i} = w \cdot \mathcal{I}_{\sigma_0 + \sigma^x}(p^x - i) 
\end{equation}

The Algorithm~\ref{algo:1}, as well as the Pytorch code~\ref{code:pytorch}, are readily adapted to these cases by following the above note.

\begin{center}
    \pgfornament[scale=.5, opacity=.85, symmetry=h, ydelta=2cm]{70}
    \pgfornament[scale=.5, opacity=.85, ydelta=2cm]{70}
\end{center}

\chapter{Learning Delays in Spiking Neural Networks using DCLS}
\label{chap:4}
\section{Disclaimer}
This chapter is strongly inspired by the article: \cite{hammouamri2023learning}. Most parts of this chapter have been taken verbatim from the paper of which we are the second author, we participated in a significant way in the research, development, experimentation, and writing of this paper.

\section{Introduction}

\begin{figure}[ht]
  \centering
  \includegraphics[width=0.92\textwidth]{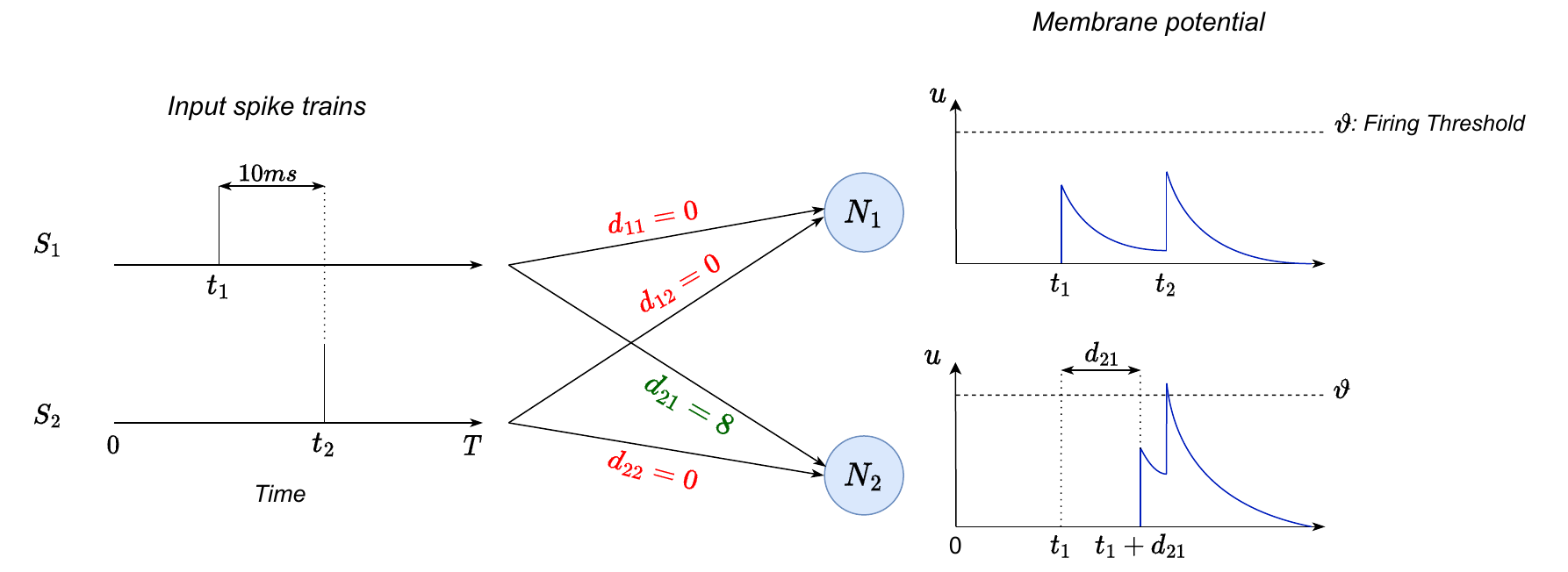}
  \caption{Coincidence detection: we consider two neurons $N_1$ and $N_2$ with the same positive synaptic weight values. $N_2$ has a delayed synaptic connection denoted $d_{21}$ of $8$ms, thus both spikes from spike train $S_1$ and $S_2$ will reach $N_2$ quasi-simultaneously. As a result, the membrane potential of $N_2$ will reach the threshold $\vartheta$ and $N_2$ will emit a spike. On the other hand, $N_1$ will not react to these same input spike trains. }
\label{fig:coincidence_detection}
\end{figure}

Spiking neurons are coincidence detectors \citep{Konig1996,Rossant2011}: they respond more when receiving synchronous, rather than asynchronous, spikes. Importantly, it is the spike arrival times that should coincide, not the spike emitting times --  these times are different because propagation is usually not instantaneous. There is a delay between spike emission and reception, called delay of connections, which can vary across connections. Thanks to these heterogeneous delays, neurons can detect complex spatiotemporal spike patterns, not just synchrony patterns \citep{Izhikevich2006} (see Figure \ref{fig:coincidence_detection}).

In the brain, the delay of a connection corresponds to the sum of the axonal, synaptic, and dendritic delays. It can reach several tens of milliseconds, but it can also be much shorter (1 ms or less) \citep{Izhikevich2006}. For example, the axonal delay can be reduced with myelination, which is an adaptive process that is required to learn some tasks (see \cite{Bowers2017a} for a review). In other words, learning in the brain can not be reduced to synaptic plasticity, delay learning is also important.

One theoretical study, in particular, has led to the same conclusion: Maass and Schmitt demonstrated, using simple spiking neuron models, that a Spiking Neural Network (SNN) with k adjustable delays can compute a much richer class of functions than a threshold circuit with k adjustable weights \citep{Maass1999}.

On most neuromorphic chips, synapses have a programmable delay. This is the case for Intel Loihi \citep{Davies2018}, IBM TrueNorth \citep{Akopyan2015}, SpiNNaker \citep{Furber2014} and SENeCA \citep{Yousefzadeh2022}.

All these points have motivated us and others (see related works in the next section) to propose delay learning rules. Here, we show that delays can be learned together with the weights, using backpropagation, in arbitrarily deep SNNs. More specifically, we first show that there is a mathematical equivalence between 1D temporal convolutions and connection delays. Thanks to this equivalence, we then demonstrate that the delays can be learned using Dilated Convolution with Learnable Spacings \citep{hassani2023dilated, khalfaouihassani2023dilated}, which was proposed for another purpose, namely to increase receptive field sizes in non-spiking 2D CNNs for computer vision. In practice, the method is fully integrated with PyTorch and leverages its automatic differentiation engine.

\section{Related Work}
\subsection{Deep Learning for Spiking Neural Networks}

Recent advances in SNN training methods like the surrogate gradient method \citep{neftci2018surrogate, slayer} and the ANN2SNN conversion methods \citep{ann2snn_1, ann2snn_2, ann2snn_3} made it possible to train increasingly deeper spiking neural networks. The surrogate gradient method defines a continuous relaxation of the non-smooth spiking nonlinearity: it replaces the gradient of the Heaviside function used in the spike-generating process with a smooth surrogate gradient that is suitable for optimization. On the other hand, the ANN2SNN methods convert conventional artificial neural networks (ANNs) into SNNs by copying the weights from ANNs while trying to minimize the conversion error. 

Other works have explored improving the spiking neurons using inspiration from biological mechanisms or techniques used in ANNs. The Parametric Leaky Integrate-and-Fire (PLIF) \citep{plif} incorporates learnable membrane time constants that could be trained jointly with synaptic weights. \citet{bellec2018} were the first to propose a method for dynamically adapting firing thresholds in deep (recurrent) SNNs, \citet{hammouamri2022mitigating} also proposes a method to dynamically adapt firing thresholds in order to improve continual learning in SNNs. Spike-Element-Wise ResNet \citep{sew} addresses the problem of vanishing/exploding gradient in the plain Spiking ResNet caused by sigmoid-like surrogate functions and successfully trained the first deep SNN with more than 150 layers. Spikformer \citep{spikformer} adapts the softmax-based self-attention mechanism of Transformers \citep{transformer} to a spike-based formulation. Other recent works like SpikeGPT \citep{spikegpt} and Spikingformer \citep{spikformer} also propose spike-based transformer architectures. These efforts have resulted in closing the gap between the performance of ANNs and SNNs on many widely used benchmarks.

\subsection{Delays in SNNs}

Few previous works considered learning delays in SNNs. \citet{Delay_Learning_Kernel} proposed a similar method to ours in which they convolve spike trains with an exponential kernel so that the gradient of the loss with respect to the delay can be calculated. However, their method is used only for a shallow SNN with no hidden layers. 

Other methods like \cite{related_delays1,related_delays1bis, related_delays2, related_delays3} also proposed learning rules developed specifically for shallow SNNs with only one layer. \citet{related_delays4} proposed to learn temporal delays with Spike Timing Dependent Plasticity (STDP) in weightless SNNs. \citet{DW_photonic} proposed a method for delay-weight supervised learning in optical spiking neural networks. \citet{iscas} proposed a method for deep feedforward SNNs that uses a set of multiple fixed delayed synaptic connections for the same two neurons before pruning them depending on the magnitude of the learned weights.

To the best of our knowledge, SLAYER \citep{slayer} and \cite{sun22, sun23, sun23-2} (which are based on SLAYER) are the only ones to learn delays and weights jointly in a deep SNN. However, unless a Spike Response Model (SRM) \citep{srm} is used, the gradient of the spikes with respect to the delays is numerically estimated using finite difference approximation, and we think that those gradients are not precise enough as we achieve similar performance in our experiments with fixed random delays (see Table~\ref{table:results} and Figure~\ref{fig:barplots}).

\label{random_init}
Furthermore, we propose a control test that was not considered by the previous works and that we deem necessary: the SNN with delay learning should outperform an equivalent SNN with fixed random and uniformly distributed delays, especially with sparse connectivity.

\section{Methods}
\subsection{Spiking Neuron Model}
\label{methods:spiking}
The spiking neuron, which is the fundamental building block of SNNs, can be simulated using various models. In this work, we use the Leaky Integrate-and-Fire model \citep{gerstnerkistler2002}, which is the most widely used for its simplicity and efficiency. The membrane potential $u_i^{(l)}$ of the $i$-th neuron in layer $l$ follows the differential equation: 

\begin{equation}
    \tau \frac{du_i^{(l)}}{dt} = -(u_i^{(l)}(t) - u_{\text{reset}}) + RI_i^{(l)}(t)
\end{equation}
where $\tau$ is the membrane time constant, $u_{reset}$ the potential at rest, $R$ the input resistance and $I_i^{(l)}(t)$ the input current of the neuron at time $t$.

In addition to the sub-threshold dynamics, a neuron emits a unitary spike $S_i^{(l)}$ when its membrane potential exceeds the threshold $\vartheta$, after which it is instantaneously reset to $u_{reset}$. Finally, the input current $I_i^{(l)}(t)$ is stateless and represented as the sum of afferent weights $W_{ij}^{(l)}$ multiplied by spikes $S_j^{(l-1)}(t)$:

\begin{equation}
I_i^{(l)}(t) = \sum_{j} W_{ij}^{(l)} S_j^{(l-1)}(t)
\end{equation}

We formulate the above equations in discrete time using Euler's method approximation, and using $u_{reset} = 0$ and $R = \tau$.

\begin{align}
    u_i^{(l)}[t] &= (1 - \frac{1}{\tau})u _i^{(l)}[t-1] + I_i^{(l)}[t] \\
    I_i^{(l)}[t] &=  \sum_{j} W_{ij}^{(l)} S_j^{(l-1)}[t] \\
    S_i^{(l)}[t] &= \Theta(u_i^{l}[t] - \vartheta)
\end{align}

We use the surrogate gradient method \citep{neftci2018surrogate} and define $\Theta'(x) \triangleq \sigma'(x)$ during the backward step, where $\sigma(x)$ is the surrogate arctangent function \citep{plif}.
\begin{figure}[!ht]
    \centering
        \includegraphics[width=0.7\textwidth]{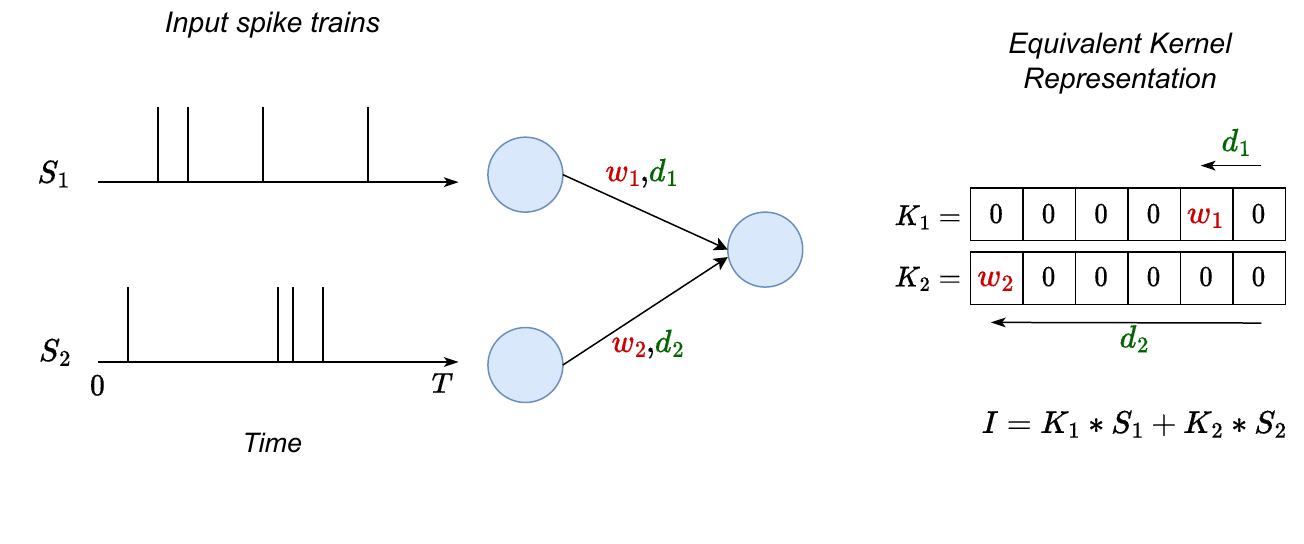}
    \caption{Example of one neuron with 2 afferent synaptic connections, convolving $K1$ and $K2$ with the zero left-padded $S_1$ and $S_2$ is equivalent to following Equation \ref{eq:Input_ff_delayd} }
    
    \label{fig:methods_fig1}
\end{figure}

\subsection{Synaptic Delays as a Temporal Convolution}
\label{methods:conv}
In the following, for clarity, we assume one synapse only between pairs of neurons (modeled with a kernel containing only one non-zero element). Generalization to multiple synapses (kernels with multiple non-zero elements) is trivial and will be explored in the experiments.

A feed-forward SNN model with delays is parameterized with $W = (w_{ij}^{(l)}) \in \mathbb{R}$ and $D = (d_{ij}^{(l)}) \in \mathbb{R}^+$, where the input of neuron $i$ at layer $l$ is 
\begin{equation}
    I_i^{(l)}[t] =  \sum_{j} w_{ij}^{(l)} S_j^{(l-1)}[t-d_{ij}^{(l)}]
    \label{eq:Input_ff_delayd}
\end{equation}

We model a synaptic connection from neuron $j$ in layer $l-1$ to neuron $i$ in layer $l$ which have a synpatic weight $w_{ij}^{(l)}$ and delay $d_{ij}^{(l)}$ as a one dimensional temporal convolution (see Figure \ref{fig:methods_fig1}) with kernel $k_{ij}^{(l)}$ as follows:

$\forall n \in \llbracket 0, ... \ T_d-1 \rrbracket  \colon$
\begin{equation}
k_{ij}^{(l)}[n] =
\begin{cases}
 w_{ij}^{(l)} & \text{if } n = T_d - d_{ij}^{(l)} - 1  \\
 0 & \text{otherwise}
\end{cases}
\label{eq:discrete_delays}
\end{equation}


where $T_d$ is the kernel size or maximum delay + 1. Thus we redefine the input $I_i^{(l)}$ in Equation \ref{eq:Input_ff_delayd} as a sum of convolutions:
\begin{equation}
I_i^{(l)} =  \sum_{j} k_{ij}^{(l)} \ast S_j^{(l-1)}
\end{equation}

We used a zero left-padding with size $T_d-1$ on the input spike trains $S$ so that $I[0]$ does correspond to $t=0$. Moreover, a zero right-padding could also be used, but it is optional; it could increase the expressivity of the learned delays with the drawback of increasing the processing time as the number of time steps after the convolution will increase.

To learn the kernel elements positions (i.e., delays), we use the 1D version of DCLS \citep{hassani2023dilated}  with a Gaussian kernel \citep{khalfaouihassani2023dilated} centered at $T_d - d_{ij}^{(l)} -1$, where $d_{ij}^{(l)}\in \llbracket 0,\ T_d-1 \rrbracket$, and of standard deviation $\sigma_{ij}^{(l)} \in \mathbb{R^*}$, thus we have:

$\forall n \in \llbracket 0, ... \ T_d-1 \rrbracket  \colon$
\begin{equation}
    k_{ij}^{(l)}[n] = \frac{w_{ij}^{(l)}}{c} \ \text{exp} \left({- \frac{1}{2}\left(\frac{n - T_d + d_{ij}^{(l)} + 1 }{\sigma_{ij}^{(l)}} \right)^2} \right)
\end{equation}

With \begin{equation} 
c = \epsilon + \sum_{n=0}^{T_d -1} \text{exp} \left({- \frac{1}{2}\left(\frac{n - T_d + d_{ij}^{(l)} + 1 }{\sigma_{ij}^{(l)}} \right)^2} \right)
\end{equation} a normalization term and $\epsilon = 1e-7$ to avoid division by zero, assuming that the tensors are in \texttt{float32} precision. During training, $d_{ij}^{(l)}$ are clamped after every batch to ensure their value stays in $\llbracket 0, ... \ T_d-1 \rrbracket$.

The learnable parameters of the 1D DCLS layer with Gaussian interpolation are the weights $w_{ij}$, the corresponding delays $d_{ij}$, and the standard deviations $\sigma_{ij}$. However, in our case, $\sigma_{ij}$ are not learned, and all kernels in our model share the same decreasing standard deviation, which will be denoted as $\sigma$. Throughout training, we exponentially decrease $\sigma$ as our end goal is to have a sparse kernel where only the delay position is non-zero and corresponds to the weight.

\begin{figure}[!ht]

  \centering
  \includegraphics[width=0.9\textwidth]{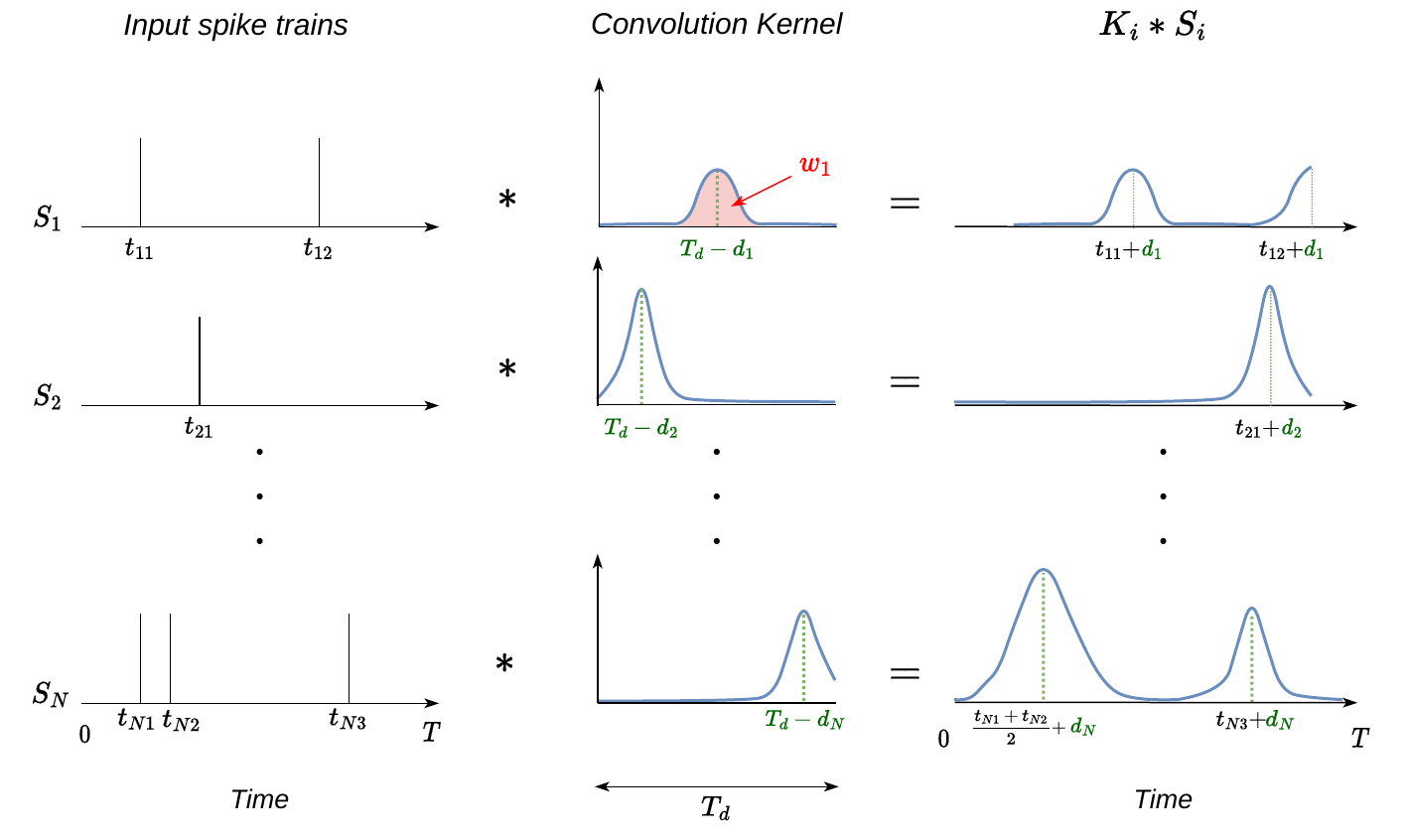}
  \caption{Gaussian convolution kernels for $N$ synaptic connections. The Gaussians are centered on the delay positions, and the area under their curves corresponds to the synaptic weights $w_i$. On the right, we see the delayed spike trains after being convolved with the kernels. (the $-1$ was omitted for figure clarity).}
  \label{fig:methods_fig2}
\end{figure}

The Gaussian kernel transforms the discrete positions of the delays into a smoother kernel (see Figure \ref{fig:methods_fig2}), which enables the calculation of the gradients $\frac{\partial L}{\partial d_{ij}^{(l)}}$.

\begin{figure}[!ht]

  \centering
  \includegraphics[width=0.7\textwidth]{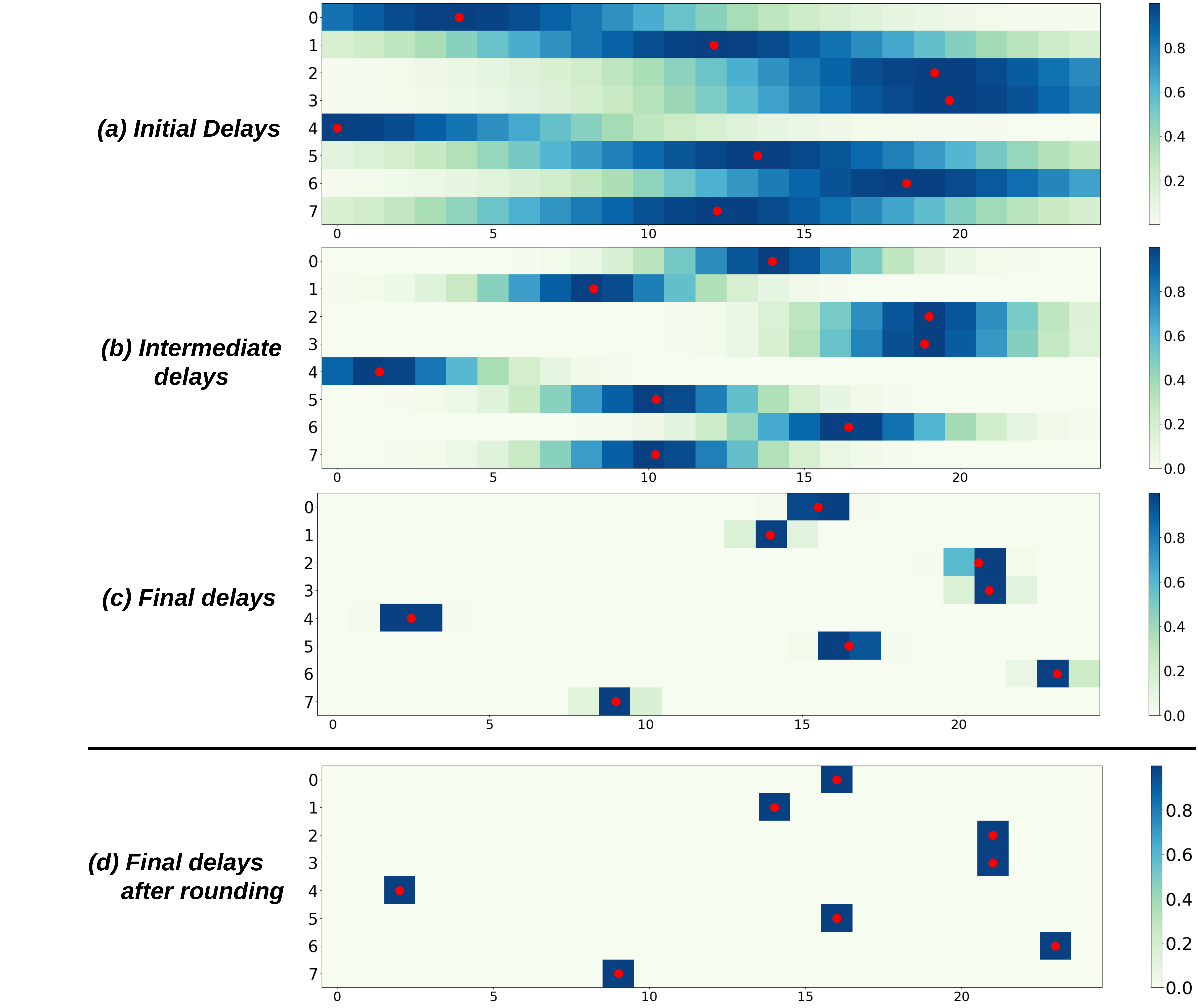}
  \caption{This figure illustrates the evolution of the same delay kernels for an example of eight synaptic connections of one neuron throughout the training process. The x-axis corresponds to time, and each kernel is of size $T_d=25$. And the y-axis is the synapse id. (a) corresponds to the initial phase where the standard deviation of the Gaussian $\sigma$ is large ($\frac{T_d}{2}$), allowing to take into consideration long temporal dependencies. (b) corresponds to the intermediate phase, (c) is taken from the final phase where $\sigma$ is at its minimum value (0.5) and weight tuning is more emphasized. Finally, (d) represents the kernel after converting to the discrete form with rounded positions.}
  \label{fig:methods_fig3}
\end{figure}

By adjusting the parameter $\sigma$, we can regulate the temporal scale of the dependencies. A small value for $\sigma$ enables the capturing of variations that occur within a brief time frame. In contrast, a larger value of $\sigma$ facilitates the detection of temporal dependencies that extend over longer durations. Thus, $\sigma$ tuning is crucial to the trade-off between short-term precision and long-term dependencies.

We start with a high $\sigma$ value and exponentially reduce it throughout the training process, after each epoch, until it reaches its minimum value of 0.5 (Fig.~\ref{fig:methods_fig3}). This approach facilitates the learning of distant long-term dependencies at the initial time. Subsequently, when $\sigma$ has a smaller value, it enables refining both weights and delays with more precision, making the Gaussian kernel more similar to the discrete kernel that is used at inference time. As we will see later in our ablation study (Section~\ref{sec:ablation}), this approach outperforms a constant $\sigma$.

Indeed, the Gaussian kernel is only used to train the model; when evaluating on the validation or test set, it is converted to a discrete kernel as described in Equation \ref{eq:discrete_delays} by rounding the delays. This permits the implementation of sparse kernels for inference which are very useful for uses on neuromorphic hardware, for example, as they correspond to only one synapse between pairs of neurons, with the corresponding weight and delay.

\section{Experiments}

\subsection{Experimental Setup}


We chose to evaluate our method on the SHD (Spiking Heidelberg Digits) and SSC (Spiking Speech Commands)/GSC (Google Speech Commands v0.02) datasets \citep{shd}, as they require leveraging temporal patterns of spike times to achieve good classification accuracy, unlike most computer vision spiking benchmarks. Both spiking datasets are constructed using artificial cochlear models to convert audio speech data to spikes; the original audio datasets are the Heidelberg Dataset (HD) and the GSC v0.02 Dataset (SC) \citep{SC} for SHD and SSC, respectively. 

The SHD dataset consists of 10k recordings of 20 different classes that consist of spoken digits ranging from zero to nine in both English and German languages. SSC and GSC are much larger datasets that consist of 100k different recordings. The task we consider on SSC and GSC is the top one classification on all 35 different classes (similar to \cite{shd, baseline}), which is more challenging than the original key-word spotting task on 12 classes, proposed in \cite{SC}.

For the two spiking datasets, we used spatio-temporal bins to reduce the input dimensions. 
Input neurons were reduced from 700 to 140 by binning every 5 neurons; as for the temporal dimension, we used a discrete time-step $\Delta t = 10$ ms and a zero right-padding to make sure all recordings in a batch have the same time duration. As for the non-spiking GSC, we used the Mel Spectrogram representation of the waveforms with 140 frequency bins and approximately 100 timesteps to remain consistent to the input sizes used in SSC.


We used a very simple architecture: a feedforward SNN with two or three hidden fully connected layers. Each feedforward layer is implemented using a DCLS module where each synaptic connection is modeled as a 1D temporal convolution with one Gaussian kernel element (as described in Section~\ref{methods:conv}), followed by batch normalization, a LIF module (as described in Section~\ref{methods:spiking}) and dropout. Table \ref{NetworkParams} lists the values of some hyperparameters used for the three datasets (for more details, refer to the code repository). 

\begin{table}[ht]
    \caption{Network parameters for different datasets}
    \label{NetworkParams}
    \centering
    \resizebox{\textwidth}{!}{
    \begin{tabular}{cccccc}
        \toprule
            Dataset  & \# Hidden Layers & \# Hidden size & $\tau$(ms) &Maximum Delay(ms) & Dropout rate  \\
        \midrule
            SHD & 2 & 256 & $10.05^*$   & 250 & 0.4\\
            SSC/GSC & 2 or 3 & 512 & 15 & 300 & 0.25\\
        \bottomrule
  \end{tabular}
  }
{\raggedright\footnotesize
*We found that a LIF with quasi-instantaneous leak $\tau=10.05$ (since  $\Delta t = 10$) is better than using a Heaviside function for SHD. \par}

\end{table}

The readout layer consists of $n_{\text{classes}}$ LIF neurons with an infinite threshold (where $n_{\text{classes}}$ is 20 or 35 for SHD and SSC/GSC, respectively). Similar to \cite{baseline}, the output $\text{out}_i[t]$ for every neuron $i$ at time $t$ is
\begin{equation}
    \text{out}_i[t] = \text{softmax}(u_i^{(r)}[t]) = \frac{e^{u_i^{(r)}[t]}}{\sum_{j=1}^{n_{\text{classes}}} e^{u_j^{(r)}[t]}}
\end{equation}
where $u_i^{(r)}[t]$ is the membrane potential of neuron $i$ in the readout layer $r$ at time $t$.\\
The final output of the model after $T$ time steps is defined as
\begin{equation}
\hat{y_i} = \sum_{t=1}^{T} \text{out}_i[t]
\end{equation}

We denote the batch size by $N$ and the ground truth by $y$. We calculate the cross-entropy loss for one batch as

\begin{equation}
\mathcal{L}= \frac{1}{N} \sum_{n=1}^{N} -\log(\text{softmax}(\hat{y}_{y_n}[n]))
\end{equation}


The Adam optimizer \citep{adam} is used for all models and groups of parameters with base learning rates $lr_{w} = 0.001$ for synaptic weights and $lr_{d} = 0.1$  for delays. We used a one-cycle learning rate scheduler \citep{one_cycle} for the weights and cosine annealing \citep{cosine_annealing} without restarts for the delays learning rates.
This work was implemented\footnote{Our code is available at: \url{https://github.com/Thvnvtos/SNN-delays} 
} using the PyTorch-based SpikingJelly \citep{SpikingJelly,Fang2023a} framework.

\subsection{Results}

We compare our method (DCLS-Delays) in Table \ref{table:results} to previous works on the SHD, SSC, and GSC-35 (35 denoting the 35 classes harder version) benchmark datasets in terms of accuracy, model size, and whether recurrent connections or delays were used.

The reported accuracy of our method corresponds to the accuracy on the test set using the best-performing model on the validation set. However, since there is no validation set provided for SHD we use the test set as the validation set (similar to \cite{baseline}). The margins of error are calculated at a 95\% confidence level using a t-distribution (we performed ten and five experiments using different random seeds for SHD and SSC/GSC, respectively).

\begin{table}[htbp]
    \caption{Classification accuracy on SHD, SSC, and GSC-35 datasets}
    \label{table:results}
    \centering
    \resizebox{\textwidth}{!}{
    \begin{tabular}{llcccl}
        \toprule
            Dataset  &   Method & Rec. & Delays &  \#Params  & Top1 Acc.  \\
        \midrule 
            \multirow{8}{4em}{\textbf{SHD}}
            & \small EventProp-GeNN \footnotesize \citep{eventprop-genn}  & \checkmark & \xmark & N/a & 84.80$\pm$1.5\%     \\
            & \small Cuba-LIF \footnotesize \citep{spikGRU} & \checkmark & \xmark  & 0.14M & 87.80$\pm$1.1\% \\
            & \small Adaptive SRNN \footnotesize \citep{Adaptive-SRNN} & \checkmark & \xmark  & N/a & 90.40\% \\
            & \small SNN+Delays \footnotesize \citep{iscas} & \xmark & \checkmark & 0.1M & 90.43\% \\
            & \small TA-SNN \footnotesize \citep{TA-SNN} & \xmark & \xmark  & N/a &  91.08\%     \\
            & \small STSC-SNN \footnotesize \citep{FFSNNattention} & \xmark & \xmark & 2.1M &  92.36\%     \\
            & \small Adaptive Delays \footnotesize \citep{sun23} & \xmark & \checkmark & 0.1M &  92.45\%      \\
            & \small DL128-SNN-Dloss \footnotesize \citep{sun23-2} & \xmark & \checkmark & 0.14M &  92.56\%      \\
            & \small Dense Conv Delays (ours)  & \xmark & \checkmark   & 2.7M &  93.44\%      \\
            & \small RadLIF \footnotesize \citep{baseline} & \checkmark & \xmark   & 3.9M &  94.62\%      \\
            & \small \textbf{DCLS-Delays (2L-1KC)}  & \xmark & \checkmark & \textbf{0.2M} &  \textbf{95.07$\pm$0.24\%} \\
        \midrule
            \multirow{3}{4em}{\textbf{SSC}}
            & \small Recurrent SNN \footnotesize \citep{shd} & \checkmark & \xmark & N/a &  50.90 $\pm$ 1.1\%      \\     
            & \small Heter. RSNN \footnotesize \citep{heterogeneity}  & \checkmark & \xmark & N/a &  57.30\% \\
            & \small SNN-CNN \footnotesize \citep{ieee_cnn} & \xmark & \checkmark  & N/a &  72.03\%     \\
            & \small Adaptive SRNN \footnotesize \citep{Adaptive-SRNN} & \checkmark & \xmark  & N/a &  74.20\%     \\
            & \small SpikGRU \footnotesize \citep{spikGRU} & \checkmark & \xmark  & 0.28M &  77.00$\pm$0.4\%     \\
            & \small RadLIF \footnotesize \citep{baseline} & \checkmark & \xmark  & 3.9M &  77.40\%      \\
            & \small Dense Conv Delays 2L (ours)  & \xmark & \checkmark   & 10.9M &  77.86\%      \\
            & \small Dense Conv Delays 3L (ours)  & \xmark & \checkmark   & 19M &  78.44\%      \\
            & \small \textbf{DCLS-Delays (2L-1KC)} & \xmark & \checkmark & \textbf{0.7M} &  \textbf{79.77$\pm$0.09\%} \\
            & \small \textbf{DCLS-Delays (2L-2KC)} & \xmark & \checkmark & \textbf{1.4M} &  \textbf{80.16$\pm$0.09\%} \\
            & \small \textbf{DCLS-Delays (3L-1KC)} & \xmark & \checkmark & \textbf{1.2M} &  \textbf{80.29$\pm$0.06\%} \\
            & \small \textbf{DCLS-Delays (3L-2KC)} & \xmark & \checkmark & \textbf{2.5M} &  \textbf{80.69$\pm$0.21\%} \\
        \midrule
            \multirow{3}{4em}{\textbf{GSC-35}}
            & \small MSAT \footnotesize \citep{msat} & \xmark & \xmark & N/a &  87.33\%      \\
            & \small Dense Conv Delays 2L (ours)  & \xmark & \checkmark   & 10.9M &  92.97\%      \\
            & \small Dense Conv Delays 3L (ours)  & \xmark & \checkmark   & 19M &  93.19\%      \\
            & \small RadLIF \footnotesize \citep{baseline} & \checkmark & \xmark  & 1.2M &  94.51\%      \\
            & \small \textbf{DCLS-Delays (2L-1KC)} & \xmark & \checkmark & \textbf{0.7M} &  \textbf{94.91$\pm$0.09\%} \\
            & \small \textbf{DCLS-Delays (2L-2KC)} & \xmark & \checkmark & \textbf{1.4M} &  \textbf{95.00$\pm$0.06\%} \\
            & \small \textbf{DCLS-Delays (3L-1KC)} & \xmark & \checkmark & \textbf{1.2M} &  \textbf{95.29$\pm$0.11\%} \\
            & \small \textbf{DCLS-Delays (3L-2KC)} & \xmark & \checkmark & \textbf{2.5M} &  \textbf{95.35$\pm$0.04\%} \\
        \bottomrule
  \end{tabular}
  }
  {\raggedright\footnotesize nL-mKC stands for a model with n hidden layers and kernel count m, where kernel count denotes the number of non-zero elements in the kernel. ``Rec.'' denotes recurrent connections. \par}
\end{table}

Our method outperforms the previous state-of-the-art accuracy on the three benchmarks (with a significant improvement on SSC and GSC) without using recurrent connections (apart from the self-recurrent connection of the LIF neuron), with a substantially lower number of parameters, and using only vanilla LIF neurons. Other methods that use delays do have a slightly lower number of parameters than we do, yet we outperform them significantly on SHD, while they didn't report any results on the harder benchmarks SSC/GSC. Finally, by increasing the number of hidden layers, we found that the accuracy plateaued after two hidden layers for SHD and three for SSC/GSC.
Furthermore, we also evaluated a model (Dense Conv Delay) that uses standard dense convolutions instead of the DCLS ones. This corresponds conceptually to having a fully connected SNN with all possible delay values as multiple synaptic connections between every pair of neurons in successive layers. This led to worse accuracy (partly due to overfitting) than DCLS. The fact that DCLS outperforms a standard dense convolution, although DCLS is more constrained and has fewer parameters, is remarkable.

\subsection{Ablation study}
\label{sec:ablation}

In this section, we conduct control experiments aimed at assessing the effectiveness of our delay learning method. The model trained using our full method will be referred to as \emph{Decreasing $\sigma$} (specifically, we use the 2L-1KC version), while \emph{Constant $\sigma$} will refer to a model where the standard deviation $\sigma$ is constant and equal to the minimum value of $0.5$ throughout the training. Additionally, \emph{Fixed random delays} will refer to a model where delays are initialized randomly and not learned, while only weights are learned. Meanwhile, \emph{Decreasing $\sigma$ - Fixed weights} will refer to a model where the weights are fixed and only delays are learned with a decreasing $\sigma$. Finally, \emph{No delays} denotes a standard SNN without delays. To ensure equal parameter counts across all models (for fair comparison), we increased the number of hidden neurons in the \emph{No delays - wider} case, and increased the number of layers instead in the \emph{No delays - deeper} case. Moreover, to make the comparison even fairer, all models have the same initialization for weights and, if required, the same initialization for delays.

\begin{figure}[!ht]
  \centering
  \begin{subfigure}[b]{0.43\textwidth}
    \includegraphics[width=\textwidth]{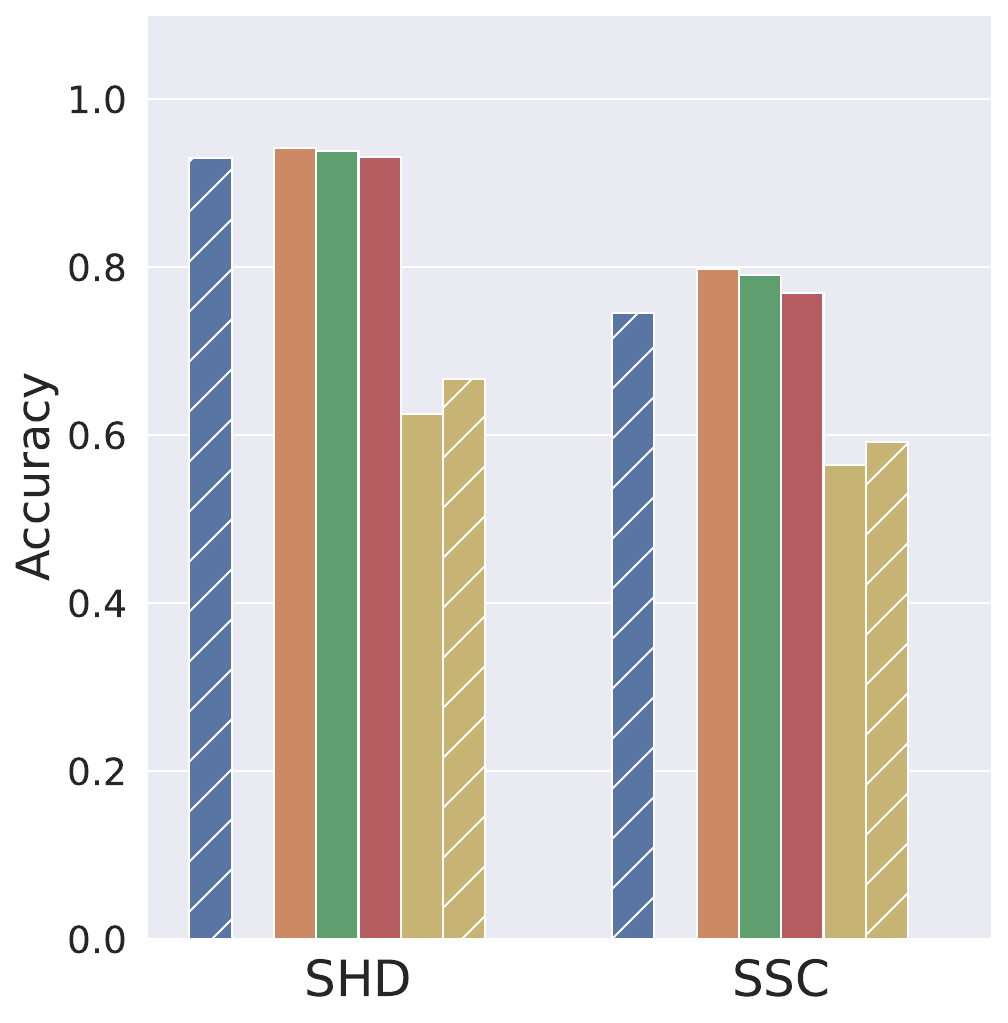}
    \caption{FC: Fully Connected}
    \label{fig:fc}
  \end{subfigure}
  \begin{subfigure}[b]{0.43\textwidth}
    \includegraphics[width=\textwidth]{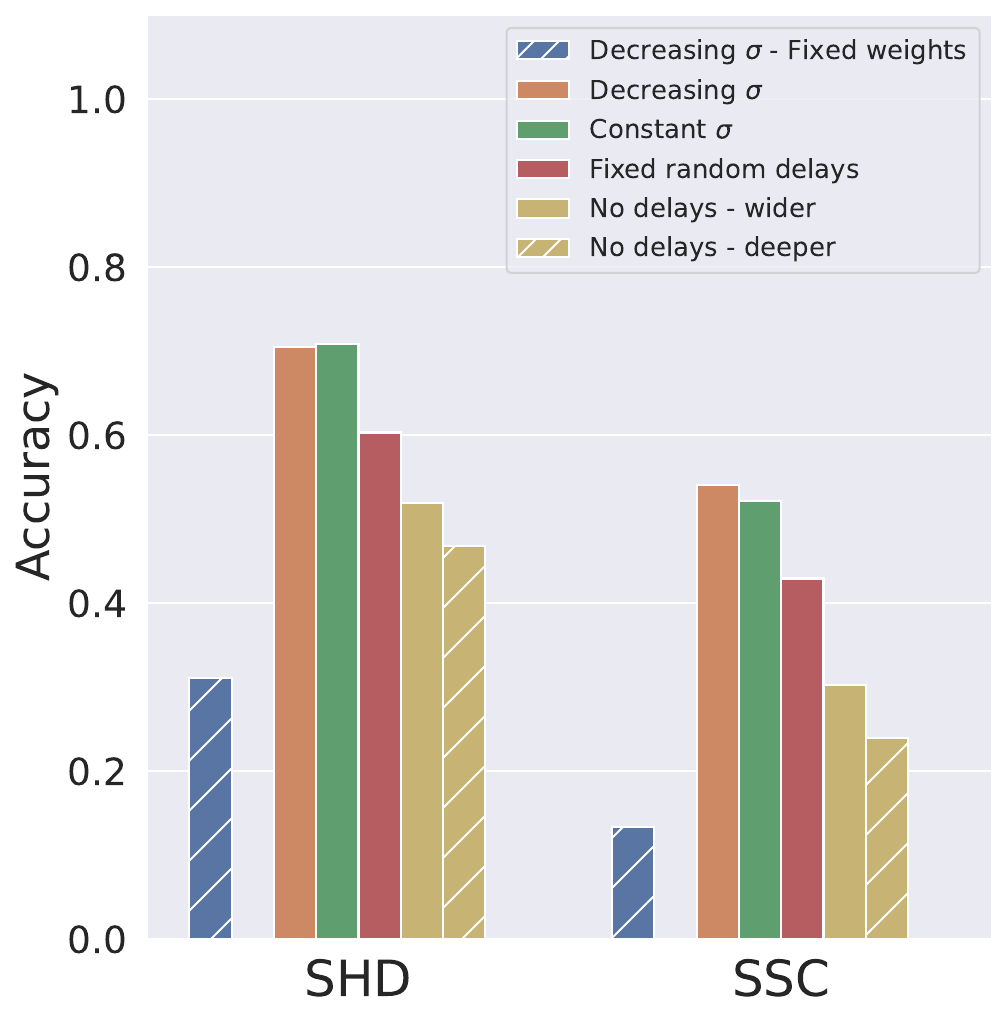}
    \caption{S: Sparse connections}
    \label{fig:S}
  \end{subfigure}
  \caption{Barplots of test accuracies on SHD and SSC datasets for different models. With (a): fully connected layers (FC) and (b): sparse synaptic connections (S). Reducing the number of synaptic connections of each neuron to ten for both SHD and SSC. }
  \label{fig:barplots}
\end{figure}
\label{random_init2}

We compared the five different models as shown in Figure \ref{fig:fc}. The models with delays (whether fixed or learned) significantly outperformed the \textit{No delays} model both on SHD (FC) and SSC (FC); for us, this was an expected outcome given the temporal nature of these benchmarks, as achieving a high accuracy necessitates learning long temporal dependencies. However, we didn't expect the Fixed random delays model to be almost on par with models where delays were trained, with Decreasing $\sigma$ model only slightly outperforming it. 

To explain this, we hypothesized that a random uniformly distributed set of delay positions would likely cover the whole temporal range. This hypothesis is plausible given the fact that the number of synaptic connections vastly outnumbers the total possible discrete delay positions for each kernel. Therefore, as the number of synaptic connections within a layer grows, the necessity of moving delay positions away from their initial state diminishes. Only tuning the weights of this set of fixed delays is enough to achieve comparable performance to delay learning.

In order to validate this hypothesis, we conducted a comparison using the same models with a significantly reduced number of synaptic connections. We applied fixed binary masks to the network's synaptic weight parameters. Specifically, for each neuron in the network, we reduced the number of its synaptic connections to ten for both datasets (except for the No delays model, which has more connections to ensure equal parameter counts). This corresponds to 96\% sparsity for SHD and 98\% sparsity for SSC. 
With the number of synaptic connections reduced, it is unlikely that the random uniform initialization of delay positions will cover most of the temporal range. Thus, specific long-term dependencies will need to be learned by moving the delays. 

The test accuracies corresponding to this control test are shown in Figure \ref{fig:S}. It illustrates the difference in performance between the Fixed random delays model and the Decreasing/Constant $\sigma$ models in the sparse case. This enforces our hypothesis and shows the need to perform this control test for delay learning methods. Furthermore, it also indicates the effectiveness of our method.

In addition, we also tested a model where only the delays are learned while the synaptic weights are fixed (Decreasing $\sigma$ - Fixed weights). It can be seen that learning only the delays gives acceptable results in the fully connected case (in agreement with \cite{beyondweights}) but not in the sparse case. To summarize, it is always preferable to learn both weights and delays (and decreasing $\sigma$ helps). If one has to choose, then learning weights is preferable, especially with sparse connectivity.

\chapter{Audio classification with Dilated Convolution with Learnable Spacings}
\section{Disclaimer}
This chapter is strongly inspired by the article: \cite{khalfaoui2023audio}. Most parts of this chapter have been taken verbatim from the paper of which we are the principal author, with the content and wording largely our own. 


\section{Introduction}

The very popular ConvNeXt model \cite{liu2022convnet}, a fully convolutional model designed for vision tasks, has been successfully adapted to audio classification on AudioSet \cite{pellegrini2023adapting} by transforming audio samples to log-mel spectrograms and adapting the stem of the ConvNeXt model to fit the input audio extracts. This has improved the state of the art of audio classification using convolutional neural networks by achieving better accuracy than PANN-type models \cite{Kong2020}, while having fewer learnable parameters. Furthermore, when used as a backbone for downstream tasks, the ConvNeXt-audio model has achieved positive, if not state-of-the-art, results for the audio captioning and audio retrieval tasks. 

Separately, the Dilated Convolution with Learnable Spacings (DCLS) method has already proven itself in several computer vision tasks \cite{hassani2023dilated}. 
Through a simple drop-in replacement of the model's Depthwise Separable Convolutions (DSC) with DCLS (which can be done automatically for all layers of a model via this script \ref{sec:appendixb}), the DCLS convolution method has empirically proven its effectiveness for several computer vision tasks using ImagNet1k \cite{deng2009imagenet} trained models as backbones. This resulted in a ConvNeXt-dcls model \cite{hassani2023dilated} and a ConvFormer-dcls model \cite{khalfaouihassani2023dilated}, depending on the model chosen in the study, by performing the replacement in the ConvNeXt and ConvFormer models, respectively.

Our aim in the present chapter is to show empirically that a drop-in replacement of the DCLS method in the same fully convolutional models can improve their accuracy for the task of audio classification on the AudioSet dataset without much effort, demonstrating once again the interest of the method not only on the reference benchmark for image classification but also on the reference benchmark for audio classification (AudioSet \cite{AudioSet}). Furthermore, we add a third test model that differs slightly from the other two in that it's a hybrid model (having both DSC layers and multi-head self-attention layers, depending on the stage to which the layer belongs): FastVit \cite{vasu2023fastvit}, and again, replacing the DSC layers by DCLS improves results. 

This chapter does not claim to be the absolute state of the art on the task of classification on AudioSet, but rather tries to provide an objective comparison between known and proven convolutional models and those equipped with a DCLS convolution that would make them more efficient.

\section{Related work}

Audio tagging systems were mainly based on convolutional neural networks until recently, with the adaptation of vision transformers to audio processing. The PANN-based models (\textit{e.g.}, CNN14), in particular, comprise blocks of plain $3 \times 3$ kernel convolution layers~\cite{Kong2020}. In~\cite{verbitskiy2022eranns}, PANN-like models were enhanced, in terms of accuracy, model size and inference speed, by adding residual connections, and by modifying the kernel sizes, the stride and padding, using a ``decreasing temporal size parameter''. Other efficient CNN architectures, such as EfficientNet~\cite{gong2021psla}, were also tested in audio tagging. In~\cite{singh2023panns}, efficient PANNs (E-PANN) were obtained by using filter pruning.  In~\cite{drossos2020sound}, DSC layers were used, which resulted in large reductions in model complexity, together with performance gains. In~\cite{pellegrini2023adapting}, doing so in PANN's CNN14 also yielded significant model size reduction (about 60\% relative), whilst observing a gain in performance. In this last study, ConvNeXt was adapted to perform the audio tagging task in AudioSet. It performed better or on par with the transformer-based architectures AST~\cite{gong2021ast} and PaSST-S~\cite{koutini2021efficient}.

\section{Methods}

\subsection{Dataset and configuration}
\textbf{Dataset.} 
In all the experiments in this chapter, we used AudioSet \cite{AudioSet}, the reference dataset in audio classification. It contains about 2 million video clips downloaded from the YouTube platform. We are only interested in the audio portion of these clips and are not using the video dataset. The audio clips available in AudioSet can vary in size, but most are 10 seconds long. If a sample is longer than that, we truncate it; if shorter, we pad it with zeros. The classification task in AudioSet consists of assigning each sample to the class or classes to which it belongs among the 527 available labels. It is thus a multi-label classification task. The majority of the excerpts correspond to one of the two classes "speech" and "music" (often both), due to their predominance on the aforementioned video hosting site. This latter fact leads to an imbalance in the dataset, with several classes being poorly represented while a few classes account for most of the dataset. 
We downloaded the data in 2018, and some of the YouTube links have been broken since then. Our AudioSet data contains 1,921,982 clips (unbalanced train), 21,022 clips (balanced train), and 19,393 clips (evaluation).

\textbf{Metrics.} \label{sec:metrics} We report the usual evaluation metric for AudioSet tagging: mean average precision (mAP) 
which is typically the metric of interest in audio tagging. 
All DCLS-equipped models studied here outperform their respective baselines using this metric.

\textbf{No weighted sampler.} Given the unbalanced nature of the dataset, many state-of-the-art models make good use of a weighted random sampler \cite{Kong2020, gong2022contrastive, huang2022mavil}, where each class in the dataset is weighted by its frequency of occurrence in the dataset. This is a classic machine learning approach to mitigate data imbalance. However, as pointed out by \cite{moore2023dataset}, these approaches based on a weighted sampler whose oversampling rate is adjusted as a training hyperparameter seem to overfit the dataset more than anything else and do not favor the rarest classes. Since in this chapter, we are only interested in the comparative study between baseline models and the same models augmented with the DCLS method, we have chosen not to include weighted samplers in our training phases, even if this means losing a few points in mAP, thereby allowing a comparison that is less noisy due to the effects of sampling. Furthermore, the naive use of Mixup augmentation \cite{zhang2018mixup} in conjunction with a weighted sampler may turn out to be a source of undesirable behavior, since proceeding in this way could destroy the weighted sampling that was originally intended, as the Mixup acts randomly by drawing two samples without taking balancing into account. Weighting-aware approaches to Mixup such as \cite{Ramasubramanian2023selmix} should be better investigated and implemented in order to better take advantage of both methods.

\textbf{Spectrogram resolution.} Many audio classification models use raw audio signals \cite{oord2016wavenet, dai2017very, akbari2021vatt}, while a growing number of state-of-the-art models use spectrograms, taking advantage of the signal's periodic aspect by using the Short-time Fourier transform \cite{Kong2020, koutini2021efficient, huang2022mavil}. We prefer this second choice in order to use computer vision baselines. Additionally, the obtained spectrograms are often filtered using the mel psychoacoustic scale \cite{stevens1937scale}. We use the latter filtering to obtain mel-frequency spectrograms, which we transform from the power/amplitude scale to the decibel scale. A comprehensive enumeration of the hyperparameters used to perform these transformations is given in Table~\ref{chap5_tab:2}.

The final spectrogram size obtained is ($F=128$, $T=1001$). In the course of our experiments, we noticed that the larger the size of the spectrograms (in both frequency and time), the greater the mAP of the models, but to the detriment of their throughput. This is a well-known phenomenon in computer vision, where higher input image resolution often leads to better vision model accuracy but also to higher computational time costs. We believe that this resolution provides a good trade-off in mAP-throughput and argue that there is a sweet spot between resolution and stem size that offers optimal performance regarding mAP-throughput.

\textbf{Adapting the stem.} The three neural networks used in this study all come from the world of computer vision. It is therefore essential to adapt the stem of these models in order to process no longer natural images made up of three channels corresponding to the RGB colors, but instead a spectrogram with a single channel and a size different from the crop images initially designed for vision tasks. To this end, we used a basic stem, common to all three studied models, namely a convolution layer with a kernel size = (2, 16) and a stride = (2, 16). This stem produces maps of size (64, 62) from input spectrograms of size (128, 1001). This type of stem is similar to that originally found in the ConvNeXt model, while the ConvFormer model featured a slightly more sophisticated stem with a kernel size larger than the stride size. The FastVit model, on the other hand, came with a much more complex stem consisting of several layers based on the MobileOne block \cite{vasu2023mobileone}. 

Imposing a common stem on all the models in the study means that, on the one hand, we can compare the models more accurately, knowing that the input resolution will be the same for all of them. On the other hand, in the absence of an optimal stem adapted to audio spectrograms, we use a coarse stem with which we can conduct our study. Note that the search for an ideal stem is a study in itself and that the stem presented here can always be refined and improved.

\textbf{Pretraining on ImageNet1k.} Using pre-trained models on ImageNet \cite{deng2009imagenet} as a better initialization to solve the tagging task on AudioSet is common practice~\cite{gong2021ast,koutini2021efficient,pellegrini2023adapting}. In the first few epochs, models initialized in this way have a clear advantage over those initialized randomly. However, this advantage is quickly regained over the course of training, and randomly initialized models often end up performing similarly to or slightly worse than pre-trained ones. We only use pre-trained models on ImageNet1k when they are available and do not cost us anything to train. Therefore, we use the symbol $\ddag$ to designate models that have not been pre-trained on ImageNet.

\textbf{Configuration.} In Table~\ref{chap5_tab:2} is a comprehensive list of the hyperparameters and augmentations used in this study. These are largely similar to the hyperparameters used in \cite{liu2022convnet}. Note that a high drop path (0.4) is used in this work to overcome the overfitting problem encountered with the tagging task on AudioSet and that large effective batch sizes were used (4096) to speed up training. However, some instabilities during training were noted, particularly for the ConvFormer model. These instabilities are known from \cite{yu2022metaformer} and were resolved by using the LAMB optimizer \cite{you2019large}, while we used AdamW \cite{adamw} for the other two models. 

\subsection{Models}

\label{sec:models}
We used three different models from computer vision that we adapted to audio inputs to corroborate our results. The first one is a fully convolutional model: ConvNeXt-tiny \cite{liu2022convnet}. The second one is also a purely convolutional model that outperforms the ConvNext model on the ImageNet classification task: ConvFormer-S18 \cite{yu2022metaformer}. The DCLS method as a replacement for DSC has already been successfully used in the latter two models for various vision tasks, including image classification on ImageNet1k \cite{hassani2023dilated, khalfaouihassani2023dilated}. The third model we used is more recent and achieves the current state-of-the-art throughput-accuracy trade-off in image classification on the ImageNet dataset: FastVit-SA24 \cite{vasu2023fastvit}. The latter is a so-called hybrid model, i.e., it contains DSC layers as well as and multi-head self-attention layers. 

\subsection{DCLS substitution}

Considering the baseline models discussed in the previous section, we carried out the following study: we trained the baselines on the concatenation of the unbalanced train and the balanced train sets of AudioSet, then evaluated them on the evaluation subset. We repeated the same process with the same models, except that this time we replaced all DSC layers having a kernel size equal to 7 with a DCLS convolution layer. In all test cases, we used exactly the same training configuration to avoid attributing performance gains to any reason other than the replacement of the DSC layers by DCLS ones. Also, to learn the positions (and standard deviations for DCLS-Gauss) of each kernel element, we followed the same training techniques as those listed in \cite{khalfaouihassani2023dilated}. This gave us 6 test cases to examine in total, for which we measure the mAP metric mentioned in Section~\ref{sec:metrics} averaged over three different seeds (seeds 0, 1, and 2). 

\section{Results}
\begin{table*}[!htbp]
\begin{center}
\resizebox{\textwidth}{!}{
$
\begin{array}{lccccc}
\toprule
\text {model}
&\begin{array}{l}
\text { ker. size } \\
\text { / count  }
\end{array} & \text {method}  & \text { \# param.} & \text { mAP }  & \begin{array}{l}
\text { throughput } \\
\text { (sample / s) }
\end{array} \\
\hline
\text {CNN14 \cite{Kong2020}} &  & \text{Conv.} & 80.7\mathrm{M} & 43.1 & 378.2 \\
\text {PaSST-S \cite{koutini2021efficient}} &  & \text{MHS. Attention.} & 87\mathrm{M} & 47.1 & 88.7 \\
\text {ConvNeXt-T \cite{pellegrini2023adapting}} & 7^2 \ / \ 49 & \text{Depth. Conv.} & 28.2\mathrm{M} & 47.1 & 153.6 \\
\hline 
\text {ConvFormer-S18}^\dag & 7^2 \ / \ 49 & \text{Depth. Conv.} & 26.8 \mathrm{M} & 43.14 \pm 0.03 & 513.3 \\
\rowcolor{lightcream}\text {ConvFormer-S18}^\dag & 23^2  \ / \  26 & \text{DCLS-Gauss} & 26.8 \mathrm{M} & 43.68 \pm 0.02 & 396.8 \\

\text {FastVIT-SA24}^\ddag & 7^2 \ / \ 49 & \text{Depth. Conv.} & \textbf{21.5} \mathrm{M} & 43.82 \pm 0.05 & \textbf{633.6}\\
\rowcolor{lightcream}\text {FastVIT-SA24}^\ddag  & 23^2  \ / \  26 & \text{DCLS-Gauss} & \textbf{21.5} \mathrm{M} & 44.4 \pm 0.07 & 551.7 \\

\text {ConvNeXt-T } & 7^2 \ / \ 49 & \text{Depth. Conv.} & 28.6 \mathrm{M} & 	44.83 \pm 0.14  & 591.4 \\
\rowcolor{lightcream}\text {ConvNeXt-T } & 23^2  \ / \  26 & \text{DCLS-Gauss} & 28.6 \mathrm{M} & \textbf{45.52} \pm 0.05  & 509.4 
\end{array}
$
}
\end{center}
\caption{\textbf{Classification mean average precision (mAP) on the evaluation set of AudioSet.} For the baselines using DSC and the DCLS-Gaussian cases, the results have been averaged over 3 distinct seeds and presented in the format mean $\pm$ standard deviation. $\dag:$ trained using LAMB, $\ddag:$ no ImageNet pretraining. The throughputs were calculated with a single NIVIDIA V100 32-GB gpu.}
\label{chap5_tab:1}
\end{table*}

The results presented in Table~\ref{chap5_tab:1} demonstrate the performance of the three models mentioned in Section \ref{sec:models} on the $128 \times 1001$ spectrograms, where the convolution method used varies. Notably, we observe that 
when comparing each baseline model with its DCLS-equipped counterpart, the use of DCLS-Gauss with a kernel size of $23^2$ and a kernel count of $26$ stands out, achieving a higher mAP (+0.6 on average) with an equal or lower number of parameters. This result highlights the effectiveness of DCLS-Gauss in enhancing classification performance. 
DCLS does, however, introduce a reduction in throughput ($13\%$ for ConvNeXt-T and FastVit-SA24 and $23\%$ for ConvFormer-S18) due to the use of larger kernels. The results of a previous study on ConvNeXt~\cite{pellegrini2023adapting} show that an mAP of $47.1$ can be achieved, but here we only reach $44.8$ for the baseline; this is since in that previous study, a higher spectrogram resolution was used ($224 \times 1001$ versus $128 \times 1001$ in this work) and that a stem size of $4 \times 4$ instead of $2 \times 16$ here was used to produce larger feature maps, which is reflected both in the large memory required to run this model and in the model's throughput.

\section{Appendix: How to replace all model's DSC by DCLS ones?}
\label{sec:appendixb}
\begin{lstlisting}[language=Python]
import copy
from torch import nn
from DCLS.construct.modules import Dcls2d

# Helper function that replaces all ".int." patterns
# by "[int]" in a character string
def replace_dots_brackets(name):
    name_split = name.split(".")
    name_split = [
        "[" + i + "]" if i.isdigit() else "." + i for i in name_split
    ]
    return "".join(name_split[:-1]), name_split[-1][1:]

# Helper function that replaces all the
# 2D depthwise separable convolution in
# a model by synchronized Dcls2d ones
def replace_depthwise_dcls(
    model, dilated_kernel_size=23, kernel_count=26, version="gauss"):
    in_channels, P, SIG = 0, None, None
    # Loop over all model modules
    for name, module in model.named_modules():
        # if the module is a depthwise separable Conv2d module
        if (isinstance(module, nn.Conv2d)
            and module.groups == module.in_channels == module.out_channels
            and module.kernel_size == (7, 7)
        ):
            name_eval, last_layer = replace_dots_brackets(name)
            dcls_conv = Dcls2d(
                module.in_channels,
                module.out_channels,
                kernel_count=kernel_count,
                stride=module.stride,
                dilated_kernel_size=dilated_kernel_size,
                padding=dilated_kernel_size // 2,
                groups=module.in_channels,
                version=version,
                bias=module.bias is not None,
            )
            nn.init.normal_(dcls_conv.weight, std=0.02)
            if module.bias is not None:
                nn.init.constant_(dcls_conv.bias, 0)

            # Synchronise positions and standard
            # deviations belonging to the same stage
            if in_channels < module.in_channels:
                in_channels = module.in_channels
                P, SIG = dcls_conv.P, dcls_conv.SIG

            dcls_conv.P, dcls_conv.SIG = P, SIG
            setattr(eval("model" + name_eval), last_layer, dcls_conv)
    return model

model = nn.Conv2d(96, 96, 7, padding=3, groups=96)
# Replace all the 2D depthwise separable convolutions
# in the model by synchronized Dcls2d ones.
model = replace_depthwise_dcls(
    copy.deepcopy(model),
    dilated_kernel_size=23,
    kernel_count=26,
    version="gauss",
)
print(model)
\end{lstlisting}

\section{Appendix: Training hyper-parameters}

\begin{table*}[!htbp]
\begin{center}
\resizebox{0.8\textwidth}{!}{
$
\begin{array}{l | c}
\toprule
\text{Configuration} & \text{AudioSet2M} \\ 
\hline \\
 \text{Optimizer} & \text{AdamW \cite{adamw}} \quad | \quad \text{LAMB \cite{you2019large}}^{\dag} \\ 
 \text{Optimizer momentum} & \beta_1=0.9, \beta_2=0.999 \\
 \text{Weight decay} & 0.05 \\
 \text{Base learning rate} & 4e-3 \\
 \text{Learning rate schedule} & \text{half-cycle cosine decay \cite{loshchilov2016sgdr}} \\
 \text{Gradient clipping} & \text{None} \\
 \text{Epochs} & 60 \\
 \text{Warm-up epochs} & 20 \\
 \text{Batch size} & 4096 \\
 \text{GPUs size} & 32 \\
 \text{Weighted sampling} & \text{False} \\
 \text{Drop path \cite{larsson2016fractalnet}} & 0.4 \\
 \text{Mixup \cite{zhang2018mixup}} & 0.8 \\
 \text{Multilabel} & \text{True} \\
  \text{Label smoothing \cite{szegedy2016rethinking}} & 0.1 \\
 \text{Loss Function} & \text{Binary Cross-Entropy} \\
 \text{Dataset Mean for Normalization} & -18.2696 \\
 \text{Dataset Std for Normalization} & 30.5735 \\
 \hline
 \multicolumn{2}{c}{\text{Spectrogram configuration}}\\\\
 \text{Number of fft} & 1024 \\
 \text{Hop length} & 320 \\
 \text{Power} & 2 \\
  \hline
 \multicolumn{2}{c}{\text{Mel scale configuration}}\\\\
 \text{Number of mels} & 128 \\
 \text{Sample rate} & 32 ~ 000 \\
 \text{Min frequency} & 50 \\
 \text{Max frequency} & 14 ~ 000 \\
 \text{Amplitude to dB} & \text{True} \\ 
  \hline
 \multicolumn{2}{c}{\text{Augmentations}}\\\\
 \text{PadOrTruncate} & 10 \times \text{sample rate} \\
 \text{RandomRoll} & [-\text{sample size}, \text{sample size}], \text{p} = 1 \\ 
 \text{SpeedPerturbation \cite{ko15_interspeech}} & \text{rates} = (0.5, 1.5),  \text{p} = 0.5 \\
 \text{RandomErasing \cite{zhong2020random}} & \text{p} = 0.25 \\

\end{array}
$
}
\end{center}
\caption{\textbf{Training hyper-parameters.}}
\label{chap5_tab:2}
\end{table*}
\chapter{Discussion and conclusion}


\section{Large kernels become essential}
\label{sec:kernel_size}


If there were only one message to \textbf{take away} from this thesis, it would be that convolutions with large kernels are important in computer vision and computer audition in order to compete with transformers. Increasing the kernel size in standard convolution is useful, but comes at a cost in terms of performance and learnable parameters. A good way to re-parameterize large kernel convolutions is DCLS, where we can define a budget of kernel elements and then have them find their optimal positions by backpropagation within a predefined limit. 

The importance of large kernels has been known for a long time \cite{peng2017large}, and has recently been the subject of more interest in the field of CNNs \cite{ding2022scaling, liu2023more}. According to \cite{ding2023unireplknet}, large kernels would be the key to unlocking the performance of CNNs in domains where they were originally not proficient, perhaps even a way to rival multi-modal transformers.

\begin{figure*}[!htbp]
     \centering
     \subfloat[][10 epochs run (underfitting)]{\includegraphics[width=1.1\textwidth]{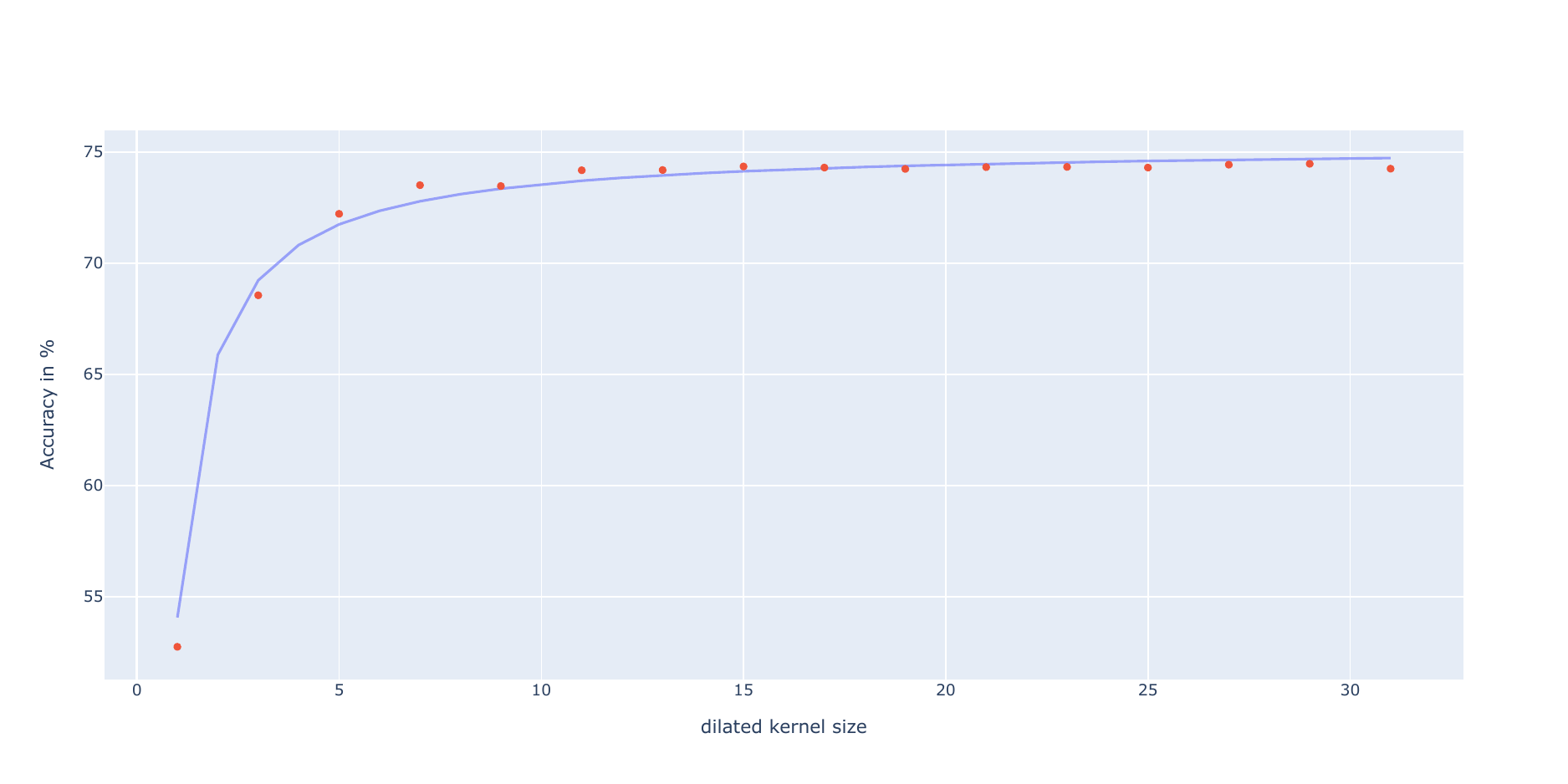}}
          \hfill
     \subfloat[][100 epochs run (overfitting)]{\includegraphics[width=1.1\textwidth]{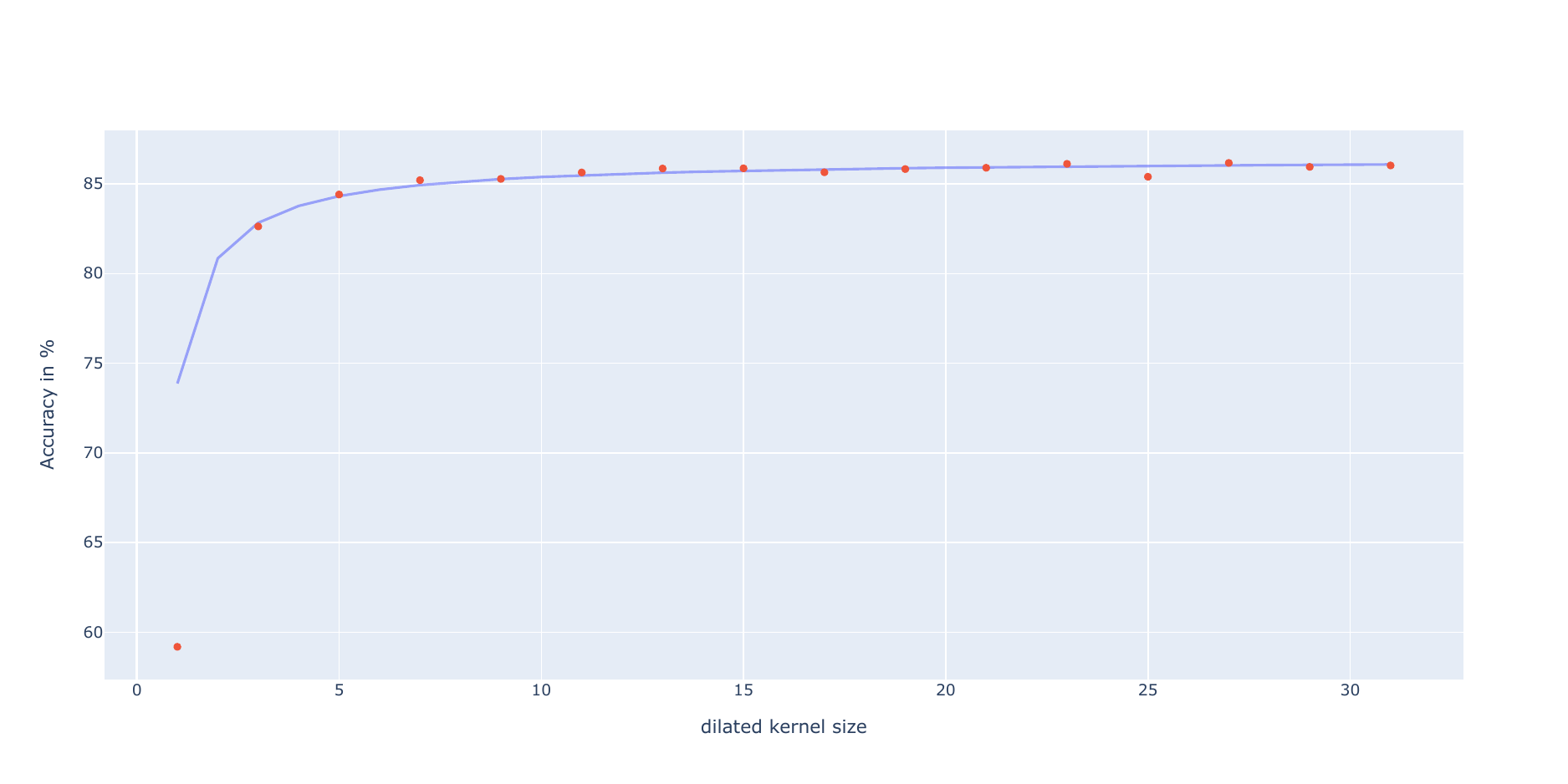}}
    \caption{Top1 accuracy as a function of dilated kernel size for a one-stage DCLS-Gauss model. {\color{blue} \rule[0.1cm]{0.5cm}{0.4pt} }: $x \mapsto \alpha + \text{ln}(e - 1/x) + \beta$ a logarithmic function that fits the experimental data {\color{red} $\bullet$ } with $\alpha = 46.23, \ \beta = 29.05$ in the underfitting case and $\alpha =27.34, \ \beta = 59.07$ in the overfitting case. $e = 2.71828...$}
    \label{fig:kernel_size_effect}    
\end{figure*}

To illustrate the effect of the large dilated kernel size in the DCLS method, we have conducted the following study for a single-label classification task on CIFAR-10 images \cite{krizhevsky2009learning}. We have taken a network consisting of a stem which is a strided convolution with a kernel of size $2$ and a stride of $2$, followed by three DCLS-Gauss convolutions that are identical in terms of hyper-parameters (this network is identical to the stem + the first stage of the  ConvNeXt-T-DCLS-Gauss model). We carried out this study twice: first in an underfitting regime by stopping the training after only 10 epochs, then a second time after 100 epochs (an overfitting regime, where train accuracy reached $100\%$ and the test loss started increasing). The results are in Figure \ref{fig:kernel_size_effect}.

From the figure \ref{fig:kernel_size_effect}, we can see a similar trend for both underfitting and overfitting regimes, i.e. as we increase the dilated kernel size of the DCLS layers, accuracy increases. This increase is logarithmic, so as the dilated kernel size increases, the gain in accuracy becomes more and more minimal. We have modeled this curve (accuracy as a function of dilated kernel size) using the following function $x \mapsto \alpha + \text{ln}(e - 1/x) + \beta$ with $\alpha = 46.23, \ \beta = 29.05$ in the underfitting case and $\alpha =27.34, \ \beta = 59.07$ in the overfitting case.

We have limited ourselves to odd dilated kernel sizes between $1$ and $31$.
Once a dilated kernel size of $7$ is reached, we can say that increasing the dilated kernel size is not cost-effective, since we achieve $99\%$ of the accuracy we could have expected with a dilated kernel size of 31, our upper bound.

This observation is matched by the calculation of the receptive field \ref{eq:rf}. With a dilated kernel size of $7$, the receptive field of the last layer of this one-stage network is $38$, whereas if we change the dilated kernel size to $5$, the receptive field of the last layer of the network becomes $26$. Given that the CIFAR-10 images are of size $32 \times 32$, this could explain why the accuracy starts to saturate at a dilated kernel size of $7$, since the last layer has a receptive field larger than the entire input images.

\section{DCLS and 2D Brownian motion}

In this section, we hypothesize that the advantage of DCLS over standard and dilated convolution may be due to the mere motion of the elements, even if this motion is random as opposed to the motion obtained by gradient descent. Therefore, we compare the convergence of DCLS (which uses gradient descent) to a 2D Brownian motion of the kernel elements.

Brownian motion is mathematically described by the Wiener process \cite{bass2011stochastic}. When compared to a kernel where all elements move following a 2D Brownian motion of standard deviation equal to one and while considering each gradient step as a step of the Wiener process, DCLS still performs way better than the random Brownian motion. This ablation proves in addition to the one where positions are randomly initialized and not learned \ref{random_init} and \ref{random_init2},  that the DCLS method does indeed work and that learning the positions of the kernel elements is significant compared to the chance level, whether in the initial state or even randomly during learning as illustrated by this comparison with Brownian motion.

Another way to confirm the last is to observe that the histograms of the positions obtained at the end of a learning process using the DCLS method \ref{convergence_positions} have a particular structure that is very different from the histograms obtained for a random motion such as Brownian motion. 

\section{Positions of convergence}
\label{convergence_positions}

Depending on the interpolation method used, the 2D-DCLS method can exhibit different convergence patterns for positions within the kernel. In the following, we'll show that these patterns are independent of the data (as we'll show that they are more or less the same for image data on ImagNet1k and audio spectrograms on AudioSet) and that they depend more on the interpolation used in the DCLS method (bilinear or Gaussian).

\begin{figure}[!htbp]
    \centering
    \includegraphics[width=1.1\textwidth]{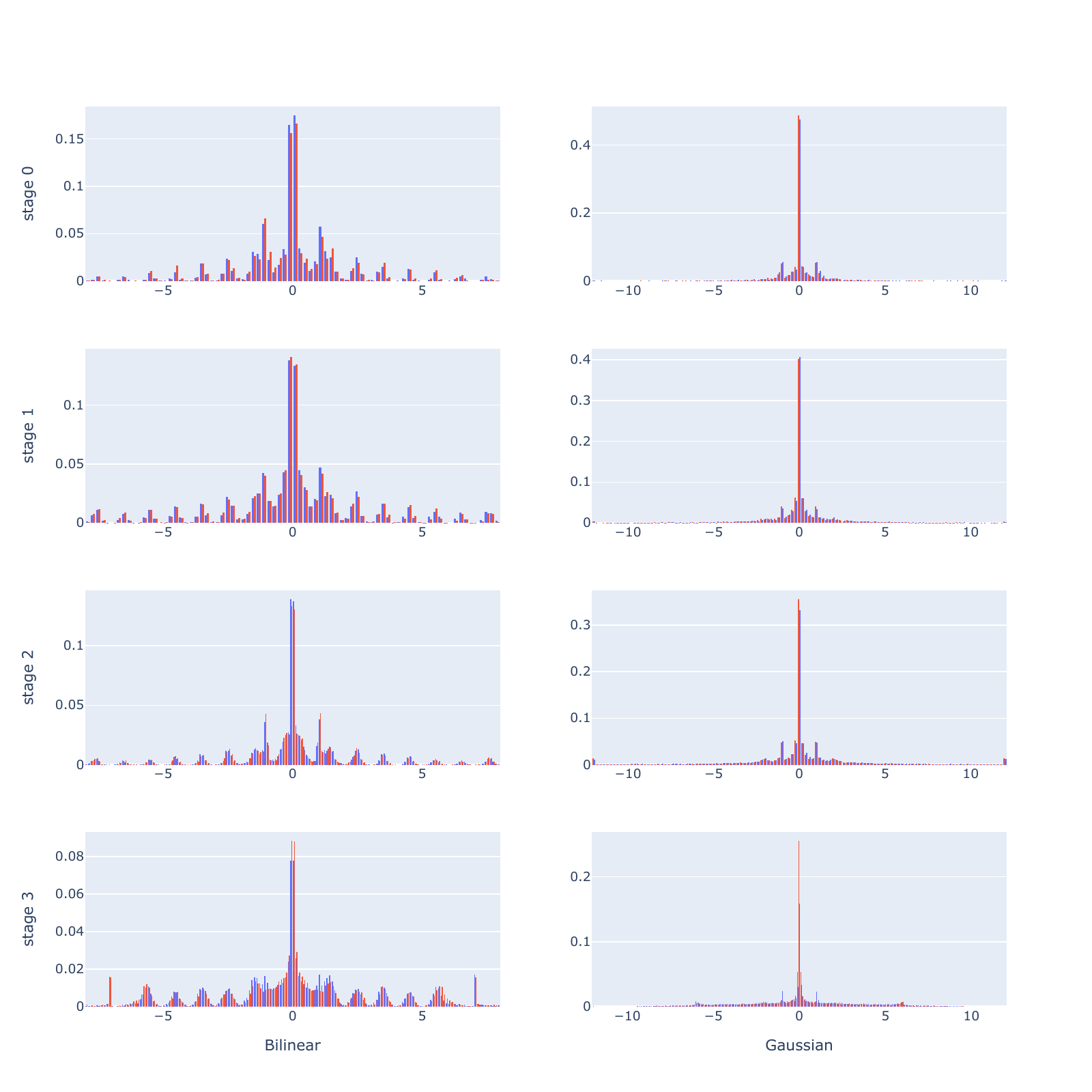}
    \caption{The distribution of the positions {\color{blue} \rule[0.1cm]{0.5cm}{0.4pt} }: along the x-axis, {\color{red} \rule[0.1cm]{0.5cm}{0.4pt} }: along the y-axis within the dilated kernel of a ConvNeXt-DCLS-Bilinear and a ConvNeXt-DCLS-Gauss models trained on ImageNet1k. The histograms are grouped by stages.}
    \label{fig:hist_images}    
\end{figure}

\begin{figure}[!htbp]
    \centering
    \includegraphics[width=1.1\textwidth]{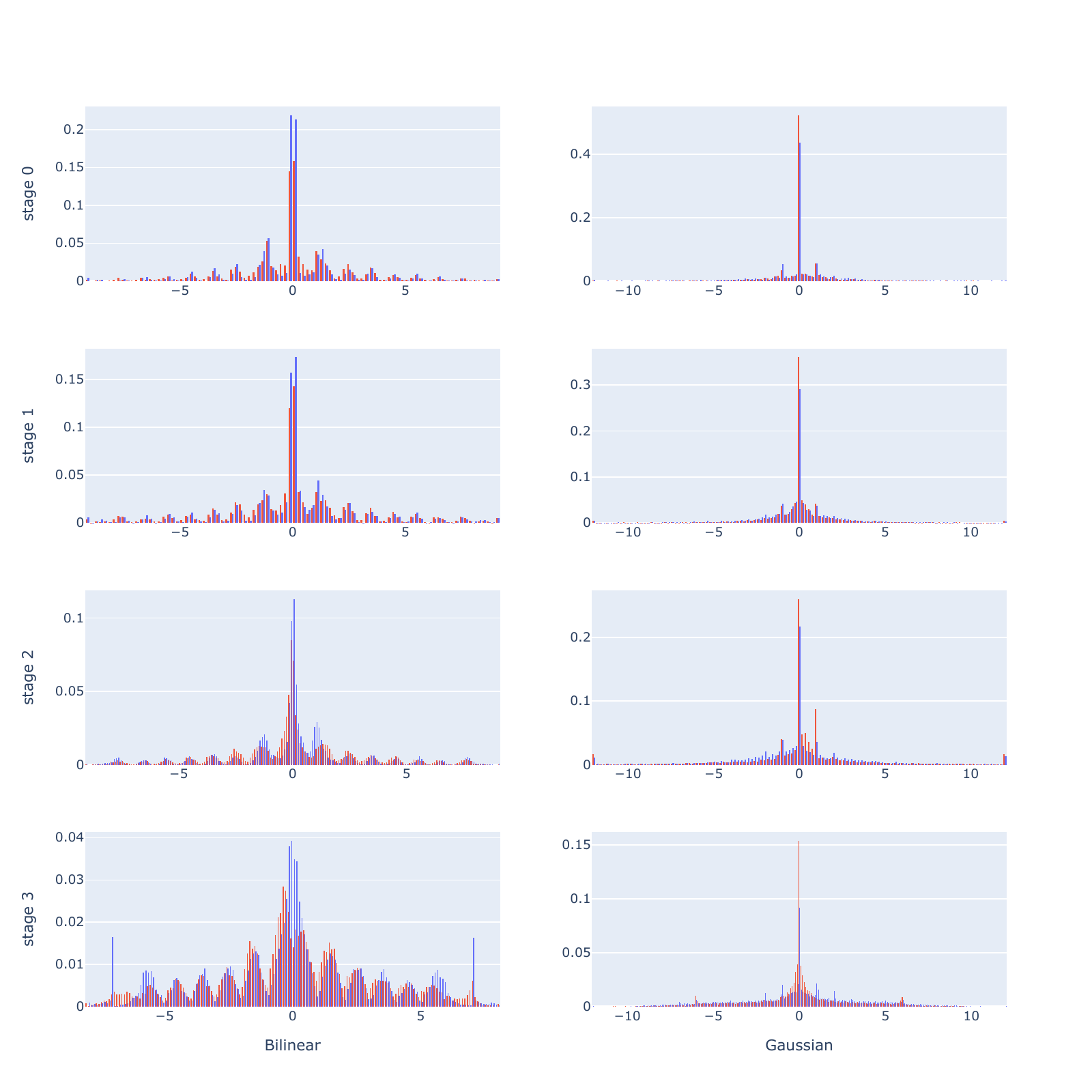}
    \caption{The distribution of the positions {\color{blue} \rule[0.1cm]{0.5cm}{0.4pt} }: along the time axis, {\color{red} \rule[0.1cm]{0.5cm}{0.4pt} }: along the frequency axis within the dilated kernel of a ConvNeXt-DCLS-Bilinear and a ConvNeXt-DCLS-Gauss models trained on AudioSet. The histograms are grouped by stages.}
    \label{fig:hist_audio}    
\end{figure}

\begin{figure}[!htbp]
    \centering
    \includegraphics[width=0.9\textwidth]{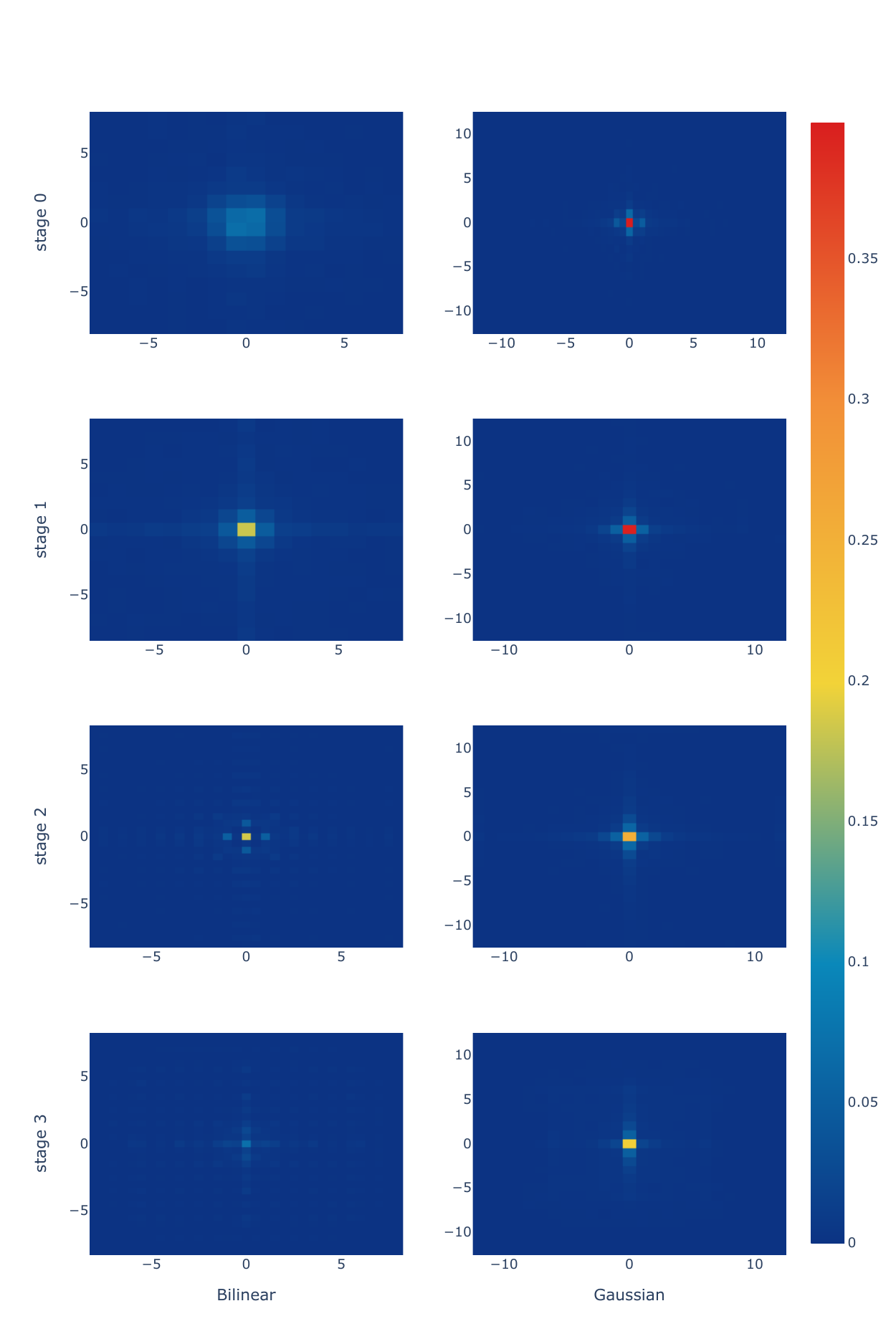}
    \caption{A 2D histogram view of the same distributions of Fig. \ref{fig:hist_images}.}
    \label{fig:hist2d_images}    
\end{figure}

\begin{figure}[!htbp]
    \centering
    \includegraphics[width=0.9\textwidth]{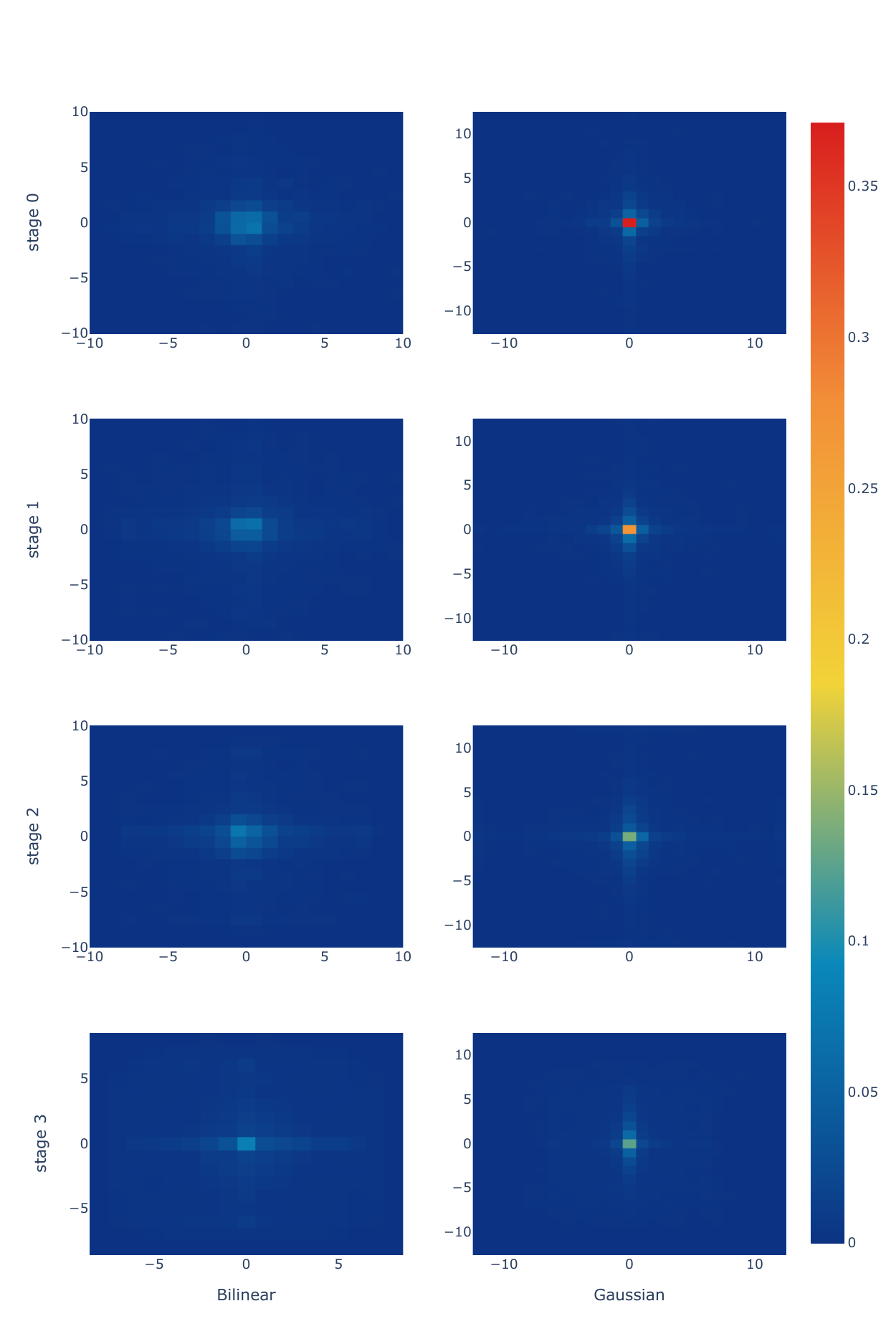}
    \caption{A 2D histogram view of the same distributions of Fig. \ref{fig:hist_audio}.}
    \label{fig:hist2d_audio}    
\end{figure}

The figures \ref{fig:hist_images} and \ref{fig:hist_audio} show that the positions learned by DCLS are distributed in a particularly remarkable way. For the DCLS-Bilinear method, the histograms show a pseudo-periodicity with a predominant central mode at $0$ followed by two secondary modes at $-1$ and $+1$ then by decreasing ripples centered on $k + \frac{1}{2}$ with $k \in \mathbb{Z}^*$.  This means that the central position is very important, and other positions are also important, but in a decreasing way. The general shape of the position distribution in the 2D-DCLS-Bilinear case is reminiscent of a squared sinc function $x \mapsto \left( \frac{sin(x)}{x} \right)^2$. The shapes of the histograms in \ref{fig:hist_images} are symmetrical and are almost identical along the x-axis {\color{blue} \rule[0.1cm]{0.5cm}{0.4pt} } and along the y-axis {\color{red} \rule[0.1cm]{0.5cm}{0.4pt} } meaning that there is no preferred direction of convergence for positions in the case of natural images. However, for the audio excerpts \ref{fig:hist_audio}, the histograms remain fairly symmetrical but still show signs of asymmetry between time axis {\color{blue} \rule[0.1cm]{0.5cm}{0.4pt} } and frequency axis {\color{red} \rule[0.1cm]{0.5cm}{0.4pt} }, especially in the last stage.

For the 2D-DCLS-Gauss method, a predominant mode appears at the central position $0$, followed by two secondary modes at $-1$ and $+1$. In contrast to the DCLS-Bilinear method, the DCLS-Gauss method shows no ripples. This is probably due to the fact that in the DCLS-Gauss method, the standard deviations are learnable and several kernel positions with large standard deviations can still be reached. The fact that the ripples in the DCLS-Bilinear method are centered at $k + \frac{1}{2}$ with $k \in \mathbb{Z}^*$ is probably caused by the fact that when a position is at $k + \frac{1}{2}$, the bilinear interpolation covers $4$ adjacent feature map pixels with the same weighting (equal to $\frac{1}{4}$ for 2D bilinear interpolation), which provides maximum area coverage of the kernel. Thus, this pseudo-periodical repartition at integer positions $+ \frac{1}{2}$.

\begin{figure}[!htbp]
    \centering
    \includegraphics[width=1\textwidth]{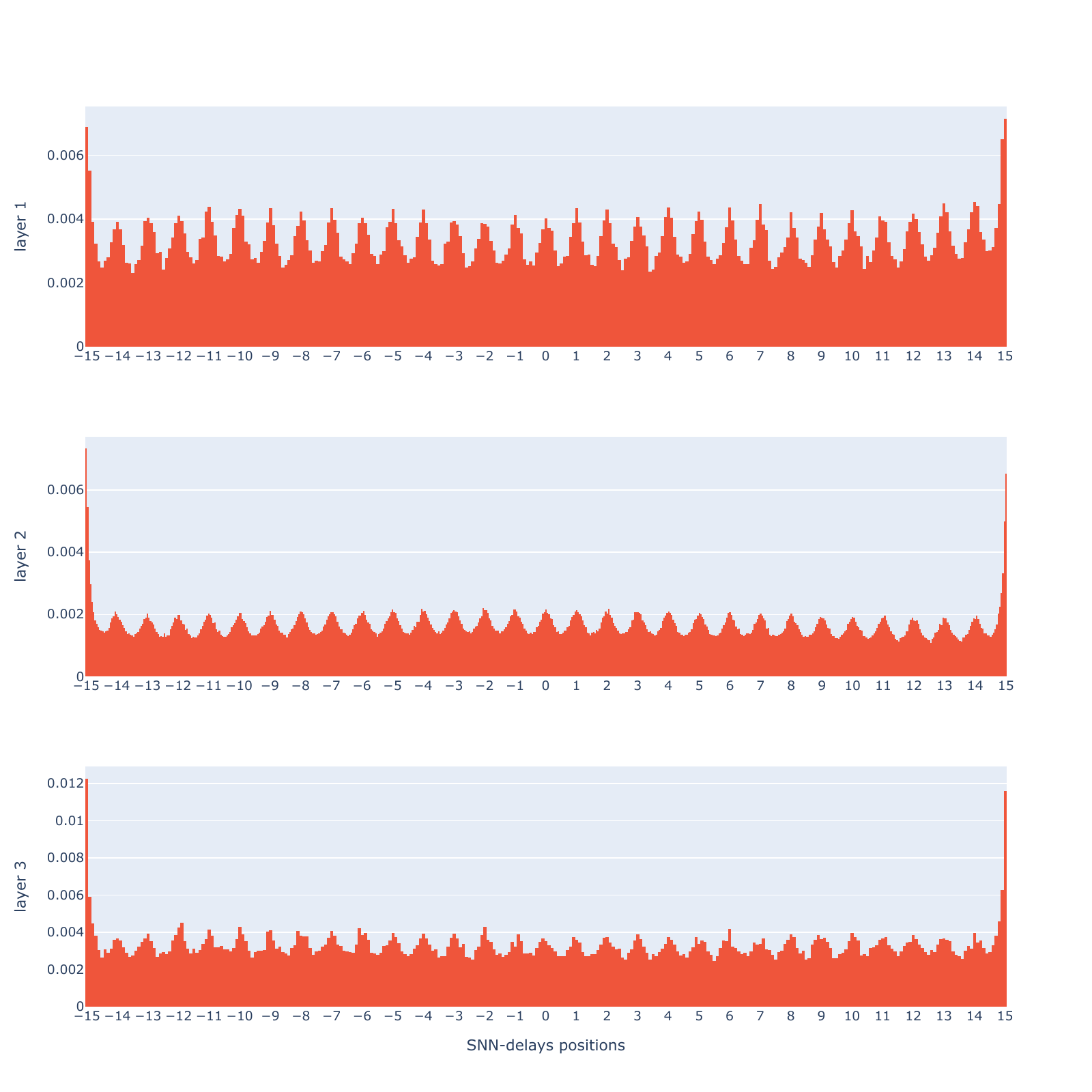}
    \caption{The distribution of the delays within the dilated kernel of an SNN-Delays model with 3 feed-forward layers, trained on the SSC dataset.}
    \label{fig:hist_snn}    
\end{figure}

Regarding the final delays obtained for the study we conducted on spiking neural networks (\ref{chap:4}),  we observe that all integer delays $k \in \mathbb{Z}^*$ are equally preferred and that a periodic pattern of period one, centered on integer delays, emerges in Figure \ref{fig:hist_snn}. This observation holds for all the networks we used in chapter \ref{chap:4} and for the three datasets we tested. Furthermore, we can clearly distinguish that this convergence pattern found for 1D-DCLS applied to learning delays in an SNN is very different from the 2D-DCLS convergence positions found for images in Figure \ref{fig:hist_images} and for audio spectrograms in  Figure \ref{fig:hist_audio}.
Admittedly, the limit delays (in this case -15 and 15) appear to be probabilistically privileged, but we think this is due to an edge phenomenon, which would indicate that the dilated kernel size (the maximum kernel size of 31) chosen here could be increased since the network seems to have a preference for certain delays that lie beyond this boundary.
The fact that integer delays are favored is an argument in support of rounding the final delays. This allows the use of fixed integer delays at convergence, which could be implemented, for example, on neuromorphic chips.

\section{Dataset limitation}

In this section, we discuss the limitations of the datasets used for the different experimental studies concerning the DCLS method. For classification, we have used large datasets (ImageNet1k \cite{deng2009imagenet}, AudioSet \cite{AudioSet}) as well as relatively small datasets such as SHD (Spiking Heidelberg Digits) \cite{shd}, SSC (Spiking Speech Commands) \cite{shd}, and GSC (Google Speech Commands v0.02). 
A spiking neural network with DCLS in chapter \ref{chap:4} outperformed the previous state-of-the-art accuracy on these last three benchmarks, with a significant improvement on SSC and GSC, and with a substantially lower number of parameters. We think that there's still considerable room for improvement in the SSC benchmark. For the GSC benchmark, non-spiking ANN models are already at $98.6\%$ accuracy \cite{chaudhary2023towards}, while our spiking neural network is currently the leader spiking model in this dataset at $95.35\%$ accuracy.

For larger datasets, we were able to show that every time we replaced the depthwise separable convolution method with DCLS, it resulted in a gain in test accuracy for all the tested models. 

Saying that a model is the state-of-the-art model on ImageNet is usually not sufficient, because one has to explain in what way. Is it in terms of absolute classification accuracy, or in terms of accuracy/throughput ratio? Is it in terms of accuracy/number of parameters ratio, or accuracy/number of labeled data used for training ratio? Or is it for some or all of these metrics? This last observation applies to audio classification on AudioSet as well. 

For example, taking the absolute accuracy as the metric (without looking at inference throughput or any other metric): is it really wise to say that ConvNeXt-T, with $82.1\%$ accuracy and a training crop of $224$, is worse for the single-label classification task than Fast-ViT-Sa24, which is at $82.6\%$ with more or less the same number of parameters and a training crop of $256$?

The metric of interest in audio classification on AudioSet is mAP instead of accuracy. Indeed, the reason AudioSet uses an mAP metric is because the task of interest for this dataset is a multi-label classification task. Multi-label classification is much more useful than single-label classification since audio excerpts can be ``noisy'', that is, the same excerpt can contain parts that satisfy at least two different labels: typically, an excerpt that contains a song with both music and sung lyrics satisfies the ``music'' label and the ``speech'' label at the same time.

ImageNet is also a noisy-label dataset. This was identified as early as the first edition of the ILSVRC2012 challenge. However, the single-label classification task on ImageNet continues to be the reference for comparing and selecting backbones for computer vision. We believe this can be improved. 

\citet{taesiri2023zoom} showed that by simply zooming in on the exact object associated with a ground truth label of an ImageNet evaluation image, most recent neural networks would achieve an accuracy of $98.91\%$ for single-label classification. ``The few hundred remaining images $(0.39\%)$ that never get correctly labeled by any model (despite optimal zooming) are mostly poorly positioned and rare images'' \citep{taesiri2023zoom}. Taking into account that a random crop is used for training neural networks on ImageNet, and that a central crop of size $224 \times 224$ or $256 \times 256$ is often used for inference, this finding does indeed call into question the reliability of the single-label classification task in this dataset. Most recent computer vision models achieve over $98\%$ accuracy if given an optimally zoomed crop. 

As stated by \citet{beyer2020we}: ``many images in the ImageNet dataset contain a clear view of a single object of interest: for these, a single label is an appropriate description of their content. However many other images contain multiple, similarly prominent objects, limiting the relevance of a single label. Some labels are overly restrictive. Others have arbitrary class distinctions''. Even images containing a single object can be subject to systematic inaccuracies due to biases in the collection procedure. Additionally, certain ImageNet classes are inherently ambiguous, making distinctions between essentially identical groups of images \citep{beyer2020we}.

Several solutions to this problem have been proposed by a variety of authors. relabeling ImageNet is one of them. Either by using a larger and more robust extra dataset machine annotator as in \cite{yun2021re} or by using feedback from human participants in a voting system as in \citep{beyer2020we}.

Another approach might be to consider a multi-label classification task with an mAP, as done for the AudioSet classification task. Or to a lesser extent, we could consider going through a multi-label classification phase before making a final choice and giving a single-label answer. \citet{yun2021re} observed that while the machine annotators are trained with single-label supervision on ImageNet, they still tend to make multi-label predictions for images with multiple categories. 

Perhaps the challenge in this kind of task should simply turn to finding the lightest (in terms of parameters, FLOPS...) and fastest (in terms of training/inference throughput...) architecture/model that reaches an acceptability threshold of $80\%$ in Top1 accuracy and $95\%$ in Top5 accuracy. 

\section{Sparse kernel convolution}

Even though DCLS-Gauss (Chapter \ref{chap:3}) makes the notion of sparse kernel irrelevant, the DCLS-Bilinear method (Chapter \ref{chap:2}) remains very sparse. The DCLS-Gauss version could be made sparse, in the context of sparse inference, by restricting the kernel elements to integer positions by performing training with decreasing non-learnable standard deviations $\sigma$ and then taking only the final integer positions as done in chapter \ref{chap:4}. However, if the goal is to perform the training with sparse kernels, the DCLS-Bilinear method is the most appropriate as it allows high sparsity while allowing the backpropagation of the gradient.

This leads us to discuss Sparse Kernel convolutions (which should not be confused with Sparse convolutions where the input is sparse and not the convolution kernel). DCLS-Bilinear is a typical case where the matrix product involved in the convolution could be seen as a product of a sparse matrix and a dense one. The cuSparse library provides this kind of sparse-matrix dense-matrix product called ``SpMM'' primitive. This primitive can be interfaced with PyTorch. We have already tested it, but we didn't see an advantage to it in the 2D case over dense matrix product as the sparsity of the DCLS kernel is $1 - \frac{4K_c}{d_1d_2}$  where $K_c$ is the kernel count (equal to $34$, or $26$ in our use cases), and $d_i \colon i \in \mathbb{N}^*$, the dilated kernel sizes (both equal to $17$ or $23$ in our use cases). According to our tests on multiple GPUs, this sparsity is still far too high for the ``SpMM''  method to be faster than the dense matrix multiplication, as this last routine is better suited to dense computation on GPUs. However, in more sparse cases, such as 3D convolution, the sparsity of the constructed kernel in DCLS-Bilinear becomes $1 - \frac{8K_c}{d_1d_2d_3}$ thus more interesting for the ``SpMM'' method. In the future, we would like to test this in addition to the implementation of large 3D kernels as explained in \ref{sec:3d}.

\section{\textit{Im2col} with learnable spacings}
\label{sec:im2colls}
An alternative to the sparse-matrix and dense-matrix product mentioned in the previous section would be to reimplement the \textsc{im2col}  function \ref{eq:image2col} so that it takes learnable positions into account. This has already been undertaken by us in the first pre-publication of DCLS \cite{hassani2021dilated} (see version v1). The problem with this \textsc{im2col} with learnable spacings method is that, for time and memory reasons,  we couldn't consider learning $C_{out}, C_{in}, K_c$ positions, and therefore must resort to learning only $C_{in}, K_c$ positions that will be shared along $C_{out}$. This is a limit for convolution, but not for local multi-head self-attention (MHSA) as we will see in the next section.

More generally, the \textsc{im2col} operation can be seen as a black-box function:
\begin{align}
  \textsc{im2col}  \colon  \quad \mathbb{R}^{C_{in}} \times \mathbb{R}^{H_{in}} \times \mathbb{R}^{W_{in}}  \to \mathbb{R}^{K_c C_{in}} \times \mathbb{R}^{H_{out}} \times \mathbb{R}^{W_{out}}
\end{align}
which can itself be approximated by a neural network, except that it is too general and lacks a suitable inductive bias. 


\subsection{Towards a dilated attention with learnable spacings module}

In a recent work, \cite{hassani2023neighborhood} introduces neighborhood attention (NA) as an efficient and scalable sliding window attention mechanism. Building upon shifted window self-attention \cite{liu2021swin}, neighborhood attention localizes attention to a neighborhood around image tokens, introducing local inductive biases, and allowing receptive field growth without needing extra operations. Inspired by dilated convolution \cite{yu2015multi}, the same authors then introduced dilated neighborhood attention (DiNA) as an extension to NA that dilates the attention window on a regular grid by a constant factor, thus expanding the receptive field at no additional cost. DilateFormer \cite{jiao2023dilateformer} has re-employed this same idea by making a simpler code in PyTorch that is just as efficient while using a multi-stage approach with different dilation factors in the MHSA heads.

It is therefore quite natural that we want to try an approach like DCLS for learning positions instead of using a regular grid, but this time for dilated neighborhood attention, which would result in a Dilated Attention with Learnable Spacings (DALS) method. Our preliminary analysis of the DALS method shows that it is possible to simply use our previous \textsc{im2col} with learnable spacings method as this time there are only $C_{in}\times K_c$ elements in the neighborhood attention grid.

\section{Convolution vs self-attention}

The last five years have seen a shift in the neural network paradigm driven by attention transformers. Starting in the field of natural language processing \cite{vaswani2017attention}, this shift has been generalized to other modalities (such as computer vision \cite{dosovitskiyimage} and audition \cite{miyazaki2020convolution, gong2021ast}). The use of multi-head self-attention blocks instead of standard convolution or its separable depthwise variant became commonplace. This transition was mainly due to the high-performance results of transformers, but can we say that they sounded the death knell for convolutional neural networks?

The answer to this question is still being debated, and recent empirical advances in the field seem to suggest that it is not. 

Firstly, \citet{cordonnier2019relationship} demonstrated that attention layers can implement convolutional layers and that a multi-head self-attention (MHSA) layer with a sufficient number of heads is at least as expressive as any convolutional layer. 

The fact that multi-head self-attention can implement any convolution under certain hypotheses is not enough to say that these two operations perform the same operation during learning, nor that they will lead to the same final result.

For \citet{cordonnier2019relationship} attention layers can perform the convolution operation and, indeed, at least in the image classification task, they often learn to do so in practice. This is shown by the conclusions of their numerical experiments on CIFAR-10 images where they show that self-attention layers attend to pixel-grid patterns similarly to CNN layers.

For \citet{park2021vision} on the other hand, multi-head self-attention layers and convolution layers exhibit opposite behaviors. A spectral analysis based on a Fourier transform of activation maps learned from ImageNet shows that convolution layers within a fully convolutional network (ResNet) act as high-pass filters, while attention layers within a fully attentional model act as low-pass filters. Showing, therefore, that MHSAs and convolutions are complementary.

This last observation motivated \cite{park2021vision} to think about how to harmonize convolution layers and MHSA layers within the same neural network. The following rules emerged:
\begin{itemize}
    \item Starting from the end of a baseline CNN model, replace convolution blocks with MHSA blocks.
    \item If the added MHSA block does not improve predictive performance, replace a convolution block
located at the end of an earlier stage with an MHSA block.
    \item Use more heads and higher hidden dimensions for MHSA blocks in the late stages.
\end{itemize}

The first of these guidelines is quite satisfactory in practice and derives from performance considerations. Placing a layer of attention (local or, even worse, global) in the first stages is much more costly in terms of memory and throughput than a convolution, due to the large size of the input feature maps. Almost all recent ``hybrid'' architectures in computer vision \cite{yu2022metaformer, xu2023parcnetv2, vasu2023fastvit}, which aim for performance in terms of accuracy, parsimony in the number of learnable parameters and throughput, use this construction rule: the first two stages contain depthwise convolution + multi-layer perceptrons (MLP) blocks, while the last two contain global MHSA + MLP blocks.

The latter construction ``divides'' the network into two parts: a first part with convolutional blocks and a second part with attention blocks. From a receptive field point of view, this is a particularly good idea, since the size of the receptive field of modern deep neural networks generally becomes larger than the input image by the end of the second stage of these networks.

Take, for example, the ConvNeXt model, which, like Resnet, is the archetype of the 4-stage computer vision model. The table \ref{tab:receptive_field} gives the size of the receptive field at each block of the ConvNeXt-T model as well as the ConvNeXt-T-DCLS with bilinear (dilated kernel size of 17)  and Gaussian (dilated kernel size of 23) interpolations.
\begin{table}[!ht]
\resizebox{\textwidth}{!}{
\centering
$
\begin{array}{c c c c c}
 &  & \text {ConvNeXt-T} & \text {ConvNeXt-T-DCLS-17} & \text {ConvNeXt-T-DCLS-23} \\
\text {Stage} & \text {Block} & \text {Receptive field size} & \text {Receptive field size} & \text {Receptive field size} \\
\hline \text{Stem} &   & 4 & 4 & 4  \\
\hline \multirow{3}{*}{0} & 0 & 28 & 68 & 92  \\
& 1 & 52 & 132 & 180  \\
& 2 & 76 & 196 & 268  \\
\hline \text{Downsampling} &  & 80 & 200 & 272 \\
\hline \multirow{3}{*}{1} & 0 & 128 & 328 & 448 \\
& 1 & 176 & 456 & 624 \\
& 2 & 224 & 584 & 800 \\
\hline \text{Downsampling} &  & 232 & 592 & 808 \\
\hline \multirow{9}{*}{2} & 0 & 328 & 848 & 1160 \\
& 1 & 424 & 1104 & 1512 \\
& 2 & 520 & 1360 & 1864 \\
& 3 & 616 & 1616 & 2216 \\
& 4 & 712 & 1872 & 2568 \\
& 5 & 808 & 2128 & 2920 \\
& 6 & 904 & 2384 & 3272 \\
& 7 & 1000 & 2640 & 3624 \\
& 8 & 1096 & 2896 & 3976 \\
\hline \text{Downsampling} &  & 1112 & 2912 & 3992 \\
\hline \multirow{3}{*}{3} & 0 & 1304 & 3424 & 4696 \\
& 1 & 1496 & 3936 & 5400 \\
& 2 & 1688 & 4448 & 6104 \\

\end{array}
$
}
\caption{The receptive field size per block of the ConvNeXt-T and ConvNeXt-T-DCLS models.}
\label{tab:receptive_field}
\end{table}

We can see that at the end of the second stage (stage 1 in the table), the receptive field size exceeds $224$. Knowing that the image crops used to train these models are of size $224$, we can argue that these modern convolutional networks may have a receptive field that is far too large in the final stages. Hence the interest in using MHSA + MLP blocks for the last two stages, since there's no need to enlarge the receptive field.

\citet{park2021vision} and \cite{cordonnier2019relationship} both showed that MHSAs flatten the loss landscape, i.e. they reduce the magnitude of the eigenvalues of the loss Hessian. However, \citet{park2021vision} suggest that the reason for the superiority of MHSAs is due to their data specificity and not because of their long-range dependency. Indeed, the most widely accepted explanation for the success of MHSAs is their weak inductive bias and capture of long-range dependencies. Still, several recent works on local attention (with a very small receptive field) outperformed global MHSAs as they reduce unnecessary degrees of freedom  \cite{liu2021swin, yang2019convolutional, chu2021twins}. The locality inductive bias reduces computational complexity and helps in optimization by convexifying the loss landscape \cite{park2021vision}.

The data specificity of transformers would be the key to their performance. However, it should not be forgotten that the architectural improvements the transformer models come with make it difficult to know what is truly improved by MHSA and what is improved with some other modification. It has been shown that the MLP-Mixer \cite{tolstikhin2021mlp} network, which does not contain any layer of self-attention, underperforms the ViT model for the same training budget. Nevertheless, MLP-Mixer still performs very well and might be sufficient for many use cases without the need for convolution or self-attention. Even more flagrantly, \citet{yu2022metaformer} showed that a simple IdentityFormer (where all spatial/token mixing layers were replaced by identity functions) was sufficient to reach 80\% in Top1 accuracy on ImageNet and that a RandFormer could go beyond 81\%!

In a recent survey comparing several fully convolutional, fully attentional, and hybrid models, \citet{nauen2023transformer} showed that ViTs were Pareto-optimal for several metrics (accuracy/throughput for example), but that hybrid models were still competitive and excelled in low-memory environments. ViTs have been known to overfit training datasets, resulting in poor predictive performance on small data regimes \citep{nauen2023transformer, park2021vision}. This belief is perhaps peculiar to the case of global attention and has been refuted by results from models using local attention, which both perform very well on large datasets and generalize better on small datasets.

However, the largest scale experiment to date comparing ViTs and ConvNets (NfNets in this case) is that of \citet{smith2023convnets} where the authors pre-trained the models on a very large image dataset (JFT-4B) and then fine-tuned them on ImageNet. They found that after fine-tuning on ImageNet, NFNets (a fully convolutional family model) matched the reported performance of Vision Transformers with comparable computational budgets. 
This last study reinforces the idea that the most important factors determining the performance of a designed model remain the compute and data available for training \cite{tolstikhin2021mlp} and that there is no strong evidence to suggest that pre-trained ViTs outperform pre-trained ConvNets when evaluated fairly. 


These efforts to harmonize convolution and attention at the architectural level are to be contrasted with methods that seek to bridge the two operations into a single one capable of performing both attention and convolution such as \cite{zhou2023interpret, dai2021coatnet, zhang2023rfaconv}.

\section{Perspectives}




\subsubsection{DCLS 3D, 3-1D and 1D versions}
\label{sec:3d}
Having demonstrated the effectiveness of the DCLS method for 2D and, to a lesser extent, 1D data, we would now like to establish its effectiveness in the 3D context. Recently, 3D large-kernel convolutional networks have been revisited by \citet{chen2022scaling}. Their results in 3D Semantic Segmentation and 3D object detection show a noticeable improvement due to large 3D kernels. The authors of \cite{chen2022scaling} were able to show that depthwise separable convolution is not beneficial for 3D tasks, and instead of using it, they created the spatial-wise partition convolution with large kernels (SW-LK). The idea of SW-LK is to share weights among spatial dimensions on convolutional kernels, instead of among channel dimensions as with the depthwise separable convolution. Thus, they were able to scale their approach to kernels of size $17 \times 17 \times 17$. Moreover, the authors use kernel-wise position encoding where the positions are learnable, similar to DCLS.

The DCLS 3D method can be used to train positions along a single dimension ( the so-called DCLS 3-1D method) or two dimensions (DCLS 3-2D) out of the three possible dimensions. In the context of video, the DCLS (3-1D) method can be very relevant since it allows us to restrict position learning to the temporal dimension and leaves dense kernels for the spatial dimensions. As we demonstrated in Chapter \ref{chap:4} for the learning of delays in an SNN, the DCLS method has great potential in the context of tasks with temporal data, which is why we want to explore this avenue of research in the future on datasets such as \cite{kay2017kinetics} and \cite{tan2022multi}.

More 3D applications such as point clouds \cite{thomas2019KPConv}, rendering \cite{kerbl20233d} and earth system forecasting \cite{gao2022earthformer} can also be envisioned.

While we have already demonstrated the usefulness of DCLS on AudioSet \cite{AudioSet} with 2D spectrograms, we would like to use DCLS 1D even more in the case of raw waveforms in audio. We wish to show that the method is also of interest in audio waveform generation, where dilated 1D convolution filters with large receptive fields are key~\citep{oord2016wavenet,yamamoto2020parallel,greshler2021catch}.



\subsubsection{Explainability measures of DCLS}

Preliminary results on the explainability measures of DCLS show that it is superior to the depthwise separable convolution, at least for the ConvNeXt model (in its three small and basic variants). The datasets used in these preliminary studies are derived from attention maps provided by human participants and collected through an online game called ClickMe \cite{linsley2018learning}. In this game, participants are presented with different images from ImageNet \cite{deng2009imagenet} and are asked to use a cursor to highlight the regions they consider important for recognizing the object in the image. Grad-CAM \cite{selvaraju2017grad} was chosen as the explaining method. Maps obtained by the Grad-CAM method were compared to those provided by human participants and a Spearman correlation score \cite{spearman1961proof} was calculated.

The DCLS method showed promising results, better than those of depthwise separable convolution. However, more extensive testing needs to be done, especially on different models (e.g., attentional and hybrid) and using other metrics (ideally a gradient-based one like Grad-CAM and a black-box one like RISE \cite{petsiuk2018rise} or SHAP \cite{lundberg2017unified}). Together, these will allow us to more reliably determine the relative superiority of DCLS in terms of explainability.

\subsubsection{DCLS, a dedicated architecture}
So far, we have used DCLS as a drop-in replacement of the depthwise separable/standard convolution in existing architectures. In the future, we would like to search for a dedicated architecture that would get the maximum benefit out of the DCLS method.

Approaches such as EfficientNet \cite{tan2019efficientnet, tan2021efficientnetv2} and RegNet \cite{radosavovic2020designing} can be a good source of inspiration as well as a good starting point. This search for the optimal architecture for DCLS is part of the Neural Architecture Search (NAS) approach, which is in itself a vast field of research \cite{Tan_2019_CVPR}. The aim is to find a near-optimal architecture with a minimum of search time. Insights gained through the study of kernel size and receptive field for the DCLS method, such as those in Section \ref{sec:kernel_size} and Table \ref{tab:receptive_field}, could be of great help in reducing the search space.

\section{Conclusion}

In this dissertation, we have examined some concepts of standard, strided, dilated, separable and advanced convolutions. We have also reviewed the notions of receptive field, and effective receptive fields and highlighted some notorious neural networks.

We then proposed DCLS, a new dilated convolution method where the positions of non-zero kernel elements are made learnable via backpropagation. The non-differentiability issue of learning discrete positions was circumvented by interpolation. We mathematically demonstrated how the gradients of positions could be calculated and we showed a competitive implementation for the method. We then proposed  Gaussian and $\Lambda$ interpolation methods as alternatives to bilinear interpolation. We demonstrated that DCLS often outperforms the standard and the dilated convolution. 
We listed several techniques that improve the learning process of DCLS, in particular sharing the positions within stages was key. We provided evidence that searching for optimal positions of weights within a dilated kernel can improve not only the accuracy in image classification but also in downstream and robustness tasks, using existing CNN architectures, without increasing their number of parameters. We reported a throughput overhead introduced by DCLS, but it remains marginal, provided that we use depthwise separable convolutions.

We also proposed a method for learning delays in feedforward spiking neural networks using DCLS. Every synaptic connection was modeled as a 1D Gaussian kernel centered on the delay position, and DCLS was used to learn the kernel positions (i.e. delays). The standard deviations of the Gaussians were decreased throughout training, such that at the end of training, we obtained an SNN model with one discrete delay per synapse, which could potentially be compatible with neuromorphic implementations. We showed that applying the DCLS method in SNNs for learning synaptic delays outperforms the state-of-the-art in the temporal spiking benchmarks SHD and SSC and the non-spiking benchmark GSC-35 while using fewer parameters than previous work. We also performed a rigorous control test that demonstrated the effectiveness of our delay learning method. 

We then demonstrated the efficacy of DCLS as a method with promising applications beyond the computer vision field. By exploiting DCLS in the audio tagging task on AudioSet, we have demonstrated tangible improvements in accuracy when compared to models employing traditional depthwise separable convolution methods.  

Finally, we discussed the limitations of the datasets used in our experimental studies, the limitations of our method, and how it could be improved. We then reviewed the current research literature, including the main differences between convolution and self-attention, and lastly gave an overview of our prospective research.



\begin{spacing}{0.9}


\bibliographystyle{plainnat} 
\cleardoublepage
\bibliography{References/references} 

\begin{thebibliography}{233}
\providecommand{\natexlab}[1]{#1}
\providecommand{\url}[1]{\texttt{#1}}
\expandafter\ifx\csname urlstyle\endcsname\relax
  \providecommand{\doi}[1]{doi: #1}\else
  \providecommand{\doi}{doi: \begingroup \urlstyle{rm}\Url}\fi

\bibitem[Akbari et~al.(2021)Akbari, Yuan, Qian, Chuang, Chang, Cui, and Gong]{akbari2021vatt}
Hassan Akbari, Liangzhe Yuan, Rui Qian, Wei-Hong Chuang, Shih-Fu Chang, Yin Cui, and Boqing Gong.
\newblock Vatt: Transformers for multimodal self-supervised learning from raw video, audio and text.
\newblock \emph{Advances in Neural Information Processing Systems}, 34:\penalty0 24206--24221, 2021.

\bibitem[Akopyan et~al.(2015)Akopyan, Sawada, Cassidy, Alvarez-Icaza, Arthur, Merolla, Imam, Nakamura, Datta, Nam, Taba, Beakes, Brezzo, Kuang, Manohar, Risk, Jackson, and Modha]{Akopyan2015}
Filipp Akopyan, Jun Sawada, Andrew Cassidy, Rodrigo Alvarez-Icaza, John Arthur, Paul Merolla, Nabil Imam, Yutaka Nakamura, Pallab Datta, Gi-Joon Nam, Brian Taba, Michael Beakes, Bernard Brezzo, Jente~B. Kuang, Rajit Manohar, William~P. Risk, Bryan Jackson, and Dharmendra~S. Modha.
\newblock {TrueNorth: Design and Tool Flow of a 65 mW 1 Million Neuron Programmable Neurosynaptic Chip}.
\newblock \emph{IEEE Transactions on Computer-Aided Design of Integrated Circuits and Systems}, 34\penalty0 (10):\penalty0 1537--1557, oct 2015.
\newblock ISSN 0278-0070.
\newblock \doi{10.1109/TCAD.2015.2474396}.
\newblock URL \url{http://ieeexplore.ieee.org/document/7229264/}.

\bibitem[Alberto Patiño-Saucedo(In press)]{iscas}
Federico~Corradi Alberto Patiño-Saucedo, Amirreza~Yousefzadeh.
\newblock Empirical study on the efficiency of spiking neural networks with axonal delays, and algorithm-hardware benchmarking.
\newblock In \emph{2023 International Symposium on Circuits and Systems}, In press.

\bibitem[Araujo et~al.(2019)Araujo, Norris, and Sim]{araujo2019computing}
André Araujo, Wade Norris, and Jack Sim.
\newblock Computing receptive fields of convolutional neural networks.
\newblock \emph{Distill}, 2019.
\newblock \doi{10.23915/distill.00021}.
\newblock https://distill.pub/2019/computing-receptive-fields.

\bibitem[Azadbakht et~al.(2022)Azadbakht, Kheradpisheh, Khalfaoui-Hassani, and Masquelier]{azadbakht2022drastically}
Alireza Azadbakht, Saeed~Reza Kheradpisheh, Ismail Khalfaoui-Hassani, and Timoth{\'e}e Masquelier.
\newblock Drastically reducing the number of trainable parameters in deep cnns by inter-layer kernel-sharing.
\newblock \emph{arXiv preprint arXiv:2210.14151}, 2022.

\bibitem[Ba et~al.(2016)Ba, Kiros, and Hinton]{ba2016layer}
Jimmy~Lei Ba, Jamie~Ryan Kiros, and Geoffrey~E Hinton.
\newblock Layer normalization.
\newblock \emph{arXiv preprint arXiv:1607.06450}, 2016.

\bibitem[Bao et~al.(2021)Bao, Dong, and Wei]{bao2021beit}
Hangbo Bao, Li~Dong, and Furu Wei.
\newblock Beit: Bert pre-training of image transformers.
\newblock \emph{arXiv preprint arXiv:2106.08254}, 2021.

\bibitem[Bass(2011)]{bass2011stochastic}
Richard~F Bass.
\newblock \emph{Stochastic processes}, volume~33.
\newblock Cambridge University Press, 2011.

\bibitem[Bekkers(2019)]{bekkers2019b}
Erik~J Bekkers.
\newblock B-spline cnns on lie groups.
\newblock \emph{arXiv preprint arXiv:1909.12057}, 2019.

\bibitem[Bellec et~al.(2018)Bellec, Salaj, Subramoney, Legenstein, and Maass]{bellec2018}
Guillaume Bellec, Darjan Salaj, Anand Subramoney, Robert Legenstein, and Wolfgang Maass.
\newblock Long short-term memory and learning-to-learn in networks of spiking neurons.
\newblock In S.~Bengio, H.~Wallach, H.~Larochelle, K.~Grauman, N.~Cesa-Bianchi, and R.~Garnett, editors, \emph{Advances in Neural Information Processing Systems}, volume~31. Curran Associates, Inc., 2018.
\newblock URL \url{https://proceedings.neurips.cc/paper_files/paper/2018/file/c203d8a151612acf12457e4d67635a95-Paper.pdf}.

\bibitem[Beyer et~al.(2020)Beyer, H{\'e}naff, Kolesnikov, Zhai, and Oord]{beyer2020we}
Lucas Beyer, Olivier~J H{\'e}naff, Alexander Kolesnikov, Xiaohua Zhai, and A{\"a}ron van~den Oord.
\newblock Are we done with imagenet?
\newblock \emph{arXiv preprint arXiv:2006.07159}, 2020.

\bibitem[Bittar and Garner(2022)]{baseline}
Alexandre Bittar and Philip~N. Garner.
\newblock A surrogate gradient spiking baseline for speech command recognition.
\newblock \emph{Frontiers in Neuroscience}, 16, 2022.
\newblock ISSN 1662-453X.
\newblock \doi{10.3389/fnins.2022.865897}.
\newblock URL \url{https://www.frontiersin.org/articles/10.3389/fnins.2022.865897}.

\bibitem[Bowers(2017)]{Bowers2017a}
Jeffrey~S. Bowers.
\newblock {Parallel Distributed Processing Theory in the Age of Deep Networks}.
\newblock \emph{Trends in Cognitive Sciences}, pages 1--12, 2017.
\newblock ISSN 13646613.
\newblock \doi{10.1016/j.tics.2017.09.013}.
\newblock URL \url{http://linkinghub.elsevier.com/retrieve/pii/S1364661317302164}.

\bibitem[Bu et~al.(2022)Bu, Fang, Ding, DAI, Yu, and Huang]{ann2snn_1}
Tong Bu, Wei Fang, Jianhao Ding, PENGLIN DAI, Zhaofei Yu, and Tiejun Huang.
\newblock Optimal {ANN}-{SNN} conversion for high-accuracy and ultra-low-latency spiking neural networks.
\newblock In \emph{International Conference on Learning Representations}, 2022.
\newblock URL \url{https://openreview.net/forum?id=7B3IJMM1k_M}.

\bibitem[Cai and Vasconcelos(2018)]{cai2018cascade}
Zhaowei Cai and Nuno Vasconcelos.
\newblock Cascade r-cnn: Delving into high quality object detection.
\newblock In \emph{Proc. IEEE/CVF Conf. Comput. Vis. Pattern Recog. (CVPR)}, pages 6154--6162, 2018.

\bibitem[Celarek et~al.(2022)Celarek, Hermosilla, Kerbl, Ropinski, and Wimmer]{celarekgaussian}
Adam Celarek, Pedro Hermosilla, Bernhard Kerbl, Timo Ropinski, and Michael Wimmer.
\newblock Gaussian mixture convolution networks.
\newblock In \emph{International Conference on Learning Representations}, 2022.

\bibitem[Chaudhary and Abrol(2023)]{chaudhary2023towards}
Aryan Chaudhary and Vinayak Abrol.
\newblock Towards on-device keyword spotting using low-footprint quaternion neural models.
\newblock In \emph{2023 IEEE Workshop on Applications of Signal Processing to Audio and Acoustics (WASPAA)}, pages 1--5. IEEE, 2023.

\bibitem[Chellapilla et~al.(2006)Chellapilla, Puri, and Simard]{chellapilla2006high}
Kumar Chellapilla, Sidd Puri, and Patrice Simard.
\newblock {High Performance Convolutional Neural Networks for Document Processing}.
\newblock In \emph{{Proc. Int. Workshop on Frontiers in Handwriting Recognition (IWFHR)}}, La Baule, 2006.
\newblock https://hal.inria.fr/inria-00112631.

\bibitem[Chen et~al.(2017)Chen, Papandreou, Kokkinos, Murphy, and Yuille]{chen2017deeplab}
Liang-Chieh Chen, George Papandreou, Iasonas Kokkinos, Kevin Murphy, and Alan~L Yuille.
\newblock Deeplab: Semantic image segmentation with deep convolutional nets, atrous convolution, and fully connected {CRF}s.
\newblock \emph{IEEE Trans. Pattern Anal. Mach. Intell.}, 40\penalty0 (4):\penalty0 834--848, 2017.

\bibitem[Chen et~al.(2018)Chen, Zhu, Papandreou, Schroff, and Adam]{chen2018encoder}
Liang-Chieh Chen, Yukun Zhu, George Papandreou, Florian Schroff, and Hartwig Adam.
\newblock Encoder-decoder with atrous separable convolution for semantic image segmentation.
\newblock In \emph{Proc. Eur. Conf. Comput. Vis. (ECCV)}, pages 801--818, 2018.

\bibitem[Chen et~al.(2020)Chen, Zhang, Zen, Weiss, Norouzi, and Chan]{chen2020wavegrad}
Nanxin Chen, Yu~Zhang, Heiga Zen, Ron~J Weiss, Mohammad Norouzi, and William Chan.
\newblock Wavegrad: Estimating gradients for waveform generation.
\newblock In \emph{International Conference on Learning Representations}, 2020.

\bibitem[Chen et~al.(2023)Chen, Li, Ning, and He]{chen2023gaussian}
Qi~Chen, Chao Li, Jia Ning, and Kun He.
\newblock Gaussian mask convolution for convolutional neural networks.
\newblock \emph{arXiv preprint arXiv:2302.04544}, 2023.

\bibitem[Chen et~al.(2022)Chen, Liu, Qi, Zhang, Sun, and Jia]{chen2022scaling}
Yukang Chen, Jianhui Liu, Xiaojuan Qi, Xiangyu Zhang, Jian Sun, and Jiaya Jia.
\newblock Scaling up kernels in 3d cnns.
\newblock \emph{arXiv preprint arXiv:2206.10555}, 2022.

\bibitem[Chollet(2017)]{chollet2017xception}
Fran{\c{c}}ois Chollet.
\newblock Xception: Deep learning with depthwise separable convolutions.
\newblock In \emph{Proc. IEEE/CVF Conf. Comput. Vis. Pattern Recog. (CVPR)}, pages 1251--1258, 2017.

\bibitem[Chu et~al.(2021)Chu, Tian, Wang, Zhang, Ren, Wei, Xia, and Shen]{chu2021twins}
Xiangxiang Chu, Zhi Tian, Yuqing Wang, Bo~Zhang, Haibing Ren, Xiaolin Wei, Huaxia Xia, and Chunhua Shen.
\newblock Twins: Revisiting the design of spatial attention in vision transformers.
\newblock \emph{Advances in Neural Information Processing Systems}, 34:\penalty0 9355--9366, 2021.

\bibitem[Conneau et~al.(2017)Conneau, Schwenk, Le~Cun, and Barrault]{conneau2017very}
Alexis Conneau, Holger Schwenk, Yann Le~Cun, and L{\"o}c Barrault.
\newblock Very deep convolutional networks for text classification.
\newblock In \emph{15th Conference of the European Chapter of the Association for Computational Linguistics, EACL 2017}, pages 1107--1116. Association for Computational Linguistics (ACL), 2017.

\bibitem[Cordonnier et~al.(2019)Cordonnier, Loukas, and Jaggi]{cordonnier2019relationship}
Jean-Baptiste Cordonnier, Andreas Loukas, and Martin Jaggi.
\newblock On the relationship between self-attention and convolutional layers.
\newblock In \emph{International Conference on Learning Representations}, 2019.

\bibitem[Cramer et~al.(2022)Cramer, Stradmann, Schemmel, and Zenke]{shd}
Benjamin Cramer, Yannik Stradmann, Johannes Schemmel, and Friedemann Zenke.
\newblock The heidelberg spiking data sets for the systematic evaluation of spiking neural networks.
\newblock \emph{IEEE Transactions on Neural Networks and Learning Systems}, 33\penalty0 (7):\penalty0 2744--2757, 2022.
\newblock \doi{10.1109/TNNLS.2020.3044364}.

\bibitem[Cuadrado et~al.(2023)Cuadrado, Ran{\c{c}}on, Cottereau, Barranco, and Masquelier]{cuadrado2023optical}
Javier Cuadrado, Ulysse Ran{\c{c}}on, Benoit~R Cottereau, Francisco Barranco, and Timoth{\'e}e Masquelier.
\newblock Optical flow estimation from event-based cameras and spiking neural networks.
\newblock \emph{Frontiers in Neuroscience}, 17:\penalty0 1160034, 2023.

\bibitem[Cubuk et~al.(2020)Cubuk, Zoph, Shlens, and Le]{cubuk2020randaugment}
Ekin~D Cubuk, Barret Zoph, Jonathon Shlens, and Quoc~V Le.
\newblock Randaugment: Practical automated data augmentation with a reduced search space.
\newblock In \emph{Proceedings of the IEEE/CVF conference on computer vision and pattern recognition workshops}, pages 702--703, 2020.

\bibitem[Dai et~al.(2017{\natexlab{a}})Dai, Qi, Xiong, Li, Zhang, Hu, and Wei]{dai2017deformable}
Jifeng Dai, Haozhi Qi, Yuwen Xiong, Yi~Li, Guodong Zhang, Han Hu, and Yichen Wei.
\newblock Deformable convolutional networks.
\newblock In \emph{Int. Conf. Comput. Vis.}, pages 764--773, 2017{\natexlab{a}}.

\bibitem[Dai et~al.(2017{\natexlab{b}})Dai, Dai, Qu, Li, and Das]{dai2017very}
Wei Dai, Chia Dai, Shuhui Qu, Juncheng Li, and Samarjit Das.
\newblock Very deep convolutional neural networks for raw waveforms.
\newblock In \emph{2017 IEEE international conference on acoustics, speech and signal processing (ICASSP)}, pages 421--425. IEEE, 2017{\natexlab{b}}.

\bibitem[Dai et~al.(2021)Dai, Liu, Le, and Tan]{dai2021coatnet}
Zihang Dai, Hanxiao Liu, Quoc~V Le, and Mingxing Tan.
\newblock Coatnet: Marrying convolution and attention for all data sizes.
\newblock \emph{Advances in neural information processing systems}, 34:\penalty0 3965--3977, 2021.

\bibitem[Dampfhoffer et~al.(2022)Dampfhoffer, Mesquida, Valentian, and Anghel]{spikGRU}
Manon Dampfhoffer, Thomas Mesquida, Alexandre Valentian, and Lorena Anghel.
\newblock Investigating current-based and gating approaches for accurate and energy-efficient spiking recurrent neural networks.
\newblock In Elias Pimenidis, Plamen Angelov, Chrisina Jayne, Antonios Papaleonidas, and Mehmet Aydin, editors, \emph{Artificial Neural Networks and Machine Learning -- ICANN 2022}, pages 359--370, Cham, 2022. Springer Nature Switzerland.
\newblock ISBN 978-3-031-15934-3.

\bibitem[Dao et~al.(2022)Dao, Fu, Ermon, Rudra, and R{\'e}]{dao2022flashattention}
Tri Dao, Dan Fu, Stefano Ermon, Atri Rudra, and Christopher R{\'e}.
\newblock Flashattention: Fast and memory-efficient exact attention with io-awareness.
\newblock \emph{Advances in Neural Information Processing Systems}, 35:\penalty0 16344--16359, 2022.

\bibitem[Davies et~al.(2018)Davies, Srinivasa, Lin, Chinya, Joshi, Lines, Wild, and Wang]{Davies2018}
Mike Davies, Narayan Srinivasa, Tsung~Han Lin, Gautham Chinya, Prasad Joshi, Andrew Lines, Andreas Wild, and Hong Wang.
\newblock {Loihi: A Neuromorphic Manycore Processor with On-Chip Learning}, 2018.
\newblock ISSN 02721732.

\bibitem[Decourt et~al.(2022)Decourt, VanRullen, Salle, and Oberlin]{decourt2022recurrent}
Colin Decourt, Rufin VanRullen, Didier Salle, and Thomas Oberlin.
\newblock A recurrent cnn for online object detection on raw radar frames.
\newblock \emph{arXiv preprint arXiv:2212.11172}, 2022.

\bibitem[Deng et~al.(2009)Deng, Dong, Socher, Li, Li, and Fei-Fei]{deng2009imagenet}
Jia Deng, Wei Dong, Richard Socher, Li-Jia Li, Kai Li, and Li~Fei-Fei.
\newblock Imagenet: A large-scale hierarchical image database.
\newblock In \emph{Proc. IEEE/CVF Conf. Comput. Vis. Pattern Recog. (CVPR)}, pages 248--255. IEEE, 2009.

\bibitem[Deng and Gu(2021)]{ann2snn_2}
Shikuang Deng and Shi Gu.
\newblock Optimal conversion of conventional artificial neural networks to spiking neural networks.
\newblock In \emph{International Conference on Learning Representations}, 2021.
\newblock URL \url{https://openreview.net/forum?id=FZ1oTwcXchK}.

\bibitem[Dhariwal et~al.(2020)Dhariwal, Jun, Payne, Kim, Radford, and Sutskever]{dhariwal2020jukebox}
Prafulla Dhariwal, Heewoo Jun, Christine Payne, Jong~Wook Kim, Alec Radford, and Ilya Sutskever.
\newblock Jukebox: A generative model for music.
\newblock \emph{arXiv preprint arXiv:2005.00341}, 2020.

\bibitem[Ding et~al.(2022)Ding, Zhang, Han, and Ding]{ding2022scaling}
Xiaohan Ding, Xiangyu Zhang, Jungong Han, and Guiguang Ding.
\newblock {Scaling up your kernels to 31x31: Revisiting large kernel design in CNNs}.
\newblock In \emph{Proc. IEEE/CVF Conf. Comput. Vis. Pattern Recog. (CVPR)}, pages 11963--11975, 2022.

\bibitem[Ding et~al.(2023)Ding, Zhang, Ge, Zhao, Song, Yue, and Shan]{ding2023unireplknet}
Xiaohan Ding, Yiyuan Zhang, Yixiao Ge, Sijie Zhao, Lin Song, Xiangyu Yue, and Ying Shan.
\newblock Unireplknet: A universal perception large-kernel convnet for audio, video, point cloud, time-series and image recognition.
\newblock \emph{arXiv preprint arXiv:2311.15599}, 2023.

\bibitem[Dosovitskiy et~al.(2020)Dosovitskiy, Beyer, Kolesnikov, Weissenborn, Zhai, Unterthiner, Dehghani, Minderer, Heigold, Gelly, et~al.]{dosovitskiy2020image}
Alexey Dosovitskiy, Lucas Beyer, Alexander Kolesnikov, Dirk Weissenborn, Xiaohua Zhai, Thomas Unterthiner, Mostafa Dehghani, Matthias Minderer, Georg Heigold, Sylvain Gelly, et~al.
\newblock An image is worth 16x16 words: Transformers for image recognition at scale.
\newblock \emph{arXiv preprint arXiv:2010.11929}, 2020.

\bibitem[Dosovitskiy et~al.(2021)Dosovitskiy, Beyer, Kolesnikov, Weissenborn, Zhai, Unterthiner, Dehghani, Minderer, Heigold, Gelly, et~al.]{dosovitskiyimage}
Alexey Dosovitskiy, Lucas Beyer, Alexander Kolesnikov, Dirk Weissenborn, Xiaohua Zhai, Thomas Unterthiner, Mostafa Dehghani, Matthias Minderer, Georg Heigold, Sylvain Gelly, et~al.
\newblock An image is worth 16x16 words: Transformers for image recognition at scale.
\newblock In \emph{International Conference on Learning Representations}, 2021.

\bibitem[Drossos et~al.(2020)Drossos, Mimilakis, Gharib, Li, and Virtanen]{drossos2020sound}
Konstantinos Drossos, Stylianos~I Mimilakis, Shayan Gharib, Yanxiong Li, and Tuomas Virtanen.
\newblock Sound event detection with depthwise separable and dilated convolutions.
\newblock In \emph{2020 International Joint Conference on Neural Networks (IJCNN)}, pages 1--7. IEEE, 2020.

\bibitem[Duchi et~al.(2011)Duchi, Hazan, and Singer]{duchi2011adaptive}
John Duchi, Elad Hazan, and Yoram Singer.
\newblock Adaptive subgradient methods for online learning and stochastic optimization.
\newblock \emph{Journal of machine learning research}, 12\penalty0 (7), 2011.

\bibitem[Fang et~al.(2021{\natexlab{a}})Fang, Yu, Chen, Masquelier, Huang, and Tian]{plif}
W.~Fang, Z.~Yu, Y.~Chen, T.~Masquelier, T.~Huang, and Y.~Tian.
\newblock Incorporating learnable membrane time constant to enhance learning of spiking neural networks.
\newblock In \emph{2021 IEEE/CVF International Conference on Computer Vision (ICCV)}, pages 2641--2651, Los Alamitos, CA, USA, oct 2021{\natexlab{a}}. IEEE Computer Society.
\newblock \doi{10.1109/ICCV48922.2021.00266}.
\newblock URL \url{https://doi.ieeecomputersociety.org/10.1109/ICCV48922.2021.00266}.

\bibitem[Fang et~al.(2020)Fang, Chen, Ding, Chen, Yu, Zhou, Masquelier, Tian, and other contributors]{SpikingJelly}
Wei Fang, Yanqi Chen, Jianhao Ding, Ding Chen, Zhaofei Yu, Huihui Zhou, Timothée Masquelier, Yonghong Tian, and other contributors.
\newblock Spikingjelly.
\newblock \url{https://github.com/fangwei123456/spikingjelly}, 2020.

\bibitem[Fang et~al.(2021{\natexlab{b}})Fang, Yu, Chen, Huang, Masquelier, and Tian]{sew}
Wei Fang, Zhaofei Yu, Yanqi Chen, Tiejun Huang, Timoth\'{e}e Masquelier, and Yonghong Tian.
\newblock Deep residual learning in spiking neural networks.
\newblock In M.~Ranzato, A.~Beygelzimer, Y.~Dauphin, P.S. Liang, and J.~Wortman Vaughan, editors, \emph{Advances in Neural Information Processing Systems}, volume~34, pages 21056--21069. Curran Associates, Inc., 2021{\natexlab{b}}.
\newblock URL \url{https://proceedings.neurips.cc/paper_files/paper/2021/file/afe434653a898da20044041262b3ac74-Paper.pdf}.

\bibitem[Fang et~al.(2023)Fang, Chen, Ding, Yu, Masquelier, Chen, Huang, Zhou, Li, and Tian]{Fang2023a}
Wei Fang, Yanqi Chen, Jianhao Ding, Zhaofei Yu, Timoth{\'{e}}e Masquelier, Ding Chen, Liwei Huang, Huihui Zhou, Guoqi Li, and Yonghong Tian.
\newblock {SpikingJelly: An open-source machine learning infrastructure platform for spike-based intelligence}.
\newblock \emph{Science Advances}, 9\penalty0 (40), oct 2023.
\newblock \doi{10.1126/sciadv.adi1480}.
\newblock URL \url{https://www.science.org/doi/10.1126/sciadv.adi1480}.

\bibitem[Franke(1982)]{franke1982scattered}
Richard Franke.
\newblock Scattered data interpolation: tests of some methods.
\newblock \emph{Mathematics of computation}, 38\penalty0 (157):\penalty0 181--200, 1982.

\bibitem[Fukushima(1975)]{fukushima1975cognitron}
Kunihiko Fukushima.
\newblock Cognitron: A self-organizing multilayered neural network.
\newblock \emph{Biological cybernetics}, 20\penalty0 (3-4):\penalty0 121--136, 1975.

\bibitem[Furber et~al.(2014)Furber, Galluppi, Temple, and Plana]{Furber2014}
Steve~B. Furber, Francesco Galluppi, Steve Temple, and Luis~A Plana.
\newblock {The SpiNNaker Project}.
\newblock \emph{Proceedings of the IEEE}, 102\penalty0 (5):\penalty0 652--665, may 2014.
\newblock \doi{10.1109/JPROC.2014.2304638}.
\newblock URL \url{https://ieeexplore.ieee.org/document/6750072/}.

\bibitem[Gao et~al.(2022)Gao, Shi, Wang, Zhu, Wang, Li, and Yeung]{gao2022earthformer}
Zhihan Gao, Xingjian Shi, Hao Wang, Yi~Zhu, Yuyang~Bernie Wang, Mu~Li, and Dit-Yan Yeung.
\newblock Earthformer: Exploring space-time transformers for earth system forecasting.
\newblock \emph{Advances in Neural Information Processing Systems}, 35:\penalty0 25390--25403, 2022.

\bibitem[Gemmeke et~al.(2017)Gemmeke, Ellis, Freedman, Jansen, Lawrence, Moore, Plakal, and Ritter]{AudioSet}
Jort~F. Gemmeke, Daniel P.~W. Ellis, Dylan Freedman, Aren Jansen, Wade Lawrence, R.~Channing Moore, Manoj Plakal, and Marvin Ritter.
\newblock Audio set: An ontology and human-labeled dataset for audio events.
\newblock In \emph{Proc. IEEE ICASSP}, New Orleans, LA, 2017.

\bibitem[Gerstner(1995)]{srm}
Wulfram Gerstner.
\newblock Time structure of the activity in neural network models.
\newblock \emph{Phys. Rev. E}, 51:\penalty0 738--758, Jan 1995.
\newblock \doi{10.1103/PhysRevE.51.738}.
\newblock URL \url{https://link.aps.org/doi/10.1103/PhysRevE.51.738}.

\bibitem[Gerstner and Kistler(2002)]{gerstnerkistler2002}
Wulfram Gerstner and Werner~M. Kistler.
\newblock \emph{Spiking Neuron Models: Single Neurons, Populations, Plasticity}.
\newblock Cambridge University Press, 2002.
\newblock \doi{10.1017/CBO9780511815706}.

\bibitem[Gong et~al.(2021{\natexlab{a}})Gong, Chung, and Glass]{gong2021ast}
Yuan Gong, Yu-An Chung, and James Glass.
\newblock {AST: Audio Spectrogram Transformer}.
\newblock In \emph{Proc. Interspeech}, pages 571--575, Brno, 2021{\natexlab{a}}.
\newblock \doi{10.21437/Interspeech.2021-698}.

\bibitem[Gong et~al.(2021{\natexlab{b}})Gong, Chung, and Glass]{gong2021psla}
Yuan Gong, Yu-An Chung, and James Glass.
\newblock {PSLA: Improving audio tagging with pretraining, sampling, labeling, and aggregation}.
\newblock \emph{IEEE/ACM Transactions on Audio, Speech, and Language Processing}, 29:\penalty0 3292--3306, 2021{\natexlab{b}}.

\bibitem[Gong et~al.(2022)Gong, Rouditchenko, Liu, Harwath, Karlinsky, Kuehne, and Glass]{gong2022contrastive}
Yuan Gong, Andrew Rouditchenko, Alexander~H Liu, David Harwath, Leonid Karlinsky, Hilde Kuehne, and James~R Glass.
\newblock Contrastive audio-visual masked autoencoder.
\newblock In \emph{The Eleventh International Conference on Learning Representations}, 2022.

\bibitem[Grappolini and Subramoney(2023)]{beyondweights}
Edoardo~W. Grappolini and Anand Subramoney.
\newblock Beyond weights: Deep learning in spiking neural networks with pure synaptic-delay training, 2023.

\bibitem[Greshler et~al.(2021)Greshler, Shaham, and Michaeli]{greshler2021catch}
Gal Greshler, Tamar Shaham, and Tomer Michaeli.
\newblock Catch-a-waveform: Learning to generate audio from a single short example.
\newblock \emph{Advances in Neural Information Processing Systems}, 34:\penalty0 20916--20928, 2021.

\bibitem[Grimaldi and Perrinet(2022)]{related_delays1}
Antoine Grimaldi and Laurent~U Perrinet.
\newblock Learning hetero-synaptic delays for motion detection in a single layer of spiking neurons.
\newblock In \emph{2022 IEEE International Conference on Image Processing (ICIP)}, pages 3591--3595, 2022.
\newblock \doi{10.1109/ICIP46576.2022.9897394}.

\bibitem[Grimaldi and Perrinet(2023)]{related_delays1bis}
Antoine Grimaldi and Laurent~U Perrinet.
\newblock Learning heterogeneous delays in a layer of spiking neurons for fast motion detection, 2023.

\bibitem[Guo et~al.(2022)Guo, Han, Wu, Tang, Chen, Wang, and Xu]{guo2022cmt}
Jianyuan Guo, Kai Han, Han Wu, Yehui Tang, Xinghao Chen, Yunhe Wang, and Chang Xu.
\newblock Cmt: Convolutional neural networks meet vision transformers.
\newblock In \emph{Proceedings of the IEEE/CVF Conference on Computer Vision and Pattern Recognition}, pages 12175--12185, 2022.

\bibitem[Hacene et~al.(2021)Hacene, Lassance, Gripon, Courbariaux, and Bengio]{hacene2021attention}
Ghouthi~Boukli Hacene, Carlos Lassance, Vincent Gripon, Matthieu Courbariaux, and Yoshua Bengio.
\newblock Attention based pruning for shift networks.
\newblock In \emph{2020 25th International Conference on Pattern Recognition (ICPR)}, pages 4054--4061, 2021.
\newblock \doi{10.1109/ICPR48806.2021.9412859}.

\bibitem[Hammouamri et~al.(2022)Hammouamri, Masquelier, and Wilson]{hammouamri2022mitigating}
Ilyass Hammouamri, Timoth{\'e}e Masquelier, and Dennis~George Wilson.
\newblock Mitigating catastrophic forgetting in spiking neural networks through threshold modulation.
\newblock \emph{Transactions on Machine Learning Research}, 2022.
\newblock ISSN 2835-8856.
\newblock URL \url{https://openreview.net/forum?id=15SoThZmtU}.

\bibitem[Hammouamri et~al.(2023)Hammouamri, Khalfaoui-Hassani, and Masquelier]{hammouamri2023learning}
Ilyass Hammouamri, Ismail Khalfaoui-Hassani, and Timoth{\'e}e Masquelier.
\newblock Learning delays in spiking neural networks using dilated convolutions with learnable spacings.
\newblock \emph{arXiv preprint arXiv:2306.17670}, 2023.

\bibitem[Han et~al.(2020)Han, Srinivasan, and Roy]{ann2snn_3}
B.~Han, G.~Srinivasan, and K.~Roy.
\newblock Rmp-snn: Residual membrane potential neuron for enabling deeper high-accuracy and low-latency spiking neural network.
\newblock In \emph{2020 IEEE/CVF Conference on Computer Vision and Pattern Recognition (CVPR)}, pages 13555--13564, Los Alamitos, CA, USA, jun 2020. IEEE Computer Society.
\newblock \doi{10.1109/CVPR42600.2020.01357}.
\newblock URL \url{https://doi.ieeecomputersociety.org/10.1109/CVPR42600.2020.01357}.

\bibitem[Han et~al.(2021)Han, Xiang, Ren, Fu, Wen, and Hao]{DW_photonic}
Yanan Han, Shuiying Xiang, Zhenxing Ren, Chentao Fu, Aijun Wen, and Yue Hao.
\newblock Delay-weight plasticity-based supervised learning in optical spiking neural networks.
\newblock \emph{Photon. Res.}, 9\penalty0 (4):\penalty0 B119--B127, Apr 2021.
\newblock \doi{10.1364/PRJ.413742}.
\newblock URL \url{https://opg.optica.org/prj/abstract.cfm?URI=prj-9-4-B119}.

\bibitem[Hassani et~al.(2023)Hassani, Walton, Li, Li, and Shi]{hassani2023neighborhood}
Ali Hassani, Steven Walton, Jiachen Li, Shen Li, and Humphrey Shi.
\newblock Neighborhood attention transformer.
\newblock In \emph{Proceedings of the IEEE/CVF Conference on Computer Vision and Pattern Recognition}, pages 6185--6194, 2023.

\bibitem[Hazan et~al.(2022)Hazan, Caby, Earl, Siegelmann, and Levin]{related_delays4}
Hananel Hazan, Simon Caby, Christopher Earl, Hava Siegelmann, and Michael Levin.
\newblock Memory via temporal delays in weightless spiking neural network, 2022.

\bibitem[He et~al.(2016)He, Zhang, Ren, and Sun]{he2016deep}
Kaiming He, Xiangyu Zhang, Shaoqing Ren, and Jian Sun.
\newblock Deep residual learning for image recognition.
\newblock In \emph{Proceedings of the IEEE conference on computer vision and pattern recognition}, pages 770--778, 2016.

\bibitem[He et~al.(2023)He, Li, Zhao, Kong, and Zeng]{msat}
Xiang He, Yang Li, Dongcheng Zhao, Qingqun Kong, and Yi~Zeng.
\newblock Msat: Biologically inspired multi-stage adaptive threshold for conversion of spiking neural networks, 2023.

\bibitem[Hendrycks and Dietterich(2019)]{hendrycks2019benchmarking}
Dan Hendrycks and Thomas Dietterich.
\newblock Benchmarking neural network robustness to common corruptions and perturbations.
\newblock \emph{arXiv preprint arXiv:1903.12261}, 2019.

\bibitem[Hendrycks et~al.(2021{\natexlab{a}})Hendrycks, Basart, Mu, Kadavath, Wang, Dorundo, Desai, Zhu, Parajuli, Guo, et~al.]{hendrycks2021many}
Dan Hendrycks, Steven Basart, Norman Mu, Saurav Kadavath, Frank Wang, Evan Dorundo, Rahul Desai, Tyler Zhu, Samyak Parajuli, Mike Guo, et~al.
\newblock The many faces of robustness: A critical analysis of out-of-distribution generalization.
\newblock In \emph{Int. Conf. Comput. Vis.}, pages 8340--8349, 2021{\natexlab{a}}.

\bibitem[Hendrycks et~al.(2021{\natexlab{b}})Hendrycks, Zhao, Basart, Steinhardt, and Song]{hendrycks2021natural}
Dan Hendrycks, Kevin Zhao, Steven Basart, Jacob Steinhardt, and Dawn Song.
\newblock Natural adversarial examples.
\newblock In \emph{Proc. IEEE/CVF Conf. Comput. Vis. Pattern Recog. (CVPR)}, pages 15262--15271, 2021{\natexlab{b}}.

\bibitem[Holschneider et~al.(1990)Holschneider, Kronland-Martinet, Morlet, and Tchamitchian]{holschneider1990real}
Matthias Holschneider, Richard Kronland-Martinet, Jean Morlet, and Ph~Tchamitchian.
\newblock A real-time algorithm for signal analysis with the help of the wavelet transform.
\newblock In \emph{Wavelets}, pages 286--297. Springer, 1990.

\bibitem[Huang et~al.(2017)Huang, Liu, Van Der~Maaten, and Weinberger]{huang2017densely}
Gao Huang, Zhuang Liu, Laurens Van Der~Maaten, and Kilian~Q Weinberger.
\newblock Densely connected convolutional networks.
\newblock In \emph{Proceedings of the IEEE conference on computer vision and pattern recognition}, pages 4700--4708, 2017.

\bibitem[Huang et~al.(2022)Huang, Sharma, Xu, Ryali, Fan, Li, Li, Ghosh, Malik, and Feichtenhofer]{huang2022mavil}
Po-Yao Huang, Vasu Sharma, Hu~Xu, Chaitanya Ryali, Haoqi Fan, Yanghao Li, Shang-Wen Li, Gargi Ghosh, Jitendra Malik, and Christoph Feichtenhofer.
\newblock Mavil: Masked audio-video learners.
\newblock \emph{arXiv preprint arXiv:2212.08071}, 2022.

\bibitem[Ioffe and Szegedy(2015)]{ioffe2015batch}
Sergey Ioffe and Christian Szegedy.
\newblock Batch normalization: Accelerating deep network training by reducing internal covariate shift.
\newblock In \emph{International conference on machine learning}, pages 448--456. pmlr, 2015.

\bibitem[Izhikevich(2006)]{Izhikevich2006}
Eugene~M Izhikevich.
\newblock {Polychronization: computation with spikes.}
\newblock \emph{Neural Comput}, 18\penalty0 (2):\penalty0 245--282, feb 2006.
\newblock \doi{10.1162/089976606775093882}.
\newblock URL \url{http://dx.doi.org/10.1162/089976606775093882}.

\bibitem[Jacobsen et~al.(2016)Jacobsen, Van~Gemert, Lou, and Smeulders]{jacobsen2016structured}
Jorn-Henrik Jacobsen, Jan Van~Gemert, Zhongyu Lou, and Arnold~WM Smeulders.
\newblock Structured receptive fields in cnns.
\newblock In \emph{Proceedings of the IEEE Conference on Computer Vision and Pattern Recognition}, pages 2610--2619, 2016.

\bibitem[Jeon and Kim(2017)]{jeon2017active}
Yunho Jeon and Junmo Kim.
\newblock Active convolution: Learning the shape of convolution for image classification.
\newblock In \emph{Proceedings of the IEEE conference on computer vision and pattern recognition}, pages 4201--4209, 2017.

\bibitem[Jeon and Kim(2018)]{jeon2018constructing}
Yunho Jeon and Junmo Kim.
\newblock Constructing fast network through deconstruction of convolution.
\newblock \emph{Advances in Neural Information Processing Systems}, 31, 2018.

\bibitem[Jeon and Kim(2020)]{jeon2020integrating}
Yunho Jeon and Junmo Kim.
\newblock Integrating multiple receptive fields through grouped active convolution.
\newblock \emph{IEEE Transactions on Pattern Analysis and Machine Intelligence}, 43\penalty0 (11):\penalty0 3892--3903, 2020.

\bibitem[Jia et~al.(2014)Jia, Shelhamer, Donahue, Karayev, Long, Girshick, Guadarrama, and Darrell]{jia2014caffe}
Yangqing Jia, Evan Shelhamer, Jeff Donahue, Sergey Karayev, Jonathan Long, Ross Girshick, Sergio Guadarrama, and Trevor Darrell.
\newblock Caffe: Convolutional architecture for fast feature embedding.
\newblock \emph{arXiv preprint arXiv:1408.5093}, 2014.

\bibitem[Jiao et~al.(2023)Jiao, Tang, Lin, Gao, Ma, Wang, and Zheng]{jiao2023dilateformer}
Jiayu Jiao, Yu-Ming Tang, Kun-Yu Lin, Yipeng Gao, Jinhua Ma, Yaowei Wang, and Wei-Shi Zheng.
\newblock Dilateformer: Multi-scale dilated transformer for visual recognition.
\newblock \emph{IEEE Transactions on Multimedia}, 2023.

\bibitem[Kalchbrenner et~al.(2016)Kalchbrenner, Espeholt, Simonyan, Oord, Graves, and Kavukcuoglu]{kalchbrenner2016neural}
Nal Kalchbrenner, Lasse Espeholt, Karen Simonyan, Aaron van~den Oord, Alex Graves, and Koray Kavukcuoglu.
\newblock Neural machine translation in linear time.
\newblock \emph{arXiv preprint arXiv:1610.10099}, 2016.

\bibitem[Kay et~al.(2017)Kay, Carreira, Simonyan, Zhang, Hillier, Vijayanarasimhan, Viola, Green, Back, Natsev, et~al.]{kay2017kinetics}
Will Kay, Joao Carreira, Karen Simonyan, Brian Zhang, Chloe Hillier, Sudheendra Vijayanarasimhan, Fabio Viola, Tim Green, Trevor Back, Paul Natsev, et~al.
\newblock The kinetics human action video dataset.
\newblock \emph{arXiv preprint arXiv:1705.06950}, 2017.

\bibitem[Kerbl et~al.(2023)Kerbl, Kopanas, Leimk{\"u}hler, and Drettakis]{kerbl20233d}
Bernhard Kerbl, Georgios Kopanas, Thomas Leimk{\"u}hler, and George Drettakis.
\newblock 3d gaussian splatting for real-time radiance field rendering.
\newblock \emph{ACM Transactions on Graphics}, 42\penalty0 (4):\penalty0 1--14, 2023.

\bibitem[Khalfaoui-Hassani et~al.(2021)Khalfaoui-Hassani, Pellegrini, and Masquelier]{hassani2021dilated}
Ismail Khalfaoui-Hassani, Thomas Pellegrini, and Timoth{\'e}e Masquelier.
\newblock Dilated convolution with learnable spacings.
\newblock \emph{arXiv preprint arXiv:2112.03740}, 2021.

\bibitem[Khalfaoui-Hassani et~al.(2023{\natexlab{a}})Khalfaoui-Hassani, Masquelier, and Pellegrini]{khalfaoui2023audio}
Ismail Khalfaoui-Hassani, Timoth{\'e}e Masquelier, and Thomas Pellegrini.
\newblock Audio classification with dilated convolution with learnable spacings.
\newblock In \emph{NeurIPS 2023-Workshop on Machine Learning for Audio}, 2023{\natexlab{a}}.

\bibitem[Khalfaoui-Hassani et~al.(2023{\natexlab{b}})Khalfaoui-Hassani, Pellegrini, and Masquelier]{hassani2023dilated}
Ismail Khalfaoui-Hassani, Thomas Pellegrini, and Timoth{\'e}e Masquelier.
\newblock Dilated convolution with learnable spacings.
\newblock In \emph{The Eleventh International Conference on Learning Representations}, 2023{\natexlab{b}}.
\newblock URL \url{https://openreview.net/forum?id=Q3-1vRh3HOA}.

\bibitem[Khalfaoui-Hassani et~al.(2023{\natexlab{c}})Khalfaoui-Hassani, Pellegrini, and Masquelier]{khalfaouihassani2023dilated}
Ismail Khalfaoui-Hassani, Thomas Pellegrini, and Timoth{\'e}e Masquelier.
\newblock Dilated convolution with learnable spacings: beyond bilinear interpolation.
\newblock In \emph{ICML 2023 Workshop on Differentiable Almost Everything: Differentiable Relaxations, Algorithms, Operators, and Simulators}, 2023{\natexlab{c}}.
\newblock URL \url{https://openreview.net/forum?id=j8FPBCltB9}.

\bibitem[Kim et~al.(2023)Kim, Choi, Jang, and Kim]{kim2023understanding}
Bum~Jun Kim, Hyeyeon Choi, Hyeonah Jang, and Sang~Woo Kim.
\newblock Understanding gaussian attention bias of vision transformers using effective receptive fields.
\newblock \emph{arXiv preprint arXiv:2305.04722}, 2023.

\bibitem[Kim and Park(2023)]{kim2023smpconv}
Sanghyeon Kim and Eunbyung Park.
\newblock Smpconv: Self-moving point representations for continuous convolution.
\newblock \emph{arXiv preprint arXiv:2304.02330}, 2023.

\bibitem[Kingma and Ba(2015)]{adam}
Diederik~P. Kingma and Jimmy Ba.
\newblock Adam: {A} method for stochastic optimization.
\newblock In Yoshua Bengio and Yann LeCun, editors, \emph{3rd International Conference on Learning Representations, {ICLR} 2015, San Diego, CA, USA, May 7-9, 2015, Conference Track Proceedings}, 2015.
\newblock URL \url{http://arxiv.org/abs/1412.6980}.

\bibitem[Ko et~al.(2015)Ko, Peddinti, Povey, and Khudanpur]{ko15_interspeech}
Tom Ko, Vijayaditya Peddinti, Daniel Povey, and Sanjeev Khudanpur.
\newblock {Audio augmentation for speech recognition}.
\newblock In \emph{Proc. Interspeech 2015}, pages 3586--3589, 2015.
\newblock \doi{10.21437/Interspeech.2015-711}.

\bibitem[Kong et~al.(2020)Kong, Cao, Iqbal, Wang, Wang, and Plumbley]{Kong2020}
Qiuqiang Kong, Yin Cao, Turab Iqbal, Yuxuan Wang, Wenwu Wang, and Mark~D. Plumbley.
\newblock Panns: Large-scale pretrained audio neural networks for audio pattern recognition.
\newblock \emph{IEEE/ACM Transactions on Audio, Speech, and Language Processing}, 28:\penalty0 2880--2894, 2020.
\newblock \doi{10.1109/TASLP.2020.3030497}.

\bibitem[K{\"{o}}nig et~al.(1996)K{\"{o}}nig, Engel, and Singer]{Konig1996}
P~K{\"{o}}nig, A~K Engel, and W~Singer.
\newblock {Integrator or coincidence detector? The role of the cortical neuron revisited}.
\newblock \emph{Trends Neurosci}, 19\penalty0 (4):\penalty0 130--7., 1996.

\bibitem[Koutini et~al.(2022)Koutini, Schl{\"u}ter, Eghbal-zadeh, and Widmer]{koutini2021efficient}
Khaled Koutini, Jan Schl{\"u}ter, Hamid Eghbal-zadeh, and Gerhard Widmer.
\newblock Efficient training of audio transformers with patchout.
\newblock In \emph{Proc. Interspeech}, pages 2753--2757, Incheon, 2022.

\bibitem[Krizhevsky et~al.(2009)Krizhevsky, Hinton, et~al.]{krizhevsky2009learning}
Alex Krizhevsky, Geoffrey Hinton, et~al.
\newblock \emph{Learning multiple layers of features from tiny images}.
\newblock PhD thesis, University of Toronto, 2009.

\bibitem[Krizhevsky et~al.(2012)Krizhevsky, Sutskever, and Hinton]{alexnet}
Alex Krizhevsky, Ilya Sutskever, and Geoffrey~E Hinton.
\newblock Imagenet classification with deep convolutional neural networks.
\newblock In F.~Pereira, C.J. Burges, L.~Bottou, and K.Q. Weinberger, editors, \emph{Advances in Neural Information Processing Systems}, volume~25. Curran Associates, Inc., 2012.
\newblock URL \url{https://proceedings.neurips.cc/paper_files/paper/2012/file/c399862d3b9d6b76c8436e924a68c45b-Paper.pdf}.

\bibitem[Kundu et~al.(2019)Kundu, Prakash, Akrami, Beerel, and Chugg]{kundu2019pre}
Souvik Kundu, Saurav Prakash, Haleh Akrami, Peter~A Beerel, and Keith~M Chugg.
\newblock A pre-defined sparse kernel based convolution for deep cnns.
\newblock \emph{arXiv preprint arXiv:1910.00724}, 2019.

\bibitem[Lagrange(1868)]{lagrange1868oeuvres}
Joseph~Louis Lagrange.
\newblock \emph{Oeuvres de Lagrange}, volume~6.
\newblock Gauthier-Villars, 1868.

\bibitem[Larsson et~al.(2016)Larsson, Maire, and Shakhnarovich]{larsson2016fractalnet}
Gustav Larsson, Michael Maire, and Gregory Shakhnarovich.
\newblock Fractalnet: Ultra-deep neural networks without residuals.
\newblock In \emph{International Conference on Learning Representations}, 2016.

\bibitem[Lavin and Gray(2016)]{lavin2016fast}
Andrew Lavin and Scott Gray.
\newblock Fast algorithms for convolutional neural networks.
\newblock In \emph{Proceedings of the IEEE conference on computer vision and pattern recognition}, pages 4013--4021, 2016.

\bibitem[LeCun et~al.(1989)LeCun, Boser, Denker, Henderson, Howard, Hubbard, and Jackel]{lecun1989handwritten}
Yann LeCun, Bernhard Boser, John Denker, Donnie Henderson, Richard Howard, Wayne Hubbard, and Lawrence Jackel.
\newblock Handwritten digit recognition with a back-propagation network.
\newblock \emph{Advances in neural information processing systems}, 2, 1989.

\bibitem[LeCun et~al.(1998)LeCun, Bottou, Bengio, and Haffner]{lecun1998gradient}
Yann LeCun, L{\'e}on Bottou, Yoshua Bengio, and Patrick Haffner.
\newblock Gradient-based learning applied to document recognition.
\newblock \emph{Proceedings of the IEEE}, 86\penalty0 (11):\penalty0 2278--2324, 1998.

\bibitem[Li et~al.(2019)Li, Wang, Hu, and Yang]{li2019selective}
Xiang Li, Wenhai Wang, Xiaolin Hu, and Jian Yang.
\newblock Selective kernel networks.
\newblock In \emph{Proceedings of the IEEE/CVF conference on computer vision and pattern recognition}, pages 510--519, 2019.

\bibitem[Lin et~al.(2014)Lin, Maire, Belongie, Hays, Perona, Ramanan, Doll{\'a}r, and Zitnick]{lin2014microsoft}
Tsung-Yi Lin, Michael Maire, Serge Belongie, James Hays, Pietro Perona, Deva Ramanan, Piotr Doll{\'a}r, and C~Lawrence Zitnick.
\newblock Microsoft coco: Common objects in context.
\newblock In \emph{Proc. Eur. Conf. Comput. Vis. (ECCV)}, pages 740--755. Springer, 2014.

\bibitem[Linsley et~al.(2018)Linsley, Shiebler, Eberhardt, and Serre]{linsley2018learning}
Drew Linsley, Dan Shiebler, Sven Eberhardt, and Thomas Serre.
\newblock Learning what and where to attend.
\newblock \emph{arXiv preprint arXiv:1805.08819}, 2018.

\bibitem[Liu et~al.(2018)Liu, Lehman, Molino, Petroski~Such, Frank, Sergeev, and Yosinski]{liu2018intriguing}
Rosanne Liu, Joel Lehman, Piero Molino, Felipe Petroski~Such, Eric Frank, Alex Sergeev, and Jason Yosinski.
\newblock An intriguing failing of convolutional neural networks and the coordconv solution.
\newblock \emph{Advances in neural information processing systems}, 31, 2018.

\bibitem[Liu et~al.(2022{\natexlab{a}})Liu, Chen, Chen, Chen, Xiao, Wu, K{\"a}rkk{\"a}inen, Pechenizkiy, Mocanu, and Wang]{liu2022more}
Shiwei Liu, Tianlong Chen, Xiaohan Chen, Xuxi Chen, Qiao Xiao, Boqian Wu, Tommi K{\"a}rkk{\"a}inen, Mykola Pechenizkiy, Decebal~Constantin Mocanu, and Zhangyang Wang.
\newblock More convnets in the 2020s: Scaling up kernels beyond 51x51 using sparsity.
\newblock In \emph{The Eleventh International Conference on Learning Representations}, 2022{\natexlab{a}}.

\bibitem[Liu et~al.(2023)Liu, Chen, Chen, Chen, Xiao, Wu, K{\"a}rkk{\"a}inen, Pechenizkiy, Mocanu, and Wang]{liu2023more}
Shiwei Liu, Tianlong Chen, Xiaohan Chen, Xuxi Chen, Qiao Xiao, Boqian Wu, Tommi K{\"a}rkk{\"a}inen, Mykola Pechenizkiy, Decebal~Constantin Mocanu, and Zhangyang Wang.
\newblock More convnets in the 2020s: Scaling up kernels beyond 51x51 using sparsity.
\newblock In \emph{The Eleventh International Conference on Learning Representations}, 2023.
\newblock URL \url{https://openreview.net/forum?id=bXNl-myZkJl}.

\bibitem[Liu et~al.(2021)Liu, Lin, Cao, Hu, Wei, Zhang, Lin, and Guo]{liu2021swin}
Ze~Liu, Yutong Lin, Yue Cao, Han Hu, Yixuan Wei, Zheng Zhang, Stephen Lin, and Baining Guo.
\newblock Swin transformer: Hierarchical vision transformer using shifted windows.
\newblock In \emph{Int. Conf. Comput. Vis.}, pages 10012--10022, 2021.

\bibitem[Liu et~al.(2022{\natexlab{b}})Liu, Mao, Wu, Feichtenhofer, Darrell, and Xie]{liu2022convnet}
Zhuang Liu, Hanzi Mao, Chao-Yuan Wu, Christoph Feichtenhofer, Trevor Darrell, and Saining Xie.
\newblock A convnet for the 2020s.
\newblock In \emph{Proc. IEEE/CVF Conf. Comput. Vis. Pattern Recog. (CVPR)}, pages 11976--11986, 2022{\natexlab{b}}.

\bibitem[Loshchilov and Hutter(2016)]{loshchilov2016sgdr}
Ilya Loshchilov and Frank Hutter.
\newblock Sgdr: Stochastic gradient descent with warm restarts.
\newblock In \emph{International Conference on Learning Representations}, 2016.

\bibitem[Loshchilov and Hutter(2017{\natexlab{a}})]{adamw}
Ilya Loshchilov and Frank Hutter.
\newblock Decoupled weight decay regularization.
\newblock In \emph{International Conference on Learning Representations}, 2017{\natexlab{a}}.

\bibitem[Loshchilov and Hutter(2017{\natexlab{b}})]{cosine_annealing}
Ilya Loshchilov and Frank Hutter.
\newblock Sgdr: Stochastic gradient descent with warm restarts, 2017{\natexlab{b}}.

\bibitem[Lu et~al.(2019)Lu, Li, Yue, Li, and Yan]{lu2019grid}
Xin Lu, Buyu Li, Yuxin Yue, Quanquan Li, and Junjie Yan.
\newblock {Grid R-CNN}.
\newblock In \emph{Proc. IEEE/CVF Conf. Comput. Vis. Pattern Recog. (CVPR)}, pages 7363--7372, 2019.

\bibitem[{\L}ukaszyk(2004)]{lukaszyk2004new}
Szymon {\L}ukaszyk.
\newblock A new concept of probability metric and its applications in approximation of scattered data sets.
\newblock \emph{Computational mechanics}, 33:\penalty0 299--304, 2004.

\bibitem[Lundberg and Lee(2017)]{lundberg2017unified}
Scott~M Lundberg and Su-In Lee.
\newblock A unified approach to interpreting model predictions.
\newblock \emph{Advances in neural information processing systems}, 30, 2017.

\bibitem[Luo et~al.(2016)Luo, Li, Urtasun, and Zemel]{luo2016understanding}
Wenjie Luo, Yujia Li, Raquel Urtasun, and Richard Zemel.
\newblock Understanding the effective receptive field in deep convolutional neural networks.
\newblock In \emph{Proc. Adv. Neural Inform. Process. Syst. (NIPS)}, pages 4905--4913, 2016.

\bibitem[Maass and Schmitt(1999)]{Maass1999}
Wolfgang Maass and Michael Schmitt.
\newblock {On the Complexity of Learning for Spiking Neurons with Temporal Coding}.
\newblock \emph{Information and Computation}, 153\penalty0 (1):\penalty0 26--46, 1999.
\newblock ISSN 08905401.
\newblock \doi{10.1006/inco.1999.2806}.

\bibitem[Megvii(2020)]{MegEngine}
Megvii.
\newblock Megengine:a fast, scalable and easy-to-use deep learning framework.
\newblock \url{https://github.com/MegEngine/MegEngine}, 2020.

\bibitem[Mintun et~al.(2021)Mintun, Kirillov, and Xie]{mintun2021interaction}
Eric Mintun, Alexander Kirillov, and Saining Xie.
\newblock On interaction between augmentations and corruptions in natural corruption robustness.
\newblock \emph{Proc. Adv. Neural Inform. Process. Syst. (NIPS)}, 34:\penalty0 3571--3583, 2021.

\bibitem[Miyazaki et~al.(2020)Miyazaki, Komatsu, Hayashi, Watanabe, Toda, and Takeda]{miyazaki2020convolution}
Koichi Miyazaki, Tatsuya Komatsu, Tomoki Hayashi, Shinji Watanabe, Tomoki Toda, and Kazuya Takeda.
\newblock Convolution-augmented transformer for semi-supervised sound event detection.
\newblock In \emph{Proc. workshop detection classification Acoust. Scenes events (DCASE)}, pages 100--104, 2020.

\bibitem[Moore et~al.(2023)Moore, Ellis, Fonseca, Hershey, Jansen, and Plakal]{moore2023dataset}
R~Channing Moore, Daniel~PW Ellis, Eduardo Fonseca, Shawn Hershey, Aren Jansen, and Manoj Plakal.
\newblock Dataset balancing can hurt model performance.
\newblock In \emph{ICASSP 2023-2023 IEEE International Conference on Acoustics, Speech and Signal Processing (ICASSP)}, pages 1--5. IEEE, 2023.

\bibitem[Nair and Hinton(2010)]{nair2010rectified}
Vinod Nair and Geoffrey~E Hinton.
\newblock Rectified linear units improve restricted boltzmann machines.
\newblock In \emph{Proceedings of the 27th international conference on machine learning (ICML-10)}, pages 807--814, 2010.

\bibitem[Nauen et~al.(2023)Nauen, Palacio, and Dengel]{nauen2023transformer}
Tobias~Christian Nauen, Sebastian Palacio, and Andreas Dengel.
\newblock Which transformer to favor: A comparative analysis of efficiency in vision transformers.
\newblock \emph{arXiv preprint arXiv:2308.09372}, 2023.

\bibitem[Neftci et~al.(2018)Neftci, Mostafa, and Zenke]{neftci2018surrogate}
Emre~O Neftci, Hesham Mostafa, and Friedemann Zenke.
\newblock Surrogate gradient learning in spiking neural networks.
\newblock \emph{IEEE SIGNAL PROCESSING MAGAZINE}, 1053\penalty0 (5888/18), 2018.

\bibitem[Nowotny et~al.(2022)Nowotny, Turner, and Knight]{eventprop-genn}
Thomas Nowotny, James~P. Turner, and James~C. Knight.
\newblock Loss shaping enhances exact gradient learning with eventprop in spiking neural networks, 2022.

\bibitem[Oord et~al.(2016)Oord, Dieleman, Zen, Simonyan, Vinyals, Graves, Kalchbrenner, Senior, and Kavukcuoglu]{oord2016wavenet}
Aaron van~den Oord, Sander Dieleman, Heiga Zen, Karen Simonyan, Oriol Vinyals, Alex Graves, Nal Kalchbrenner, Andrew Senior, and Koray Kavukcuoglu.
\newblock {WaveNet: A generative model for raw audio}.
\newblock \emph{arXiv preprint arXiv:1609.03499}, 2016.

\bibitem[Park and Kim(2021)]{park2021vision}
Namuk Park and Songkuk Kim.
\newblock How do vision transformers work?
\newblock In \emph{International Conference on Learning Representations}, 2021.

\bibitem[Paszke et~al.(2019)Paszke, Gross, Massa, Lerer, Bradbury, Chanan, Killeen, Lin, Gimelshein, Antiga, Desmaison, Kopf, Yang, DeVito, Raison, Tejani, Chilamkurthy, Steiner, Fang, Bai, and Chintala]{PyTorch}
Adam Paszke, Sam Gross, Francisco Massa, Adam Lerer, James Bradbury, Gregory Chanan, Trevor Killeen, Zeming Lin, Natalia Gimelshein, Luca Antiga, Alban Desmaison, Andreas Kopf, Edward Yang, Zachary DeVito, Martin Raison, Alykhan Tejani, Sasank Chilamkurthy, Benoit Steiner, Lu~Fang, Junjie Bai, and Soumith Chintala.
\newblock Pytorch: An imperative style, high-performance deep learning library.
\newblock In \emph{Advances in Neural Information Processing Systems 32}, pages 8024--8035. Curran Associates, Inc., 2019.
\newblock URL \url{http://papers.neurips.cc/paper/9015-pytorch-an-imperative-style-high-performance-deep-learning-library.pdf}.

\bibitem[Pavllo et~al.(2019)Pavllo, Feichtenhofer, Grangier, and Auli]{pavllo20193d}
Dario Pavllo, Christoph Feichtenhofer, David Grangier, and Michael Auli.
\newblock 3d human pose estimation in video with temporal convolutions and semi-supervised training.
\newblock In \emph{Proceedings of the IEEE/CVF conference on computer vision and pattern recognition}, pages 7753--7762, 2019.

\bibitem[Pellegrini et~al.(2023)Pellegrini, Khalfaoui-Hassani, Labbé, and Masquelier]{pellegrini2023adapting}
Thomas Pellegrini, Ismail Khalfaoui-Hassani, Etienne Labbé, and Timothée Masquelier.
\newblock {Adapting a ConvNeXt Model to Audio Classification on AudioSet}.
\newblock In \emph{Proc. INTERSPEECH 2023}, pages 4169--4173, 2023.
\newblock \doi{10.21437/Interspeech.2023-1564}.

\bibitem[Peng et~al.(2017)Peng, Zhang, Yu, Luo, and Sun]{peng2017large}
Chao Peng, Xiangyu Zhang, Gang Yu, Guiming Luo, and Jian Sun.
\newblock Large kernel matters--improve semantic segmentation by global convolutional network.
\newblock In \emph{Proceedings of the IEEE conference on computer vision and pattern recognition}, pages 4353--4361, 2017.

\bibitem[Perez-Nieves et~al.(2021)Perez-Nieves, Leung, Dragotti, and Goodman]{heterogeneity}
Nicolas Perez-Nieves, Vincent C.~H. Leung, Pier~Luigi Dragotti, and Dan F.~M. Goodman.
\newblock Neural heterogeneity promotes robust learning.
\newblock \emph{Nature Communications}, 12\penalty0 (1):\penalty0 5791, Oct 2021.
\newblock ISSN 2041-1723.
\newblock \doi{10.1038/s41467-021-26022-3}.
\newblock URL \url{https://doi.org/10.1038/s41467-021-26022-3}.

\bibitem[Petsiuk et~al.(2018)Petsiuk, Das, and Saenko]{petsiuk2018rise}
Vitali Petsiuk, Abir Das, and Kate Saenko.
\newblock Rise: Randomized input sampling for explanation of black-box models.
\newblock \emph{arXiv preprint arXiv:1806.07421}, 2018.

\bibitem[Pintea et~al.(2021)Pintea, T{\"o}men, Goes, Loog, and van Gemert]{pintea2021resolution}
Silvia~L Pintea, Nergis T{\"o}men, Stanley~F Goes, Marco Loog, and Jan~C van Gemert.
\newblock Resolution learning in deep convolutional networks using scale-space theory.
\newblock \emph{IEEE Transactions on Image Processing}, 30:\penalty0 8342--8353, 2021.

\bibitem[Polyak and Juditsky(1992)]{polyak1992acceleration}
Boris~T Polyak and Anatoli~B Juditsky.
\newblock Acceleration of stochastic approximation by averaging.
\newblock \emph{SIAM journal on control and optimization}, 30\penalty0 (4):\penalty0 838--855, 1992.

\bibitem[Qi et~al.(2017)Qi, Zhang, Xiao, Hu, Cheng, Wei, and Dai]{qi2017deformable}
Haozhi Qi, Zheng Zhang, Bin Xiao, Han Hu, Bowen Cheng, Yichen Wei, and Jifeng Dai.
\newblock Deformable convolutional networks--coco detection and segmentation challenge 2017 entry.
\newblock In \emph{Proc. ICCV COCO Challenge Workshop}, volume~15, page~1, 2017.

\bibitem[Qiao et~al.(2021)Qiao, Chen, and Yuille]{qiao2021detectors}
Siyuan Qiao, Liang-Chieh Chen, and Alan Yuille.
\newblock Detectors: Detecting objects with recursive feature pyramid and switchable atrous convolution.
\newblock In \emph{Proceedings of the IEEE/CVF conference on computer vision and pattern recognition}, pages 10213--10224, 2021.

\bibitem[Radosavovic et~al.(2020)Radosavovic, Kosaraju, Girshick, He, and Doll{\'a}r]{radosavovic2020designing}
Ilija Radosavovic, Raj~Prateek Kosaraju, Ross Girshick, Kaiming He, and Piotr Doll{\'a}r.
\newblock Designing network design spaces.
\newblock In \emph{Proceedings of the IEEE/CVF conference on computer vision and pattern recognition}, pages 10428--10436, 2020.

\bibitem[Ramasubramanian et~al.(2023)Ramasubramanian, Rangwani, Takemori, Samanta, Umeda, and Venkatesh~Babu]{Ramasubramanian2023selmix}
Shrinivas Ramasubramanian, Harsh Rangwani, Sho Takemori, Kunal Samanta, Yuhei Umeda, and R.~Venkatesh~Babu.
\newblock Selmix: Selective mixup fine tuning for optimizing non-decomposable metrics.
\newblock In \emph{The Differentiable Almost Everything Workshop of the 40 th International Conference on Machine Learning}, 2023.

\bibitem[Ren et~al.(2020)Ren, Hu, Tan, Qin, Zhao, Zhao, and Liu]{ren2020fastspeech}
Yi~Ren, Chenxu Hu, Xu~Tan, Tao Qin, Sheng Zhao, Zhou Zhao, and Tie-Yan Liu.
\newblock Fastspeech 2: Fast and high-quality end-to-end text to speech.
\newblock In \emph{International Conference on Learning Representations}, 2020.

\bibitem[Riad et~al.(2021)Riad, Teboul, Grangier, and Zeghidour]{riad2021learning}
Rachid Riad, Olivier Teboul, David Grangier, and Neil Zeghidour.
\newblock Learning strides in convolutional neural networks.
\newblock In \emph{International Conference on Learning Representations}, 2021.

\bibitem[Rippel et~al.(2015)Rippel, Snoek, and Adams]{rippel2015spectral}
Oren Rippel, Jasper Snoek, and Ryan~P Adams.
\newblock Spectral representations for convolutional neural networks.
\newblock \emph{Advances in neural information processing systems}, 28, 2015.

\bibitem[Romero et~al.(2021{\natexlab{a}})Romero, Bruintjes, Tomczak, Bekkers, Hoogendoorn, and van Gemert]{romero2021flexconv}
David~W Romero, Robert-Jan Bruintjes, Jakub~M Tomczak, Erik~J Bekkers, Mark Hoogendoorn, and Jan~C van Gemert.
\newblock Flexconv: Continuous kernel convolutions with differentiable kernel sizes.
\newblock \emph{arXiv preprint arXiv:2110.08059}, 2021{\natexlab{a}}.

\bibitem[Romero et~al.(2021{\natexlab{b}})Romero, Kuzina, Bekkers, Tomczak, and Hoogendoorn]{romero2021ckconv}
David~W Romero, Anna Kuzina, Erik~J Bekkers, Jakub~M Tomczak, and Mark Hoogendoorn.
\newblock Ckconv: Continuous kernel convolution for sequential data.
\newblock \emph{arXiv preprint arXiv:2102.02611}, 2021{\natexlab{b}}.

\bibitem[Romero et~al.(2022{\natexlab{a}})Romero, Bruintjes, Bekkers, Tomczak, Hoogendoorn, and van Gemert]{romero2022flexconv}
David~W Romero, R~Bruintjes, Erik~J Bekkers, Jakub~M Tomczak, Mark Hoogendoorn, and JC~van Gemert.
\newblock Flexconv: Continuous kernel convolutions with differentiable kernel sizes.
\newblock In \emph{10th International Conference on Learning Representations}, 2022{\natexlab{a}}.

\bibitem[Romero et~al.(2022{\natexlab{b}})Romero, Kuzina, Bekkers, Tomczak, and Hoogendoorn]{romero2022ckconv}
David~W. Romero, Anna Kuzina, Erik~J Bekkers, Jakub~Mikolaj Tomczak, and Mark Hoogendoorn.
\newblock {CKC}onv: Continuous kernel convolution for sequential data.
\newblock In \emph{International Conference on Learning Representations}, 2022{\natexlab{b}}.
\newblock URL \url{https://openreview.net/forum?id=8FhxBtXSl0}.

\bibitem[Rossant et~al.(2011)Rossant, Leijon, Magnusson, and Brette]{Rossant2011}
Cyrille Rossant, Sara Leijon, Anna~K Magnusson, and Romain Brette.
\newblock {Sensitivity of noisy neurons to coincident inputs.}
\newblock \emph{The Journal of Neuroscience}, 31\penalty0 (47):\penalty0 17193--206, nov 2011.
\newblock ISSN 1529-2401.
\newblock \doi{10.1523/JNEUROSCI.2482-11.2011}.
\newblock URL \url{http://www.ncbi.nlm.nih.gov/pubmed/22114286}.

\bibitem[Sadovsky et~al.(2023)Sadovsky, Jakubec, and Jarina]{ieee_cnn}
Erik Sadovsky, Maros Jakubec, and Roman Jarina.
\newblock Speech command recognition based on convolutional spiking neural networks.
\newblock In \emph{2023 33rd International Conference Radioelektronika (RADIOELEKTRONIKA)}, pages 1--5, 2023.
\newblock \doi{10.1109/RADIOELEKTRONIKA57919.2023.10109082}.

\bibitem[Sandler et~al.(2018)Sandler, Howard, Zhu, Zhmoginov, and Chen]{sandler2018mobilenetv2}
Mark Sandler, Andrew Howard, Menglong Zhu, Andrey Zhmoginov, and Liang-Chieh Chen.
\newblock Mobilenetv2: Inverted residuals and linear bottlenecks.
\newblock In \emph{Proceedings of the IEEE conference on computer vision and pattern recognition}, pages 4510--4520, 2018.

\bibitem[Selvaraju et~al.(2017)Selvaraju, Cogswell, Das, Vedantam, Parikh, and Batra]{selvaraju2017grad}
Ramprasaath~R Selvaraju, Michael Cogswell, Abhishek Das, Ramakrishna Vedantam, Devi Parikh, and Dhruv Batra.
\newblock Grad-cam: Visual explanations from deep networks via gradient-based localization.
\newblock In \emph{Proceedings of the IEEE international conference on computer vision}, pages 618--626, 2017.

\bibitem[Shelhamer et~al.(2019)Shelhamer, Wang, and Darrell]{shelhamer2019blurring}
Evan Shelhamer, Dequan Wang, and Trevor Darrell.
\newblock Blurring the line between structure and learning to optimize and adapt receptive fields.
\newblock \emph{arXiv preprint arXiv:1904.11487}, 2019.

\bibitem[Shensa(1992)]{shensa1992discrete}
Mark~J Shensa.
\newblock The discrete wavelet transform: wedding the a trous and mallat algorithms.
\newblock \emph{IEEE Trans. Signal Process.}, 40\penalty0 (10):\penalty0 2464--2482, 1992.

\bibitem[Shepard(1968)]{shepard1968two}
Donald Shepard.
\newblock A two-dimensional interpolation function for irregularly-spaced data.
\newblock In \emph{Proceedings of the 1968 23rd ACM national conference}, pages 517--524, 1968.

\bibitem[Shrestha and Orchard(2018)]{slayer}
Sumit~Bam Shrestha and Garrick Orchard.
\newblock {SLAYER}: Spike layer error reassignment in time.
\newblock In S.~Bengio, H.~Wallach, H.~Larochelle, K.~Grauman, N.~Cesa-Bianchi, and R.~Garnett, editors, \emph{Advances in Neural Information Processing Systems 31}, pages 1419--1428. Curran Associates, Inc., 2018.
\newblock URL \url{http://papers.nips.cc/paper/7415-slayer-spike-layer-error-reassignment-in-time.pdf}.

\bibitem[Sifre and Mallat(2014)]{sifre2014rigid}
Laurent Sifre and St{\'e}phane Mallat.
\newblock Rigid-motion scattering for texture classification.
\newblock \emph{arXiv preprint arXiv:1403.1687}, 2014.

\bibitem[Simonyan and Zisserman(2014)]{simonyan2014very}
Karen Simonyan and Andrew Zisserman.
\newblock Very deep convolutional networks for large-scale image recognition.
\newblock \emph{arXiv preprint arXiv:1409.1556}, 2014.

\bibitem[Singh et~al.(2023)Singh, Liu, and Plumbley]{singh2023panns}
Arshdeep Singh, Haohe Liu, and Mark~D Plumbley.
\newblock {E-PANNs: Sound Recognition Using Efficient Pre-trained Audio Neural Networks}.
\newblock In \emph{Proc. Inter Noise}, Chiba, 2023.

\bibitem[Smith and Topin(2018)]{one_cycle}
Leslie~N. Smith and Nicholay Topin.
\newblock Super-convergence: Very fast training of neural networks using large learning rates, 2018.

\bibitem[Smith et~al.(2023)Smith, Brock, Berrada, and De]{smith2023convnets}
Samuel~L Smith, Andrew Brock, Leonard Berrada, and Soham De.
\newblock Convnets match vision transformers at scale.
\newblock \emph{arXiv preprint arXiv:2310.16764}, 2023.

\bibitem[Sosnovik et~al.(2019)Sosnovik, Szmaja, and Smeulders]{sosnovik2019scale}
Ivan Sosnovik, Micha{\l} Szmaja, and Arnold Smeulders.
\newblock Scale-equivariant steerable networks.
\newblock \emph{arXiv preprint arXiv:1910.11093}, 2019.

\bibitem[Sosnovik et~al.(2021{\natexlab{a}})Sosnovik, Moskalev, and Smeulders]{sosnovik2021disco}
Ivan Sosnovik, Artem Moskalev, and Arnold Smeulders.
\newblock Disco: accurate discrete scale convolutions.
\newblock \emph{arXiv preprint arXiv:2106.02733}, 2021{\natexlab{a}}.

\bibitem[Sosnovik et~al.(2021{\natexlab{b}})Sosnovik, Moskalev, and Smeulders]{sosnovik2021transform}
Ivan Sosnovik, Artem Moskalev, and Arnold Smeulders.
\newblock How to transform kernels for scale-convolutions.
\newblock In \emph{Proceedings of the IEEE/CVF International Conference on Computer Vision}, pages 1092--1097, 2021{\natexlab{b}}.

\bibitem[Spearman(1961)]{spearman1961proof}
Charles Spearman.
\newblock The proof and measurement of association between two things.
\newblock \emph{American Journal of Psychology}, 1961.

\bibitem[Stevens et~al.(1937)Stevens, Volkmann, and Newman]{stevens1937scale}
Stanley~Smith Stevens, John Volkmann, and Edwin~Broomell Newman.
\newblock A scale for the measurement of the psychological magnitude pitch.
\newblock \emph{The journal of the acoustical society of america}, 8\penalty0 (3):\penalty0 185--190, 1937.

\bibitem[Sukhbaatar et~al.(2019)Sukhbaatar, Grave, Bojanowski, and Joulin]{sukhbaatar2019adaptive}
Sainbayar Sukhbaatar, {\'E}douard Grave, Piotr Bojanowski, and Armand Joulin.
\newblock Adaptive attention span in transformers.
\newblock In \emph{Proceedings of the 57th Annual Meeting of the Association for Computational Linguistics}, pages 331--335, 2019.

\bibitem[Sun et~al.(2022)Sun, Zhu, and Botteldooren]{sun22}
Pengfei Sun, Longwei Zhu, and Dick Botteldooren.
\newblock Axonal delay as a short-term memory for feed forward deep spiking neural networks.
\newblock In \emph{ICASSP 2022 - 2022 IEEE International Conference on Acoustics, Speech and Signal Processing (ICASSP)}, pages 8932--8936, 2022.
\newblock \doi{10.1109/ICASSP43922.2022.9747411}.

\bibitem[Sun et~al.(2023{\natexlab{a}})Sun, Chua, Devos, and Botteldooren]{sun23-2}
Pengfei Sun, Yansong Chua, Paul Devos, and Dick Botteldooren.
\newblock Learnable axonal delay in spiking neural networks improves spoken word recognition.
\newblock \emph{Frontiers in Neuroscience}, 17, 2023{\natexlab{a}}.
\newblock ISSN 1662-453X.
\newblock \doi{10.3389/fnins.2023.1275944}.
\newblock URL \url{https://www.frontiersin.org/articles/10.3389/fnins.2023.1275944}.

\bibitem[Sun et~al.(2023{\natexlab{b}})Sun, Eqlimi, Chua, Devos, and Botteldooren]{sun23}
Pengfei Sun, Ehsan Eqlimi, Yansong Chua, Paul Devos, and Dick Botteldooren.
\newblock Adaptive axonal delays in feedforward spiking neural networks for accurate spoken word recognition, 2023{\natexlab{b}}.

\bibitem[Szegedy et~al.(2016)Szegedy, Vanhoucke, Ioffe, Shlens, and Wojna]{szegedy2016rethinking}
Christian Szegedy, Vincent Vanhoucke, Sergey Ioffe, Jon Shlens, and Zbigniew Wojna.
\newblock Rethinking the inception architecture for computer vision.
\newblock In \emph{Proceedings of the IEEE conference on computer vision and pattern recognition}, pages 2818--2826, 2016.

\bibitem[Tabernik et~al.(2016)Tabernik, Kristan, Wyatt, and Leonardis]{tabernik2016towards}
Domen Tabernik, Matej Kristan, Jeremy~L Wyatt, and Ale{\v{s}} Leonardis.
\newblock Towards deep compositional networks.
\newblock In \emph{2016 23rd international conference on pattern recognition (ICPR)}, pages 3470--3475. IEEE, 2016.

\bibitem[Tabernik et~al.(2018)Tabernik, Kristan, and Leonardis]{tabernik2018spatially}
Domen Tabernik, Matej Kristan, and Ale{\v{s}} Leonardis.
\newblock Spatially-adaptive filter units for deep neural networks.
\newblock In \emph{Proceedings of the IEEE Conference on Computer Vision and Pattern Recognition}, pages 9388--9396, 2018.

\bibitem[Tabernik et~al.(2020)Tabernik, Kristan, and Leonardis]{tabernik2020spatially}
Domen Tabernik, Matej Kristan, and Ale{\v{s}} Leonardis.
\newblock Spatially-adaptive filter units for compact and efficient deep neural networks.
\newblock \emph{International Journal of Computer Vision}, 128\penalty0 (8):\penalty0 2049--2067, 2020.

\bibitem[Taesiri et~al.(2023)Taesiri, Nguyen, Habchi, Bezemer, and Nguyen]{taesiri2023zoom}
Mohammad~Reza Taesiri, Giang Nguyen, Sarra Habchi, Cor-Paul Bezemer, and Anh Nguyen.
\newblock Zoom is what you need: An empirical study of the power of zoom and spatial biases in image classification.
\newblock \emph{arXiv preprint arXiv:2304.05538}, 2023.

\bibitem[Taherkhani et~al.(2015)Taherkhani, Belatreche, Li, and Maguire]{related_delays3}
Aboozar Taherkhani, Ammar Belatreche, Yuhua Li, and Liam~P. Maguire.
\newblock Dl-resume: A delay learning-based remote supervised method for spiking neurons.
\newblock \emph{IEEE Transactions on Neural Networks and Learning Systems}, 26\penalty0 (12):\penalty0 3137--3149, 2015.
\newblock \doi{10.1109/TNNLS.2015.2404938}.

\bibitem[Tan et~al.(2022)Tan, Wang, Han, Cao, Wu, and Zha]{tan2022multi}
Ganchao Tan, Yang Wang, Han Han, Yang Cao, Feng Wu, and Zheng-Jun Zha.
\newblock Multi-grained spatio-temporal features perceived network for event-based lip-reading.
\newblock In \emph{Proceedings of the IEEE/CVF Conference on Computer Vision and Pattern Recognition}, pages 20094--20103, 2022.

\bibitem[Tan and Le(2019{\natexlab{a}})]{tan2019efficientnet}
Mingxing Tan and Quoc Le.
\newblock Efficientnet: Rethinking model scaling for convolutional neural networks.
\newblock In \emph{International conference on machine learning}, pages 6105--6114. PMLR, 2019{\natexlab{a}}.

\bibitem[Tan and Le(2021)]{tan2021efficientnetv2}
Mingxing Tan and Quoc Le.
\newblock Efficientnetv2: Smaller models and faster training.
\newblock In \emph{International Conference on Machine Learning}, pages 10096--10106. PMLR, 2021.

\bibitem[Tan and Le(2019{\natexlab{b}})]{tan2019mixconv}
Mingxing Tan and Quoc~V Le.
\newblock Mixconv: Mixed depthwise convolutional kernels.
\newblock \emph{arXiv preprint arXiv:1907.09595}, 2019{\natexlab{b}}.

\bibitem[Tan et~al.(2019)Tan, Chen, Pang, Vasudevan, Sandler, Howard, and Le]{Tan_2019_CVPR}
Mingxing Tan, Bo~Chen, Ruoming Pang, Vijay Vasudevan, Mark Sandler, Andrew Howard, and Quoc~V. Le.
\newblock Mnasnet: Platform-aware neural architecture search for mobile.
\newblock In \emph{Proceedings of the IEEE/CVF Conference on Computer Vision and Pattern Recognition (CVPR)}, June 2019.

\bibitem[Thomas et~al.(2019)Thomas, Qi, Deschaud, Marcotegui, Goulette, and Guibas]{thomas2019KPConv}
Hugues Thomas, Charles~R. Qi, Jean-Emmanuel Deschaud, Beatriz Marcotegui, Fran{\c{c}}ois Goulette, and Leonidas~J. Guibas.
\newblock Kpconv: Flexible and deformable convolution for point clouds.
\newblock \emph{Int. Conf. Comput. Vis.}, 2019.

\bibitem[Tolstikhin et~al.(2021)Tolstikhin, Houlsby, Kolesnikov, Beyer, Zhai, Unterthiner, Yung, Steiner, Keysers, Uszkoreit, et~al.]{tolstikhin2021mlp}
Ilya~O Tolstikhin, Neil Houlsby, Alexander Kolesnikov, Lucas Beyer, Xiaohua Zhai, Thomas Unterthiner, Jessica Yung, Andreas Steiner, Daniel Keysers, Jakob Uszkoreit, et~al.
\newblock Mlp-mixer: An all-mlp architecture for vision.
\newblock \emph{Advances in neural information processing systems}, 34:\penalty0 24261--24272, 2021.

\bibitem[Trockman and Kolter(2022)]{trockman2022patches}
Asher Trockman and J~Zico Kolter.
\newblock Patches are all you need?
\newblock \emph{arXiv preprint arXiv:2201.09792}, 2022.

\bibitem[Vasu et~al.(2023{\natexlab{a}})Vasu, Gabriel, Zhu, Tuzel, and Ranjan]{vasu2023fastvit}
Pavan Kumar~Anasosalu Vasu, James Gabriel, Jeff Zhu, Oncel Tuzel, and Anurag Ranjan.
\newblock Fastvit: A fast hybrid vision transformer using structural reparameterization.
\newblock \emph{arXiv preprint arXiv:2303.14189}, 2023{\natexlab{a}}.

\bibitem[Vasu et~al.(2023{\natexlab{b}})Vasu, Gabriel, Zhu, Tuzel, and Ranjan]{vasu2023mobileone}
Pavan Kumar~Anasosalu Vasu, James Gabriel, Jeff Zhu, Oncel Tuzel, and Anurag Ranjan.
\newblock Mobileone: An improved one millisecond mobile backbone.
\newblock In \emph{Proceedings of the IEEE/CVF Conference on Computer Vision and Pattern Recognition}, pages 7907--7917, 2023{\natexlab{b}}.

\bibitem[Vaswani et~al.(2017{\natexlab{a}})Vaswani, Shazeer, Parmar, Uszkoreit, Jones, Gomez, Kaiser, and Polosukhin]{transformer}
Ashish Vaswani, Noam Shazeer, Niki Parmar, Jakob Uszkoreit, Llion Jones, Aidan~N Gomez, \L~ukasz Kaiser, and Illia Polosukhin.
\newblock Attention is all you need.
\newblock In I.~Guyon, U.~Von Luxburg, S.~Bengio, H.~Wallach, R.~Fergus, S.~Vishwanathan, and R.~Garnett, editors, \emph{Advances in Neural Information Processing Systems}, volume~30. Curran Associates, Inc., 2017{\natexlab{a}}.
\newblock URL \url{https://proceedings.neurips.cc/paper_files/paper/2017/file/3f5ee243547dee91fbd053c1c4a845aa-Paper.pdf}.

\bibitem[Vaswani et~al.(2017{\natexlab{b}})Vaswani, Shazeer, Parmar, Uszkoreit, Jones, Gomez, Kaiser, and Polosukhin]{vaswani2017attention}
Ashish Vaswani, Noam Shazeer, Niki Parmar, Jakob Uszkoreit, Llion Jones, Aidan~N Gomez, {\L}ukasz Kaiser, and Illia Polosukhin.
\newblock Attention is all you need.
\newblock \emph{Advances in neural information processing systems}, 30, 2017{\natexlab{b}}.

\bibitem[Verbitskiy et~al.(2022)Verbitskiy, Berikov, and Vyshegorodtsev]{verbitskiy2022eranns}
Sergey Verbitskiy, Vladimir Berikov, and Viacheslav Vyshegorodtsev.
\newblock Eranns: Efficient residual audio neural networks for audio pattern recognition.
\newblock \emph{Pattern Recognition Letters}, 161:\penalty0 38--44, 2022.

\bibitem[Wang et~al.(2019{\natexlab{a}})Wang, Ge, Lipton, and Xing]{wang2019learning}
Haohan Wang, Songwei Ge, Zachary Lipton, and Eric~P Xing.
\newblock Learning robust global representations by penalizing local predictive power.
\newblock \emph{Proc. Adv. Neural Inform. Process. Syst. (NIPS)}, 32, 2019{\natexlab{a}}.

\bibitem[Wang et~al.(2022)Wang, Dai, Chen, Huang, Li, Zhu, Hu, Lu, Lu, Li, et~al.]{wang2022internimage}
Wenhai Wang, Jifeng Dai, Zhe Chen, Zhenhang Huang, Zhiqi Li, Xizhou Zhu, Xiaowei Hu, Tong Lu, Lewei Lu, Hongsheng Li, et~al.
\newblock Internimage: Exploring large-scale vision foundation models with deformable convolutions.
\newblock \emph{arXiv preprint arXiv:2211.05778}, 2022.

\bibitem[Wang et~al.(2019{\natexlab{b}})Wang, Lin, and Dang]{Delay_Learning_Kernel}
Xiangwen Wang, Xianghong Lin, and Xiaochao Dang.
\newblock A delay learning algorithm based on spike train kernels for spiking neurons.
\newblock \emph{Frontiers in Neuroscience}, 13, 2019{\natexlab{b}}.
\newblock ISSN 1662-453X.
\newblock \doi{10.3389/fnins.2019.00252}.
\newblock URL \url{https://www.frontiersin.org/articles/10.3389/fnins.2019.00252}.

\bibitem[Wang and Ji(2018)]{wang2018smoothed}
Zhengyang Wang and Shuiwang Ji.
\newblock Smoothed dilated convolutions for improved dense prediction.
\newblock In \emph{Proceedings of the 24th ACM SIGKDD International Conference on Knowledge Discovery \& Data Mining}, pages 2486--2495, 2018.

\bibitem[Warden(2018)]{SC}
Pete Warden.
\newblock Speech commands: A dataset for limited-vocabulary speech recognition, 2018.

\bibitem[Wei et~al.(2022)Wei, Tian, Wang, Liang, and Chen]{wei2022optimized}
Tao Wei, Yonghong Tian, Yaowei Wang, Yun Liang, and Chang~Wen Chen.
\newblock Optimized separable convolution: Yet another efficient convolution operator.
\newblock \emph{AI Open}, 3:\penalty0 162--171, 2022.

\bibitem[Wightman et~al.(2021)Wightman, Touvron, and J{\'e}gou]{wightman2021resnet}
Ross Wightman, Hugo Touvron, and Herv{\'e} J{\'e}gou.
\newblock Resnet strikes back: An improved training procedure in timm.
\newblock \emph{arXiv preprint arXiv:2110.00476}, 2021.

\bibitem[Worrall and Welling(2019)]{worrall2019deep}
Daniel Worrall and Max Welling.
\newblock Deep scale-spaces: Equivariance over scale.
\newblock \emph{Advances in Neural Information Processing Systems}, 32, 2019.

\bibitem[Wu et~al.(2018)Wu, Wan, Yue, Jin, Zhao, Golmant, Gholaminejad, Gonzalez, and Keutzer]{wu2018shift}
Bichen Wu, Alvin Wan, Xiangyu Yue, Peter Jin, Sicheng Zhao, Noah Golmant, Amir Gholaminejad, Joseph Gonzalez, and Kurt Keutzer.
\newblock Shift: A zero flop, zero parameter alternative to spatial convolutions.
\newblock In \emph{Proceedings of the IEEE conference on computer vision and pattern recognition}, pages 9127--9135, 2018.

\bibitem[Xiao et~al.(2018)Xiao, Liu, Zhou, Jiang, and Sun]{xiao2018unified}
Tete Xiao, Yingcheng Liu, Bolei Zhou, Yuning Jiang, and Jian Sun.
\newblock Unified perceptual parsing for scene understanding.
\newblock In \emph{Proc. Eur. Conf. Comput. Vis. (ECCV)}, pages 418--434, 2018.

\bibitem[Xie et~al.(2016)Xie, Girshick, Doll{\'{a}}r, Tu, and He]{resnext}
Saining Xie, Ross~B. Girshick, Piotr Doll{\'{a}}r, Zhuowen Tu, and Kaiming He.
\newblock Aggregated residual transformations for deep neural networks.
\newblock \emph{CoRR}, abs/1611.05431, 2016.
\newblock URL \url{http://arxiv.org/abs/1611.05431}.

\bibitem[Xu et~al.(2023)Xu, Zhang, Hu, Zhang, and Wang]{xu2023parcnetv2}
Ruihan Xu, Haokui Zhang, Wenze Hu, Shiliang Zhang, and Xiaoyu Wang.
\newblock Parcnetv2: Oversized kernel with enhanced attention.
\newblock In \emph{Proceedings of the IEEE/CVF International Conference on Computer Vision}, pages 5752--5762, 2023.

\bibitem[Yamamoto et~al.(2020)Yamamoto, Song, and Kim]{yamamoto2020parallel}
Ryuichi Yamamoto, Eunwoo Song, and Jae-Min Kim.
\newblock {Parallel WaveGAN: A fast waveform generation model based on generative adversarial networks with multi-resolution spectrogram}.
\newblock In \emph{Proc. ICASSP}, pages 6199--6203. IEEE, 2020.

\bibitem[Yang et~al.(2019)Yang, Wang, Wong, Chao, and Tu]{yang2019convolutional}
Baosong Yang, Longyue Wang, Derek~F Wong, Lidia~S Chao, and Zhaopeng Tu.
\newblock Convolutional self-attention networks.
\newblock In \emph{Proceedings of the 2019 Conference of the North American Chapter of the Association for Computational Linguistics: Human Language Technologies, Volume 1 (Long and Short Papers)}, pages 4040--4045, 2019.

\bibitem[Yao et~al.(2022)Yao, Wang, Hu, Xing, and Wang]{yao2022adcnn}
Jie Yao, Dongdong Wang, Hao Hu, Weiwei Xing, and Liqiang Wang.
\newblock Adcnn: Towards learning adaptive dilation for convolutional neural networks.
\newblock \emph{Pattern Recognition}, 123:\penalty0 108369, 2022.

\bibitem[Yao et~al.(2021)Yao, Gao, Zhao, Wang, Lin, Yang, and Li]{TA-SNN}
M.~Yao, H.~Gao, G.~Zhao, D.~Wang, Y.~Lin, Z.~Yang, and G.~Li.
\newblock Temporal-wise attention spiking neural networks for event streams classification.
\newblock In \emph{2021 IEEE/CVF International Conference on Computer Vision (ICCV)}, pages 10201--10210, Los Alamitos, CA, USA, oct 2021. IEEE Computer Society.
\newblock \doi{10.1109/ICCV48922.2021.01006}.
\newblock URL \url{https://doi.ieeecomputersociety.org/10.1109/ICCV48922.2021.01006}.

\bibitem[Yin et~al.(2021)Yin, Corradi, and Bohte]{Adaptive-SRNN}
Bojian Yin, Federico Corradi, and Sander~M. Bohte.
\newblock Accurate and efficient time-domain classification with adaptive spiking recurrent neural networks.
\newblock \emph{Nature Machine Intelligence}, 2021.
\newblock \doi{10.1038/s42256-021-00397-w}.
\newblock URL \url{https://doi.org/10.1038/s42256-021-00397-w}.

\bibitem[You et~al.(2019)You, Li, Reddi, Hseu, Kumar, Bhojanapalli, Song, Demmel, Keutzer, and Hsieh]{you2019large}
Yang You, Jing Li, Sashank Reddi, Jonathan Hseu, Sanjiv Kumar, Srinadh Bhojanapalli, Xiaodan Song, James Demmel, Kurt Keutzer, and Cho-Jui Hsieh.
\newblock Large batch optimization for deep learning: Training bert in 76 minutes.
\newblock \emph{arXiv preprint arXiv:1904.00962}, 2019.

\bibitem[Yousefzadeh et~al.(2022)Yousefzadeh, van Schaik, Tahghighi, Detterer, Traferro, Hijdra, Stuijt, Corradi, Sifalakis, and Konijnenburg]{Yousefzadeh2022}
Amirreza Yousefzadeh, Gert-Jan van Schaik, Mohammad Tahghighi, Paul Detterer, Stefano Traferro, Martijn Hijdra, Jan Stuijt, Federico Corradi, Manolis Sifalakis, and Mario Konijnenburg.
\newblock {SENeCA: Scalable Energy-efficient Neuromorphic Computer Architecture}.
\newblock In \emph{2022 IEEE 4th International Conference on Artificial Intelligence Circuits and Systems (AICAS)}, pages 371--374. IEEE, jun 2022.
\newblock ISBN 978-1-6654-0996-4.
\newblock \doi{10.1109/AICAS54282.2022.9870025}.
\newblock URL \url{https://ieeexplore.ieee.org/document/9870025/}.

\bibitem[Yu et~al.(2022{\natexlab{a}})Yu, Gu, Li, Wang, Wang, and Li]{FFSNNattention}
Chengting Yu, Zheming Gu, Da~Li, Gaoang Wang, Aili Wang, and Erping Li.
\newblock Stsc-snn: Spatio-temporal synaptic connection with temporal convolution and attention for spiking neural networks.
\newblock \emph{Frontiers in Neuroscience}, 16, 2022{\natexlab{a}}.
\newblock ISSN 1662-453X.
\newblock \doi{10.3389/fnins.2022.1079357}.
\newblock URL \url{https://www.frontiersin.org/articles/10.3389/fnins.2022.1079357}.

\bibitem[Yu and Koltun(2015)]{yu2015multi}
Fisher Yu and Vladlen Koltun.
\newblock Multi-scale context aggregation by dilated convolutions.
\newblock \emph{arXiv preprint arXiv:1511.07122}, 2015.

\bibitem[Yu et~al.(2017)Yu, Koltun, and Funkhouser]{yu2017dilated}
Fisher Yu, Vladlen Koltun, and Thomas Funkhouser.
\newblock Dilated residual networks.
\newblock In \emph{Proc. IEEE/CVF Conf. Comput. Vis. Pattern Recog. (CVPR)}, pages 472--480, 2017.

\bibitem[Yu et~al.(2022{\natexlab{b}})Yu, Luo, Zhou, Si, Zhou, Wang, Feng, and Yan]{yu2022poolformer}
Weihao Yu, Mi~Luo, Pan Zhou, Chenyang Si, Yichen Zhou, Xinchao Wang, Jiashi Feng, and Shuicheng Yan.
\newblock Metaformer is actually what you need for vision.
\newblock In \emph{Proceedings of the IEEE/CVF conference on computer vision and pattern recognition}, pages 10819--10829, 2022{\natexlab{b}}.

\bibitem[Yu et~al.(2022{\natexlab{c}})Yu, Si, Zhou, Luo, Zhou, Feng, Yan, and Wang]{yu2022metaformer}
Weihao Yu, Chenyang Si, Pan Zhou, Mi~Luo, Yichen Zhou, Jiashi Feng, Shuicheng Yan, and Xinchao Wang.
\newblock Metaformer baselines for vision.
\newblock \emph{arXiv preprint arXiv:2210.13452}, 2022{\natexlab{c}}.

\bibitem[Yu et~al.(2023)Yu, Zhou, Yan, and Wang]{yu2023inceptionnext}
Weihao Yu, Pan Zhou, Shuicheng Yan, and Xinchao Wang.
\newblock Inceptionnext: When inception meets convnext.
\newblock \emph{arXiv preprint arXiv:2303.16900}, 2023.

\bibitem[Yun et~al.(2019)Yun, Han, Oh, Chun, Choe, and Yoo]{yun2019cutmix}
Sangdoo Yun, Dongyoon Han, Seong~Joon Oh, Sanghyuk Chun, Junsuk Choe, and Youngjoon Yoo.
\newblock Cutmix: Regularization strategy to train strong classifiers with localizable features.
\newblock In \emph{Proceedings of the IEEE/CVF international conference on computer vision}, pages 6023--6032, 2019.

\bibitem[Yun et~al.(2021)Yun, Oh, Heo, Han, Choe, and Chun]{yun2021re}
Sangdoo Yun, Seong~Joon Oh, Byeongho Heo, Dongyoon Han, Junsuk Choe, and Sanghyuk Chun.
\newblock Re-labeling imagenet: from single to multi-labels, from global to localized labels.
\newblock In \emph{Proceedings of the IEEE/CVF Conference on Computer Vision and Pattern Recognition}, pages 2340--2350, 2021.

\bibitem[Zhang et~al.(2018)Zhang, Cisse, Dauphin, and Lopez-Paz]{zhang2018mixup}
Hongyi Zhang, Moustapha Cisse, Yann~N Dauphin, and David Lopez-Paz.
\newblock mixup: Beyond empirical risk minimization.
\newblock In \emph{International Conference on Learning Representations}, 2018.

\bibitem[Zhang et~al.(2020)Zhang, Wu, Belatreche, Pan, Xie, Chua, Li, Qu, and Li]{related_delays2}
Malu Zhang, Jibin Wu, Ammar Belatreche, Zihan Pan, Xiurui Xie, Yansong Chua, Guoqi Li, Hong Qu, and Haizhou Li.
\newblock Supervised learning in spiking neural networks with synaptic delay-weight plasticity.
\newblock \emph{Neurocomputing}, 409:\penalty0 103--118, 2020.
\newblock ISSN 0925-2312.
\newblock \doi{https://doi.org/10.1016/j.neucom.2020.03.079}.
\newblock URL \url{https://www.sciencedirect.com/science/article/pii/S0925231220304665}.

\bibitem[Zhang et~al.(2023)Zhang, Liu, Yang, Song, Ye, Li, and Song]{zhang2023rfaconv}
Xin Zhang, Chen Liu, Degang Yang, Tingting Song, Yichen Ye, Ke~Li, and Yingze Song.
\newblock Rfaconv: Innovating spatial attention and standard convolutional operation.
\newblock \emph{arXiv preprint arXiv:2304.03198}, 2023.

\bibitem[Zhong et~al.(2020)Zhong, Zheng, Kang, Li, and Yang]{zhong2020random}
Zhun Zhong, Liang Zheng, Guoliang Kang, Shaozi Li, and Yi~Yang.
\newblock Random erasing data augmentation.
\newblock In \emph{Proceedings of the AAAI conference on artificial intelligence}, volume~34, pages 13001--13008, 2020.

\bibitem[Zhou et~al.(2019)Zhou, Zhao, Puig, Xiao, Fidler, Barriuso, and Torralba]{zhou2019semantic}
Bolei Zhou, Hang Zhao, Xavier Puig, Tete Xiao, Sanja Fidler, Adela Barriuso, and Antonio Torralba.
\newblock Semantic understanding of scenes through the ade20k dataset.
\newblock \emph{Int. J. Comput. Vis.}, 127\penalty0 (3):\penalty0 302--321, 2019.

\bibitem[Zhou et~al.(2023{\natexlab{a}})Zhou, Loy, and Dai]{zhou2023interpret}
Chong Zhou, Chen~Change Loy, and Bo~Dai.
\newblock Interpret vision transformers as convnets with dynamic convolutions.
\newblock \emph{arXiv preprint arXiv:2309.10713}, 2023{\natexlab{a}}.

\bibitem[Zhou et~al.(2023{\natexlab{b}})Zhou, Zhu, He, Wang, YAN, Tian, and Yuan]{spikformer}
Zhaokun Zhou, Yuesheng Zhu, Chao He, Yaowei Wang, Shuicheng YAN, Yonghong Tian, and Li~Yuan.
\newblock Spikformer: When spiking neural network meets transformer.
\newblock In \emph{The Eleventh International Conference on Learning Representations}, 2023{\natexlab{b}}.
\newblock URL \url{https://openreview.net/forum?id=frE4fUwz_h}.

\bibitem[Zhu et~al.(2023)Zhu, Zhao, and Eshraghian]{spikegpt}
Rui-Jie Zhu, Qihang Zhao, and Jason~K. Eshraghian.
\newblock Spikegpt: Generative pre-trained language model with spiking neural networks.
\newblock \emph{arXiv preprint arXiv:2302.13939}, 2023.

\bibitem[Zhu et~al.(2019{\natexlab{a}})Zhu, Qiu, Calderbank, Sapiro, and Cheng]{zhu2019scaling}
Wei Zhu, Qiang Qiu, Robert Calderbank, Guillermo Sapiro, and Xiuyuan Cheng.
\newblock Scaling-translation-equivariant networks with decomposed convolutional filters.
\newblock \emph{arXiv preprint arXiv:1909.11193}, 2019{\natexlab{a}}.

\bibitem[Zhu et~al.(2019{\natexlab{b}})Zhu, Hu, Lin, and Dai]{zhu2019deformable}
Xizhou Zhu, Han Hu, Stephen Lin, and Jifeng Dai.
\newblock Deformable convnets v2: More deformable, better results.
\newblock In \emph{Proc. IEEE/CVF Conf. Comput. Vis. Pattern Recog. (CVPR)}, pages 9308--9316, 2019{\natexlab{b}}.

\end{thebibliography}



\end{spacing}


\begin{appendices} 




\end{appendices}

\printthesisindex 

\newpage

\includepdf[pages=-]{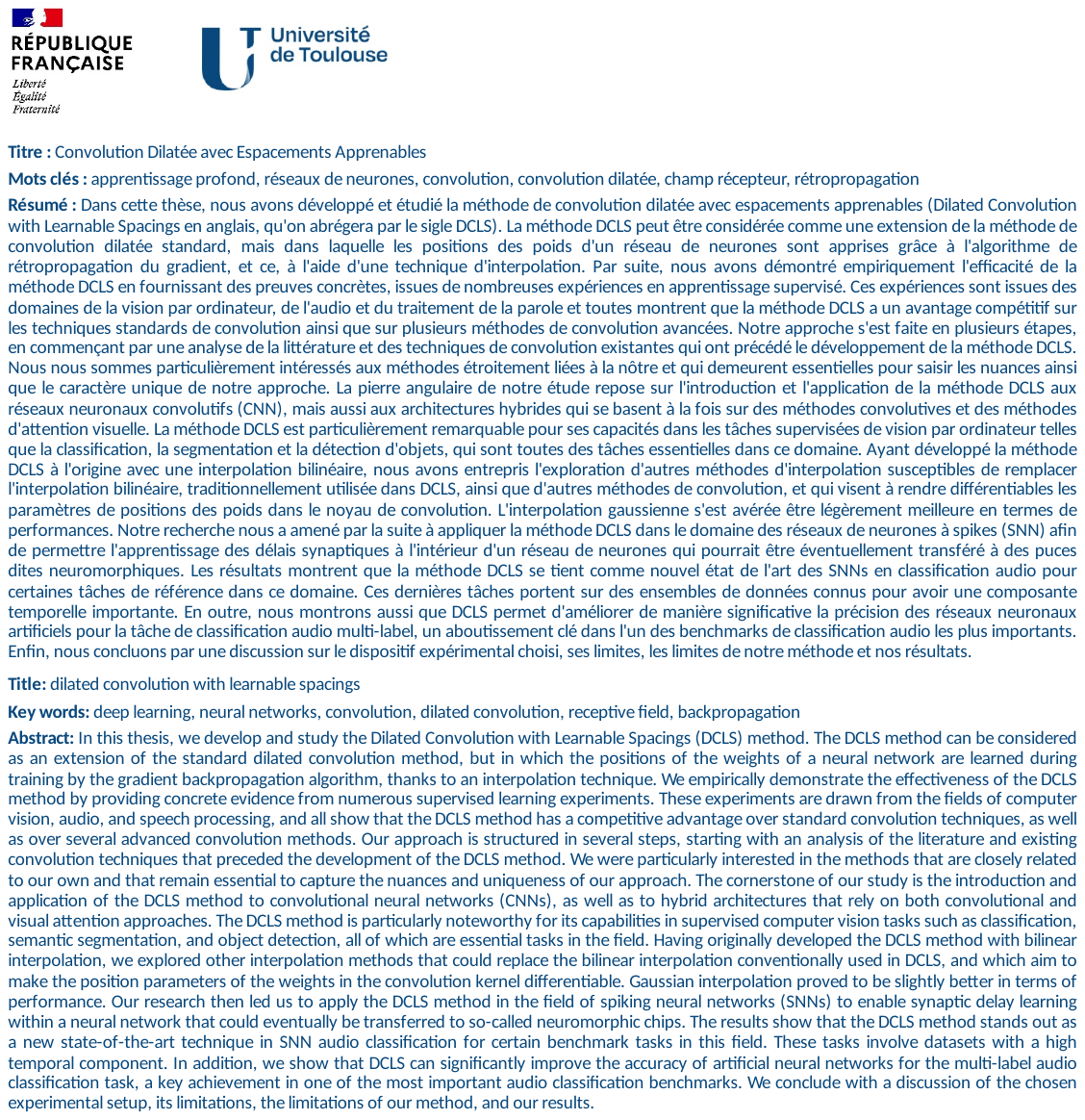}
\end{document}